\documentclass[final,3p,times]{elsarticle}

\usepackage{amssymb}
\usepackage{bm}
\usepackage{amsmath}
\usepackage[ruled]{algorithm2e}
\usepackage[usenames,dvipsnames]{color}
\usepackage{array}
\usepackage{subcaption}
\usepackage{graphicx}
\usepackage{lineno}
\usepackage{tikz}
\usepackage{booktabs}
\usepackage{float}
\usepackage{caption}
\usepackage{placeins}
\usepackage{longtable}
\usepackage{tcolorbox}

\usetikzlibrary{shapes, arrows.meta, positioning, calc}
\graphicspath{{Figures/}}
\biboptions{sort&compress}
\newcommand{\doi}[1]{\href{https://doi.org/#1}{\nolinkurl{#1}}}

\usepackage{hyperref}
\hypersetup{
    colorlinks,
    linkcolor={magenta},
    citecolor={blue},
    urlcolor={blue!80!black},
    breaklinks=true,
    plainpages=true
}

\newcommand{\cref}[3]{\hyperref[#2]{#1~\ref*{#2}{#3}}}
\newcommand{\crefs}[3]{\hyperref[#2]{#1~\ref*{#2}-\ref*{#3}}}
\newcommand{\colref}[2]{\hyperref[#2]{#1~\ref*{#2}}}
\newcommand{\eqnref}[1]{\colref{Eq.}{#1}}
\newcommand{\figref}[1]{\colref{Figure}{#1}}

\newcommand{\secref}[1]{\colref{Section}{#1}}

\newcommand{\tabref}[1]{\colref{Table}{#1}}

\begin{document}

\begin{frontmatter}

\title{A Vision-Language Model (VLM)-based Pipeline for End-to-End Procedural Modeling of Field-Grown Maize from Point Clouds}

\author[inst1]{Mozhgan Hadadi}
\author[inst3]{Talukder Z. Jubery}
\author[inst1,inst3,inst4]{Adarsh Krishnamurthy\texorpdfstring{\corref{cor1}}{}}
\ead{adarsh@iastate.edu}
\author[inst1,inst3,inst4]{Baskar Ganapathysubramanian\texorpdfstring{\corref{cor1}}{}}
\ead{baskarg@iastate.edu}

\affiliation[inst1]{organization={Department of Mechanical Engineering},
            addressline={Iowa State University},
            city={Ames},
            postcode={50011},
            state={Iowa},
            country={USA}}

\affiliation[inst3]{organization={Translational AI Center},
            addressline={Iowa State University},
            city={Ames},
            postcode={50011},
            state={Iowa},
            country={USA}}

\affiliation[inst4]{organization={Plant Science Institute},
            addressline={Iowa State University},
            city={Ames},
            postcode={50011},
            state={Iowa},
            country={USA}}
\cortext[cor1]{Corresponding Authors}

\begin{abstract}
Editable 3D models of field-grown crops support high-throughput phenotyping and \emph{in silico} breeding trials, but building them from scanned point clouds requires organ-level segmentation and fitting. Procedural generators can turn an organ-level parameter set into an analysis-suitable 3D model, but obtaining that set requires hours of manual tuning per plant or segmentation models trained on species-specific labels. We present an automated pipeline that reconstructs procedural maize models from raw 3D point clouds without manual tuning or species-specific training data. A multimodal vision-language model (VLM) annotates leaf midlines in rendered orthographic views. Deterministic geometric algorithms back-project the annotations onto the point cloud, merge them into 3D leaves by cross-view consensus, and grow the midlines to full blades on an orientation-weighted surface graph. Measured organ parameters populate a plant descriptor for a Non-Uniform Rational B-Spline (NURBS)-based procedural model generator. Each leaf surface is then refined against its scan points by differentiable NURBS fitting. The pipeline reached a median whole-plant Chamfer distance of $5.4$\,mm on $100$ genotypically diverse field-grown maize plants from the MaizeField3D dataset. The reconstructions were closer to the scans than those of an earlier semi-automated pipeline based on manual annotations. The pipeline recovered $1{,}017$ of $1{,}023$ ($99.4\%$) curated reference leaves at an intersection-over-union of at least $0.5$ without using those labels as input. These results show that VLM annotations become usable organ-level measurements when downstream geometric stages can correct them. This makes automated generation of editable 3D plant assets feasible at the scale of modern phenotyping experiments.
\end{abstract}

\begin{graphicalabstract}
    \centering
    \vspace{0.5em}
    \includegraphics[width=\textwidth]{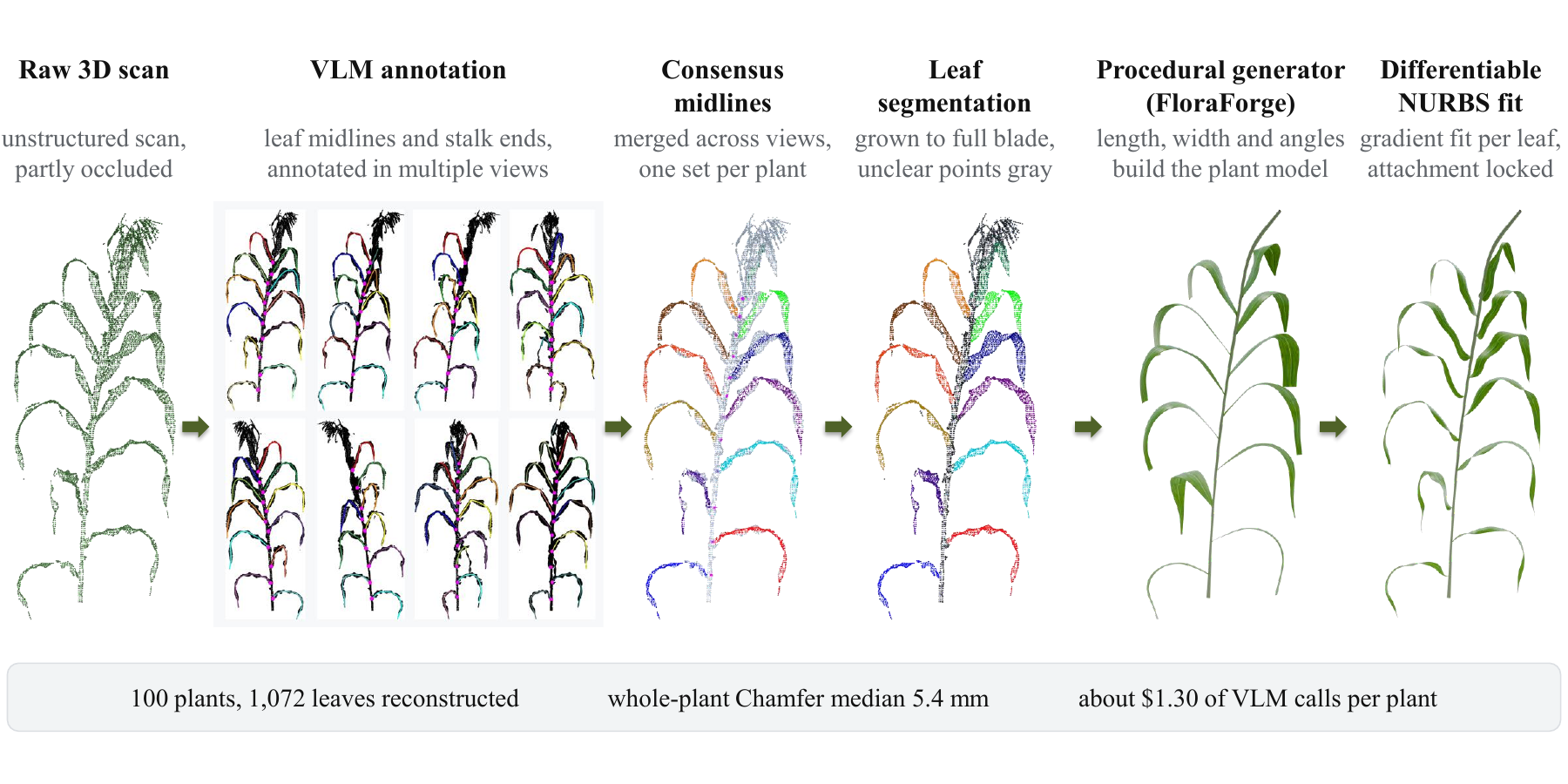}
\end{graphicalabstract}

\begin{highlights}
\item Zero-shot VLM annotates leaf midlines in rendered point-cloud views
\item Geometry-corrected zero-shot recovery detects 99.4\% of curated reference leaves
\item Measured organ traits populate an editable 3D procedural model without manual tuning
\item Differentiable NURBS refinement fits each leaf surface to its own 3D scan points
\item Median whole-plant Chamfer distance of 5.4 mm over 100 field-grown maize plants
\end{highlights}

\end{frontmatter}


\section{Introduction}
\label{sec:intro}

Accurate three-dimensional models of crop plants enable high-throughput phenotyping of morphological traits such as leaf area, canopy architecture, and phyllotaxy~\citep{paulus2019measuring,gibbs2016approaches}. These models also support functional-structural plant modeling (FSPM) of light interception, photosynthesis, and carbon partitioning~\citep{vos2010functional,louarn2020two,FunctionalStructuralModeling}, and they allow forecasting of plant stress state and exploration of what-if scenarios. In addition, 3D crop models underpin emerging agricultural digital twins for \emph{in silico} breeding trials and agronomic optimization~\citep{bailey2019helios,mitsanis2024digital}. Yet constructing such models from field-acquired sensor data remains a significant challenge in crop improvement~\citep{furbank2011phenomics,yang2020crop}, particularly for species such as maize (\textit{Zea mays})~\citep{schnable2009b73}.

Point clouds captured using terrestrial 3D scanning and Structure-from-Motion (SfM) photogrammetry~\citep{berger2017survey,westoby2012structure} provide rich 3D spatial data at the individual-plant scale. More recently, neural radiance field (NeRF)-based methods have emerged as an image-based route for reconstructing detailed plant geometry, including under challenging field conditions~\citep{arshad2024nerf}. Subsequent work has explored stationary-camera configurations aimed at making NeRF-based reconstruction more practical for high-throughput agricultural phenotyping~\citep{ku2025scnerf}. These advances make dense 3D plant geometry increasingly accessible. However, the resulting geometry, whether acquired directly by 3D scanning or reconstructed from imagery using SfM or NeRF, does not by itself provide a structured, organ-level representation of plant architecture. Such geometric data are inherently unstructured and can be noisy, occluded, and variable in point density~\citep{Li2019,Guerrero2018}. Extracting phenotypically meaningful parameters from these data (for example, node positions, internode lengths, leaf curvature profiles, attachment angles, and widths) requires a structured geometric representation of the plant's organ-level architecture~\citep{paulus2019measuring,jin2018stem}.

Procedural 3D modeling addresses this need by encoding plant geometry through a compact set of biologically interpretable parameters that can be algorithmically expanded into complete surface meshes~\citep{prusinkiewicz2012algorithmic,lindenmayer1968mathematical}. Neural reconstruction methods can recover detailed plant geometry and appearance, but their representations generally do not encode biologically meaningful organ parameters explicitly or permit controlled editing of individual traits such as node position, leaf angle, width, or curvature~\citep{cheng2025demeter,yang2025neuraleaf,liu2025neural}. In contrast, procedural models offer explicit parametric control, editability, and compatibility with downstream physics-based simulations~\citep{vos2010functional,bailey2019helios}. Recent procedural-crop pipelines obtain this structure by one of two routes: a search route that recovers it directly from sensor data, or a parameter-generation route that builds it from a specified set of organ-level parameters. CropCraft~\citep{zhai2024cropcraft} takes the search route, combining inverse procedural modeling~\citep{stava2014inverse,guo2020inverse} with neural radiance fields and Bayesian optimization at canopy scale. At the organ level, Non-Uniform Rational B-Spline (NURBS; \citep{piegl2012nurbs}) pipelines take the same route, achieving high-fidelity leaf surface fitting through hierarchical optimization on point clouds of field-grown maize~\citep{hadadi2025procedural}. Phytomer-based geometric models, constructed for maize from organ-level measurements~\citep{3DPhytomerBased}, follow the parameter-generation route. So do procedural generators such as FloraForge~\citep{hadadi2025floraforge}, which accept a configuration file (called the plant descriptor, PD) that specifies per-organ geometric parameters and produce analysis-ready 3D meshes for radiation transport, fluid dynamics, and finite-element analysis. However, the PD parameters are often specified \emph{manually} through iterative trial-and-error comparison against a target point cloud, a process reported to take 2 to 6~hours per plant~\citep{hadadi2025floraforge}. Manual specification is therefore impractical at the scale of modern phenotyping experiments involving hundreds or thousands of individual plants.

The core barrier to automating PD parameter adjustment is the lack of organ-level segmentation. To populate the PD, one must first identify the leaves, locate their midlines, find the positions of the stalk nodes, and measure the geometric attributes of the organs. Prior work on 3D plant point cloud segmentation has used deep learning architectures such as PlantNet~\citep{li2022plantnet}, PSegNet~\citep{li2022psegnet}, and PointNet-based maize pipelines~\citep{li2022maizeseg}, as well as Transformer-based networks~\citep{li2023pst} and annotation-efficient methods~\citep{luo2023eff3dpseg}. These methods are effective within their training distributions, but they require substantial labeled data that must be collected again whenever the species, growth stage, or acquisition setup changes~\citep{Zarei2024}. Collecting these labels for each new configuration is itself a bottleneck for fully automated phenotyping.

Multimodal vision-language models (VLMs) such as Gemini~\citep{gemini2024}, GPT-4o~\citep{openai2024gpt4o}, and Claude~\citep{anthropic2024claude} combine image understanding with instruction following. VLMs have shown strong zero-shot generalization across diverse visual domains~\citep{li2024foundation_ag,arshad2025leveraging}, including agricultural image analysis~\citep{xing2026zero}. Similarly, the Segment Anything Model (SAM)~\citep{kirillov2023sam} and Grounding DINO~\citep{liu2024grounding} have shown that promptable, open-vocabulary 2D segmentation is achievable without task-specific training. In the 3D domain, several recent works have shown that 2D foundation models can be lifted to three dimensions by rendering point clouds or NeRF scenes into 2D views, applying 2D segmentation or feature extraction, and back-projecting the results into 3D~\citep{peng2023openscene,cen2023segment,zhang2022pointclip,T_2024_CVPR}. This \emph{render-segment-back-project} paradigm has proven effective for general indoor and outdoor scenes. However, the paradigm has not been applied to plant phenotyping, where thin overlapping organs, fine-grained instance boundaries, and the need for metric parameter extraction require specialized treatment.

A VLM applied to rendered views of plant point clouds can annotate the visible portions of the leaves. For each supplied view, however, the VLM returns only a midline curve per detected leaf and candidate markers for stalk nodes. Real leaves are much wider than their midlines. In addition, no single view shows all leaves and nodes, because consecutive leaves often attach at nearby heights, two leaves may cross in projection, and the tassel hides the upper leaves. The VLM also fails in a few recurring ways: it draws two crossing leaves as one continuous stroke, breaks one leaf into disconnected fragments, draws more strokes than there are leaves in a crowded canopy, and omits the nodes that it cannot see. We start from per-view annotations of uncertain quality and recover one organ-level model of the plant in which each leaf identity, leaf point, and stalk attachment has either passed a stated geometric test or been reported as unresolved.

Converting the recovered organs into an editable surface model is a second challenge. In earlier work~\citep{hadadi2025procedural}, we approximated the leaves with NURBS surfaces fitted first by Particle Swarm Optimization (PSO)~\citep{pyswarmsJOSS2018} and then refined using a differentiable NURBS stage (NURBS-Diff; \citep{prasad2022nurbs}). This two-step procedure is accurate but computationally intensive, since the search dominates the runtime. In addition, the control points returned by the procedure cannot be interpreted as the quantities that an agronomist works with, such as leaf pitch, azimuth, node height, and width along the blade. In this work, we start the refinement from a surface built from measured organ parameters, so the PSO search is not required. The control points of this starting surface also correspond to organ features, so the leaves can be held attached to the stalk while their shapes are refined.

The methods closest to ours address only one of these two challenges. The supervised segmentation networks cited above return organ labels without organ parameters, and they require species-, stage-, and sensor-specific annotation. DeMeter~\citep{cheng2025demeter} and neural leaf-surface models~\citep{yang2025neuraleaf,liu2025neural} return geometry in a form that cannot be edited parameter by parameter. CropCraft~\citep{zhai2024cropcraft} recovers procedural crop structure by Bayesian optimization over a neural radiance field, but it operates at canopy scale and does not produce a per-organ descriptor of a single plant. These methods also do not report whether the attachment of a returned organ to the plant is geometrically plausible. Our pipeline does not require task-specific training on the target species, measures all parameters on the corrected per-leaf point clouds, and reports the outcome of the geometric tests for each leaf and node that it returns.

We validate organ recovery against the independently curated MaizeField3D leaf labels, but those labels are not supplied to the pipeline and are not used for task-specific training. This comparison tests whether the zero-shot annotation and geometric reconciliation recover the same physical organs without target-domain annotation. We evaluate end-to-end procedural reconstruction separately against PSO+NURBS-Diff, the prior workflow that maps the same type of point-cloud input to fitted NURBS plant surfaces. The claim of this work is that the organ segmentation required for procedural reconstruction can be obtained without first collecting target-domain organ labels. 

\begin{figure}[t!]
    \centering
    \includegraphics[width=\textwidth]{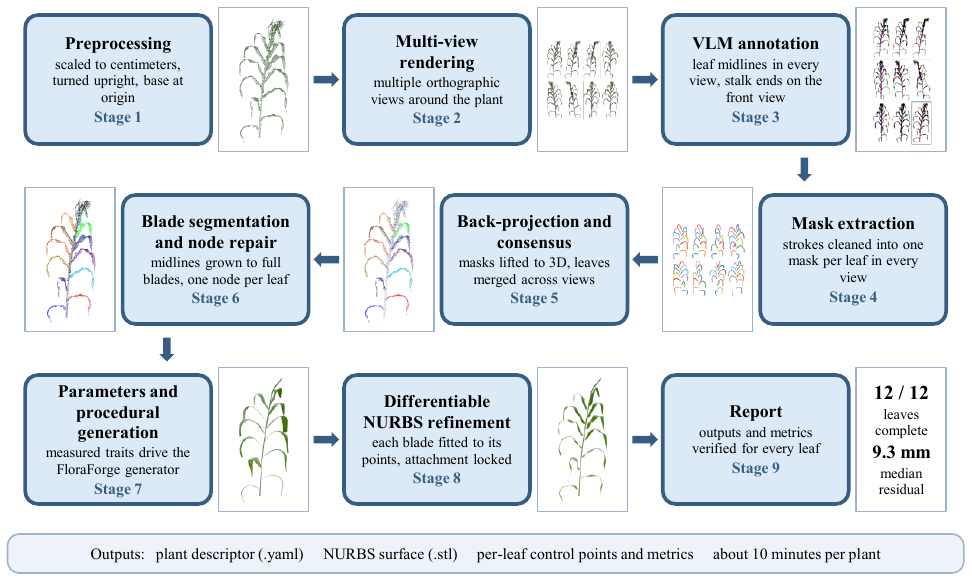}
    \caption{The nine pipeline stages (preprocess, render, VLM, masks, consensus, blade-level segmentation and node repair, parameters plus procedural plant generation (PPG), NURBS-Diff, and report), read along each row and turning back at the end of each row. Each stage is shown beside its own output for one of the evaluated plants (Kui2021). Rendering, annotation, masking, and consensus share one ordered set of views. Sparse per-view evidence is expanded to full blades, and misassigned blade material is corrected before parameters are measured. Node positions, leaf lengths, B-spline midline control points, width profiles, azimuths, pitches, and stalk diameters are measured from the corrected blade clouds and reconstructed stalk. These values automatically populate the FloraForge plant descriptor, from which the initial procedural model is generated. NURBS-Diff then uses gradient-based optimization to fit each attached blade surface more closely to its segmented points.}
    \label{fig:pipeline}
\end{figure}

In this work, we present an automated pipeline that converts a raw 3D scan of a maize plant into an editable procedural model (\figref{fig:pipeline}). The pipeline is designed to prevent any single unreliable input from determining the result. We first normalize the plant to a common scale and orientation, and then render it from multiple viewpoints. A VLM then annotates the leaf midlines in the rendered views (and also the stalk in one view). We re-cut the strokes where two leaves were drawn as one and rejoin them where one leaf was drawn as two. We then back-project the annotations onto the 3D points using the camera that rendered the corresponding view. Because the views disagree with one another, we reconcile the per-view leaves into a single set of leaves for the plant. Next, we grow the leaf labels from their midlines out to the full leaf surfaces. The leaves are attached to the reconstructed stalk, and the attachments are tested against stated geometric limits and then accepted, recomputed, or reported as unresolved. We measure the plant descriptor parameters (for example, leaf length, width profile, azimuth, pitch, and node height) on the corrected leaves, generate the procedural surface from the descriptor, and refine the leaf surfaces against their own points using NURBS-Diff. Finally, the outcomes of all tests are saved per leaf.

The key contributions of this work are:
\begin{enumerate}
    \item A zero-shot multi-view annotation protocol in which a vision-language model marks leaf midlines and leaf-stalk junctions in a set of orthographic views.
    We validate the returned images geometrically before accepting them, and a failing image triggers a new request. An annotation that still fails its checks is recorded in the validation log for human evaluation.

    \item A geometric reconstruction method that back-projects these annotations through the rendering cameras and reconciles them across views into globally consistent 3D leaves. The method grows the sparse midlines out to full blades, and it corrects merged or misassigned blade material only when the correction passes its own test.

    \item A constrained node rule that tests leaf-to-stalk attachments against stated limits on their distance to the leaf base and their spacing from neighboring nodes. Attachments that pass are kept, and those that fail are recomputed by projecting the leaf base onto the reconstructed stalk. 

    \item An automatic scan-to-descriptor stage that measures the stalk centerline, stalk diameters, node positions, leaf lengths, midline control points, width profiles, azimuths, and pitches from the corrected blades. These values are written into the plant descriptor used by the procedural generator, with no manual parameter tuning.

    \item A differentiable NURBS refinement that is initialized from the measured procedural surface and fits each blade to its own points while the attachment control row stays fixed. The measured initialization replaces the stochastic search that our earlier pipeline required~\citep{hadadi2025procedural}.
\end{enumerate}

The rest of the paper is organized as follows. \secref{sec:methods} details the nine stages of the pipeline, from preprocessing to reporting. \secref{sec:dataset} describes the dataset. The main algorithmic contributions are VLM-based annotation (\secref{sec:vlm}), back-projection and cross-view consensus (\secref{sec:backproject}), blade-level segmentation and node repair (\secref{sec:leafseg}), parameter extraction and procedural generation (\secref{sec:params_ppg}), and differentiable NURBS refinement (\secref{sec:nurbsdiff}). \secref{sec:results} presents the results, including the comparison with the earlier PSO+NURBS-Diff workflow, the unresolved cases, and future directions. We conclude in \secref{sec:conclusion}.

\section{Methods}
\label{sec:methods}

Our pipeline reconstructs an editable 3D procedural model from a single maize point cloud in nine stages (\figref{fig:pipeline}), namely (1)~preprocessing, (2)~rendering, (3)~VLM annotation, (4)~mask extraction, (5)~cross-view consensus, (6)~blade-level segmentation and node repair, (7)~parameter extraction plus FloraForge~\citep{hadadi2025floraforge} procedural plant generation (PPG), (8)~NURBS-Diff~\citep{prasad2022nurbs} refinement, and (9)~reporting. All algorithmic thresholds are fixed before the run and applied unchanged to all plants. These thresholds are specified partly in the pipeline's published parameter profile and otherwise by defaults in the source code. The values quoted in this section are those of the reported run, and \tabref{tab:parameters} summarizes the principal values with the stage each governs and where each is set. Only stage~3 calls an AI foundation model. Stages 4 to 9 are deterministic functions of their inputs. Several downstream stages can correct the annotations that they receive. Specifically, the mask stage re-cuts strokes that the model drew as one, consensus discards leaves supported by only one view, blade-level segmentation uses the consensus midlines as seeds to recover the full blades and then rebuilds the final midlines from the segmented blade clouds, and the node rule can replace invalid or missing annotated node positions with projections onto the reconstructed stalk.

\paragraph{Reproducibility}
Each run saves the configuration, inputs, and outputs of all stages. Any number reported below can therefore be traced to the stage that produced it. A stage whose inputs or outputs no longer match the saved versions is recomputed. A stage that cannot produce its required downstream artifacts stops the run for that plant. VLM image-validation failures are handled separately as described in \secref{sec:vlm}: if no attempt passes, the best attempt is retained and its failure is logged. The model queried in stage~3 is not seeded. We therefore claim reproducibility of stages 4 to 9 given the same saved annotations. The annotations themselves may differ between runs.

\subsection{Dataset}
\label{sec:dataset}

\begin{figure}[t!]
    \centering
    \newlength{\pcdwidth}
    \setlength{\pcdwidth}{0.19\linewidth}
    \newcommand{\pcdcell}[2]{%
        \begin{subfigure}[b]{\pcdwidth}
            \centering
            \includegraphics[width=\linewidth]{#1}
            \caption{#2}
        \end{subfigure}}
    \pcdcell{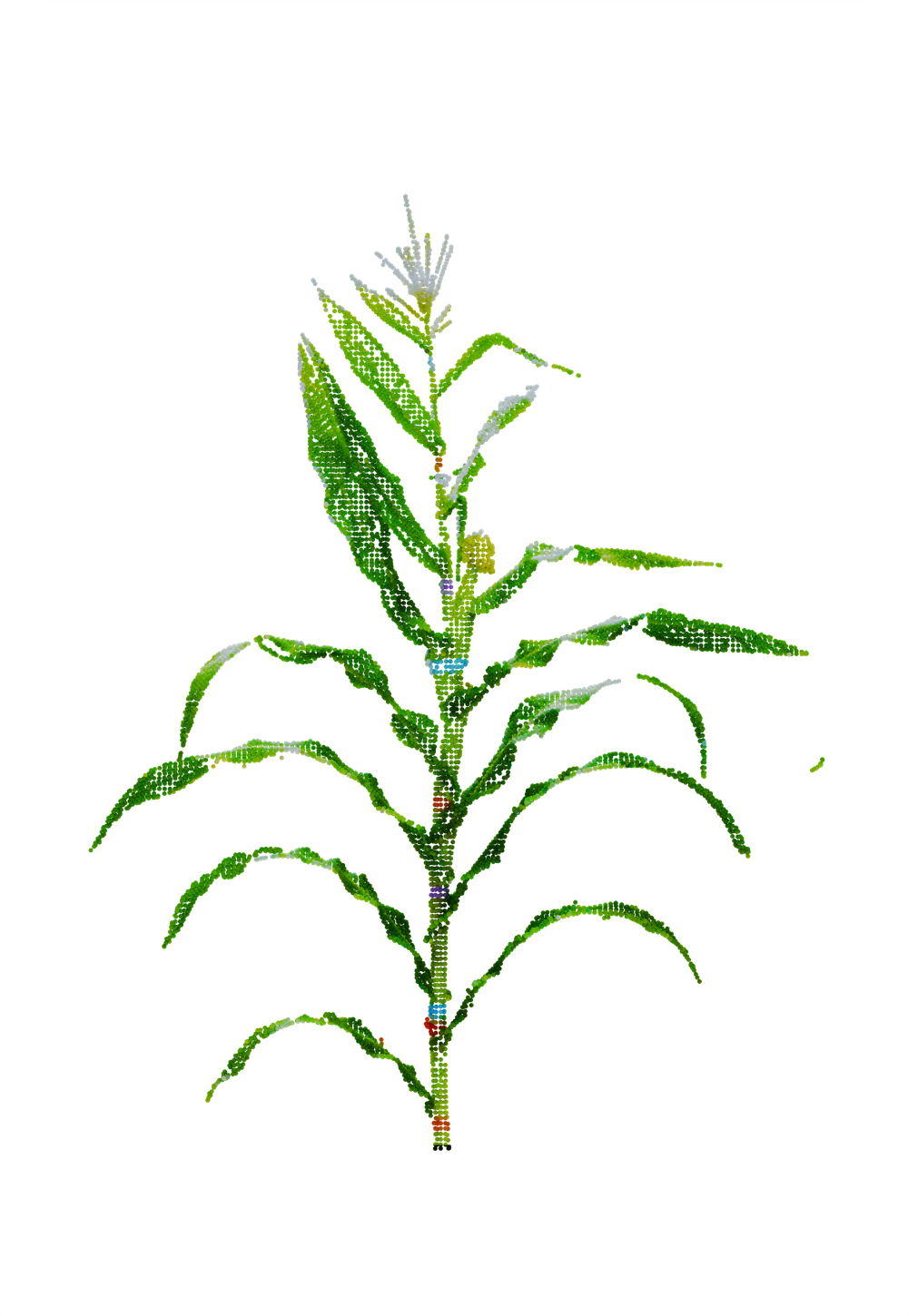}{M162W}\hfill \pcdcell{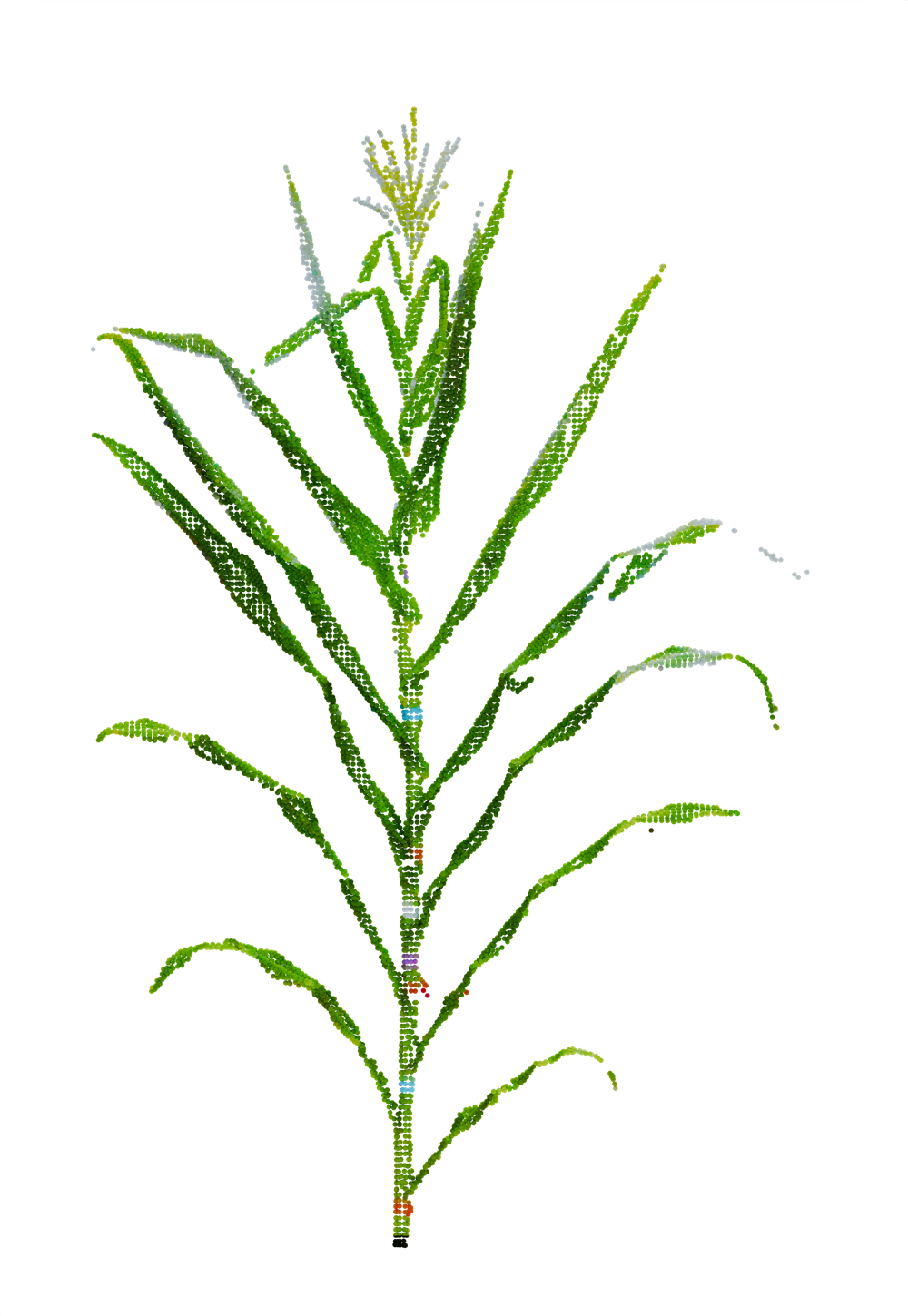}{CML69}\hfill \pcdcell{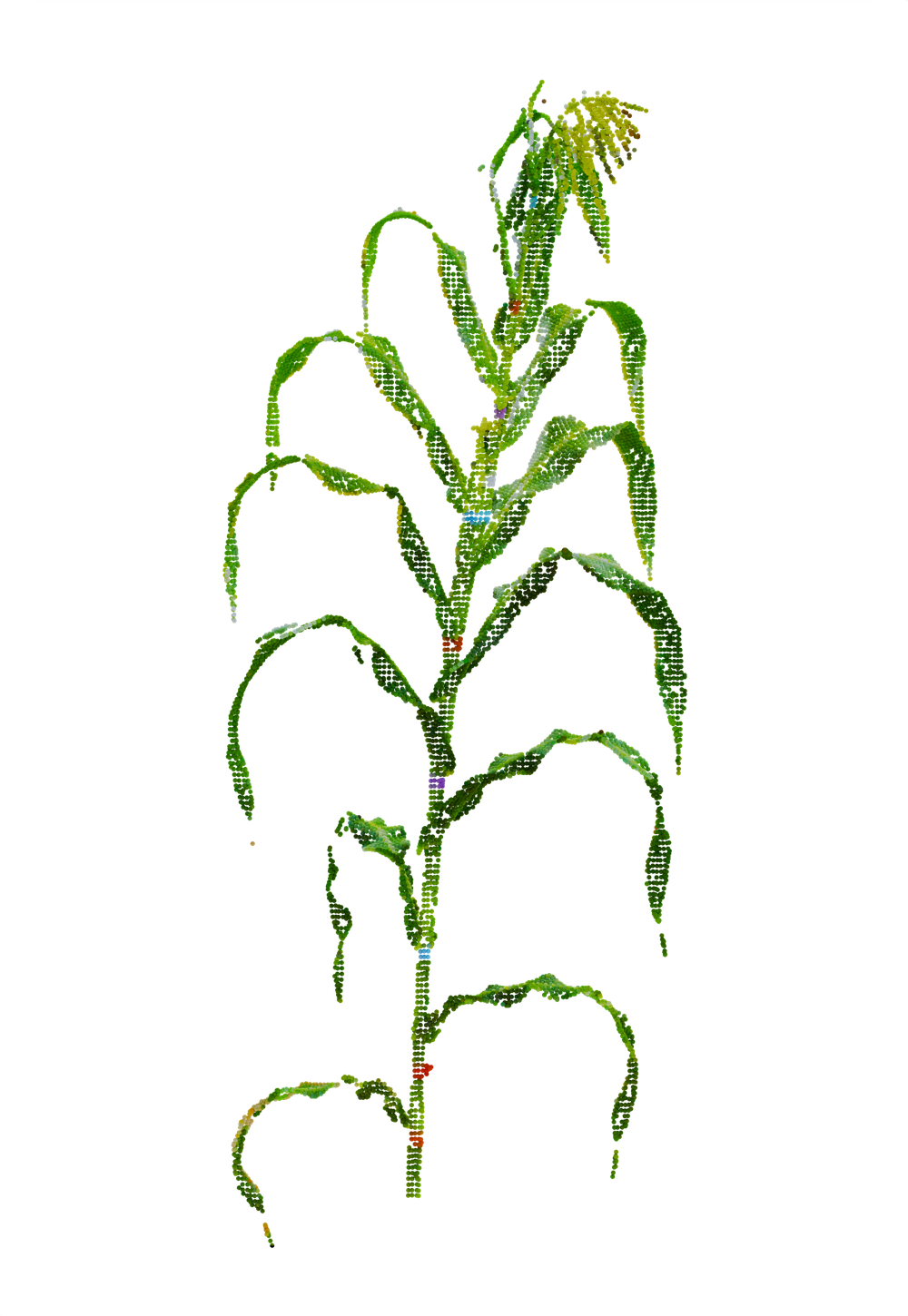}{Kui2021}\hfill \pcdcell{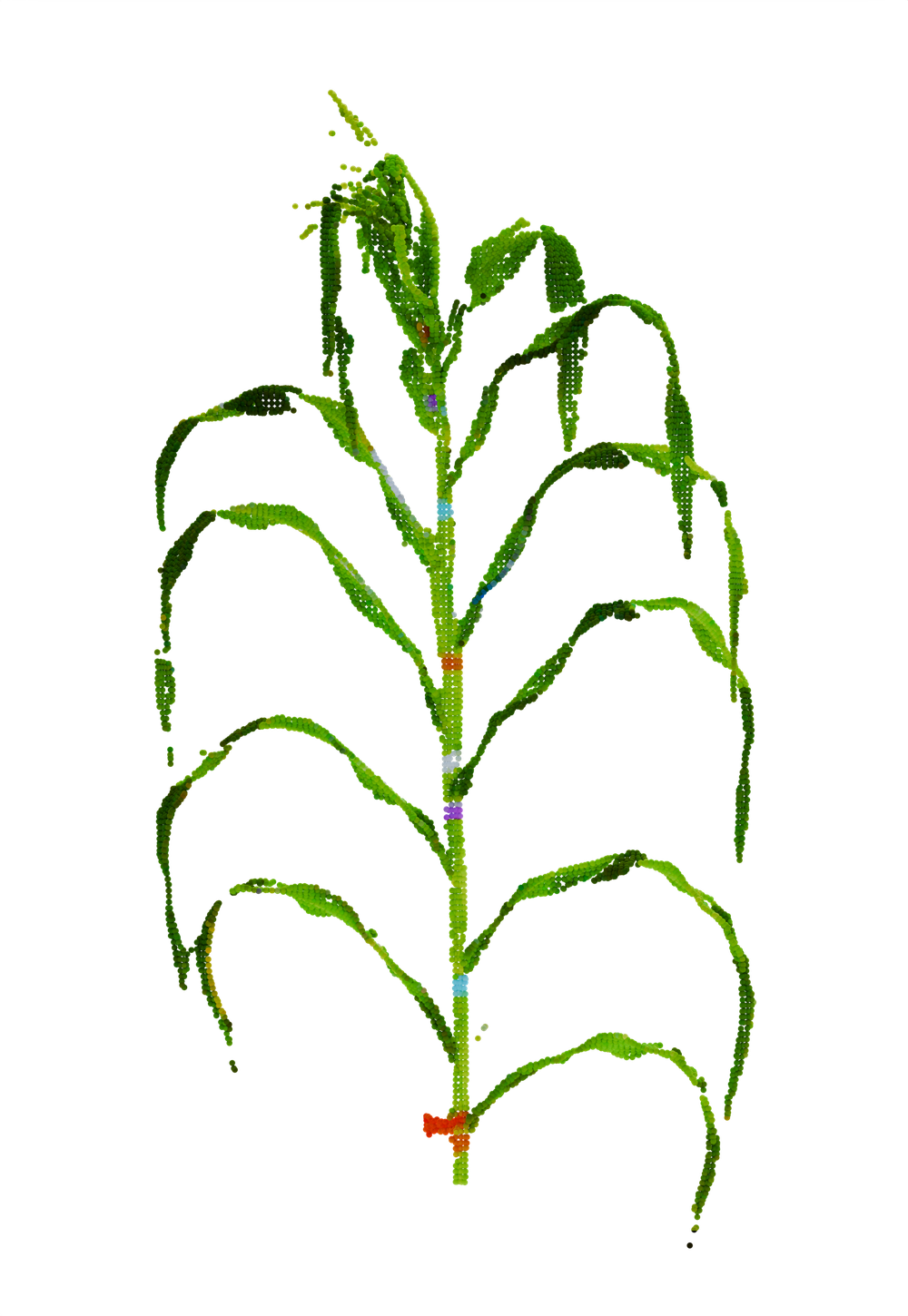}{CML238}\hfill \pcdcell{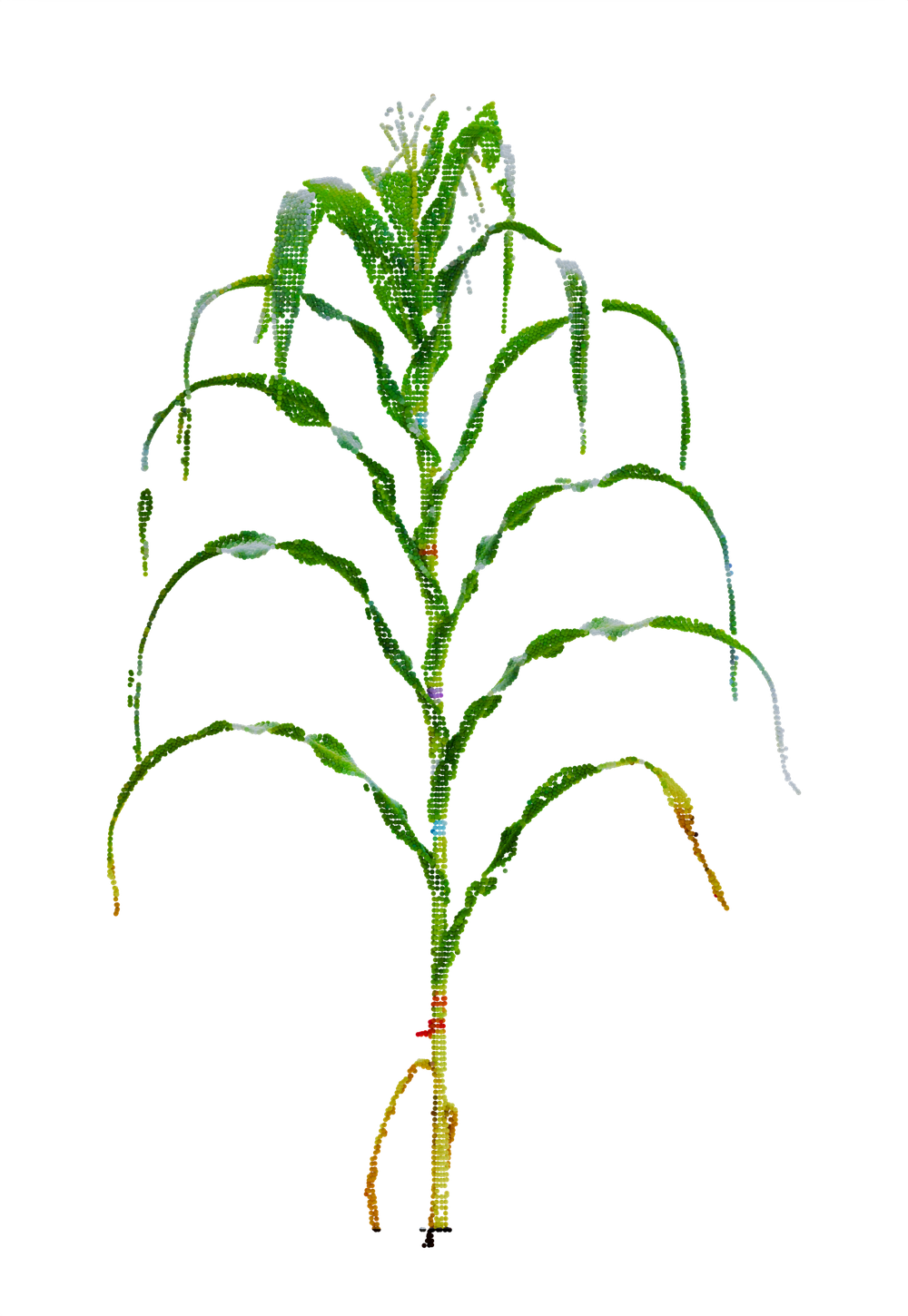}{CML331}

    \vspace{0.6em}
    \pcdcell{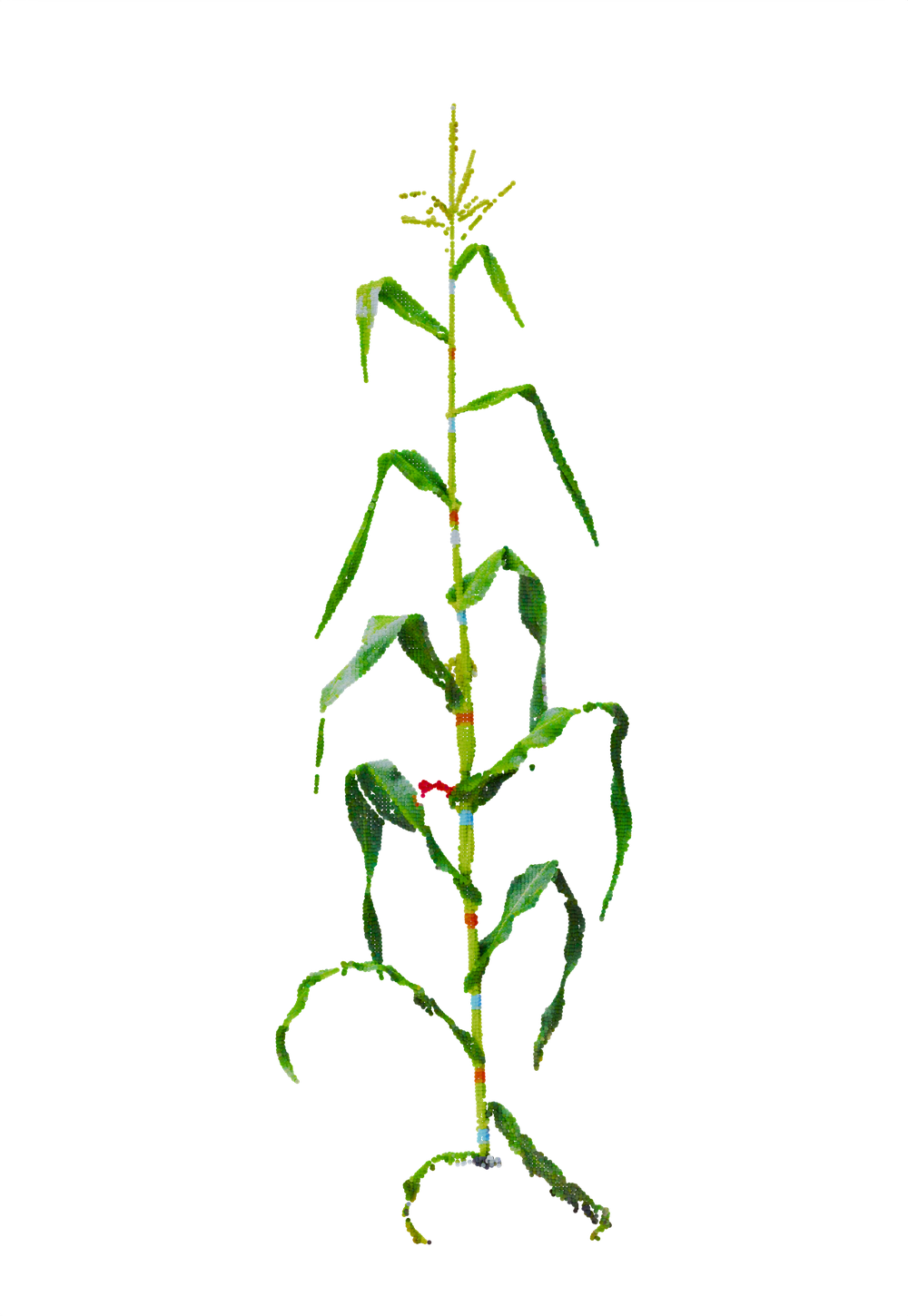}{4226}\hfill \pcdcell{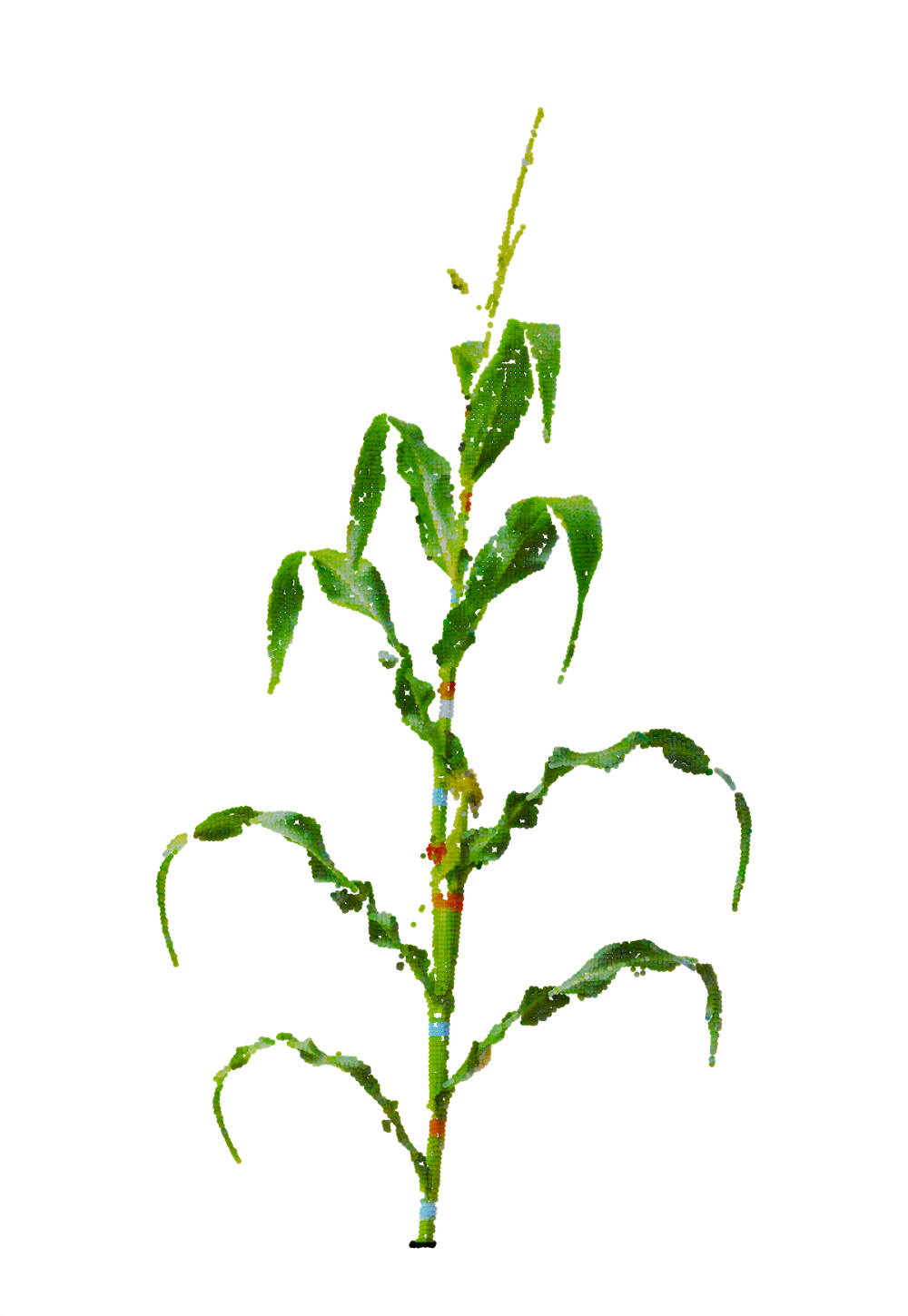}{78551S}\hfill \pcdcell{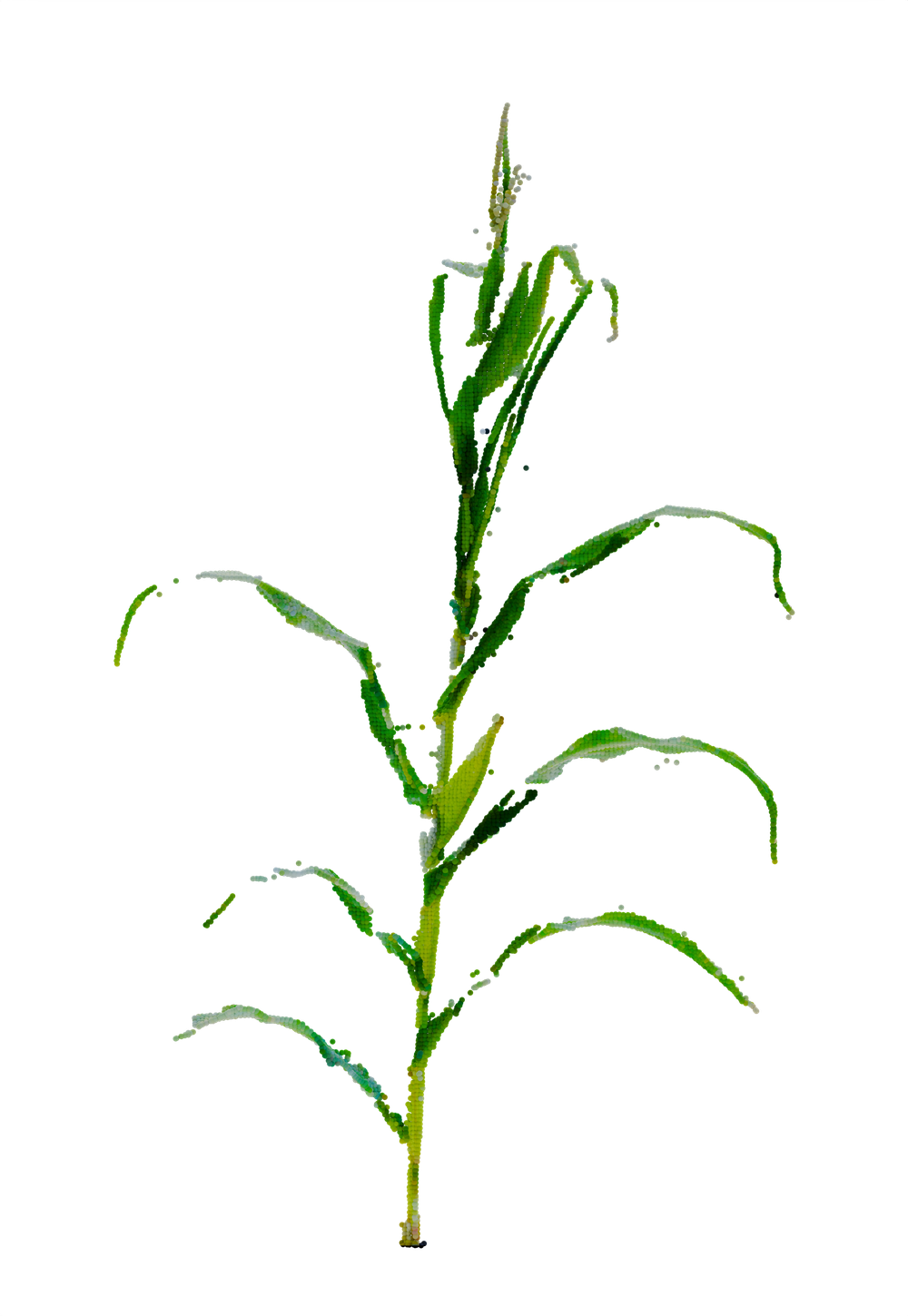}{A679}\hfill \pcdcell{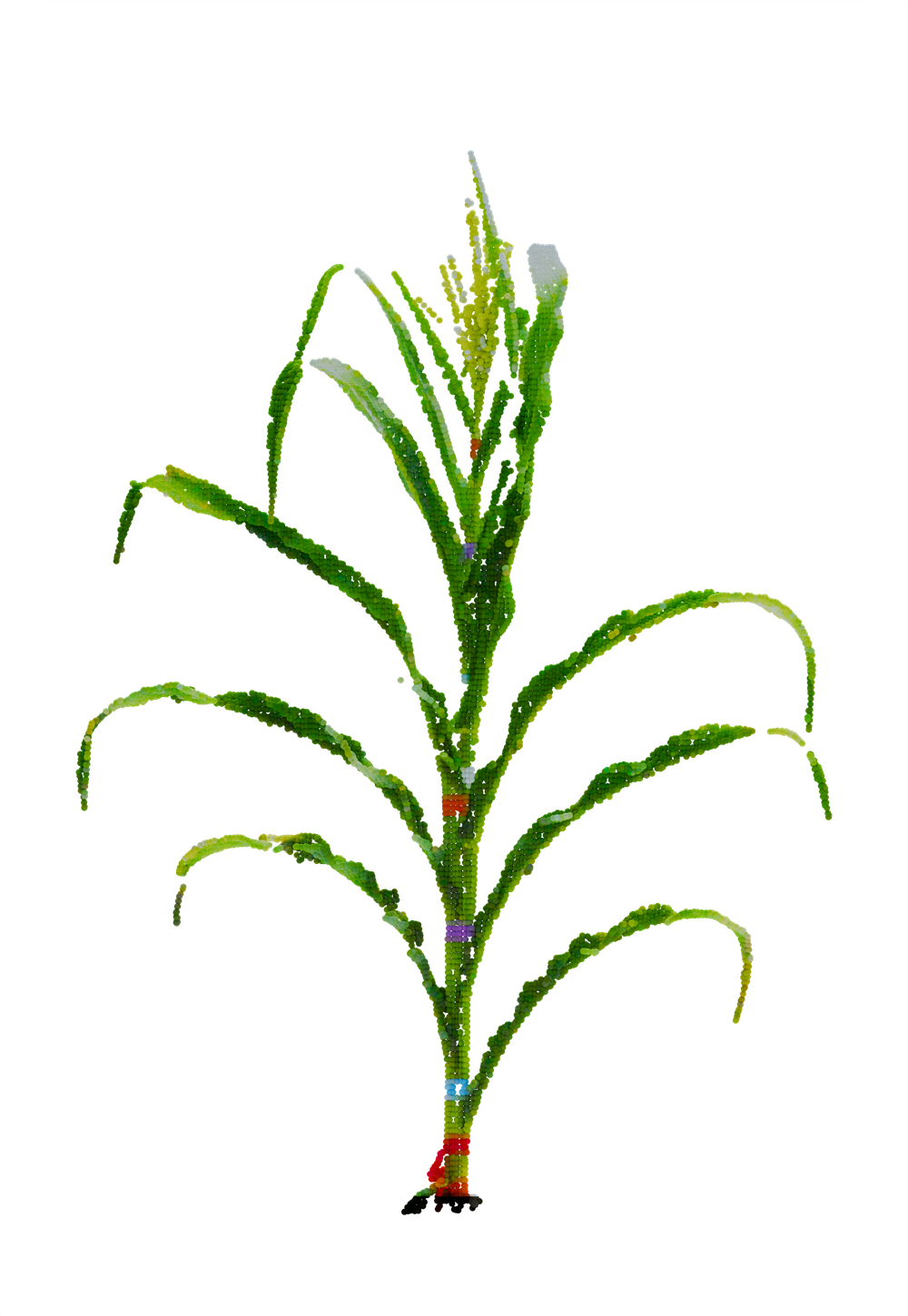}{CI90C}\hfill \pcdcell{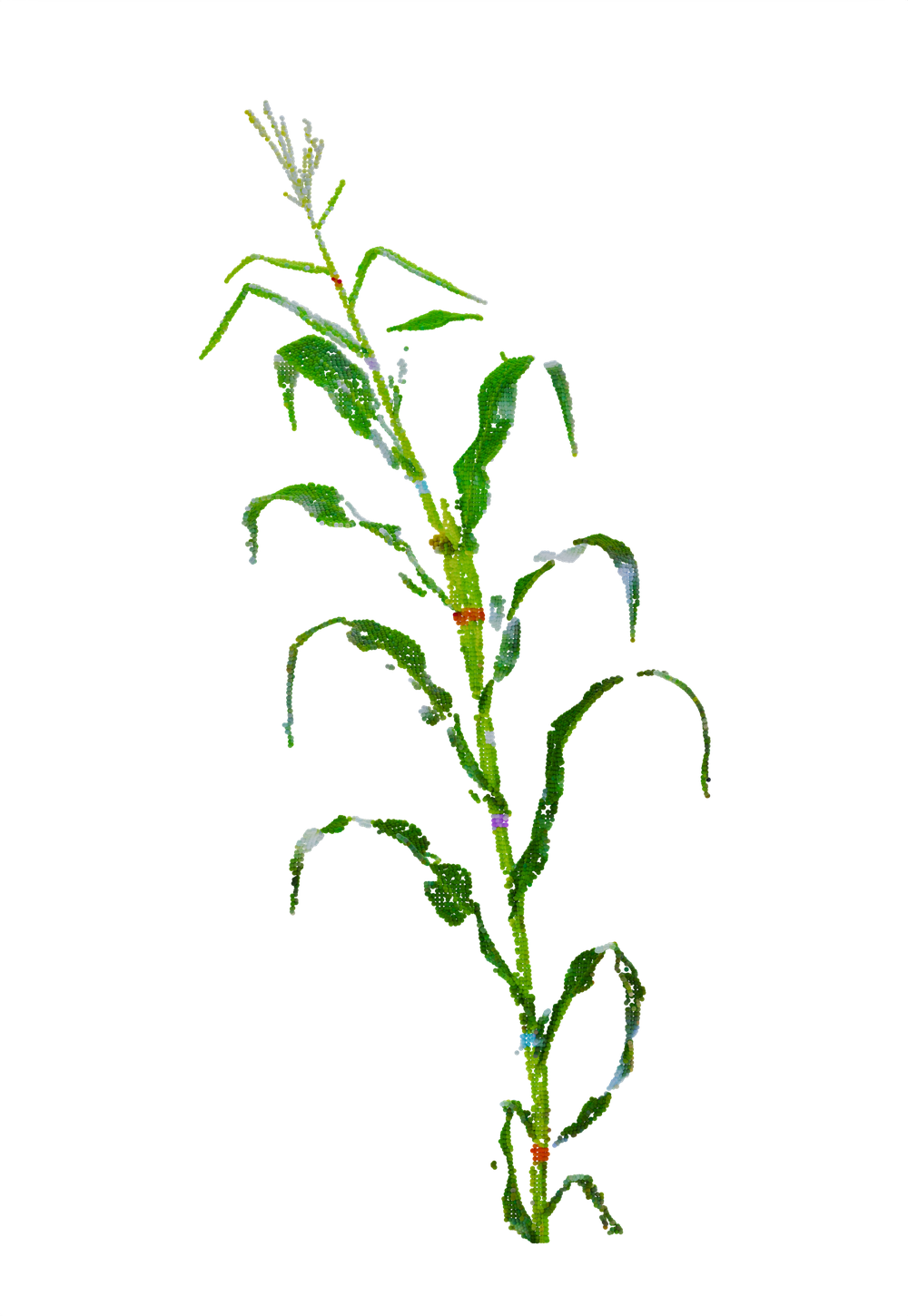}{T8}
    \caption{The ten plants reported leaf by leaf (\tabref{tab:plants}), shown as the orthographic front view of each canonicalized point cloud in the colors recorded by the scanner. Each cloud is the input from which the pipeline of \secref{sec:methods} reconstructs a procedural model.}
    \label{fig:dataset_pcds}
\end{figure}

We evaluate the pipeline on $100$ field-grown maize plants from the MaizeField3D dataset~\citep{maizefield3d}, a curated collection of terrestrial point cloud scans spanning multiple genotypes of the Shoot Apical Meristem diversity panel~\citep{thompson2015diversity,leiboff2015genetic}. All plants are processed end to end with a single fixed parameter set and no per-plant tuning. Ten of these plants, each a distinct genotype (\tabref{tab:plants}, \figref{fig:dataset_pcds}), are also reported leaf by leaf in \secref{sec:results}. We selected these ten plants to span the morphological range of the panel in plant height, leaf angle distribution, and canopy compactness. This lets us examine the per-organ behavior of the pipeline on diverse architectures. The plants were grown during the summer of 2021 on Iowa State University's Curtis Farm in Ames, IA, under agronomically relevant crop management. The plants were harvested around anthesis (reproductive stage) by cutting at ground level and brought indoors for 3D scanning. At anthesis, leaf morphology, phyllotaxy, and canopy architecture are fully expressed, and leaf angles, dimensions, and curvatures are well defined. This growth stage is therefore well suited to phenotyping and digital reconstruction.

\begin{table}[b!]
    \centering
    \caption{The ten plants reported leaf by leaf, identified by genotype, with the number of leaves recovered by the pipeline. All scans are from the MaizeField3D release~\citep{maizefield3d}. These ten plants account for $118$ of the $1{,}072$ leaves reconstructed across the $100$ plants.}
    \label{tab:plants}
    \setlength{\extrarowheight}{2pt}
    \footnotesize
    \begin{tabular}{@{}clc@{\hspace{2.2em}}clc@{}}
        \toprule
        \# & Genotype & Leaves & \# & Genotype & Leaves \\
        \midrule
        1 & M162W   & 15 & 6  & 4226   & 12 \\
        2 & CML69   & 15 & 7  & 78551S & 8 \\
        3 & Kui2021 & 12 & 8  & A679   & 8 \\
        4 & CML238  & 12 & 9  & CI90C  & 10 \\
        5 & CML331  & 12 & 10 & T8     & 14 \\
        \bottomrule
    \end{tabular}
\end{table}

The 3D point clouds were acquired with a Faro Focus S350 terrestrial laser scanner. The MaizeField3D study reports an angular resolution of $0.011^\circ$ and an approximate point spacing of $1.5$\,mm at a range of $10$\,m for this scanner~\citep{maizefield3d}. Each cloud is stored as a colored 3D PLY file in the scanner's metric frame, and the pipeline normalizes each cloud to a canonical scale and orientation before rendering (\secref{sec:preprocess}). Each scan is decimated to a working resolution of approximately $10^4$ points before it enters the pipeline. Across the $100$ plants, the median nearest-neighbor spacing at this working resolution is $4.1$\,mm. The reduction keeps a whole plant within the memory and runtime limits of the per-point graph constructions in \secref{sec:leafseg}. At this resolution, blade width and curvature remain well sampled at the organ scale of the plant descriptor. Gaps and occlusions still occur due to leaf overlap, scanning angle, and variations in leaf reflectance. Leaves with sharp twists or extreme curvature are also challenging to reconstruct. The recovered leaf counts range from $8$ to $15$ across the $100$ plants ($1{,}072$ reconstructed leaves in total), and the ten plants reported leaf by leaf span the same range. The source point clouds are publicly available through the MaizeField3D release~\citep{maizefield3d}.

\subsection{Preprocessing}
\label{sec:preprocess}

Field scans are recorded in the scanner's own frame and units, so they are preprocessed before rendering. Coordinates are first multiplied by a fixed factor of $100$ to convert meters to centimeters, the unit used for downstream geometric quantities such as leaf width, node spacing, and stalk radius, and for the plant descriptor. The horizontal axes are then canonicalized by two-dimensional principal-component analysis of the $xy$ coordinates. We rotate the dominant horizontal direction onto $+x$ to present the plant's widest profile to the front camera, and resolve the residual $180^\circ$ ambiguity of the principal axis by requiring the mean $x$ coordinate to be positive. We rotate the cloud instead of only swapping its axes, because an axis swap cannot align an arbitrarily oriented canopy with $+x$. The canonicalization rotates the cloud only about the vertical axis and leaves the $z$ coordinates unchanged. The height origin is determined by the separate pass described next.

A separate pass places the origin at the stalk base. It takes the lowest $20\%$ of the points by height and computes the median horizontal position of this slice. This median defines the center of a vertical core column, which consists of the points of the whole cloud whose horizontal distance from the center falls within the $20$th percentile of all such distances. The \emph{lowest} point of this column is taken as the stalk base and translated to $(0,0,0)$. The median locates only the core and is not used as the base, because the median height of a low slice lies above the ground and the attachment heights measured later are referenced to the estimated stalk base. The output is a set of PLY files in a canonical, centimeter-scale, base-aligned frame assumed by all subsequent stages.

\subsection{Multi-view orthographic rendering}
\label{sec:render}

\begin{figure}[t!]
    \centering
    \newlength{\viewwidth}
    \setlength{\viewwidth}{0.20\linewidth}
    \newcommand{\viewcell}[2]{%
        \begin{subfigure}[b]{\viewwidth}
            \centering
            \includegraphics[width=\linewidth]{#1}
            \caption{#2}
        \end{subfigure}}
    \viewcell{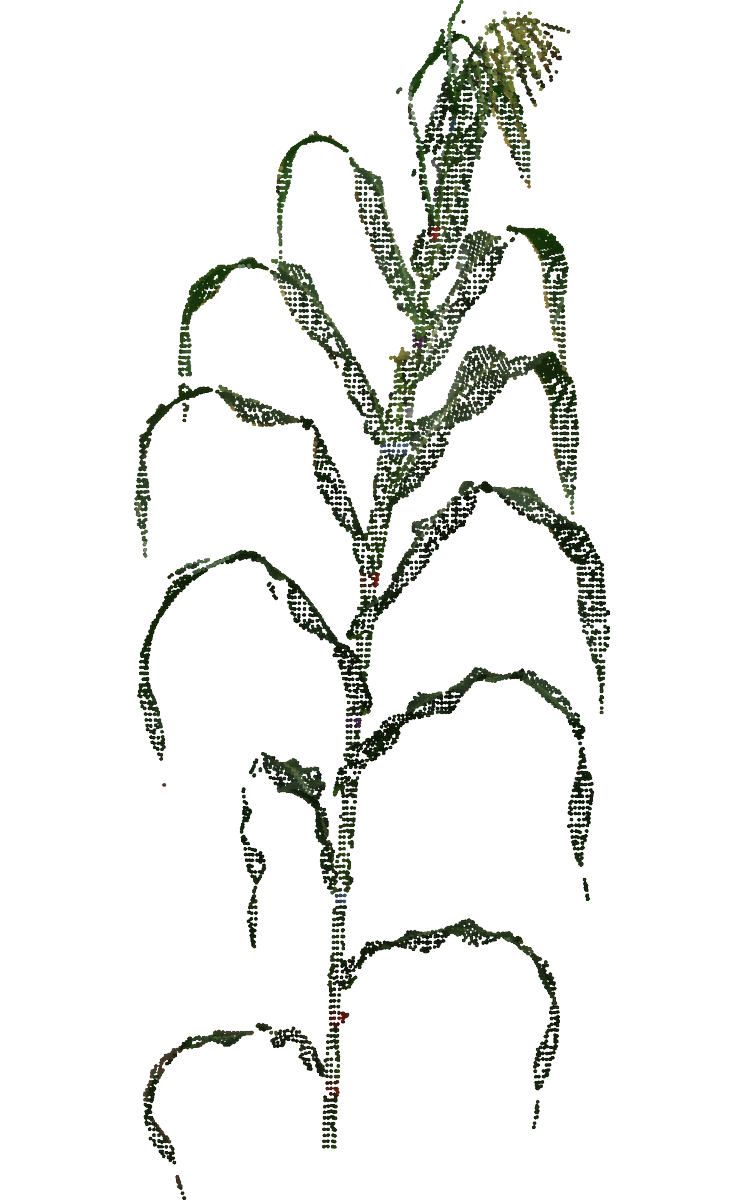}{front ($0^\circ$)}\hfill \viewcell{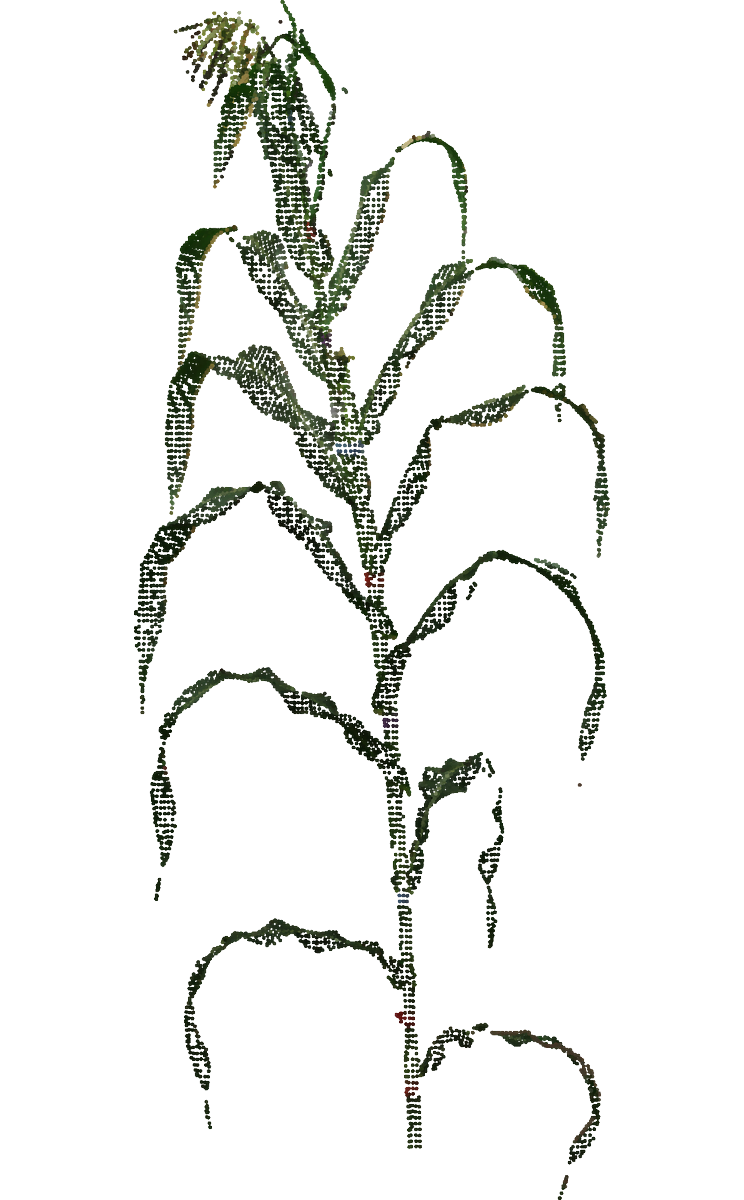}{back ($180^\circ$)}\hfill \viewcell{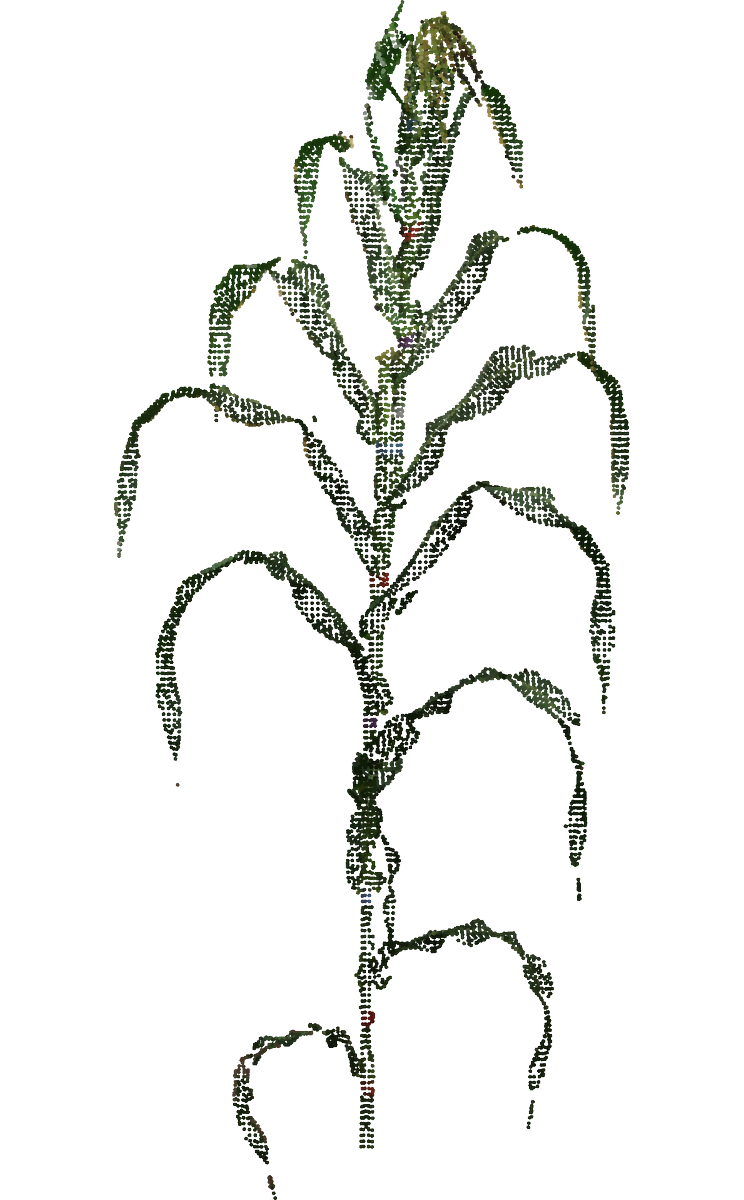}{$+30^\circ$}\hfill \viewcell{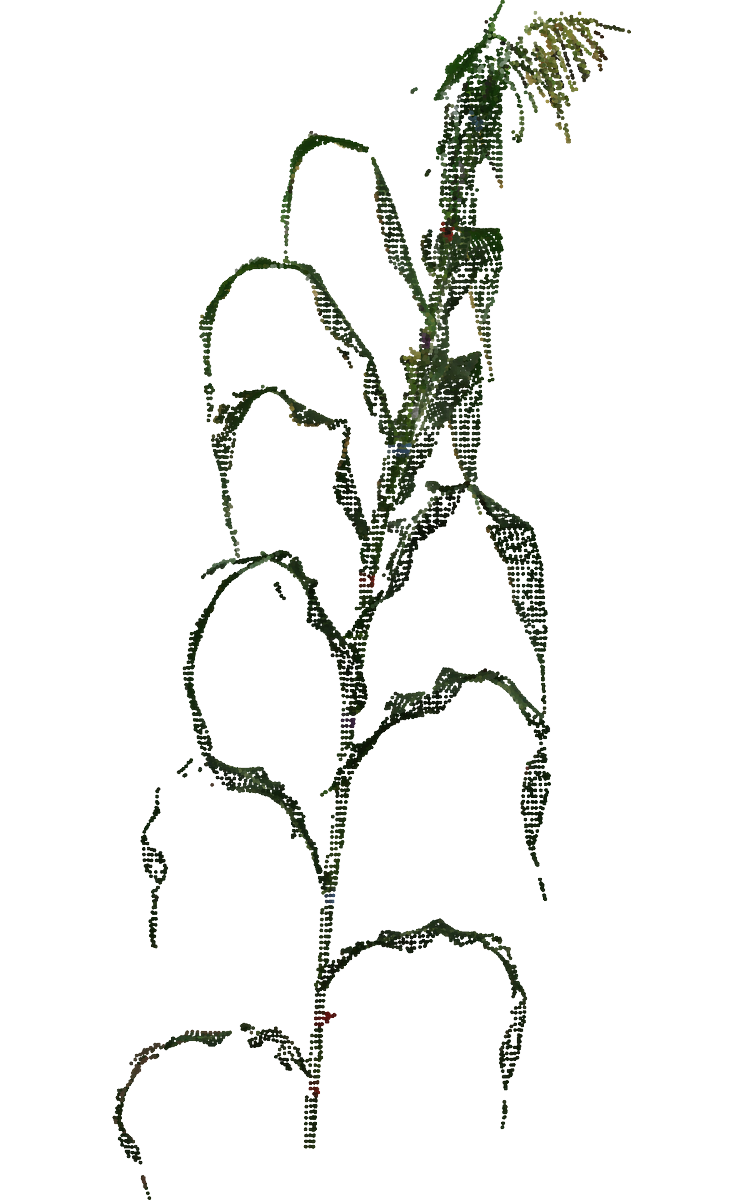}{$-30^\circ$}

    \vspace{0.25em}
    \viewcell{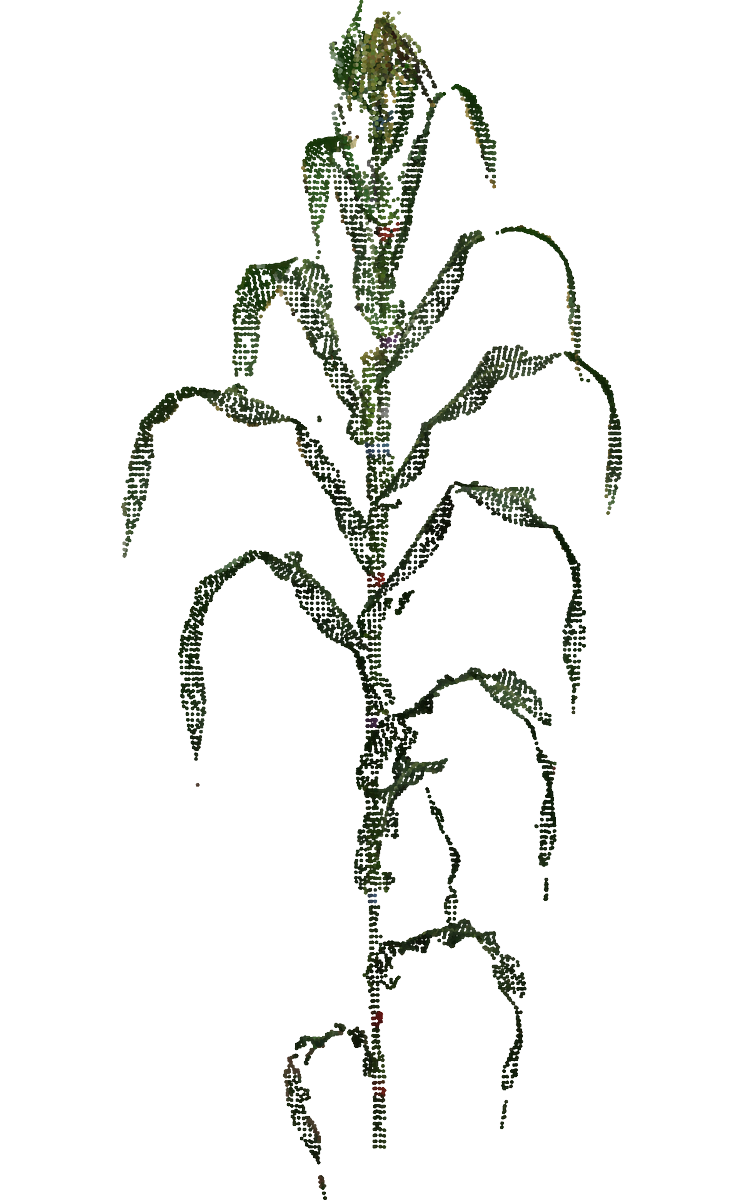}{$+45^\circ$}\hfill \viewcell{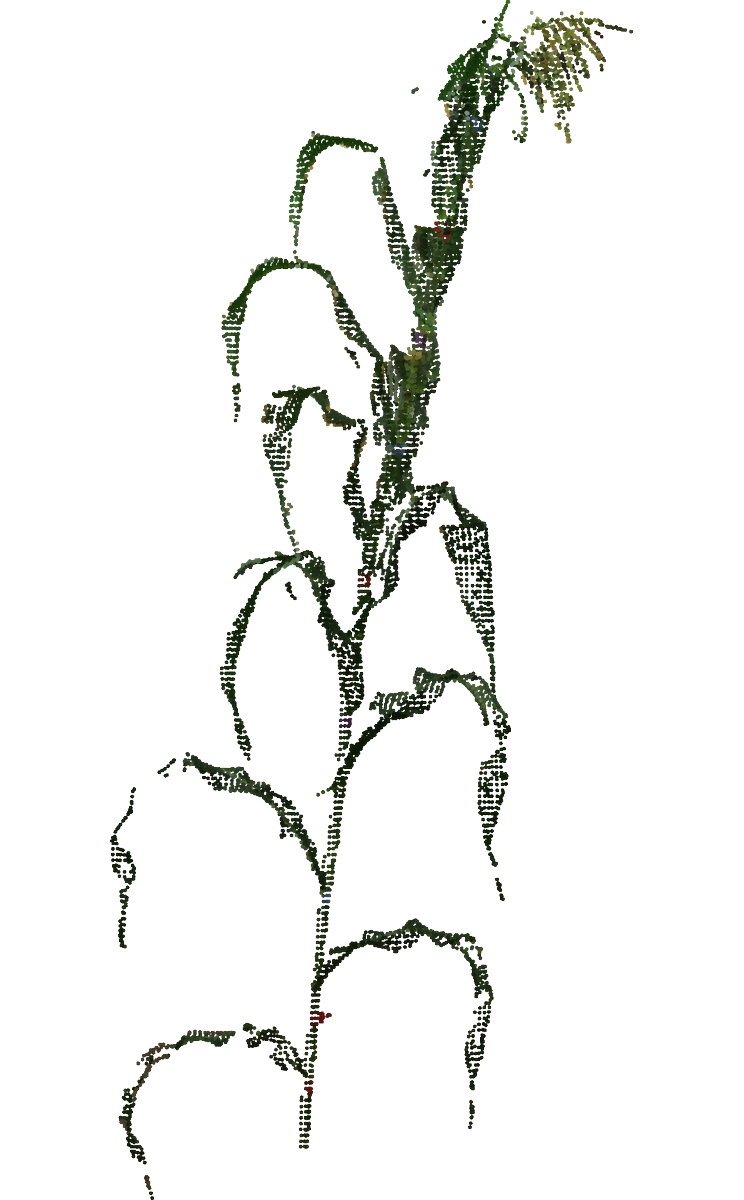}{$-45^\circ$}\hfill \viewcell{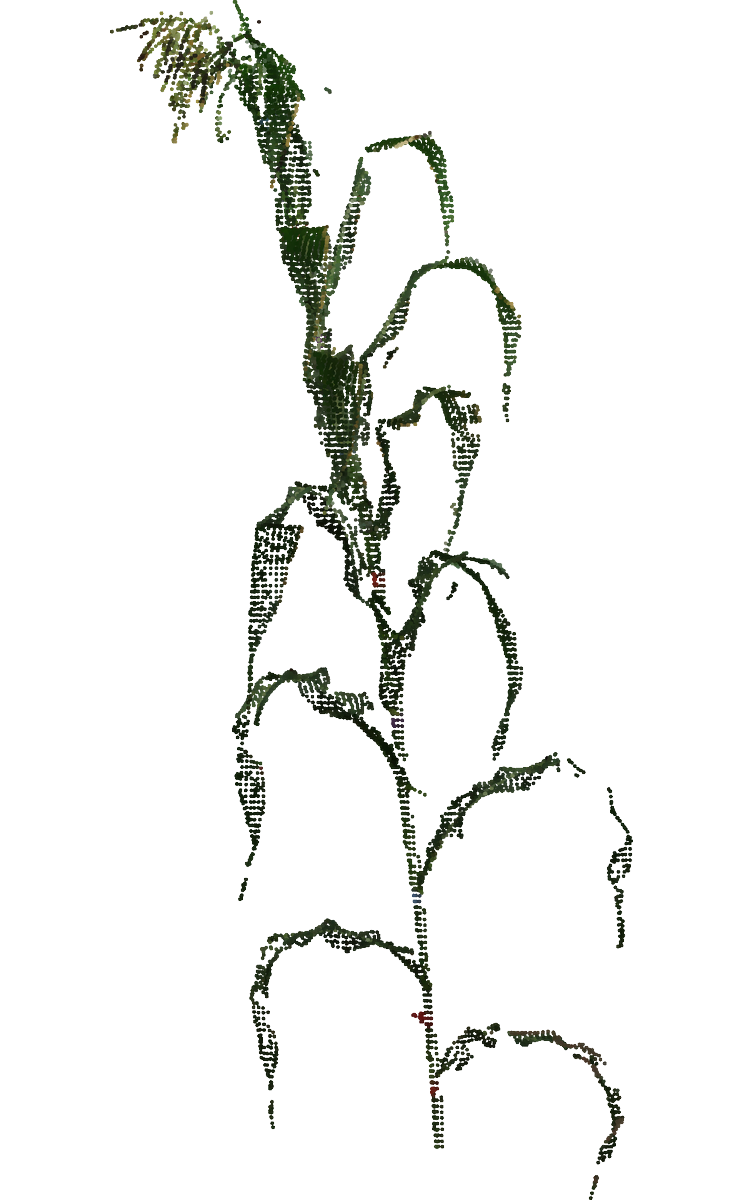}{$+135^\circ$}\hfill \viewcell{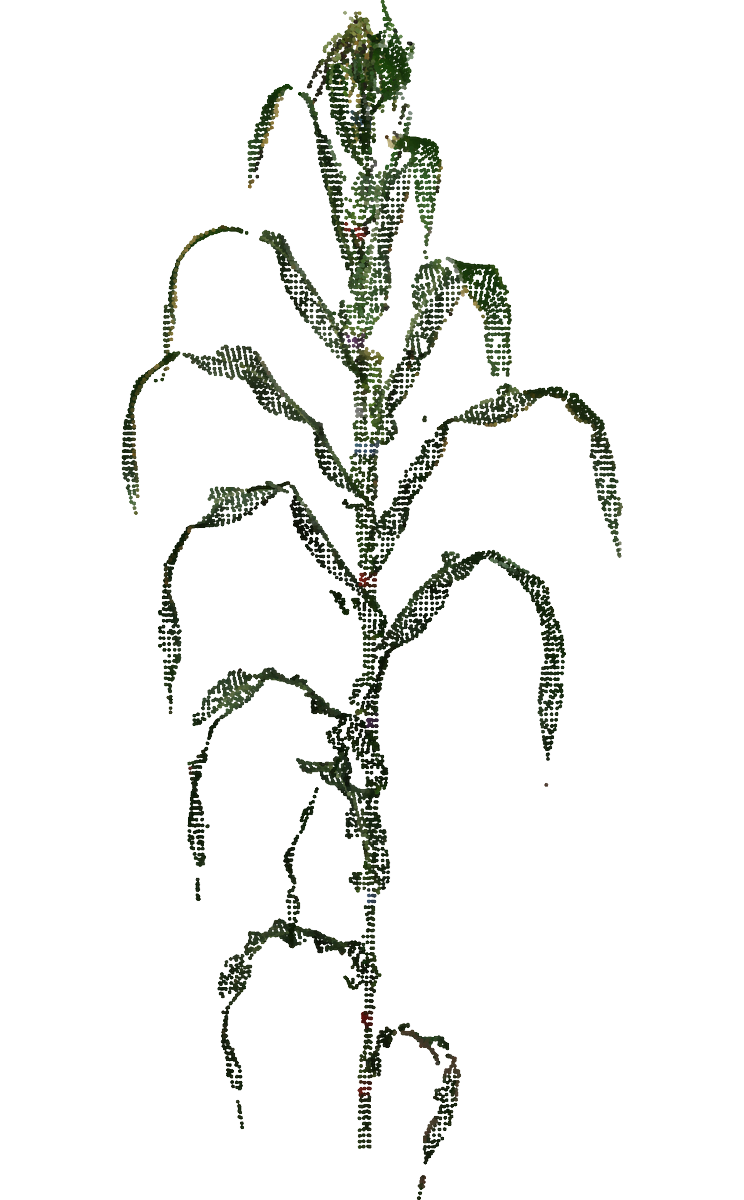}{$-135^\circ$}
    \caption{The eight orthographic renderings of genotype Kui2021 used as input to the VLM stage. The view set covers the front, back, and six side viewpoints, so a leaf that is self-occluded in one view is generally visible in several others. The panels are cropped to the plant. The full renders also include the three corner overlays of \secref{sec:render}.}
    \label{fig:multiview_grid}
\end{figure}

The VLM operates on 2D images, so we render the preprocessed cloud (\secref{sec:preprocess}) into orthographic (parallel projection) images using PyVista~\citep{sullivan2019pyvista}. Since preprocessing has already canonicalized the horizontal axes, no per-plant alignment is required. The render, VLM, mask, and consensus stages use the same ordered set of $N_v=8$ views. We identify each view by its camera azimuth, measured counter-clockwise about $+z$ from the $+y$ axis, so the front camera is at $0^\circ$ and the back camera at $180^\circ$. In pipeline order, the views are $0^\circ$, $180^\circ$, $+30^\circ$, $-30^\circ$, $+45^\circ$, $-45^\circ$, $+135^\circ$, and $-135^\circ$ (\figref{fig:multiview_grid}). The saved output files name these \texttt{front}, \texttt{back}, \texttt{side\_30-}, \texttt{side\_30}, \texttt{side\_45-}, \texttt{side\_45}, \texttt{side\_135-}, and \texttt{side\_135}, respectively. The trailing minus only distinguishes the two members of a pair and does not mark a negative angle, so the figures are labeled by azimuth. Parallel projection preserves organ scale and gives the pixel-to-world mapping reused during back-projection.

Views are rendered at $2400\times2400$ pixels on a white background, with points drawn as unlit spheres to present the scan's own colors unshaded to the annotator. Each image includes only three overlays: a text label identifying the view, a world-axis indicator, and a vertical scale bar of fixed $10$\,cm length. The overlay set is kept deliberately small. The annotation prompt of \secref{sec:vlm} asks the model to leave the corner overlays untouched, and each additional mark is another region where a spurious colored stroke could be drawn and then back-projected.

\begin{figure}[t!]
    \centering
    \newlength{\annotwidth}
    \setlength{\annotwidth}{0.19\linewidth}
    \newcommand{\annotcell}[2]{%
        \begin{subfigure}[b]{\annotwidth}
            \centering
            \includegraphics[width=\linewidth]{#1}
            \caption{#2}
        \end{subfigure}}
    \annotcell{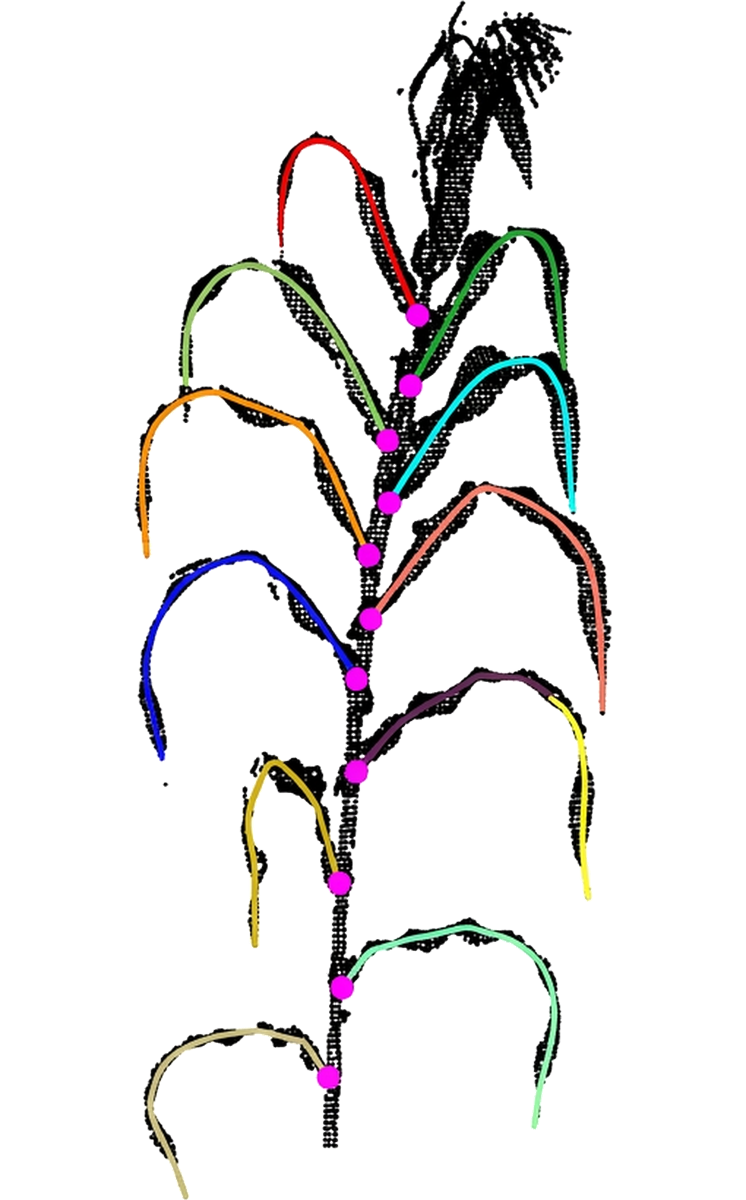}{front ($0^\circ$)}\hfill \annotcell{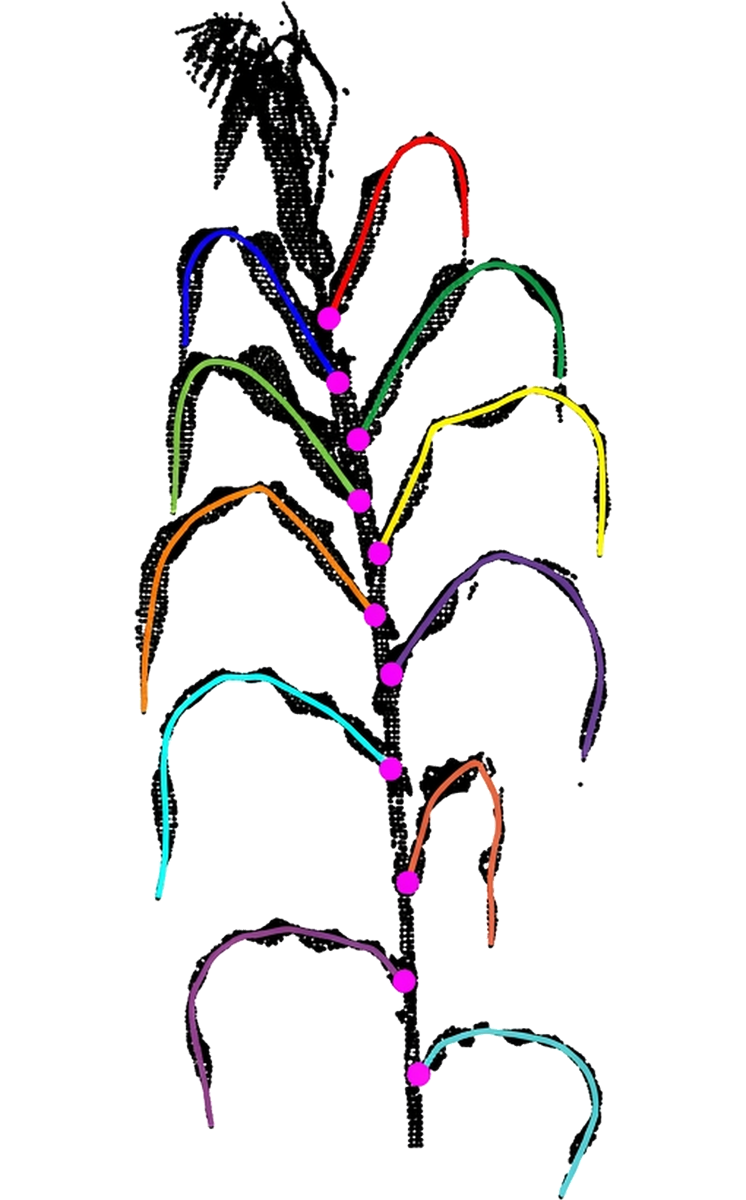}{back ($180^\circ$)}\hfill \annotcell{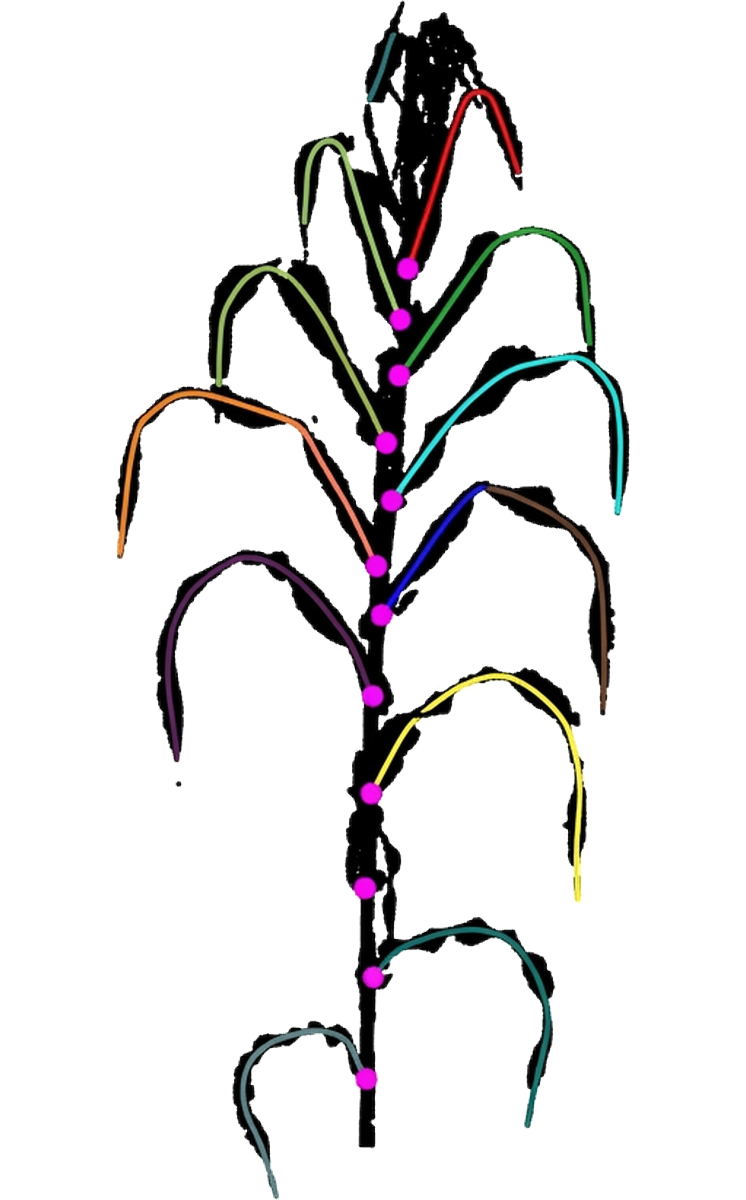}{$+30^\circ$}\hfill \annotcell{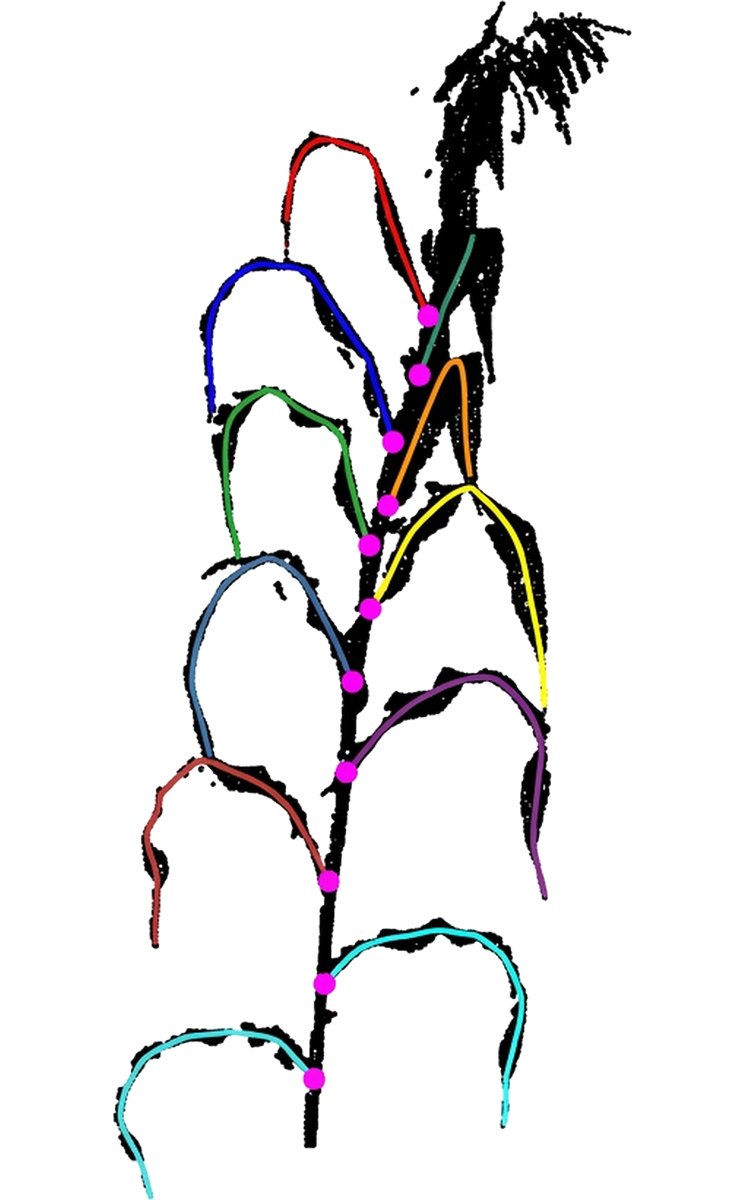}{$-30^\circ$}\hfill \annotcell{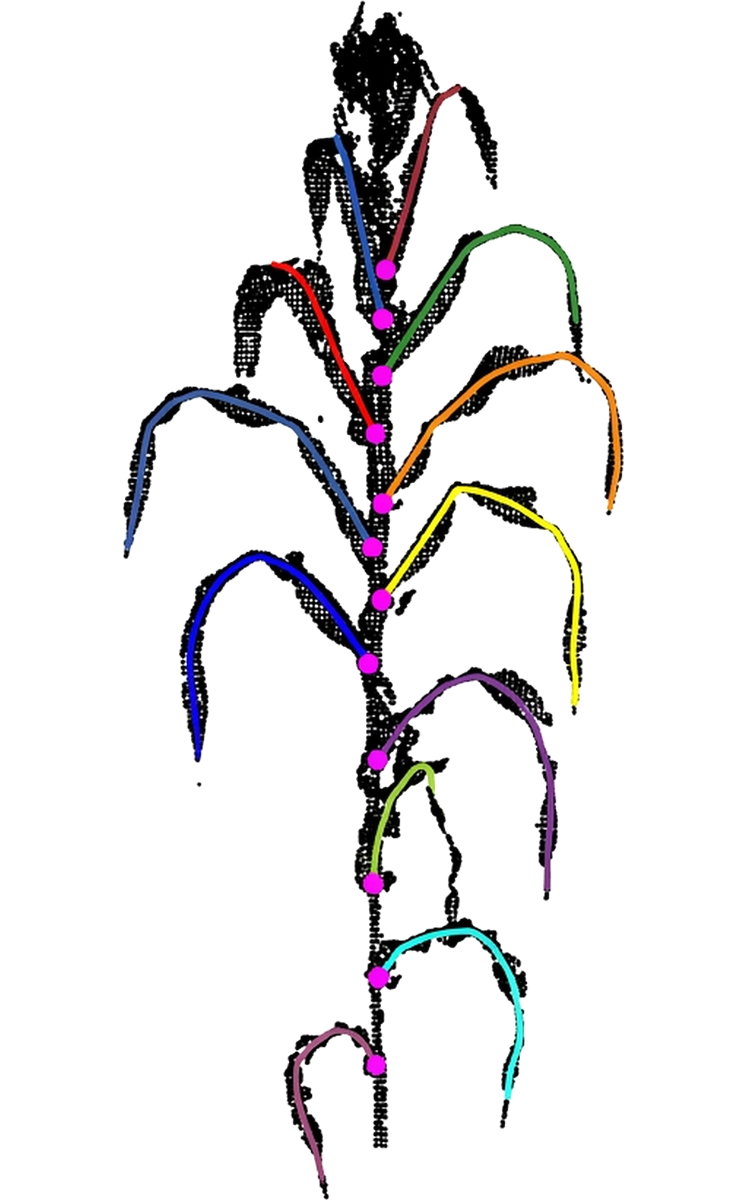}{$+45^\circ$}

    \vspace{0.6em}
    \annotcell{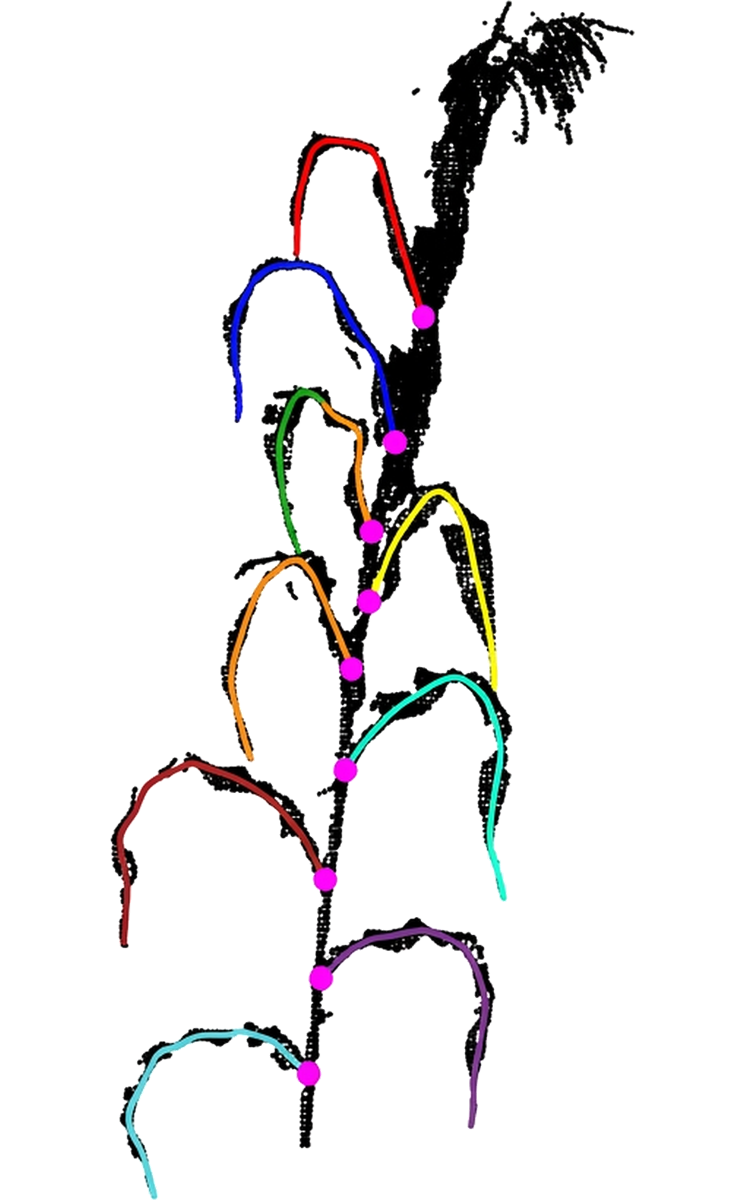}{$-45^\circ$}\hfill \annotcell{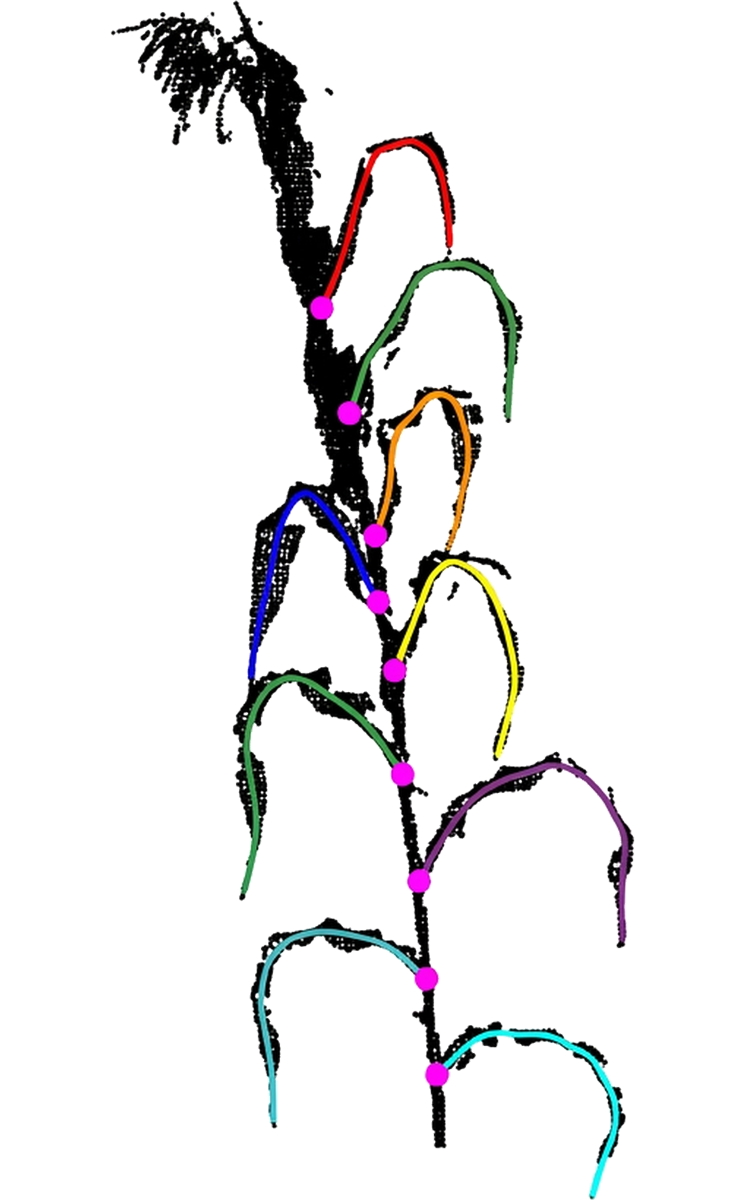}{$+135^\circ$}\hfill \annotcell{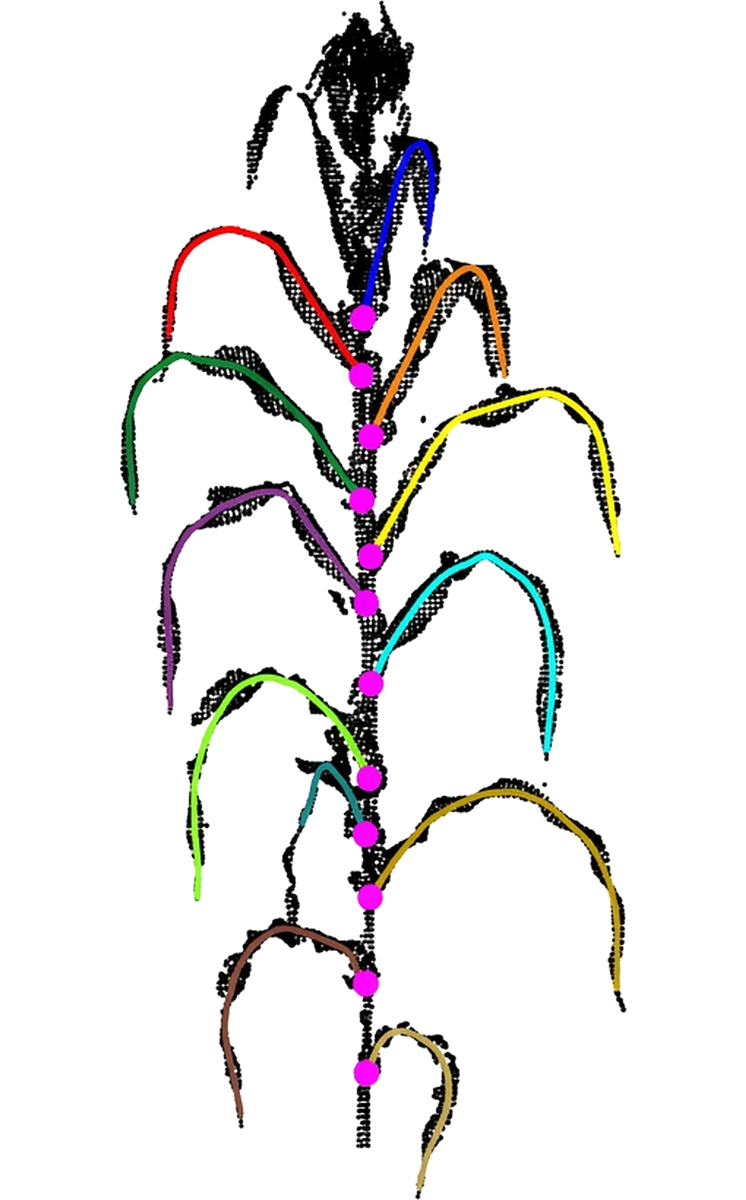}{$-135^\circ$}\hfill \annotcell{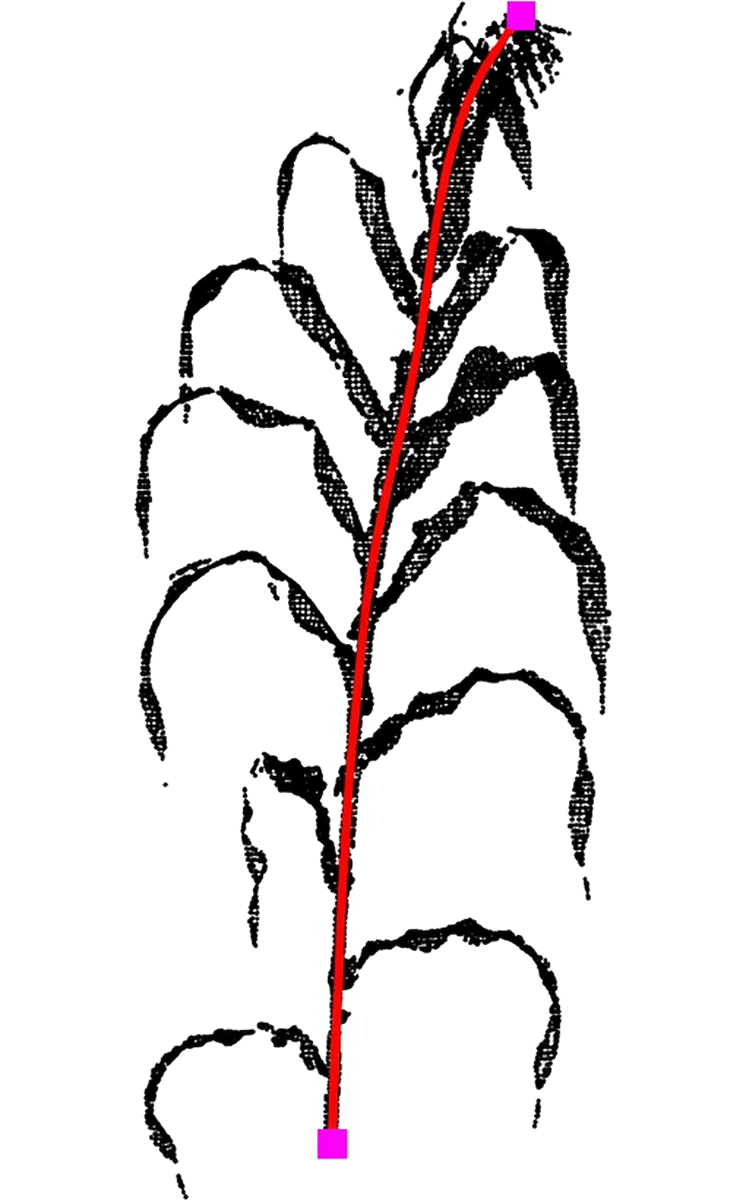}{front, stalk pass}\hfill
    \hspace{\annotwidth}
    \caption{The annotations returned for Kui2021, one per rendered view of \figref{fig:multiview_grid}. Panels (a) to (h) show the leaf pass of Prompt~1, with one uniquely colored midline per visible leaf and magenta dots at the stalk-leaf junctions. Panel~(i) shows the separate stalk pass of Prompt~2, which is applied to the front view only and returns the red stalk stroke and the two magenta endpoint squares used to reconstruct the stalk centerline (\secref{sec:leafseg}). The mask stage reassigns the colors, and consensus reconciles the view-local stroke identities and counts across views (\secref{sec:backproject}). In this example, the leaf count differs between views, a disagreement that consensus resolves.}
    \label{fig:vlm_output}
\end{figure}

\subsection{Per-view VLM-based segmentation}
\label{sec:vlm}

We annotate the $N_v$ rendered views \emph{independently} with a multimodal foundation model that generates an annotation image conditioned on each input view. The generated image is not guaranteed to preserve the input plant geometry or pixel registration, so its canvas shape and plant silhouette are checked before it is used. The reported runs use Gemini, model \texttt{gemini-3-pro-image-preview}~\citep{gemini3proimage}. An image output is required because it provides pixel-level locations for leaf midlines and node markers, and a text-only response does not provide these locations. Both prompts (\figref{fig:prompt}) are issued zero-shot, with no annotated exemplars, and Prompt~1 uses identical text across views. The VLM stage also supports optional paired few-shot examples, but none were supplied in the reported runs. The leaf-color palette is not synchronized between views, since consensus reconciles view-local identities geometrically (\secref{sec:backproject}) without color matching. Natural-language descriptions can optionally be requested through \texttt{--vlm-describe}; they were not requested in the reported runs and are not used by the geometric stages.

The front view also receives a second, complementary prompt that asks the model to trace the stalk with a red stroke and to place one magenta square at the stalk base and one at the top of the tassel. The two squares define the 3D anchors for shortest-path centerline reconstruction, so an unconstrained curve fit is not required (\secref{sec:leafseg}).

\paragraph{Annotation validation and retry}
Since the VLM output can be unreliable, we check each returned image using geometric criteria before accepting it. Three failure modes are screened. First, a canvas-shape check requires the returned aspect ratio to match the original within a tolerance of $0.05$. This check rejects the multi-panel and letterboxed images that a generative model occasionally produces but accepts a benign change of pixel resolution. Second, a silhouette check requires an intersection-over-union of at least $0.60$ between the plant region of the annotation and the dilated plant region of the original. This check rejects large changes in plant position, scale, or silhouette, but it does not establish pixel-level preservation within the silhouette. Third, for the stalk prompt only, the annotation must contain precisely two magenta connected components and one red component and must reach an intersection-over-union of at least $0.40$. The red stroke is checked for presence but is not fitted; after an annotation is accepted, the path of the red stroke does not affect centerline reconstruction when the endpoint squares are recovered. A failing leaf annotation is re-requested up to three times, and a failing stalk annotation is re-requested at most twice. If no attempt passes, we keep the best attempt by these criteria and record the failure in a validation log. The validation score is not used by consensus; a retained leaf annotation is processed under the same cross-view support rules as the other views. Accepted annotations were reviewed before the geometric stages were run. Where a view showed a midline that stopped short of the leaf tip or a leaf that had been left unannotated, the annotation was re-requested with the same prompt or corrected by hand. This review used the rendered views alone and did not consult the curated organ labels of \secref{sec:seg_validation}.

\paragraph{Overlay extraction}
When necessary, the accepted annotation is resized to the original image dimensions before overlay extraction. Overlays are then separated from the plant image in HSV space. Colored strokes are isolated by saturation and value thresholds ($S \ge 60$, $V \ge 60$), and magenta node dots are isolated by a narrow hue window ($H \in [145, 162]$ on the $0$ to $179$ scale) with high saturation and value ($S, V \ge 165$). Morphological opening and closing remove small artifacts. Candidate dots are then filtered by shape to retain only compact blobs (circularity~$\ge 0.55$, aspect ratio~$\le 1.8$, extent~$\ge 0.50$). This shape filter separates a node dot from the end of a colored midline that falls in the magenta hue window. The leaf-midline mask is the color overlay minus the dilated magenta mask. This extraction is applied to each view independently and is used by the leaf-separation stage described next.

\subsection{Mask extraction and skeleton-based leaf separation}
\label{sec:mask}

The per-view VLM annotation (\secref{sec:vlm}) gives a leaf-midline mask and a set of magenta stalk-node dots per view. The VLM occasionally paints two crossing leaves as a single continuous stroke, and conversely breaks one leaf into two visually disconnected fragments (e.g., when a leaf passes behind another and reappears). A plain connected-component (CC) analysis of the raw mask fails in both cases, since it merges two crossing leaves into one blob and keeps the two halves of a fragmented leaf apart. This stage addresses both failure modes in each view before the 3D consensus is computed. Its output is a uniquely colored per-view mask in which curve fragments assigned to the same candidate leaf share one identifier.

\paragraph{Skeletonization and junction-based splitting}
For each view, we first skeletonize the leaf-midline mask, that is, reduce its colored midline strokes to one-pixel-wide curves. Junction pixels (skeleton pixels with three or more 8-connected neighbors) mark where two leaves cross or a stroke branches. Subtracting the junction set, dilated with a radius of 2 pixels, splits the skeleton into disjoint ``arm'' segments with no internal junction. Skeleton arms shorter than 5 pixels and label regions smaller than 280 pixels are discarded as noise. Each surviving arm is a candidate curve segment belonging to a single underlying leaf.

\paragraph{Tangent-based curve continuation and re-merging}
At a crossing, four arms meet at the deleted junction zone, and the correct grouping places together the arms whose outward tangents continue smoothly through that zone. For each arm, we detect its degree-1 skeleton endpoints and estimate outward-pointing tangents at these endpoints by local principal-component analysis over a short window along the skeleton. For a candidate pair of arms $(A,B)$, the continuation cost combines (i) endpoint gap, (ii) penalties for tangent misalignment in both directions, and (iii) a low-weight CIELAB distance between the arms' mean colors. The tangent terms require $A$'s outward tangent to point toward $B$ and $B$'s outward tangent to point toward $A$. Pairs that fail the distance, tangent-alignment, or color thresholds are rejected. A greedy union-find from lowest to highest continuation cost then merges arm pairs, with each endpoint used in at most one merge because a leaf midline is a linear (non-branching) curve. The merge assigns a common identifier to disconnected fragments judged to belong to the same leaf. At a crossing, the merge pairs the arms whose tangents continue smoothly through the junction and thereby recovers both underlying leaves from the four incident arms.

\paragraph{Body reassignment}
Once the arms are grouped into leaf midlines, the non-skeleton pixels of the original mask are assigned to the merged arm with the closest skeleton. The per-view leaves thus retain the full stroke width of the VLM mask.

\paragraph{Node-dot anchoring and final per-view leaves}
The resulting curve groups are matched one-to-one to nearby magenta node dots. A group without a matched dot is treated as a possible distal fragment and is merged into an anchored leaf only when the geometric continuation test supports the merge; otherwise, it remains separate.

Each recovered leaf is repainted with a unique output color, producing a per-view segmentation image and metadata mapping output colors to per-view leaf identifiers. The per-leaf mask files and their metadata are used by the back-projection stage, and no auxiliary structured VLM description is required. \figref{fig:vlm_to_3d}(a) shows the masks derived from the annotations in \figref{fig:vlm_output}.

\subsection{Per-view back-projection and cross-view consensus}
\label{sec:backproject}

To transfer the per-view 2D annotations into 3D, we first back-project each view independently onto the input point cloud, giving a per-view set of local (leaf-id, 3D points) tuples. We then reconcile these per-view leaf identifiers into a single, globally consistent leaf set by cross-view consensus. For example, a leaf annotated as ``L2'' in the front view and as ``L4'' in the $+45^\circ$ side view is recognized as the same physical leaf.

\begin{figure}[t!]
    \centering
    \begin{subfigure}[b]{0.58\linewidth}
        \centering
        \includegraphics[width=\linewidth]{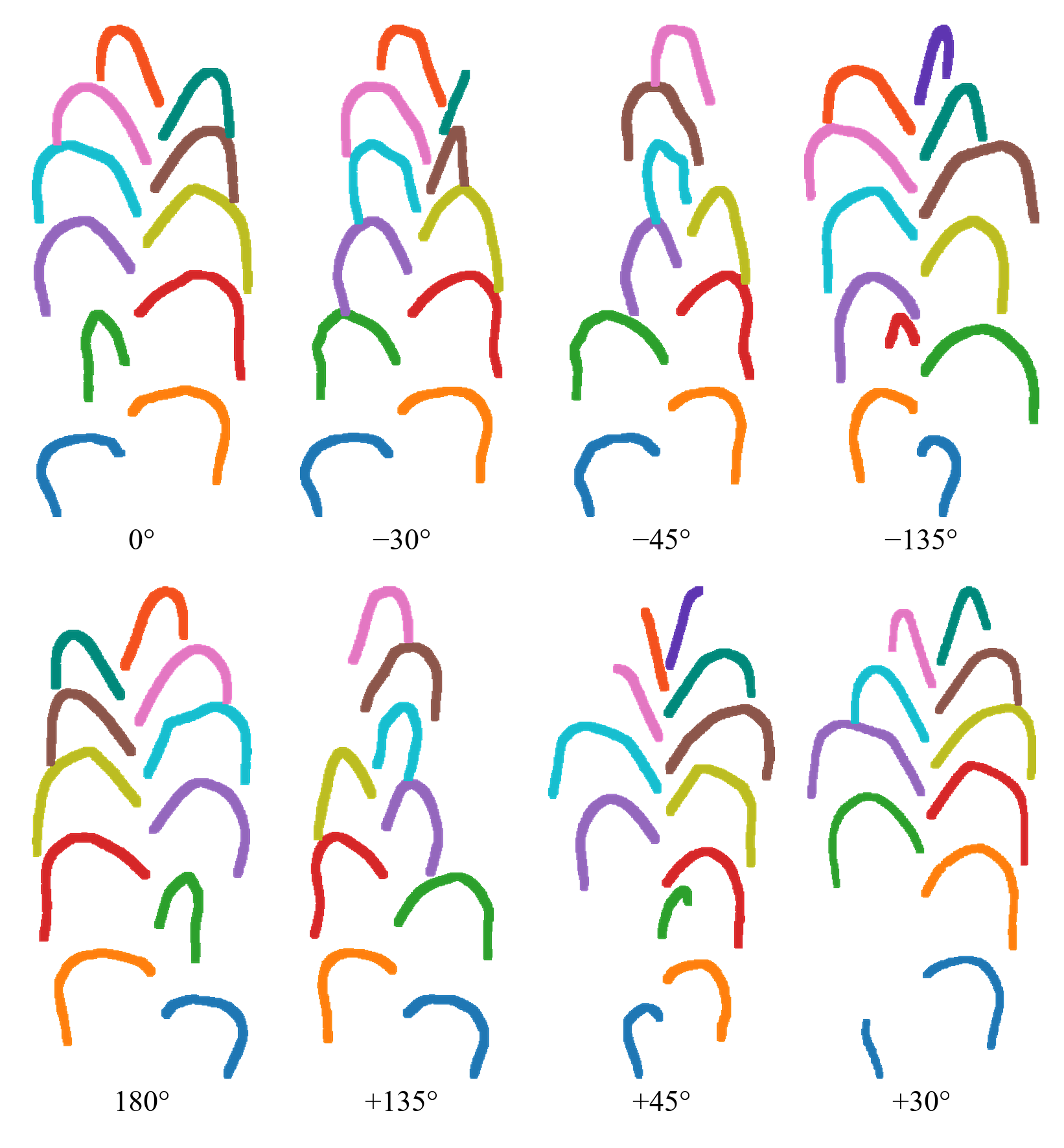}
        \caption{Per-leaf masks in all eight views}
    \end{subfigure}\hfill
    \begin{subfigure}[b]{0.38\linewidth}
        \centering
        \includegraphics[width=\linewidth]{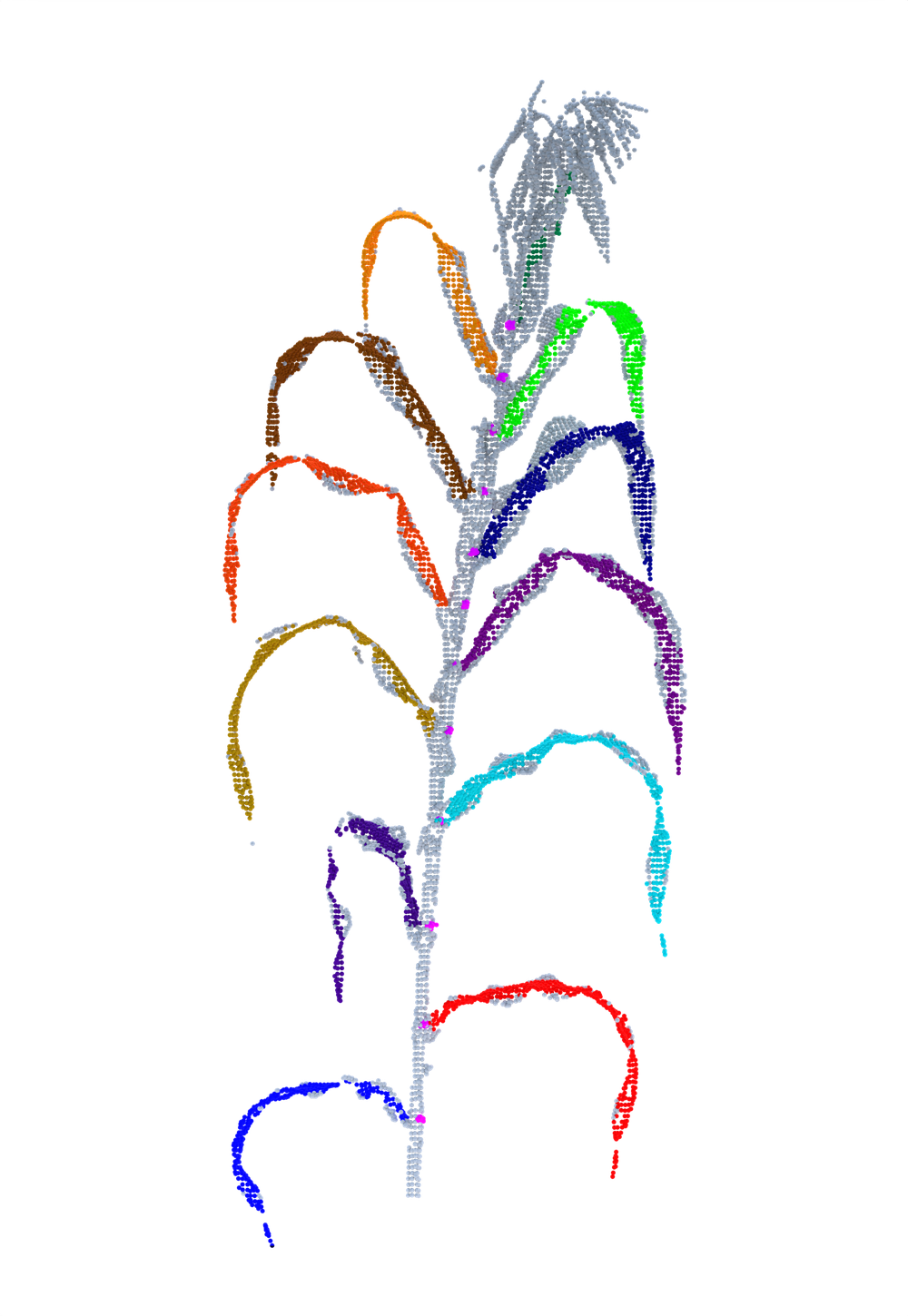}
        \caption{Consensus leaves in 3D}
    \end{subfigure}
    \caption{From per-view masks to a single leaf set in 3D, on Kui2021.
    (a)~The mask stage applied to all rendered views. Each annotation stroke is skeletonized, cut at its junctions, regrouped by tangent continuation, and repainted with one color per view-local leaf. This regrouping can separate two leaves drawn as one stroke and assign disconnected fragments to one leaf identity. The mask sheets are labeled by view, in the order listed in \secref{sec:render}. The number of masks differs between views because a leaf occluded in one view is visible in another. The colors are view-local identifiers and do not correspond between views. (b)~The masks of all eight views, back-projected onto the point cloud with the camera that rendered them and reconciled by cross-view consensus into one numbered leaf set, drawn as one midline curve per leaf over the cloud (gray). Only in this panel do the colors identify the same physical leaf across views, as established by the consensus rule of \secref{sec:backproject}.}
    \label{fig:vlm_to_3d}
\end{figure}

\paragraph{Per-view back-projection}
For each view $v$, we reuse the camera configuration of the rendering stage and project the 3D points to screen coordinates with VTK's world-to-display transform. Before label lookup, a depth-buffer test rejects points that lie behind the rendered front surface by more than the view-specific sphere-splat offset plus a fixed normalized-depth slack of $0.008$. For each remaining projected point, we search a small pixel neighborhood (default radius~$= 1$~pixel) in that view's leaf midline mask. If a match is found, the point receives the corresponding \emph{per-view leaf identifier} from the skeleton-based leaf separation of \secref{sec:mask}. Otherwise, it remains unassigned in that view. The stored camera transform does not require additional calibration; the remaining error comes from pixel discretization and residual misregistration in the accepted generated annotation. \figref{fig:vlm_to_3d} shows the per-view masks~(a) and the single leaf set~(b) obtained after back-projection and cross-view reconciliation. For each of the $N_v$ views, the output is a mapping from the points of the input cloud to a (view, per-view-leaf-id) label, or to ``unassigned in this view''.

\paragraph{Cross-view leaf consensus}
Consensus is built on a mutual-best-match rule that unifies view-local leaf identifiers into a global set. For each pair of views $(v,w)$ and each pair of view-local leaves $(L_v,L_w)$, we compute the point overlap $s(L_v,L_w)=|P(L_v)\cap P(L_w)|$, where $P(\cdot)$ is the set of back-projected 3D points. A pair is linked only when the overlap reaches at least $40$ shared points \emph{and} each leaf is the other's best match in the opposite view. The admitted links form a union-find graph whose connected components become the global leaves. The $40$-point floor suppresses links from small accidental overlaps, and mutual matching reduces the risk that one broad projected stroke is linked to several nearby leaves. A global leaf supported by fewer than two views is discarded (the two-view floor). This floor removes any annotation seen in only one view, including a hallucinated stroke, but can also discard a genuine leaf visible in only one view. After the identifiers are reconciled, each 3D point is assigned to the leaf receiving the most per-view votes and is retained only when at least two views vote for that leaf.

A global leaf with one or more contributing magenta dots receives a provisional 3D anchor. Each contributing dot is back-projected to the front-most visible cloud point within a radius of $8$ pixels, and the per-view candidates are reduced to a single anchor by a coordinate-wise median. The anchor is used during consensus cleanup and passed to blade-level segmentation as an intermediate attachment reference. It is not the final attachment node.

\paragraph{Two corrections applied to the raw consensus}
Two systematic failure modes of the raw per-view masks are corrected at this point. First, the VLM occasionally paints a single midline mask across two vertically stacked leaves, with a stray ``bridging'' node dot connecting them. A naive consensus preserves this merge. We cluster the associated magenta dots by their $z$-coordinates and flag a global leaf as \emph{bridged} when its dots resolve into two or more $z$-clusters separated by more than $6$\,cm. The merged global leaf is split into separate global leaves only if each resulting sub-cluster is supported by at least three distinct views and the back-projected points do not form a smooth continuation across the candidate boundary at the split point. A single misplaced dot in the middle of an otherwise correct leaf would defeat the $z$-gap test. We therefore make the clustering robust to outliers. When one cluster spans at least $8$\,cm, we drop the single most isolated interior dot, re-cluster, and accept the split only if this exposes two view-supported sub-clusters. The dropped dot is then re-attached to the nearest surviving cluster, and no dot position is synthesized.

Second, the mutual-best-match rule occasionally still admits a small ``duplicate'' leaf whose back-projected points lie almost entirely on top of a larger neighboring leaf. We merge a smaller leaf into a larger host only when at least $50\%$ of its points are within $2$\,cm of points assigned to other leaves and at least $70\%$ of those nearby points are closest to the same host. After the merge, the duplicate provisional anchor of the smaller leaf is removed, and the surviving leaves are renumbered. After per-point voting, points farther than $80$\,cm from their provisional anchor are removed for leaves that have such an anchor. Connected-component analysis then always retains the largest component of each leaf. Any additional component is discarded if it contains fewer than $50$ points. For a leaf with a provisional anchor, an additional component is also discarded if its centroid is more than $45$\,cm from that anchor.

\paragraph{Node handling after consensus}
After blade segmentation and fragment correction, final attachment nodes are determined by the constrained node rule of \secref{sec:leafseg}. The rule preserves an existing node when that node satisfies both geometric conditions. It projects a node when the node is missing or the leaf geometry was changed by a split, and it replaces an unsafe existing node only when projection resolves the violation without invalidating another node. Any remaining violation is reported.

\paragraph{Preliminary global leaf numbering}
After the consensus corrections, the surviving global leaves are renumbered, and the mapping from per-view leaf identifiers to global identifiers is saved for traceability. This ordering is preliminary. Blade-level segmentation can subsequently modify the leaf set and its ordering, and later stages use the corrected blade-level identifiers.

\subsection{Blade-level segmentation refinement and node repair}
\label{sec:leafseg}

The consensus labels of \secref{sec:backproject} give a coherent set of global leaves. However, these labels cover only the midline of each leaf. Three problems remain. First, the midline labels do not span the full blade width, so parameters that depend on blade extent cannot yet be measured reliably. Second, the provisional dot-derived anchors can be missing or misplaced and are not the final attachment nodes. Third, after the labels are propagated to the full blades, a blade fragment can remain assigned to the wrong leaf. Where two leaves touch or cross in 3D, propagation can attach a lobe of one blade to its neighbor. Where a scan gap splits a blade, the distal fragment can remain with whichever leaf is geodesically closer. This stage propagates the midline labels to the full blades, determines the final nodes, and corrects supported cases of misassigned blade material. Parameter extraction then operates on the accepted per-leaf full-blade clouds. Unsupported points remain unlabeled.

The procedural parameters of \secref{sec:params_ppg} must be measured on the corrected full-blade clouds and referenced to the corrected attachment nodes. We therefore run this stage \emph{before} parameter extraction and procedural generation. The stalk centerline pre-pass described next is built first, and the final-node rule and procedural assembly share this centerline.

\paragraph{Stalk centerline reconstruction}
The final-node rule below depends on the geometry of the stalk centerline. We therefore reconstruct the centerline before blade propagation, using the two magenta endpoint squares from the stalk annotation (\secref{sec:vlm}). The squares are back-projected with the camera transform used for the leaf masks. Within each square, cloud points farther than $4$\,cm from the initial square position are discarded, and the median of the remaining points defines the corresponding endpoint. If either endpoint cannot be recovered, the endpoint-based extraction returns no centerline and the geometric fallback described below is used. Only the two endpoint squares determine the centerline; the accuracy of the red stroke's path is therefore immaterial once they have been recovered.

The centerline is then traced as a shortest path between the two anchors over a $k$-nearest-neighbor graph ($k=100$) on the cloud itself. The edge weights include a radial penalty. To compute this penalty, we first fit the dominant stalk axis by five iterations of principal-component analysis, each restricted to the innermost $1.5\%$ of points by radial distance. The fitted axis gives the radial distance $r_i$ of each point. Edges are weighted as $\|\mathbf{p}_i-\mathbf{p}_j\|\big(1+\alpha\,\bar{r}_{ij}/\tilde{r}\big)$, where $\bar{r}_{ij}=(r_i+r_j)/2$ is the mean radius of the two edge endpoints, $\tilde{r}$ is the median radius over the cloud, and $\alpha=0.05$. Normalizing by the median makes the penalty strength independent of plant width. The radial term favors paths near the fitted axis while allowing the path to follow stalk curvature.  The traced path is fitted with a cubic B-spline and resampled to $400$ points uniformly in arc length. Large clouds are subsampled to $10^5$ points before the graph is built, a guard against undecimated input that is not triggered at the $10^4$ working resolution of \secref{sec:dataset}.  If endpoint-based extraction fails, a geometric centerline is estimated from the full point cloud. The resulting centerline seeds the stalk during blade propagation and defines the curve onto which nodes are projected.

\paragraph{Blade propagation on an orientation-weighted surface graph}
Each leaf label is then grown outward from its consensus midline. A first pass separates the stalk from the foreground (the non-stalk points) by holding out the points within a bounded radius of the reconstructed stalk centerline. This radius is $3$\,cm over the lower $70\%$ of plant height. Higher up, it adapts to the plant as the $0.7$ quantile of the perpendicular distances of nearby points, smoothed over seven height samples and floored at $0.5$\,cm, to avoid assigning too many points to the narrowing upper stalk. This pre-pass determines the stalk/foreground separation and an initial dense per-leaf labeling. The midline-guided stage retains these dense labels as a prior instead of discarding them. The consensus midlines provide the seed evidence and geometric costs used to test coherent changes to those labels.

At each foreground point, we estimate a local frame by principal-component analysis over its $k = 20$ nearest neighbors, taking the leading eigenvector as the surface tangent and the trailing one as the surface normal. We then build a $k$-nearest-neighbor graph ($k = 20$) over the foreground points, discard edges longer than four times the median nearest-neighbor distance, and weight the remaining edges by their length times an orientation penalty
\begin{equation}
\label{eq:surface_edge_weight}
    w_{ij} = \|\mathbf{p}_i - \mathbf{p}_j\|\,\Big(1 + \lambda_t \big(1 - |\mathbf{t}_i \cdot \mathbf{t}_j|\big)^2 + \lambda_n \big(1 - |\mathbf{n}_i \cdot \mathbf{n}_j|\big)^2\Big),
\end{equation}
with $\lambda_t = 5$ and $\lambda_n = 8$. The orientation penalty in \eqnref{eq:surface_edge_weight} discourages graph transitions between touching leaves. Nearby points on the same blade tend to have similar tangent and normal directions, whereas points on contacting blades can differ in orientation. The resulting path is therefore biased toward remaining on one blade.

The consensus midline points serve as seeds, but not all of them are reliable. A midline point becomes an \emph{immutable} seed only if at least $60\%$ of its twelve nearest seed neighbors carry the same leaf label. This gate suppresses the contradictory seeds that occur where two annotated midlines cross in 3D. If the gate leaves a leaf with fewer than eight seeds, all of that leaf's seeds are restored. This ensures that the gate cannot remove a leaf's midline support entirely. For each leaf, we run Dijkstra's algorithm~\citep{dijkstra1959} from all its immutable seeds simultaneously over the weighted graph. A point's cost for that leaf is its geodesic cost, plus $0.25$ times its Euclidean distance to the nearest seed, plus tangent- and normal-agreement penalties weighted by $2$\,cm each. Immutable seeds are pinned to their own leaf at zero cost and to all other leaves at infinite cost. For each foreground point, the minimum-cost leaf and the gap to the second-lowest cost are computed. These values guide the boundary-stabilization decisions applied to the initial dense labels.

\paragraph{Unlabeled points and boundary stabilization}
Points left unlabeled by the initial dense pass remain unlabeled in the reported configuration and are excluded from downstream measurements. They are not assigned merely because a consensus midline can reach them. This restriction avoids forcing background, tassel, or unsupported material into a leaf. For initially labeled points, a hysteresis pass compares the cost of the current label with that of the minimum-cost alternative. A connected group of at least eight points changes labels only when the alternative improves the cost by at least $8$\,cm. An immutable seed assigned away from its own leaf is reset to that leaf. The output is a set of per-leaf full-blade point clouds that provides the observed blade extent and later serves as the target point cloud in \secref{sec:nurbsdiff}.

\paragraph{Constrained rule for final node placement}
Each leaf requires one attachment node on the reconstructed stalk. An initial node is derived from the consensus midline by selecting its point nearest the stalk centerline and projecting it to the nearest centerline sample. After full-blade segmentation and fragment correction, a separate projection candidate is computed from the full-blade point nearest the centerline. Replacing the initial nodes with this candidate would discard attachment information provided by the consensus midline and could introduce error when the scanned blade base is incomplete.

The node rule is therefore constrained by two explicit conditions on the final node set. A node is admissible only if (a)~its leaf base lies within $6$\,cm of it, and (b)~it is at least $4$\,cm from its nearest neighboring node. The first condition rejects a node that has drifted away from the blade to which it should attach. The second rejects coincident node pairs, which the procedural generator cannot resolve into distinct internodes. A leaf with a missing node is projected outright. Both leaves involved in a split are also projected outright, namely the source leaf from which the split was taken, which has lost material, and the new leaf that the split created, which has no node at all. A move is treated differently. It reassigns material between two leaves that both already have a node, and neither node is discarded. Both nodes are tested against the two conditions like any other node.

All other leaves keep their existing node if it satisfies both conditions. A node that fails either condition triggers a fresh projection, which is adopted only if it makes that leaf admissible \emph{without} making any currently admissible leaf inadmissible. This safeguard is required because the spacing condition couples neighbors, and moving one node can push a valid neighbor below the spacing floor. A repair that trades one violation for another is rejected. If no admissible replacement exists, the original node is retained and reported as unresolved. Each leaf thus ends with one of four outcomes: preserved, repaired, projected because required, or unresolved. The before-and-after node positions, gap, spacing, and failure reasons of each leaf are saved to the run report. A plant with an unresolved node is still reconstructed. \secref{sec:seg_results} identifies the one node retained as unresolved. \tabref{tab:nodes} reports all outcomes, including one required projection that still violates the gap condition.

\begin{figure}[t!]
    \centering
    \includegraphics[width=\textwidth]{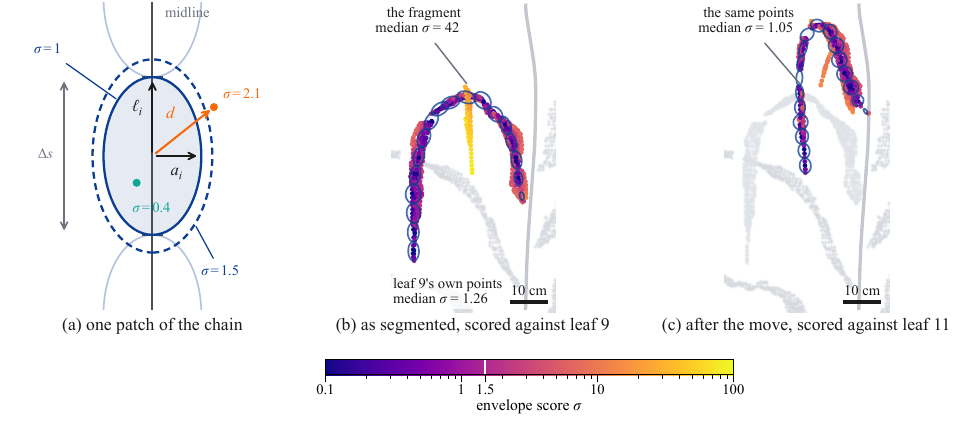}
    \caption{Ellipsoid-envelope diagnostic and correction of a misassigned
    blade fragment, on CML331. \textbf{(a)} A blade is tiled by a chain of ellipsoids, one per patch of equal arc length $\Delta s$, each with an axial half-length $\ell_i$, a half-width $a_i$ measured across the blade, and a half-thickness $b_i$ through it. The score of \eqnref{eq:envelope_score} divides a point's offset $\mathbf{d}$ from the patch center by these half-axes. Thus $\sigma = 1$ is the boundary of a padded ellipsoid, and $\sigma \le 1.5$ is the slack allowed by the acceptance test. \textbf{(b)} Leaf~$9$ after the initial assignment, drawn in the plant's front view with the rest of the plant in gray and with points colored by the score of their own leaf's envelope. The blade sits at a median score of $\sigma = 1.3$, while the part of the fragment lying on this leaf reaches $\sigma = 42$, placing those points far outside this leaf's envelope. The correction moved $90$ points in total, taken from this leaf and from leaf~$7$. \textbf{(c)} The same points after the accepted \textsc{move}, scored against leaf~$11$'s envelope. Here the fragment sits at $1.1$, within leaf~$11$'s acceptance envelope.}
    \label{fig:envelopes}
\end{figure}

\paragraph{Ellipsoidal blade envelopes}
The assignment and stabilization passes above act point by point, and the orientation-weighted graph accounts only for local surface continuity. Neither pass determines whether an entire blade fragment belongs to its current leaf. A fragment connected through a sequence of individually low-cost edges is accepted even when, as a whole, it lies well outside its assigned blade. To test whole-fragment membership, we build an explicit, compact geometric model of each leaf as a chain of ellipsoids that tiles the blade from base to tip. For each leaf, we recompute a base-to-tip midline on its own point cloud (a $k$-nearest-neighbor graph, a geodesic farthest-point search for the tip, and a spline fit). This midline is resampled into $n = 18$ patches of equal arc length $\Delta s = L/n$. The patch spacing thus adapts to leaf length, while all leaves are described by the same number of patches. A rotation-minimizing frame $(\mathbf{t}, \mathbf{n}, \mathbf{b})$ is attached at each patch center. We use this frame instead of a Frenet frame because the latter flips abruptly at inflection points of a curved blade. Each frame is then rotated about $\mathbf{t}$ to align $\mathbf{n}$ with the principal in-plane direction of the patch points, i.e., with the estimated local blade-width direction. The rotation angles are unwrapped and smoothed along the chain. This prevents a single noisy patch from twisting its neighbors.

For patch $i$, centered at $\mathbf{c}_i$, we determine the ellipsoid half-axes as follows. The axial half-length $\ell_i$ along $\mathbf{t}$ is $\Delta s / 2$, which makes consecutive ellipsoids share a boundary and tiles the blade without overshooting it. The half-width $a_i$ along $\mathbf{n}$ and the half-thickness $b_i$ along $\mathbf{b}$ are the $90$th percentiles of $|\mathbf{d}\cdot\mathbf{n}|$ and $|\mathbf{d}\cdot\mathbf{b}|$ over the patch points, where $\mathbf{d}$ is the offset from the patch center. These percentiles are computed after two rounds of iterative outlier trimming. The trimming limits the influence of foreign points that would otherwise enlarge the envelope around the fragment being tested. The per-patch widths are then capped at $1.5$ times their median along the chain, and both profiles are smoothed by a three-patch moving median. This prevents one noisy patch from determining the envelope of its neighbors. The thickness is also capped at $b_i \le 0.5\,a_i$, since a real maize blade is a thin sheet. Finally, all three axes are tapered by a square-root profile over the last $15\%$ of the arc length, which narrows the terminal ellipsoids toward the tip.

Scoring uses slightly padded axes, $\tilde{a}_i = a_i + 0.5$\,cm and $\tilde{b}_i = \max(b_i,\,1.5\,\text{cm})$, with the axial half-length kept at $\ell_i$ and the same tip taper applied. The padding accommodates variation in the percentile-based axis estimates. The floor on $\tilde{b}_i$ prevents the slightly off-surface points of a blade only a few millimeters thick from scoring many envelope radii away in the thin direction. A point $\mathbf{p}$ is then scored against the leaf by the minimum over patches of the Mahalanobis-like quantity
\begin{equation}
\label{eq:envelope_score}
    \sigma_i(\mathbf{p}) =
    \left(\frac{\mathbf{d}\cdot\mathbf{n}_i}{\tilde{a}_i}\right)^{\!2}
    + \left(\frac{\mathbf{d}\cdot\mathbf{b}_i}{\tilde{b}_i}\right)^{\!2}
    + \left(\frac{\mathbf{d}\cdot\mathbf{t}_i}{\ell_i}\right)^{\!2},
    \qquad \mathbf{d}=\mathbf{p}-\mathbf{c}_i .
\end{equation}
The leaf score is $\sigma(\mathbf{p})=\min_i \sigma_i(\mathbf{p})$. A value of $\sigma=1$ lies on the boundary of a padded ellipsoid, and larger values indicate increasing distance outside it. The test treats $\sigma\leq1.5$ as inside, providing additional allowance while limiting the inclusion of clearly separated fragments. A rendering of the ellipsoid chain with the per-leaf points colored by $\sigma$ serves as a per-plant diagnostic. In this rendering, a misassigned fragment appears as a lobe of points outside its assigned leaf's envelope but deep inside a neighbor's. \figref{fig:envelopes} shows one such fragment with its scores under both leaves, and the repair below uses the same score.

\paragraph{Correcting misassigned blade fragments}
We decompose each leaf's point cloud into connected components and use these envelopes to test eligible components other than the main blade as candidates for reassignment. A candidate is moved to another leaf only when it is a demonstrably better continuation of that leaf than of its current one. The test combines the component's distance to the destination midline and surface, the agreement between its principal direction and the destination's local tangent, and its envelope score under both leaves. It also uses the alignment of the offset from the destination's nearest midline endpoint with that endpoint's outward tangent. We distinguish two regimes. A component adjacent to the destination blade must continue it. A component separated from the destination by a scan gap must also lie within a narrow continuation corridor extrapolated from the destination's endpoint tangent. The corridor prevents a gap from being bridged merely because two leaves happen to be close. A sufficiently small component that is inconsistent with its current leaf and has no credible alternative is left unassigned. Larger or ambiguous cases are retained and reported for review.

A separate case arises when a single component contains both genuine blade of its current leaf \emph{and} a lobe belonging to a neighbor. This case arises where two blades touch along a seam, and the propagation fuses them into one component. Moving the entire component would also reassign genuine points of its current leaf. Instead, we locate the destination midline endpoint nearest the component, take its outward tangent as a continuation direction, and move only the points inside a $25^\circ$ three-dimensional cone about that direction. The rest of the component stays with its original leaf. The test is geometric and uses no plant-specific or genotype-specific constants. The results of all operations are written to a parallel set of files, so the corrected and uncorrected segmentations can be compared.

\paragraph{Verifying each correction before it is kept}
The operations run in a fixed order. Midline-guided propagation is followed by local geometric repair and candidate detection, then by the \textsc{move} and \textsc{split} corrections, and finally by verification. Candidates are scored from the current plant geometry alone, with no list of known-bad leaves and no per-plant correction input. A repair triggered on one plant would thus be triggered on any plant with the same geometry. A \textsc{move} reassigns a coherent fragment to a uniquely supported continuation of another leaf, and a \textsc{split} creates a new leaf from independently corroborated geometry. All plants share the same thresholds. When the geometric support is ambiguous, which is the common case, no correction is made.

Since these operations change leaf identity, we verify each one before keeping it. A correction is tied to the set of points against which it was scored. It therefore cannot be applied to a segmentation that has changed since it was proposed. The corrections are then checked as a set. They must account for all points of the input, leave unaffected leaves untouched, leave no source or destination leaf empty, and preserve the bottom-to-top order of the attachments. Corrected outputs are promoted only after the leaf partition, nodes, and rebuilt midlines have been validated. The \textsc{move} operations are applied transactionally and rolled back as a set if their validation fails. A selected \textsc{split} must pass its own validation, or the stage stops. When validation fails, the previous canonical outputs therefore remain unchanged and no partial correction is published.

\paragraph{Per-leaf midline rebuild from the full-blade clouds}
The operations above can move blade material between leaves and can create an entirely new leaf. As a result, the per-view midlines used as input to this stage no longer describe the leaves in its output. We therefore recompute the midlines of the accepted leaves from their own final full-blade clouds. For each leaf, we take the full-blade point nearest its final node as the base and build a distance-weighted $k$-nearest-neighbor graph on the blade points ($k = 12$). Edges longer than five times the leaf's median nearest-neighbor distance are discarded to keep the path from jumping across a fold. If this cap disconnects the blade graph, its components are reconnected by the minimum set of bridge edges required to obtain a base-to-tip path. The midline is the geodesic path from the base to the farthest reachable blade point. It is evaluated with an interpolating spline whose smoothing parameter is zero and sampled at a minimum of $200$ ordered points, which retains the base endpoint recovered from the blade cloud. The rebuild is rejected outright if the resulting curve loses its base endpoint, is non-finite, or has a degenerate arc length. The accepted output for each leaf therefore consists of a full-blade cloud, a final stalk node, and a midline derived from that cloud whose first sample is the blade point nearest the node.

\paragraph{Final bottom-to-top leaf renumbering}
The consensus stage assigns preliminary leaf identifiers. After final-node resolution, the leaves are re-indexed in ascending order of the attachment node's arc-length position along the stalk centerline. Downstream stages and all reported tables and figures use this ordering, and the mapping from consensus identifiers to final identifiers is saved for traceability.

\subsection{Geometric parameter extraction and procedural model generation}
\label{sec:params_ppg}

This stage extracts the geometric descriptors of the organs and assembles them into a plant descriptor for the procedural plant generator. We describe parameter extraction and procedural mesh generation as two logical operations for clarity. However, they are implemented as a single pipeline step. This keeps the descriptor and the mesh generated from it consistent.

All plant-specific geometric quantities in the descriptor are derived from the accepted output of \secref{sec:leafseg}, namely the full-blade point clouds, the midlines rebuilt from those clouds, the final stalk nodes, and the reconstructed stalk centerline. The sparse consensus clouds are retained only to document the annotation input and are not substituted for a missing full-blade cloud. A leaf without its required full-blade cloud is omitted instead of reconstructed from the sparse midline alone.

\paragraph{Leaf base and node assignment}
In the reported configuration, each final attachment node is read from the corrected node record and matched to its leaf by identifier. The node's $z$-coordinate is written to the descriptor as \texttt{node-z}. Its arc-length coordinate along the centerline, \texttt{base-s}, is retained in the intermediate control-point output for ordering and traceability. FloraForge locates each attachment by mapping \texttt{node-z} to the nearest height on the explicit stalk centerline.

\paragraph{Geodesic path reconstruction}
The path along which parameters are sampled is reconstructed from the midline rebuilt on the corrected full blade. A distance-weighted $k$-nearest-neighbor graph ($k=50$) is built on the rebuilt midline samples, and Dijkstra's algorithm~\citep{dijkstra1959} computes distances from the sample nearest the final node. The farthest reachable sample defines the tip, and the shortest predecessor path defines the ordered midline.

Two refinements make the path reconstruction robust on real blades. First, a plain $k$-nearest-neighbor graph on a folded or self-approaching leaf contains shortcut edges that jump across the fold, and the geodesic path then cuts the corner and leaves the surface. We therefore cap edge length adaptively by discarding edges longer than six times the leaf's median nearest-neighbor distance, a cap that scales with local point density. Second, such a cap can disconnect the graph at a genuine scan gap. The resulting components are reconnected by the minimum number of bridge edges required to restore connectivity. The reconnection keeps the blade material beyond a gap in the path, and the shortcuts within a fold remain removed. If the graph cannot be constructed or a predecessor path cannot be recovered, we use a greedy nearest-neighbor walk that balances local proximity with progress toward the Euclidean-farthest point. The resulting polyline is smoothed by a spline fit before control points are sampled.

\paragraph{Checking the start of the midline path}
The reconstructed path should begin near the leaf attachment because the procedural generator grafts the first transverse rows of the leaf surface to the stalk. Spline smoothing can displace the start of the path from the trusted leaf base derived from the corrected full-blade cloud.

A repair is applied only when the start-to-base gap is at least $2$\,cm and the initial backtracking is at least $1$\,cm. The repair can remove a leading tail, but it never removes more than the first $20\%$ of the path. It then places the first sample at the trusted leaf base and tapers the displacement to zero over the first $5$\,cm using a smoothstep profile. The repair outcome is saved with the control-point results. Only a leaf for which this repair was applied is eligible for the post-fit collar repair in \secref{sec:nurbsdiff}.

\paragraph{Control point extraction}
Along the reconstructed midline path, we sample $N$ control points (default $N = 8$, configurable) at uniformly spaced normalized arc-length values $u \in \{0, \frac{1}{N-1}, \frac{2}{N-1}, \ldots, 1\}$. The control points define an interpolating B-spline curve (degree~3) that represents the leaf midline in the plant descriptor. We report the B-spline reconstruction RMSE against the original path as a quality metric.

\paragraph{Width and orientation}
At each control point, we measure leaf width from the corrected full-blade cloud using points whose displacement along the local tangent is less than $1$\,cm from the transverse plane. The blade points in this window are projected onto the plane normal to the local tangent, and the width is twice their maximum perpendicular distance from the midline axis. The resulting ordered $(u,\texttt{width-cm})$ profile is written with the control points, copied into the plant descriptor with all its samples, and used by the generator. Missing local measurements are interpolated onto the procedural rows. Finite measured values are not replaced by smoothing or by a scalar taper. The scalar maximum is kept as a summary and is used for the analytical taper only when no measured profile exists.

Leaf orientation is taken from the midline tangent over a short proximal window in world coordinates, i.e., from the direction in which the blade leaves the stalk. A leaf that arches over thus does not take the orientation of its own tip. The tangent $\mathbf{t}$ is converted to the two angles that the generator expects. The azimuth is $(\operatorname{atan2}(t_y,t_x)+90^\circ) \bmod 360^\circ$, measured counter-clockwise about $+z$ from the $+y$ axis, and the pitch is $\operatorname{atan2}(|t_z|,\|(t_x,t_y)\|)$, the unsigned elevation above the horizontal plane. The pitch is unsigned, but the three-dimensional control points retain whether the reconstructed midline rises or descends. One stalk-diameter value per internode is measured at the internode midpoint from a horizontal cross-section around a separately estimated geometric stalk trajectory.

\paragraph{Plant descriptor}
The extracted parameters are assembled into a YAML plant descriptor for FloraForge~\citep{hadadi2025floraforge}. The descriptor contains the explicit three-dimensional stalk centerline and the \texttt{node-z} values used to locate attachments on that centerline, together with one midpoint-derived stalk-diameter value for each internode. For each leaf, it contains the azimuth, pitch, length, scalar width summary, B-spline control points, and control-point width profile. A provenance field records whether the width profile was measured or derived from the scalar summary.

\paragraph{Mesh generation}
FloraForge forms the stalk surface by sweeping a B-spline ring of interpolated radius along the stalk centerline. Each leaf is generated as a NURBS surface whose control net has $11$ rows along the blade length (degree~3) and $3$ columns across the blade width (degree~2). The midline is reconstructed from the extracted control points by degree-3 B-spline interpolation. The measured width profile is interpolated from its recorded $u$ positions onto the $11$ longitudinal rows and determines their transverse extent. Only a leaf with no measured profile uses the analytical taper from the scalar maximum. The transverse profile is shaped by fixed generator parameters for camber, twist, and fold. In the reported configuration, the first two transverse control rows are grafted to the stalk, and the rows are checked for consistent orientation. The per-organ control nets and the assembled surface are exported, with the leaf parameter axes represented as $3\times11$ for differentiable refinement.

\subsection{Differentiable NURBS refinement from procedural initialization}
\label{sec:nurbsdiff}

The procedural mesh produced by FloraForge is a structurally complete initial model. Its leaves follow the extracted midlines and measured width profiles and remain attached to the procedural stalk. However, its low-dimensional transverse construction cannot reproduce all local undulations, twists, and edge deviations in the scan. We therefore use NURBS-Diff~\citep{prasad2022nurbs} to refine each leaf against its corrected full-blade cloud, with the procedural surface as the initial guess.

We adopt the NURBS parameterization and loss structure of \citet{hadadi2025procedural} with three changes: (a) each leaf is initialized from parameters plus PPG instead of PSO~\citep{pyswarmsJOSS2018}, (b) the node-side control row is locked throughout refinement, and (c) candidate prealignments and post-fit corrections are retained only when they pass their stated tests.

\paragraph{NURBS representation}
Each leaf is represented as a tensor-product NURBS surface
\begin{equation}
\mathbf{S}(u,v)=\frac{\sum\limits_{i=0}^{n}\sum\limits_{j=0}^{m}N_{i,p}(u)N_{j,q}(v)w_{i,j}\mathbf{P}_{i,j}}{\sum\limits_{i=0}^{n}\sum\limits_{j=0}^{m}N_{i,p}(u)N_{j,q}(v)w_{i,j}}, \quad 0\leq u,v\leq 1,
\end{equation}
where $\mathbf{P}_{i,j}$ and $w_{i,j}$ are the control points and weights, and $N_{i,p}$ and $N_{j,q}$ are the B-spline basis functions of degrees $p$ and $q$. Following \citet{hadadi2025procedural}, we hold the weights and knot vectors fixed and optimize only the control-point positions. The standard basis recursion and the control-point derivative used by NURBS-Diff are given in \ref{sec:appendix:nurbs}.

\paragraph{Refinement from the procedural surface}
For each leaf, we start from its procedural NURBS surface (\secref{sec:params_ppg}). Its control net is $C_u\times C_v=3\times11$, with $u$ across the blade and $v$ running from node to tip, and it is evaluated on an $8\times32$ parametric grid. The target is the corresponding corrected full-blade cloud, subsampled to at most $4000$ points. Both geometries are rescaled from centimeters to meters to retain the loss weights of the earlier workflow. We use Adam~\citep{kingma2014adam} to optimize the control-point positions for $500$ iterations. The learning rate starts at $10^{-2}$ and is halved at iterations $50$, $100$, $150$, $200$, $250$, and $300$, and the final two hundred iterations run at $1.6\times10^{-4}$. The reported configuration uses the pure-PyTorch surface evaluator~\citep{paszke2019pytorch}. The node-side control row is restored after each step. A leaf whose optimization fails numerically is restarted from its initialization, for at most three attempts in total. If all attempts fail, the leaf is recorded as failed and no fitted geometry is promoted.

\paragraph{Node-preserving rigid prealignment}
The procedural surface is placed using the extracted midline, pitch, and azimuth, but a residual orientation mismatch can remain on a strongly twisted leaf. We therefore test a rigid prealignment before optimization. The target base is the mean of the $2\%$ of target points nearest the procedural node, and the target tip is the mean of the $2\%$ farthest from that estimated base, with a minimum of three points used for each estimate. The proposed transform aligns the procedural and target node-to-tip directions and, when a nondegenerate transverse direction can be estimated from the target points, also aligns the blade-width directions. The node-side control row is then restored to its original procedural coordinates.

The prealignment is treated as a candidate. We compute the symmetric Chamfer metric before and after the transform and accept the candidate only when it reduces that metric by at least $8\%$. Otherwise, the original control net is retained. This comparison is used only for prealignment selection; the fidelity reported in \secref{sec:results} is measured separately on the exported surface.

The orientation guard is applied selectively. A leaf whose prealignment passes the $8\%$ gate is fitted from the prealigned control net with the orientation guard described below. A leaf whose candidate is rejected uses the original, unguarded optimization. A leaf eligible for the post-fit collar repair also uses the original optimization before that local repair is tested. In every case, the attachment row is restored after each optimizer step.

\paragraph{Loss function}
The refinement objective keeps the structure of the loss of \citet{hadadi2025procedural}. It combines a one-sided Chamfer term with two smoothness regularizers on the evaluated surface and one regularizer on the control net:
\begin{equation}
\label{eq:loss_NURBS-Diff}
    \mathcal{L} = d_{CD}^{\text{one-sided}}(X, Y) + \lambda_{1} \mathcal{L}_{1} + \lambda_{2} \mathcal{L}_{2} + \lambda_{3} \mathcal{L}_{3},
\end{equation}
where $Y$ is the target blade cloud and $X=\{\mathbf{S}_{k,l}\}$ are the surface samples on the $8\times32$ evaluation grid, indexed by $k$ across the blade and $l$ along it. The regularizers use second differences of this grid and squared separations of control points in place of analytic derivatives:
\begin{gather}
    \label{eq:loss_curv1}
    \mathcal{L}_{1}=\sum_{k,l}\big\|2\mathbf{S}_{k,l}-\mathbf{S}_{k-1,l}-\mathbf{S}_{k+1,l}\big\|^2 , \\
    \label{eq:loss_curv2}
    \mathcal{L}_{2}=\sum_{k,l}\big\|2\mathbf{S}_{k,l}-\mathbf{S}_{k-1,l}-\mathbf{S}_{k,l+1}\big\|^2 , \\
    \label{eq:loss_transverse}
    \mathcal{L}_{3}=\sum_{j=0}^{C_v-1}\big(\|\mathbf{P}_{2,j} - \mathbf{P}_{1,j}\|_2^2+\|\mathbf{P}_{2,j} - \mathbf{P}_{0,j}\|_2^2+\|\mathbf{P}_{1,j} - \mathbf{P}_{0,j}\|_2^2\big),
\end{gather}
The sums in \eqnref{eq:loss_curv1} and \eqnref{eq:loss_curv2} are evaluated at $k=1,\ldots,6$ and $l=1,\ldots,30$, the interior slice used by the implementation on the $8\times32$ grid.

The data term penalizes the distance from the target points to the evaluated surface. We choose a one-sided term because the target is a partial observation. The term takes the mean squared distance from each $y\in Y$ to its nearest evaluated surface sample (the observed-target-to-surface direction) and gives the reverse direction zero weight. A self-occluded band of a blade is missing from the cloud, and a symmetric term would pull the surface into that hole to avoid being penalized for covering unobserved area. As a result, surface regions with no supporting points are not constrained by the data, and the regularizers and checks below are required to constrain them.

The first regularizer, \eqnref{eq:loss_curv1}, is a centered second difference across the blade. It suppresses the cross-blade wrinkling that appears where the cloud is sparse, and it has the largest weight. The second, \eqnref{eq:loss_curv2}, is a mixed difference that steps backwards across the blade and forwards along it. Unlike \eqnref{eq:loss_curv1}, it does not vanish on a uniformly parameterized plane, where it reduces to the difference $\mathbf{u}-\mathbf{v}$ between the two parametric step vectors. It thus acts as an undirected shear penalty whose exact minimizer, $\mathbf{u}=\mathbf{v}$, is a degenerate surface collapsed onto a line. We retain this term at the weight used in~\citep{hadadi2025procedural}, two orders of magnitude below $\lambda_{1}$, which avoids re-tuning the objective between the two initialization strategies. The third acts on the control net. Since $C_u=3$, the points $\mathbf{P}_{0,j}, \mathbf{P}_{1,j}, \mathbf{P}_{2,j}$ are the three transverse control points at length station $j$, and $\mathcal{L}_{3}$ sums the squared pairwise separations of the blade cross-section over all $C_v$ stations. It is thus a transverse-compactness prior that penalizes sideways spreading of the cross-section into a neighboring organ or into noise beyond the blade edge, where the one-sided Chamfer term provides the weakest constraint. The implementation names this weight for a tip-proximity role, because the same expression addresses the last three control indices. At $C_u=3$ the two readings coincide, and the term acts transversely. The node-side station is included in the sum, but its row is unaffected because it is restored to its initial coordinates after each step. The weights $\lambda_{1}=10^{-2}$, $\lambda_{2}=10^{-4}$, and $\lambda_{3}=8\times 10^{-4}$ are retained from~\citep{hadadi2025procedural}.

\paragraph{Orientation check against self-folding}
For fits selected by the prealignment gate, we evaluate the surface on a dense $25 \times 101$ grid, compute the cross product of the two finite-difference cell directions, and compare it with the corresponding cell of the guarded initialization. For each cell whose initial area exceeds the numerical tolerance, the candidate cell is invalid when its orientation cosine is below $0.02$ or its area is less than $2\%$ of the corresponding initial cell area. If any cell is invalid, the proposed Adam step is retried at successively halved fractions of its displacement, up to twelve times, and the first fraction with no invalid cells is retained. If no fraction passes, the step is rejected and the control net returns to its pre-step state. The Adam first-moment estimate is scaled by the accepted fraction to reduce immediate movement in the rejected direction. This step halving implements a line search along the proposed update under the two cell-level constraints. The per-step orientation check is disabled for the original baseline fits; their eligible post-fit corrections are instead assessed by the collar and middle-fold tests below.

\paragraph{Repairing a folded base collar}
The fixed node-side row preserves the attachment, but optimization can still fold the downstream rows beneath it when the extracted path begins behind the trusted blade base. We therefore test a local post-fit repair only for leaves whose path-start repair in \secref{sec:params_ppg} was applied. The candidate is retained only when its row progression, transverse orientation, and sampled collar-cell orientation pass the stated geometric checks; otherwise, the original fit is retained. \ref{sec:appendix:collar} gives the candidate construction and acceptance criteria.

\paragraph{Repairing a mid-blade fold}
The second post-fit diagnostic addresses a fold near the middle of the blade. Such a fold can occur when the procedural cross-section is rotated about the midline relative to the observed blade over part of its length.

The diagnostic uses three tests against the procedural initialization. First, on control rows $2$ through $C_v-3$ whose initial width is at least $20\%$ of the maximum initial width, it flags an angle of at least $90^\circ$ between the fitted and initial end-to-end transverse vectors. Second, on control rows $1$ through $C_v-3$, it flags a centerline step whose projection onto the corresponding initial step direction is less than $-0.002$\,m. Third, it evaluates the surface on a $17\times61$ grid, excludes the outer $20\%$ of the grid at each blade end, and flags the surface when more than $3\%$ of the cells with nondegenerate initial area are collapsed or have an orientation cosine below $-0.05$. A violation of any test flags a middle fold.

The mid-blade repair is tested only for a baseline fit that was not already selected for guarded prealignment or collar repair. When a fold is diagnosed, we return to the procedural initialization and rotate its transverse control rows about the unchanged midline toward the principal directions of nearby target points. The node row remains unchanged, and the roll angles are smoothed along the blade before the leaf is refit. A fold-free candidate is accepted only when its target-to-surface fit ratio relative to the existing fit is at most $1.02$. If a residual fold remains, the candidate can be accepted only under three conditions. The counts of flagged transverse rows and backstep rows and the fraction of flagged surface cells must not increase, at least one of these three measures must decrease, and the target-to-surface fit must strictly improve while remaining within the $1.02$ ratio. The baseline measures use the original procedural surface as their reference, whereas the candidate measures use the locally rolled procedural surface. Otherwise, the existing fit is retained.

\paragraph{Residuals associated with the exported surface}
Let $Y_{\mathrm{fit}}$ be the fitting target in meters, subsampled to at most $4000$ blade points, and let $S_{\mathrm{grid}}$ be the $8\times32$ samples from the final selected surface. The fitting stage records
\begin{equation}
  d_{\mathrm{grid}} =
  \frac{1}{|Y_{\mathrm{fit}}|}\sum_{y\in Y_{\mathrm{fit}}}
  \min_{s\in S_{\mathrm{grid}}}\lVert y-s\rVert_2^2 ,
\end{equation}
in m$^2$, with RMS form $1000\sqrt{d_{\mathrm{grid}}}$ in millimeters. This sampled-grid quantity is retained in the run metrics for guard decisions and auditing. Because its value includes the spacing between the grid samples, it is not the blade-level residual reported in \secref{sec:quantitative}. For that result, the selected STL surface is treated as a triangle mesh $\mathcal{T}_{\mathrm{export}}$, and the full corrected blade cloud, denoted by $Y_{\mathrm{full}}$, is used to compute
\begin{equation}
  r_{\mathrm{export}} =
  1000\sqrt{
  \frac{1}{|Y_{\mathrm{full}}|}\sum_{y\in Y_{\mathrm{full}}}
  d\!\left(y,\mathcal{T}_{\mathrm{export}}\right)^2
  },
\end{equation}
where $d(y,\mathcal{T}_{\mathrm{export}})$ is the exact point-to-triangle distance in meters. All blade-level residuals reported in \secref{sec:results} use $r_{\mathrm{export}}$.

\subsection{Full-plant assembly and export}
\label{sec:report}
After per-leaf refinement, the selected leaf surfaces are reassembled with the remaining procedural surfaces. The node-side row remains fixed in all selected leaf surfaces, so refinement retains the procedural attachment positions. The export contains one ordered DAT surface for each input surface split, together with a combined STL. When a stalk split is present, it remains unchanged procedural geometry and is scaled to the meter units used by the fitted leaves. The NURBS output of a plant is accepted as complete only when all expected leaves have succeeded, the required fidelity metrics are finite and unit-consistent, the guard-decision record identifies the selected-geometry metric, and the referenced per-leaf and full-plant output files pass the completeness and checksum tests. The report stage checks the required outputs and metrics for each of the $100$ evaluated plants before the cohort-level results are published. A missing required leaf output or fidelity metric therefore prevents inclusion of an incomplete plant result.

\section{Results and Discussion}
\label{sec:results}

Unless stated otherwise, all numbers in this section come from runs of the nine stages of \secref{sec:methods} on the $100$ plants of \secref{sec:dataset}, under a single fixed parameter set applied unchanged to all plants. Each number is taken from the per-leaf metrics and segmentation reports that these runs wrote. The principal exceptions are identified where they appear: (a) the annotation timing in \tabref{tab:runtime}, which is pooled over a larger batch, (b) the published baseline figures of \secref{sec:comparison}, which were measured on the hardware of the earlier study, and (c) the manual count and label comparison of \secref{sec:seg_validation}, which are external references not produced by the pipeline. Additional controlled reruns are identified in their respective subsections. \secref{sec:qualitative} to \secref{sec:quantitative} report organ-level results on the ten plants of \tabref{tab:plants}, which together contain $118$ of the $1{,}072$ recovered leaves, and \secref{sec:hundred} gives the same quantities over all $100$ plants. We report, in turn, organ-level recovery, the fidelity of the fitted surfaces to the scan, scaling to the full set, how often the geometric checks engage and their effect when they do, the comparison with the earlier PSO+NURBS-Diff workflow, computational cost, and agreement with an external reference.

\subsection{Qualitative results}
\label{sec:qualitative}

\paragraph{Per-view annotation and back-projection}
\figref{fig:vlm_output} and \figref{fig:vlm_to_3d} show the annotation, mask, and back-projection stages for all eight views of one plant. The leaf annotations traced curves along the visible blades and placed magenta dots near the apparent junctions between the leaves and the stalk. Because each accepted annotation is registered to its rendered input view, the stored camera transform maps the extracted overlays back onto the cloud without additional calibration. A recurring difficulty occurred near the top of the canopy, where consecutive leaves attach at nearby heights and the tassel occludes them. In these views, a small upper leaf was sometimes omitted, or two leaves were represented by one stroke. The mask and consensus stages corrected some of these cases when curve continuity or other views provided sufficient support. \secref{sec:seg_results} reports the resulting view support and organ-level recovery.

\begin{figure*}[t!]
    \centering
    \setlength{\tabcolsep}{1.5pt}
    \renewcommand{\arraystretch}{0.9}
    \newcommand{\rescell}[1]{%
        \raisebox{-0.5\height}{\includegraphics[height=0.215\textwidth]{#1}}}
    \newcommand{\rowname}[1]{\scriptsize #1}
    \begin{tabular}{@{}>{\raggedright\arraybackslash}m{0.135\textwidth}
                     @{\hspace{3pt}}ccccc@{}}
        & \scriptsize CML69 & \scriptsize CML238 & \scriptsize CML331
        & \scriptsize Kui2021 & \scriptsize 4226 \\[1pt]
        \rowname{Input scan (stage 1)}
        & \rescell{dataset_9005.png} & \rescell{dataset_9006.png}
        & \rescell{dataset_9007.png} & \rescell{dataset_9008.png}
        & \rescell{dataset_9000.png} \\
        \rowname{VLM leaf annotation (stage 3)}
        & \rescell{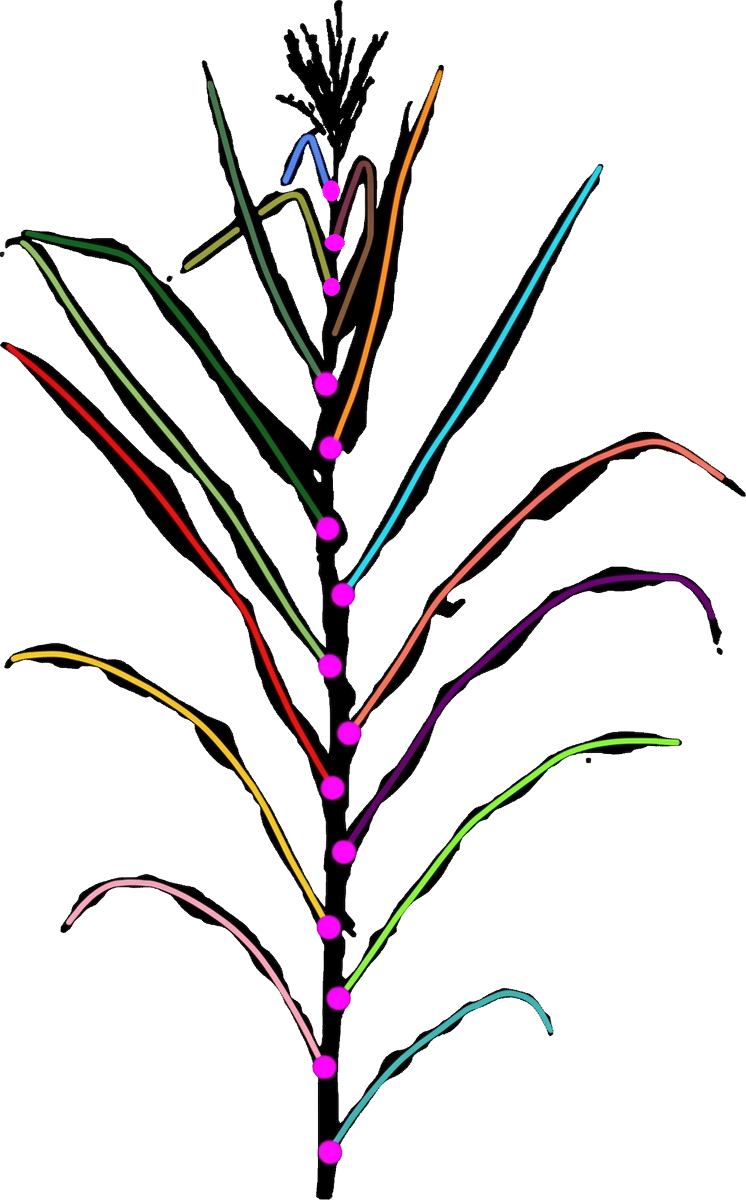} & \rescell{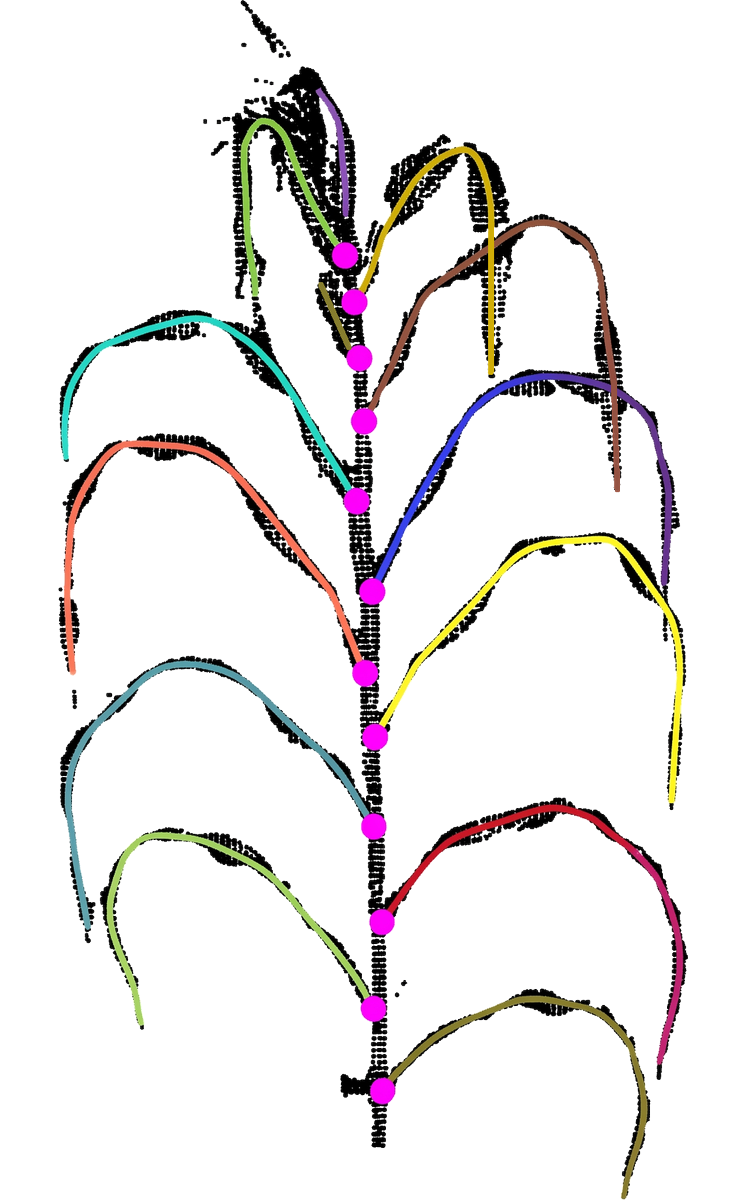}
        & \rescell{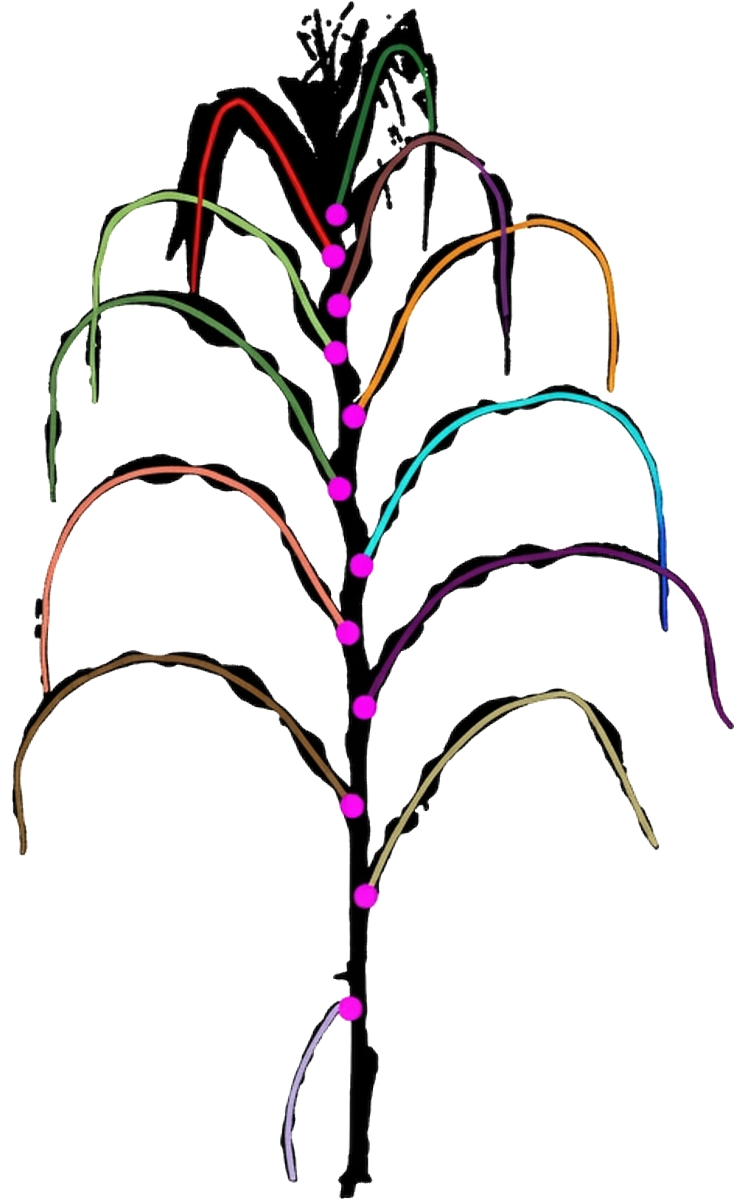} & \rescell{annot_9008_front.png}
        & \rescell{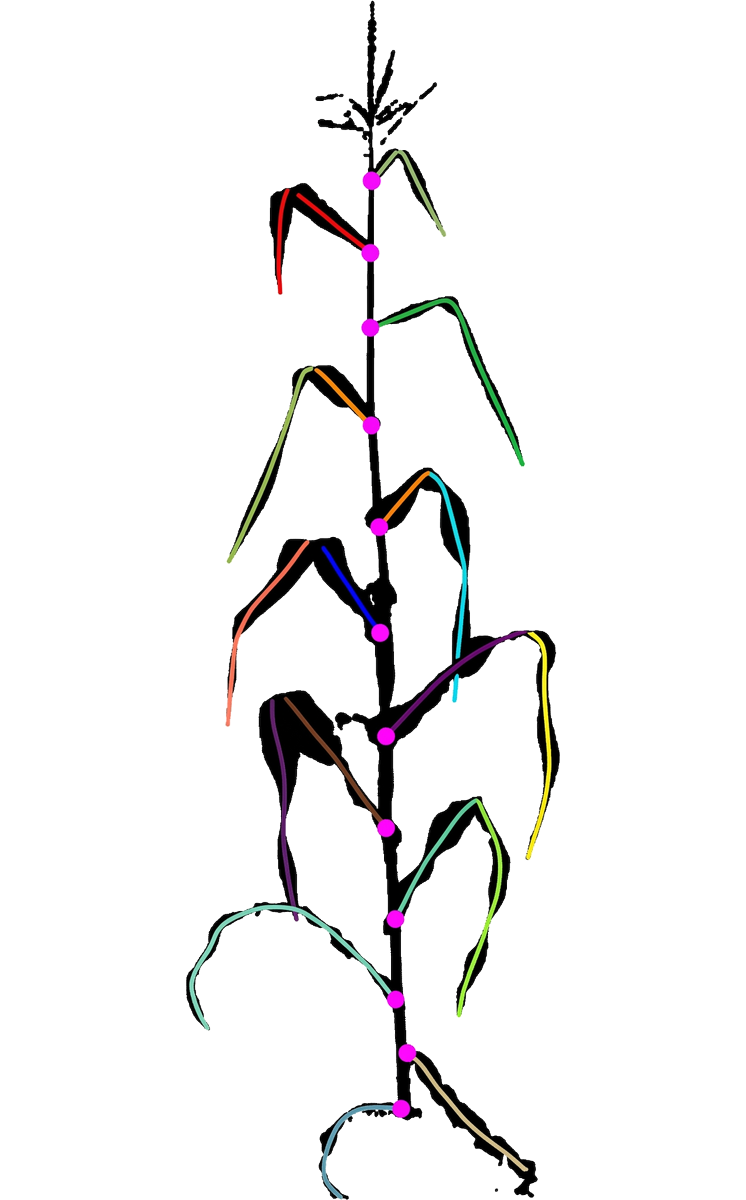} \\
        \rowname{Blade segmentation (stage 6)}
        & \rescell{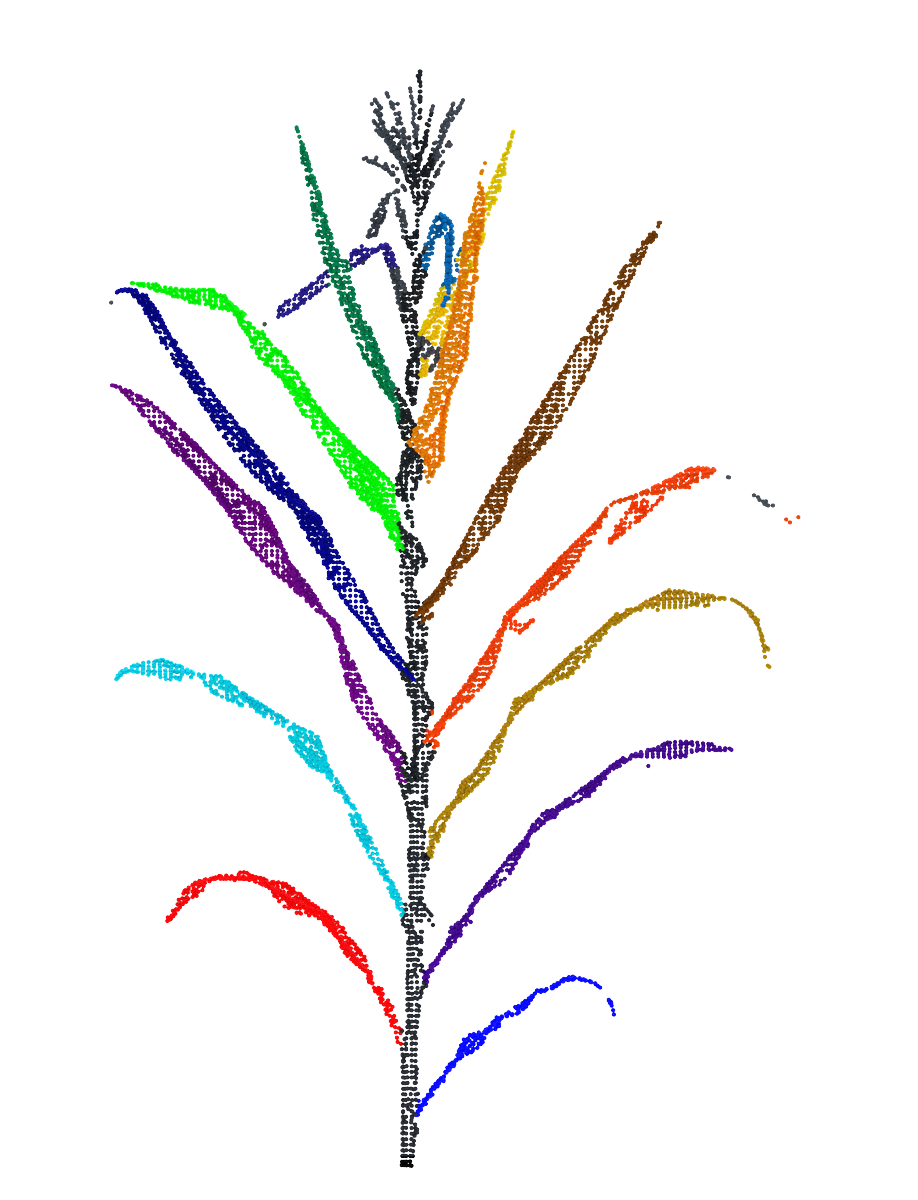} & \rescell{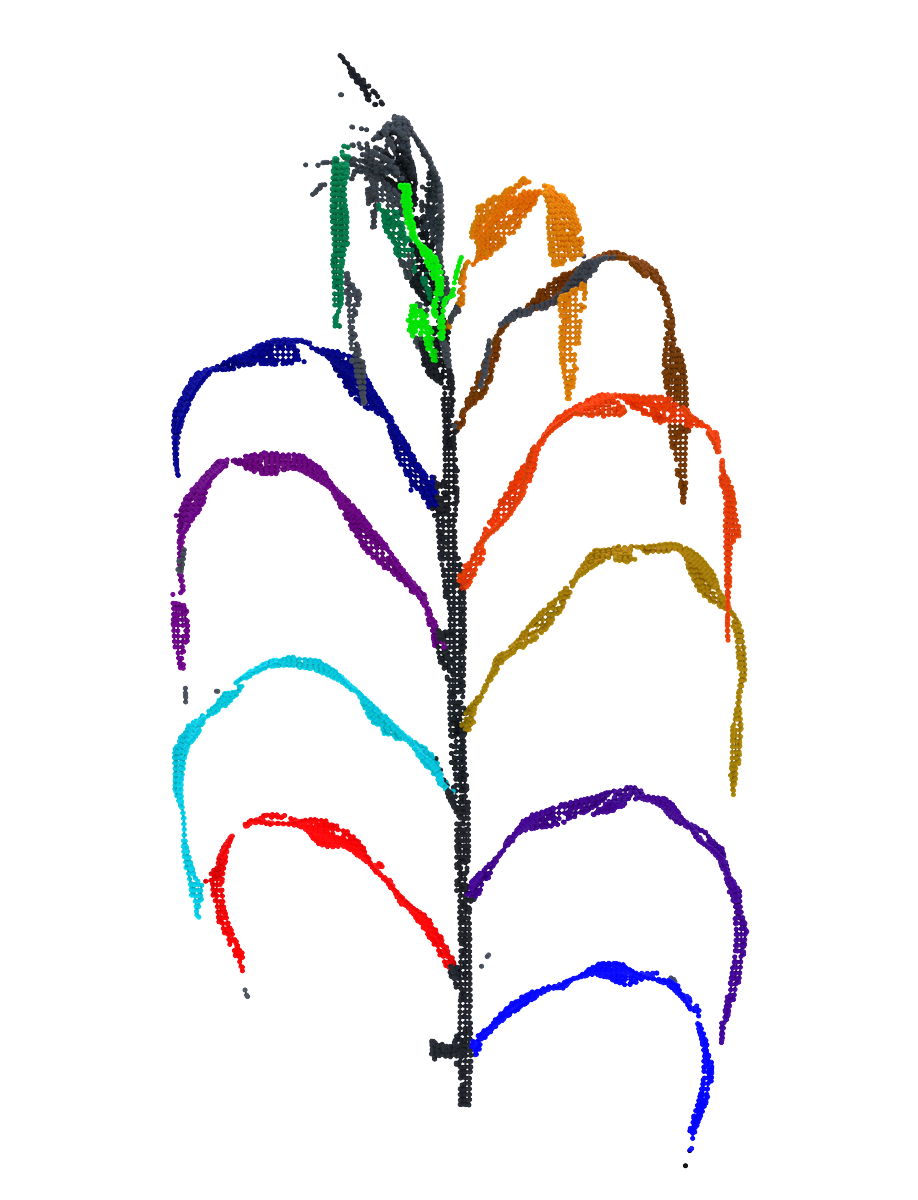}
        & \rescell{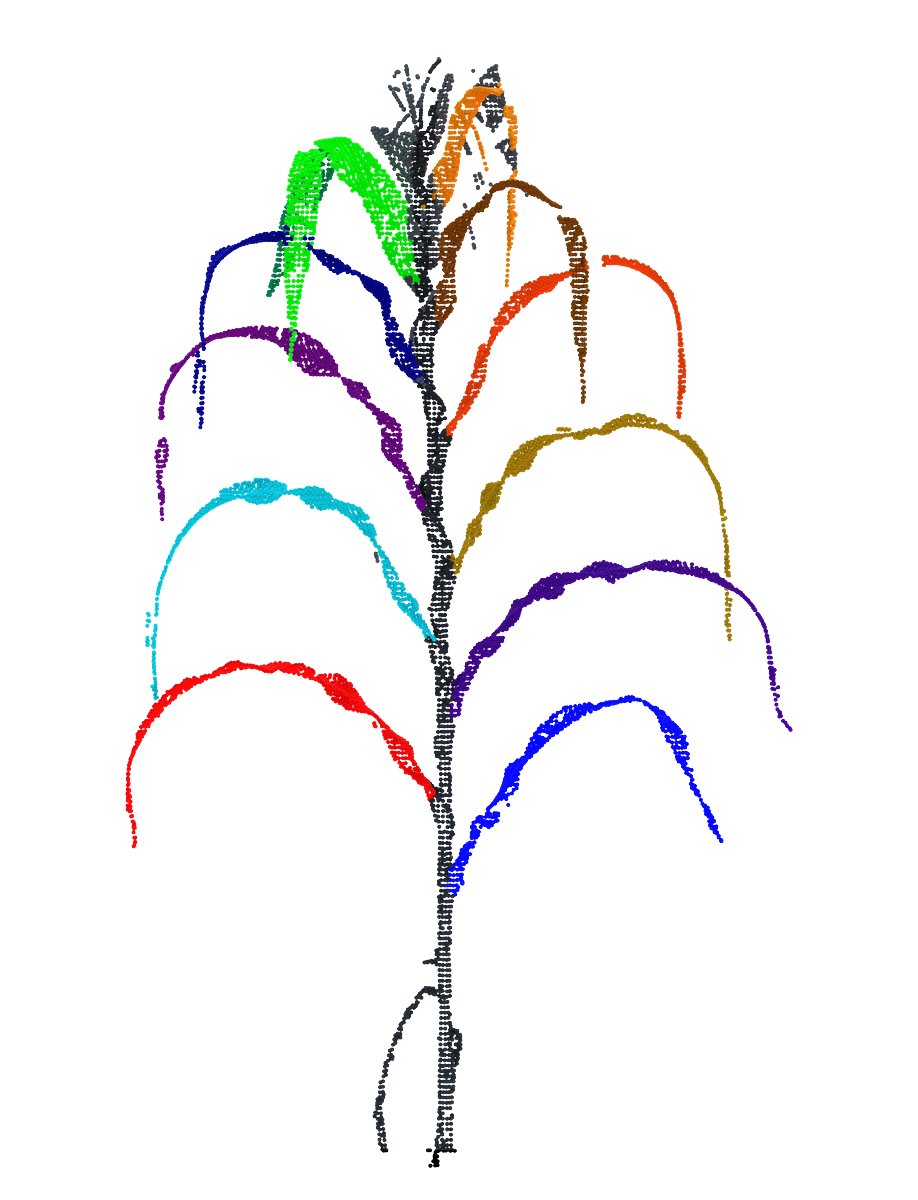} & \rescell{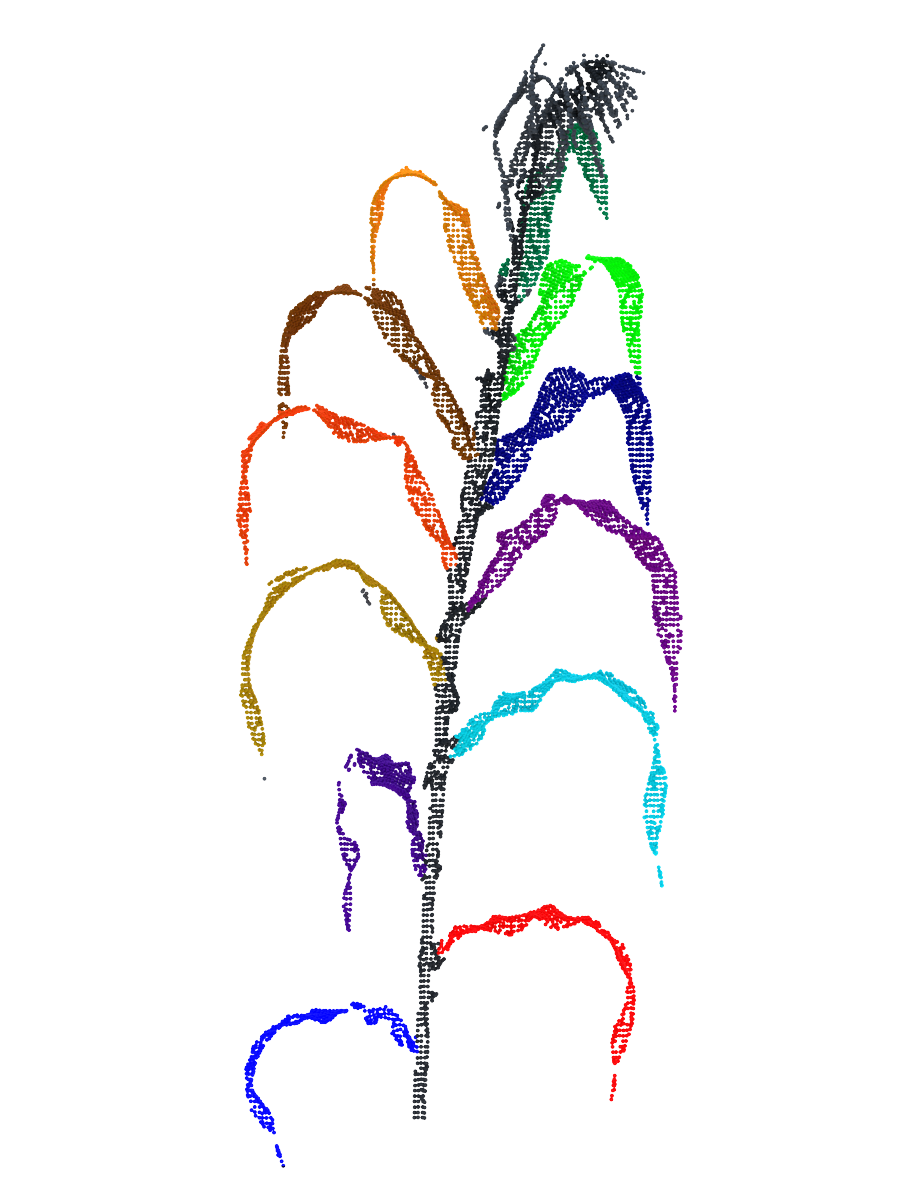}
        & \rescell{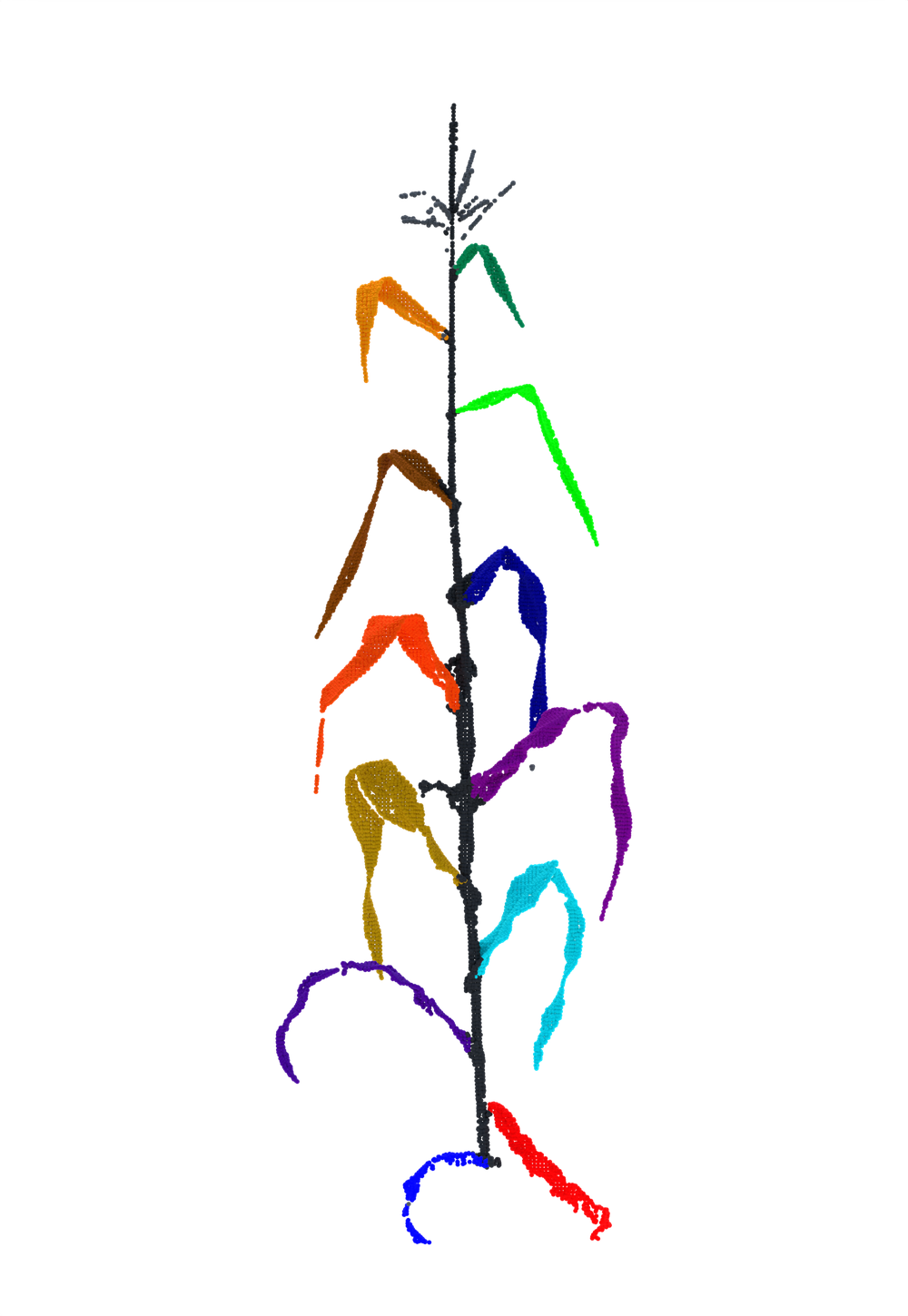} \\
        \rowname{Initial procedural model (stage 7)}
        & \rescell{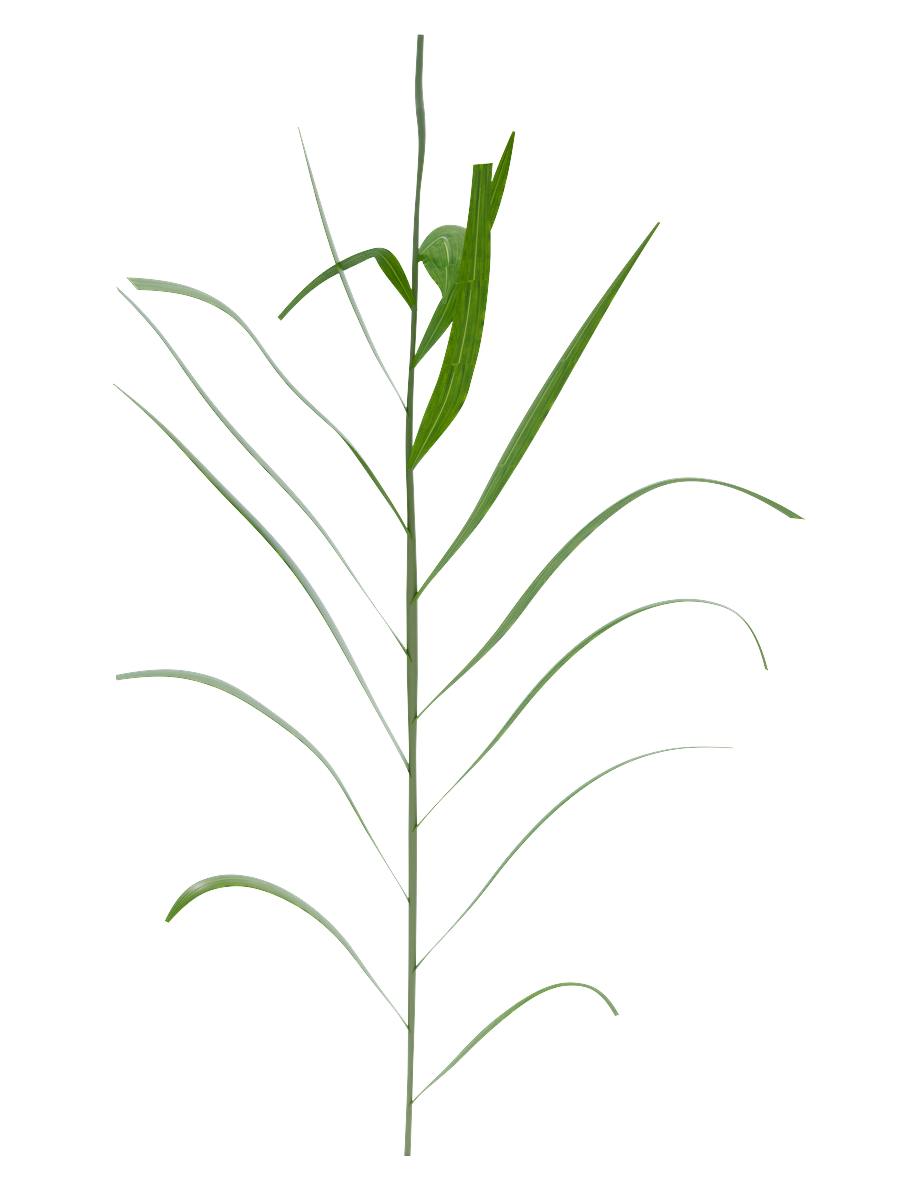} & \rescell{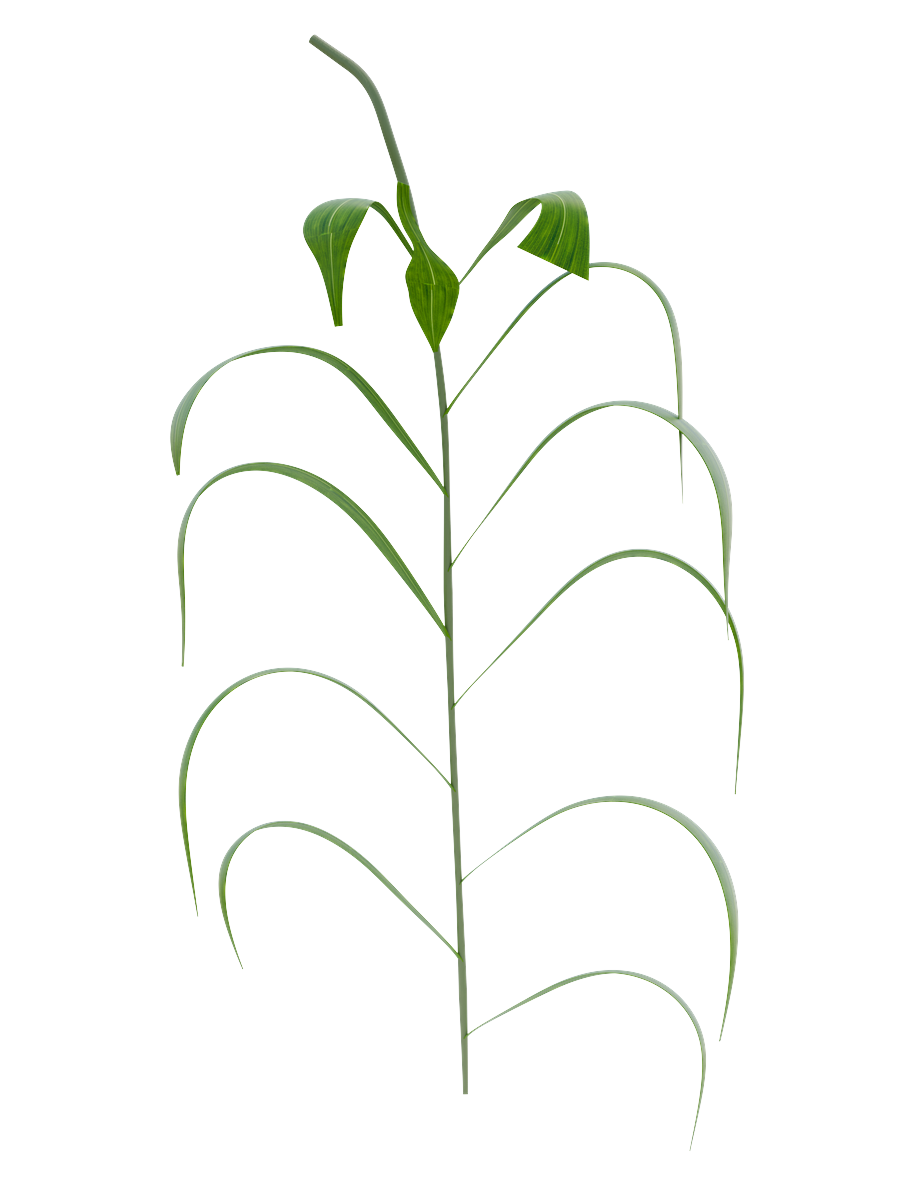}
        & \rescell{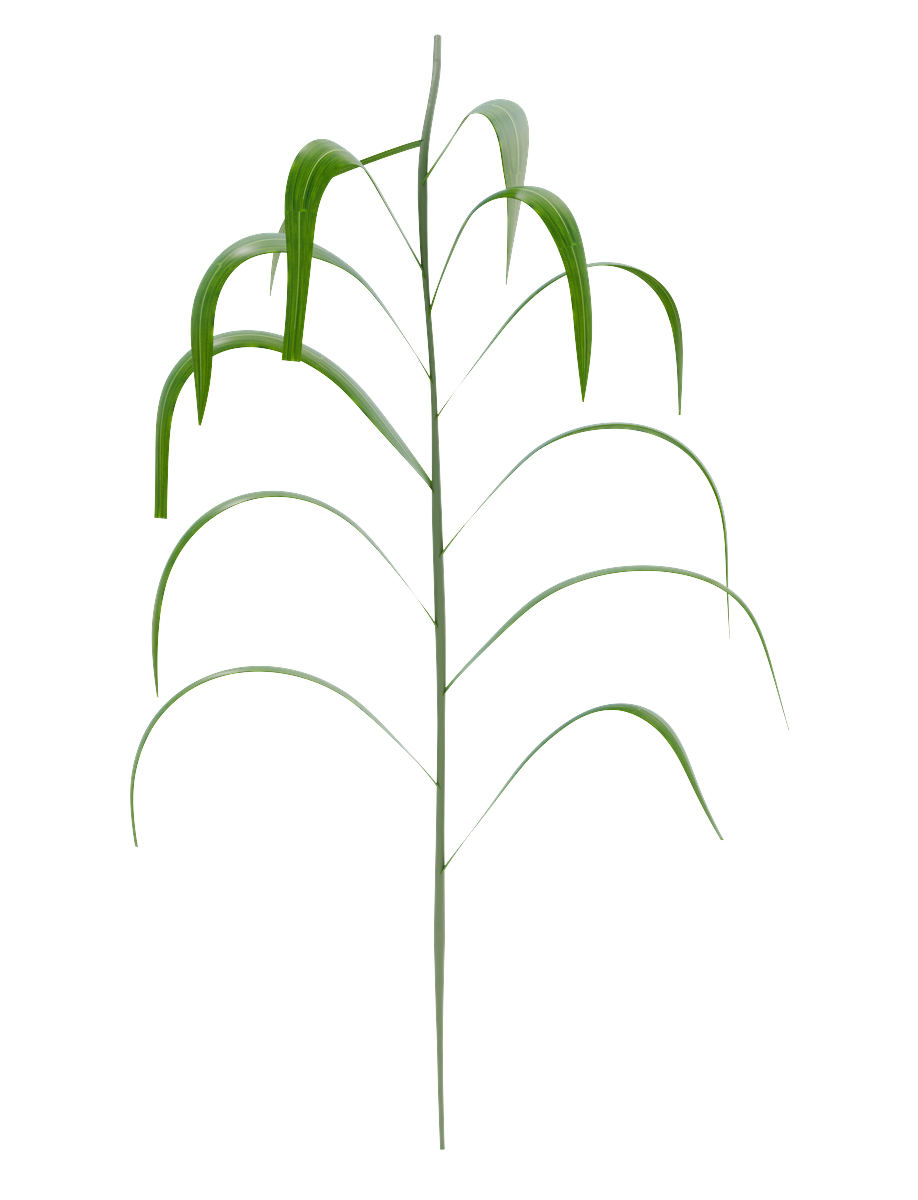} & \rescell{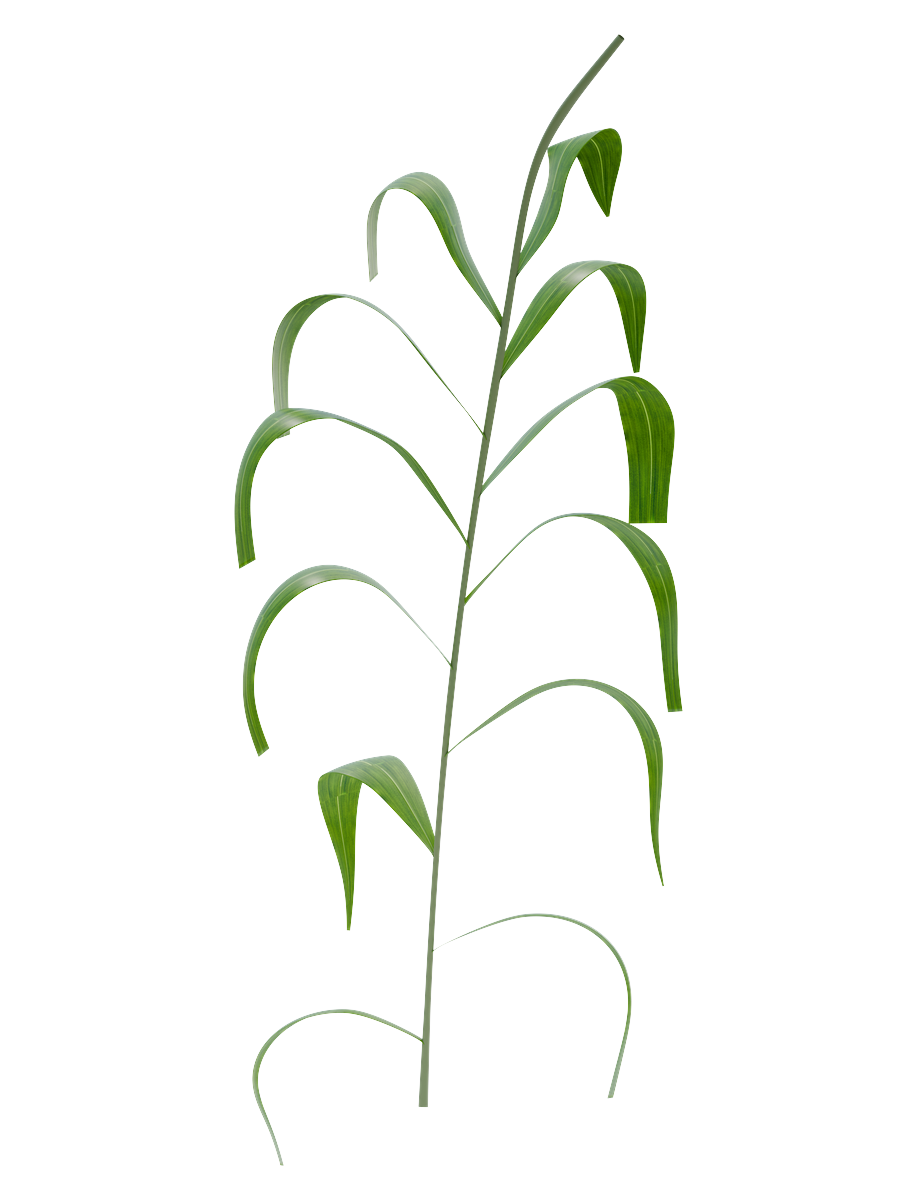}
        & \rescell{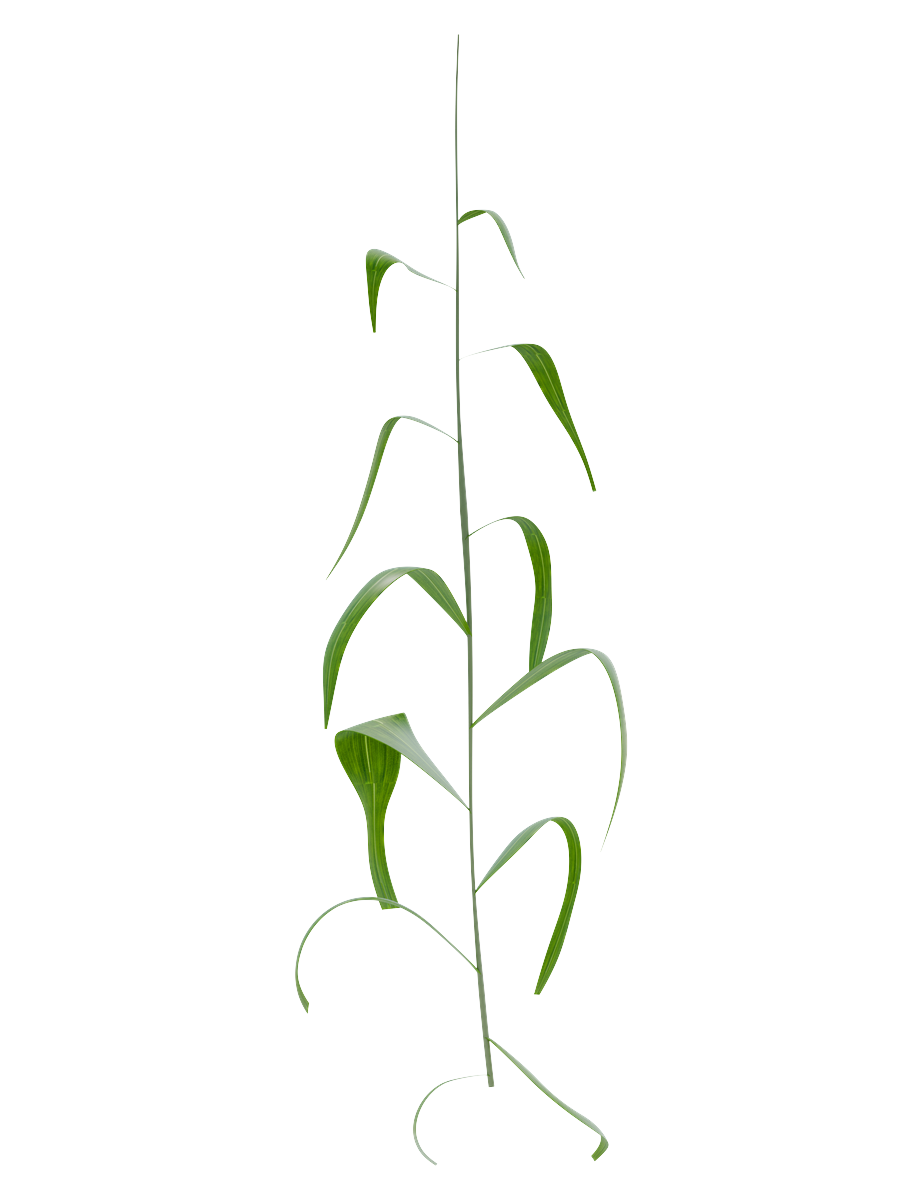} \\
        \rowname{Refined procedural model (stage 8)}
        & \rescell{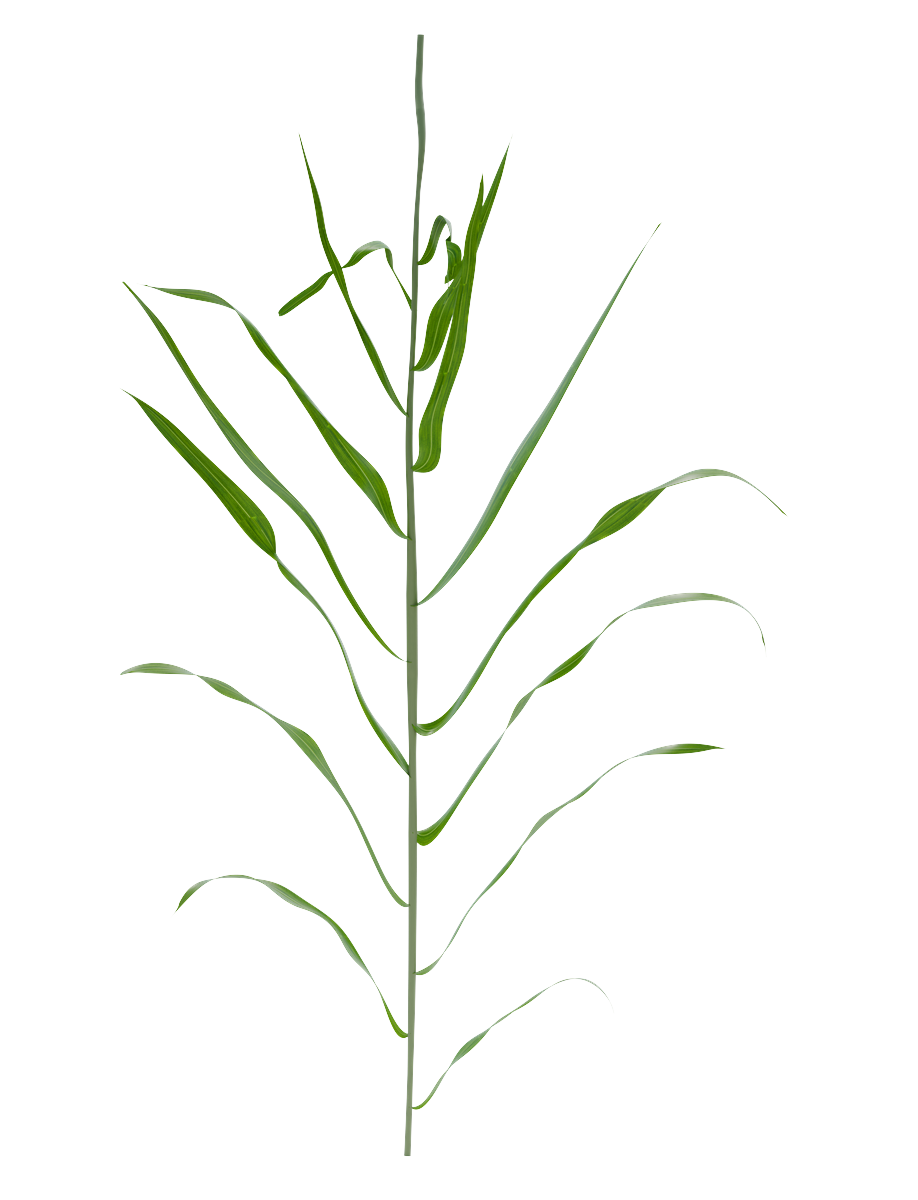} & \rescell{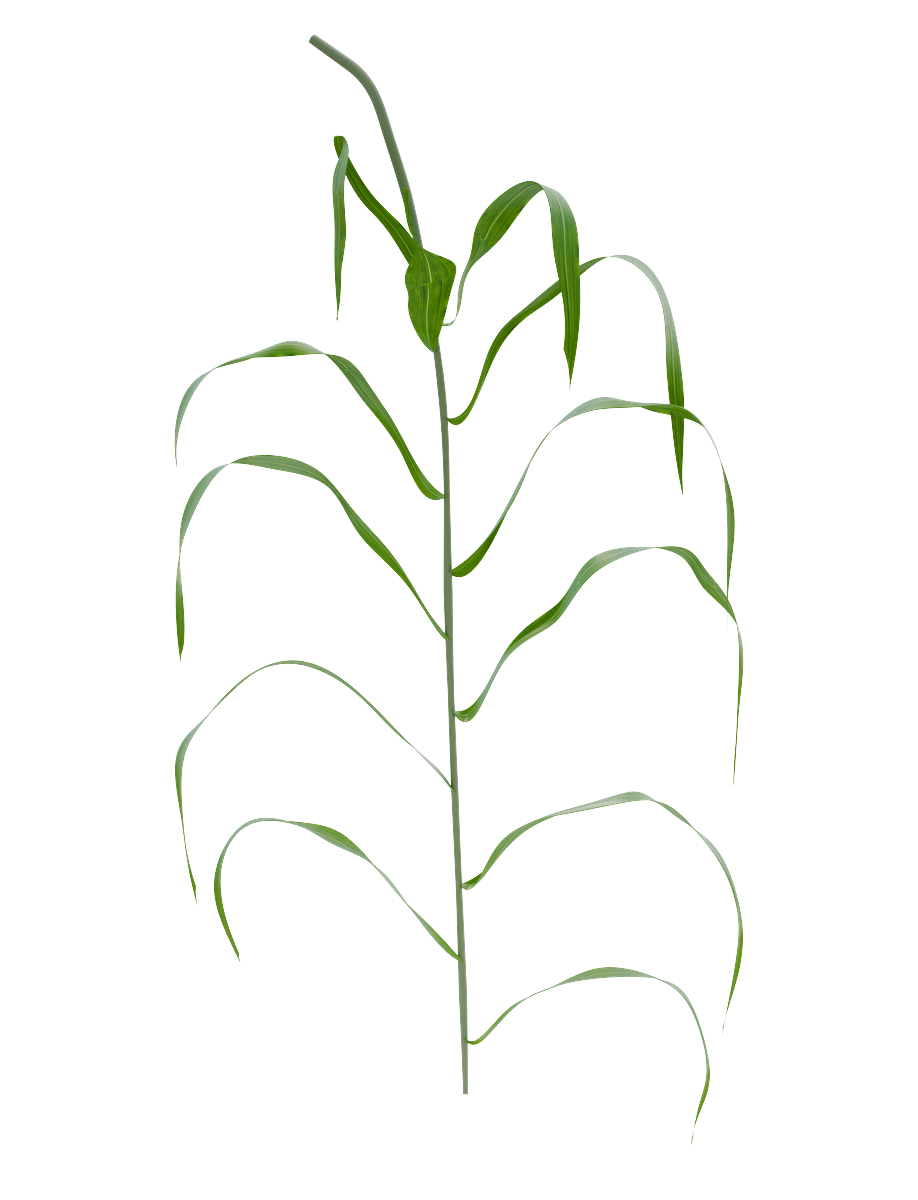}
        & \rescell{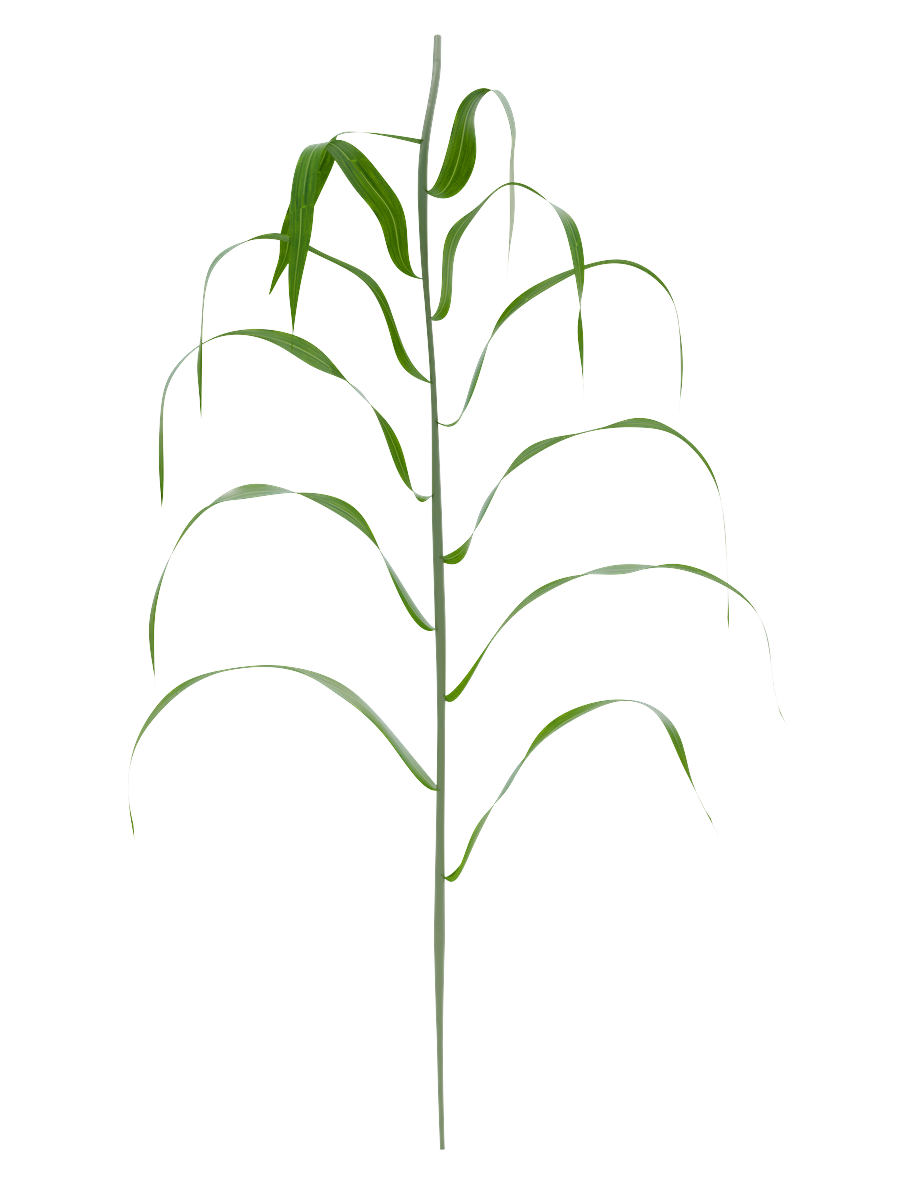} & \rescell{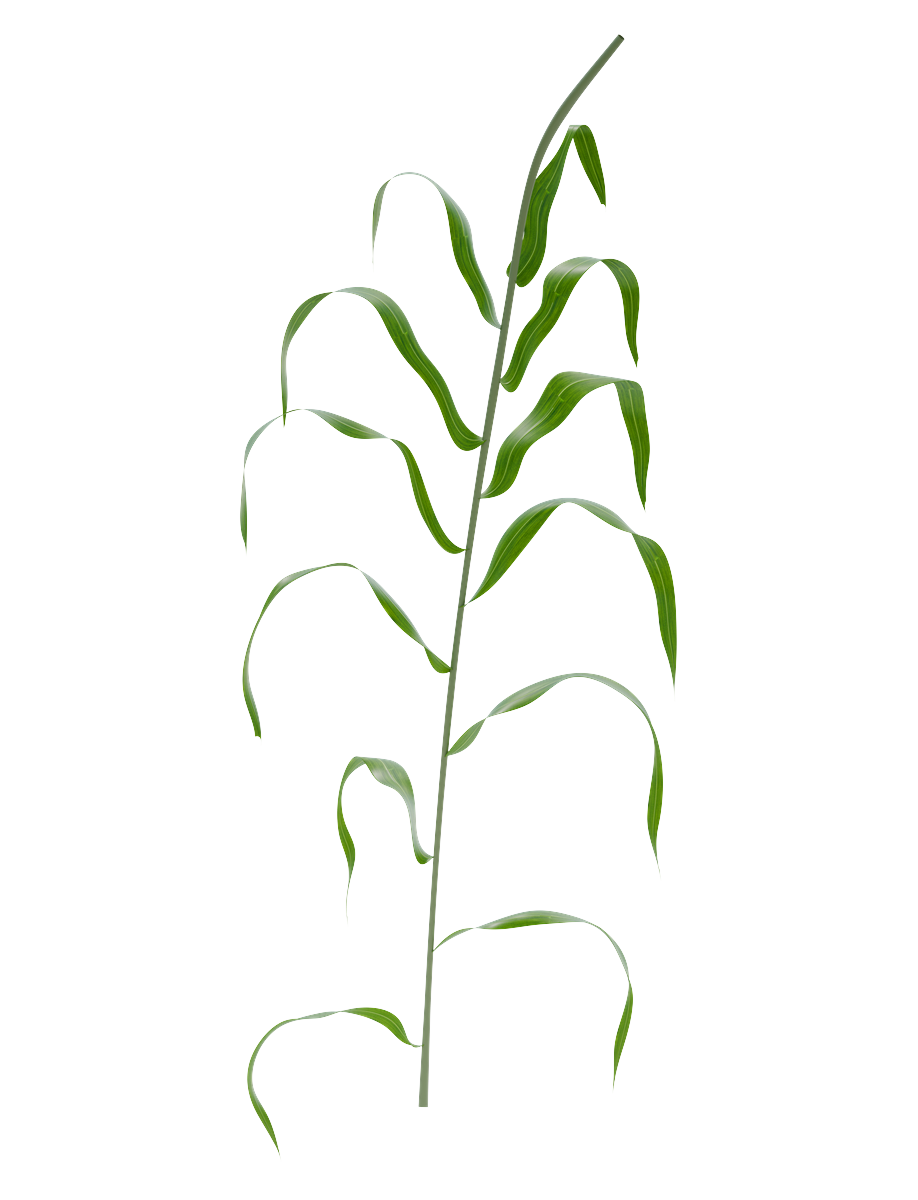}
        & \rescell{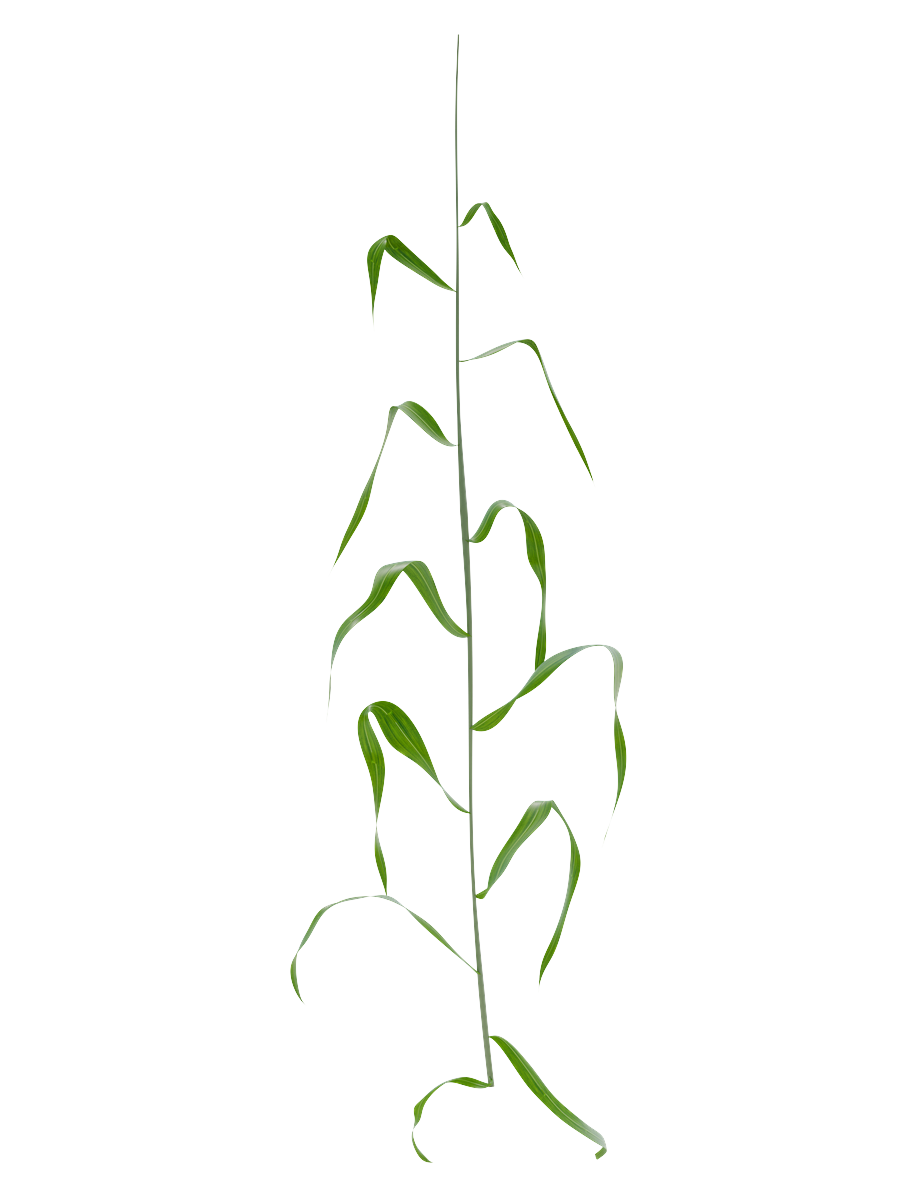} \\
    \end{tabular}
    \caption{End-to-end reconstruction of five of the $100$ evaluated plants, with the genotype named above each column. From top to bottom, the rows show the input scan in the colors recorded by the scanner, the front-view leaf annotation used for reconstruction (one of the eight leaf-view annotations), the blade-level segmentation of \secref{sec:leafseg}, the initial procedural model generated by FloraForge~\citep{hadadi2025floraforge} from the measured plant descriptor, and the same procedural model after NURBS-Diff refinement of its leaf surfaces against the segmented blade points. Each row is labeled with the stage of \figref{fig:pipeline} that produced it. In the segmentation row, each leaf is grown from its consensus midline out to the full blade, and ambiguous points are left unlabeled (gray). All rows except the annotation are rendered with the same camera within a column, so a leaf can be followed from the scan to the refined surface.}
    \label{fig:reconstruction}
\end{figure*}

\paragraph{From per-view annotations to a refined procedural model}
\figref{fig:reconstruction} shows five evaluated plants at five stages, from the input scan to the refined procedural model. In these examples, the refined meshes preserved the overall plant architecture, including stalk shape and leaf arrangement, and represented the curvature of individual blades and their attachment order. This assessment was visual. \secref{sec:quantitative} reports the corresponding surface-to-blade distances and agreement between traits measured on the exported surfaces and on the segmented blade clouds.

\paragraph{NURBS-Diff refinement}
The last two rows of \figref{fig:reconstruction} show the outputs before and after NURBS-Diff. NURBS-Diff fitted each leaf surface against its segmented blade points. Because the node-side control row was restored to its position in the initial procedural model after every optimization step, attachment was preserved by construction. \secref{sec:quantitative} reports the fidelity of the final fitted surfaces.

\subsection{Organ-level recovery}
\label{sec:seg_results}

\paragraph{Contribution of the eight views}
The mask stage of \secref{sec:mask} produced $890$ per-view leaf components over the $80$ rendered views of the ten plants. Consensus reconciled these into $117$ global leaves, and the segmentation stage of \secref{sec:leafseg} recovered one more, giving $118$. Of the $936$ possible leaf-view pairs, $835$ ($89.2\%$) were annotated. Thus, an average view contained $10.4$ of a plant's leaves (mean $11.7$ per plant), i.e., a single viewpoint missed roughly one leaf in ten (\tabref{tab:consensus_support}).

The main benefit of the eight views is redundancy, which leaf count alone does not show. On nine of the ten plants, the best single view already contained all the consensus leaves, and on T8 it missed one. The per-view component count was also a poor proxy for leaf count. For example, 4226 was oversegmented into $15$ components in one view but has $12$ leaves. Redundancy helped in three ways (\tabref{tab:consensus_support}). First, consensus excluded $41$ components that did not enter a supported global leaf and reconciled $14$ additional same-view fragments with supported leaves. Together, these exclusions and reconciliations account for the $55$-component difference ($6.2\%$ of the total) between the mask output and the observed leaf-view pairs. Second, no fixed viewpoint was sufficient. The front view alone resolved $108$ of the $117$ consensus leaves, and on T8 the weakest single view resolved only $8$ of the $14$ leaves. Third, support was unevenly distributed. Of the $117$ leaves, $80$ ($68.4\%$) were seen from all eight views. Four leaves were supported by only two views (the two-view floor) and would have been lost had that floor been raised. These four leaves had the weakest accepted view support and are discussed below.

\begin{table}[t!]
    \centering
    \caption{Per-view visibility of the recovered leaves. ``Leaves'' is the
    number of global leaves established by cross-view consensus. ``Front'' is the number of these leaves resolved by the front view alone. ``Mean'' and ``Fewest'' give the mean and the minimum of this count over the eight views. ``All eight'' counts the leaves resolved in all eight views, and ``Two only'' counts the leaves resolved in exactly two views, i.e., at the two-view support floor. ``Components'' is the total number of per-view leaf components produced by the mask stage over the eight views.}
    \label{tab:consensus_support}
    \setlength{\extrarowheight}{2pt}
    \footnotesize
    \begin{tabular}{@{}lrrrrrr r@{}}
        \toprule
        Genotype & Leaves & Front & Mean & Fewest & All eight
        & Two only & Components \\
        \midrule
        M162W   & 15 & 14 & $13.0$ & 11 & 9  & 1 & 109 \\
        CML69   & 15 & 14 & $13.2$ & 11 & 11 & 0 & 113 \\
        Kui2021 & 12 & 11 & $10.5$ & 9  & 7  & 1 & 85 \\
        CML238  & 12 & 11 & $11.0$ & 10 & 9  & 0 & 97 \\
        CML331  & 11 & 11 & $9.9$  & 9  & 8  & 0 & 84 \\
        4226    & 12 & 12 & $11.4$ & 10 & 9  & 0 & 96 \\
        78551S  & 8  & 8  & $7.5$  & 6  & 5  & 0 & 74 \\
        A679    & 8  & 7  & $7.1$  & 6  & 6  & 0 & 60 \\
        CI90C   & 10 & 9  & $9.4$  & 9  & 8  & 0 & 76 \\
        T8      & 14 & 11 & $11.4$ & 8  & 8  & 2 & 96 \\
        \midrule
        All ten & 117 & 108 & $10.4$ & 6 & 80 & 4 & 890 \\
        \bottomrule
    \end{tabular}
\end{table}

\paragraph{Unlabeled points}
The blade propagation of \secref{sec:leafseg} left some points unlabeled. The ten segmentation reports account for $99{,}633$ points: $18{,}481$ classified as stalk, $74{,}645$ assigned to a leaf in the final segmentation, and $6{,}507$ left unlabeled. The unlabeled share among the $81{,}152$ non-stalk points was $8.0\%$, ranging from $2.2\%$ on 78551S to $15.1\%$ on Kui2021. We use this fraction as a diagnostic. The unlabeled fraction was highest on Kui2021 ($15.1\%$) and A679 ($14.7\%$). These two plants also had the largest cloud-to-mesh component of the whole-plant Chamfer distance (\secref{sec:quantitative}). This pattern is consistent with many unlabeled points belonging to tassel, sheath, and contact regions instead of blade tissue. The two readouts are not independent, since unlabeled material is excluded from the descriptor and has no corresponding surface in the model. Assigning all unlabeled points to leaves would risk distorting the measured widths and lengths. The organ-count and point-level comparisons in \secref{sec:seg_validation} evaluate missed leaves and the recovery of labeled blade tissue.

\paragraph{Outcome of the automatic fragment correction}
Across the ten plants, the \textsc{move} and \textsc{split} tests of \secref{sec:leafseg} rejected $579$ candidate corrections and accepted two, both on genotype CML331. Of the rejected candidates, $523$ failed the geometric support tests and $56$ failed the additional tests for moving a primary component. The accepted \textsc{move} reassigned $90$ points from leaves~$7$ and~$9$ to leaf~$11$, which was the only supported endpoint continuation. A \textsc{split} then moved $289$ points from leaf~$8$ into a new twelfth leaf; both connected-component and directional-continuation evidence supported the split. Both corrections passed their point-conservation checks, neither had to be reverted, and the leaf ordering remained monotonic. No plant-specific correction rules or labels were supplied.

The gates are conservative by design and accepted nothing on nine of the ten plants. This result shows only that no residual error on those plants passed the gates, which is weaker than showing that the plants are error-free. Such errors are most likely in the upper canopy discussed below. In addition, a leaf recovered by a \textsc{split} has no annotation-derived collar, and the stage flagged that its attachment was inferred from the fragment geometry. The new CML331 leaf had a base-to-node gap of $18.3$\,cm, compared with a median of $2.0$\,cm for leaves whose bases were observed. The constrained node rule therefore reported a gap violation. CML331 leaf~$12$ was nonetheless retained in the reported model and underwent parameter extraction and NURBS-Diff refinement. The node report records the violation.

\paragraph{Leaves with the weakest view support}
With only ten plants, we can list the leaves whose recovery had the weakest support within the run itself. (\secref{sec:seg_validation} compares the segmentation leaf by leaf against an independent reference.) The weakest support was usually in the upper canopy, where blades attach within a few centimeters of one another and the views disagreed on whether a stroke belonged to one leaf or to two. Four leaves were supported by exactly two views, the two-view support floor of \secref{sec:backproject}. Three of these were the topmost leaf of their plant (M162W leaf~$15$, Kui2021 leaf~$12$, T8 leaf~$14$). The fourth, T8 leaf~$1$, was the lowest leaf on that plant, at a height of $8.3$\,cm. All four were among the six smallest consensus clusters of the $117$. Weak view support therefore coincided here with small consensus clusters at both canopy extremes. One more leaf, CML331 leaf~$12$, was not established by consensus and entered the reconstruction only through the fragment correction of \secref{sec:leafseg} (\tabref{tab:consensus_support}). Raising the support floor would discard the four two-view leaves, whereas relaxing the correction gates would increase the risk of splitting a single blade into spurious leaves. We therefore retain the two-view floor and the conservative correction gates.

\paragraph{Attachment nodes under the constrained node rule}
\tabref{tab:nodes} summarizes the outcomes of the constrained node rule of \secref{sec:leafseg} for the $118$ attachments. The same pair of conditions was used for all plants, a base-to-node gap of at most $6$\,cm and a spacing of at least $4$\,cm between nearest nodes. The annotation-derived node was preserved on $114$ leaves. The rule is designed to preserve such a node because the node is supported by annotations from several views, whereas projection produces a geometric estimate. One node violated the gap condition and was repaired by projecting the leaf base onto the stalk centerline, which reduced the base-to-node gap of T8 leaf~$11$ from $8.7$ to $1.4$\,cm. The replacement was adopted only after checking that it did not push a neighboring valid node below the spacing floor. Two leaves on CML331 were projected outright because of the split of \secref{sec:leafseg}, namely the leaf that lost material and the new leaf, which had no node. Finally, no projection for CML69 leaf~$14$ satisfied both conditions, and this leaf kept its original node with the violation reported.

Overall, $116$ of the $118$ attachments ($98.3\%$) satisfied both conditions. The two that did not were CML69 leaf~$14$, at a gap of $8.2$\,cm, and CML331 leaf~$12$, at $18.3$\,cm. We prefer this behavior to the two alternatives. Projecting all nodes would replace the $114$ valid annotation-derived nodes with geometric estimates and would not guarantee the spacing condition. Never projecting would leave one leaf without a node and two others with nodes that violate the stated conditions. The segmentation stage flagged CML69 leaf~$14$ and CML331 leaf~$12$ because their measured base-to-node gaps exceeded the $6$\,cm limit. Visual review indicated that points between the segmented leaf base and the node were not assigned to the leaf, and this omission increased the measured gap. The pipeline outputs were retained unchanged, and no manual node correction was applied.

\begin{table}[t!]
    \centering
    \caption{Outcome of the constrained node rule for final node placement over the
    $118$ leaves of the ten plants. The two conditions require a base-to-node gap of at most $6$\,cm and a spacing of at least $4$\,cm to the nearest node. ``Conditions met'' is the number of leaves in each group whose final node satisfies both. Two leaves fail the base-to-node gap condition while satisfying the spacing condition.}
    \label{tab:nodes}
    \setlength{\extrarowheight}{2pt}
    \footnotesize
    \begin{tabular}{@{}lrrl@{}}
        \toprule
        Outcome & Leaves & Conditions met & Leaves affected \\
        \midrule
        Existing node preserved       & 114 & 114 & n/a \\
        Unsafe node repaired by projection & 1 & 1
            & T8 L11 \\
        Projected because required    & 2   & 1   & CML331 L8, L12 \\
        Unresolved                    & 1   & 0   & CML69 L14 \\
        \midrule
        Total                         & 118 & 116 & \\
        \bottomrule
    \end{tabular}
\end{table}

\subsection{Geometric fidelity}
\label{sec:quantitative}

\paragraph{B-spline reconstruction accuracy}
\tabref{tab:recon} reports the per-leaf B-spline reconstruction root-mean-square error (RMSE). This metric measures how closely the eight-point B-spline reproduces the geodesic midline path used to construct it and therefore quantifies the midline approximation written to the plant descriptor. Over the $118$ leaves, the median residual was $1.7$\,mm, with an interquartile range of $1.1$ to $2.8$\,mm. This median indicates that eight control points represented the maize midlines well within the $4.1$\,mm median point spacing of the working clouds (\secref{sec:dataset}). The median residual was also $0.24\%$ of the median blade length of $70.2$\,cm. The distribution had a long upper tail, with a 90th percentile of $3.9$\,mm and a single worst leaf at $13.1$\,mm, some $70\%$ above the next largest residual. The upper tail contained cases in which the fixed eight-point representation reproduced the sampled path less accurately; blade lengths in this set spanned $20.9$ to $102.7$\,cm. Increasing the control-point count could reduce this approximation error, but this option was not tested. This residual describes the procedural midline approximation. It does not describe the final fitted surface evaluated below.

\begin{table}[t!]
    \centering
    \caption{B-spline midline reconstruction RMSE across the $118$
    reconstructed leaves. Each leaf midline is represented by 8 control points with a degree-3 interpolating B-spline, and the residual is measured against the geodesic midline path from which the control points were sampled.}
    \label{tab:recon}
    \setlength{\extrarowheight}{2pt}
    \footnotesize
    \begin{tabular}{@{}lr@{}}
        \toprule
        Metric & Value (mm) \\
        \midrule
        Median RMSE     & $1.68$ \\
        Mean RMSE       & $2.17$ \\
        Std.\ dev.\     & $1.73$ \\
        Interquartile range & $1.10$ to $2.79$ \\
        90th percentile & $3.93$ \\
        Max             & $13.10$ \\
        \bottomrule
    \end{tabular}
\end{table}

\paragraph{Residual on points withheld from fitting}
The published fits used all corrected blade points because no blade reached the $4{,}000$-point target cap. The largest blade cloud had $1{,}338$ points and the median had $653$. The resulting residual is an in-sample fitting error. It measures how closely the surface reproduces its fitting targets, not its agreement with points excluded from optimization. To assess this distinction, we reran NURBS-Diff for all $118$ leaves after withholding a random fifth of each blade cloud under a fixed seed. This holdout withheld $14{,}925$ of $74{,}645$ points and left $59{,}720$ for the optimization, with all other parameters unchanged. We then recomputed both residuals with a single routine, as the exact distance from a point set to the exported triangulated surface. Using one routine ensured that neither residual depended on the internal records of the fitting stage. The surfaces reproduced unseen points nearly as well as fitted points. The mean RMS was $5.64$\,mm on the withheld points and $5.30$\,mm on the fitted points, a mean penalty of $0.34$\,mm and a median of $0.23$\,mm, or $8\%$ in the mean per-leaf ratio. As expected, the withheld residual was larger on $80$ of the $118$ leaves. The largest penalty, $2.9$\,mm, occurred on a leaf with only $22$ withheld points. Overall, this random holdout test provided little evidence of overfitting to individual sampled points. Because the withheld points came from the same segmented clouds, the holdout remains an internal check and not independent validation of blade geometry.

\begin{figure}[t!]
    \centering
    \includegraphics[width=\textwidth]{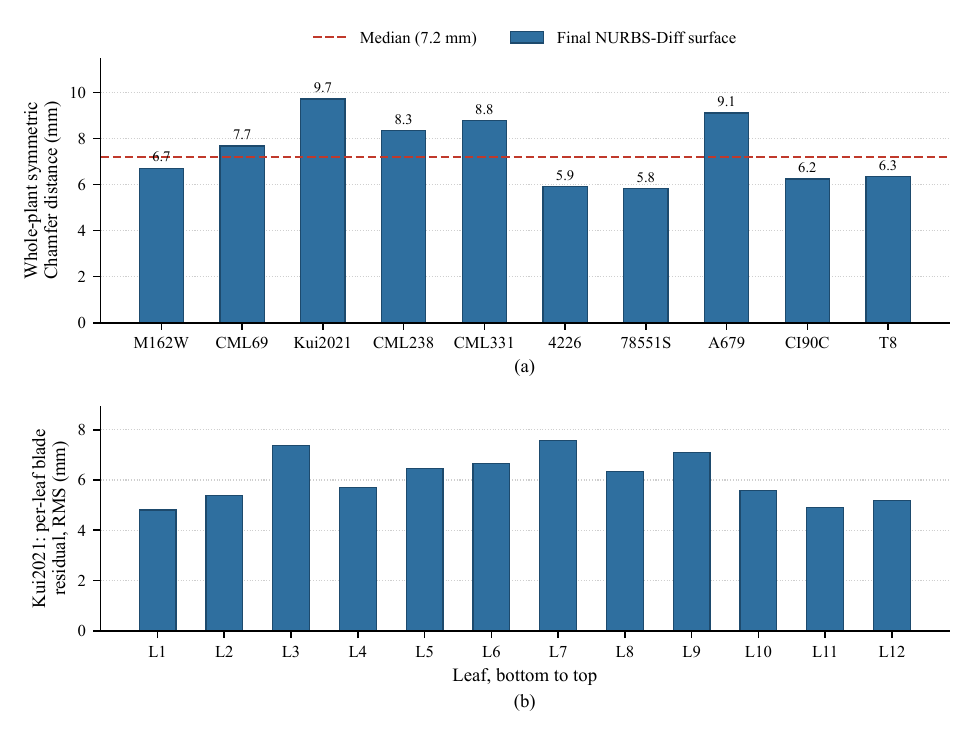}
    \caption{Fidelity of the final refined surfaces at two scales.
    \textbf{(a)} Whole-plant symmetric Chamfer distance between the input point cloud and the assembled full-plant mesh for the ten genotypes of \tabref{tab:plants}. The dashed line marks the median. \textbf{(b)} Blade-level residual between each final fitted surface and its observed blade points on Kui2021, computed as the exact point-to-surface distance of \secref{sec:quantitative}, with leaves numbered bottom-to-top.}
    \label{fig:cd_meter}
\end{figure}

\paragraph{Blade-level distance to the fitted surface}
We computed each per-leaf value as the root mean square of the exact distance from the corrected observed blade points to the selected post-repair surface, as exported and triangulated. We report this quantity in millimeters throughout. Over the $118$ leaves, the median was $5.20$\,mm, with an interquartile range of $4.21$ to $6.26$\,mm and extremes of $1.46$ and $11.82$\,mm. Per-plant medians ranged from $3.8$\,mm on A679 to $6.1$\,mm on CML331 and are listed in \tabref{tab:AllPlantsCD}. \tabref{tab:cd_per_leaf} gives the final per-leaf values for 78551S, and \figref{fig:cd_meter}(b) gives those for Kui2021.

\begin{table}[t!]
    \centering
    \caption{Per-leaf distance from the observed blade points to the fitted
    surface for genotype 78551S on the final selected geometry. Leaves are numbered bottom-to-top. Values are the exact point-to-surface distance of \secref{sec:quantitative}, in millimeters. \tabref{tab:AllPlantsCD} gives a per-plant summary.}
    \label{tab:cd_per_leaf}
    \setlength{\extrarowheight}{2pt}
    \footnotesize
    \begin{tabular}{@{}lrr@{}}
        \toprule
        Leaf & Final RMS (mm) & Blade points \\
        \midrule
        L1  & $4.78$ & 615 \\
        L2  & $5.89$ & 931 \\
        L3  & $7.60$ & 1030 \\
        L4  & $8.24$ & 1161 \\
        L5  & $5.95$ & 1113 \\
        L6  & $5.39$ & 1093 \\
        L7  & $5.09$ & 1045 \\
        L8  & $3.75$ & 687 \\
        \bottomrule
    \end{tabular}
\end{table}

\paragraph{Whole-plant Chamfer distance}
Following~\citep{hadadi2025procedural}, we also evaluated whole-plant fidelity using the symmetric mean Chamfer distance~\citep{fan2017point}. This distance is the average of the mean nearest-neighbor distances from each point set to the other. In contrast to the per-leaf quantity above, the Chamfer distance is a mean \emph{distance} and not a mean squared distance, and is therefore reported as a length. For each plant, we sampled $200{,}000$ points from the assembled full-plant mesh, uniformly by triangle area. This sampling density provided roughly twenty sampled mesh points per scan point. We then computed the bidirectional Chamfer distance against the input point cloud, including the stalk.

\figref{fig:cd_meter}(a) shows the per-plant symmetric Chamfer distance of the final refined surfaces. The median was $7.2$\,mm, with a range of $5.8$ to $9.7$\,mm.

The whole-plant Chamfer distance and blade-level residual use different point sets and aggregations. The whole-plant value compares the assembled mesh with the full scan, including the stalk, whereas the blade-level residual uses only corrected blade points. Their medians were $7.2$ and $5.2$\,mm, respectively. The mesh-to-cloud direction averaged $6.1$\,mm, and the cloud-to-mesh direction averaged $8.9$\,mm. This asymmetry was largest on A679, at $12.7$\,mm cloud-to-mesh and $5.5$\,mm mesh-to-cloud, and on Kui2021, at $12.8$ and $6.6$\,mm. The larger cloud-to-mesh values were consistent with scanned structures such as tassels and sheaths that have no corresponding generated surface, but the directional Chamfer distance does not identify which structures produce each residual. Comparisons with methods that reconstruct additional shoot organs should account for this difference in modeled anatomy.

\begin{figure}[t!]
    \centering
    \begin{subfigure}[t]{0.48\textwidth}
        \centering
        \includegraphics[width=\textwidth]{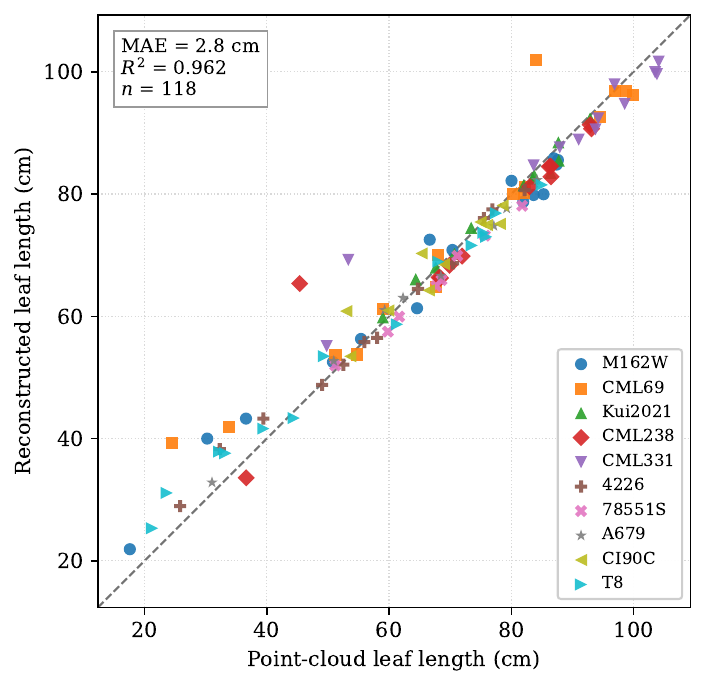}
        \caption{Blade length}
        \label{fig:trait_length}
    \end{subfigure}
    \hfill
    \begin{subfigure}[t]{0.48\textwidth}
        \centering
        \includegraphics[width=\textwidth]{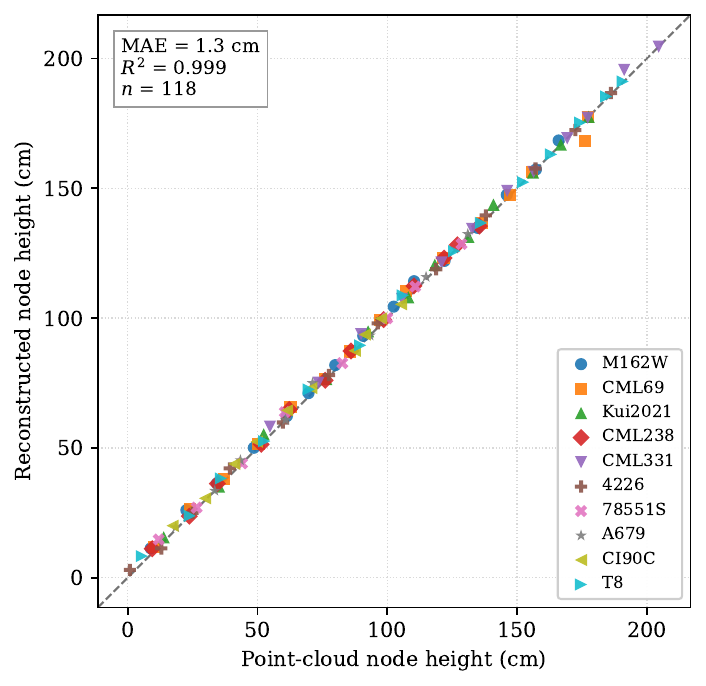}
        \caption{Node height}
        \label{fig:trait_node}
    \end{subfigure}

    \vspace{1.2em}

    \begin{subfigure}[t]{0.48\textwidth}
        \centering
        \includegraphics[width=\textwidth]{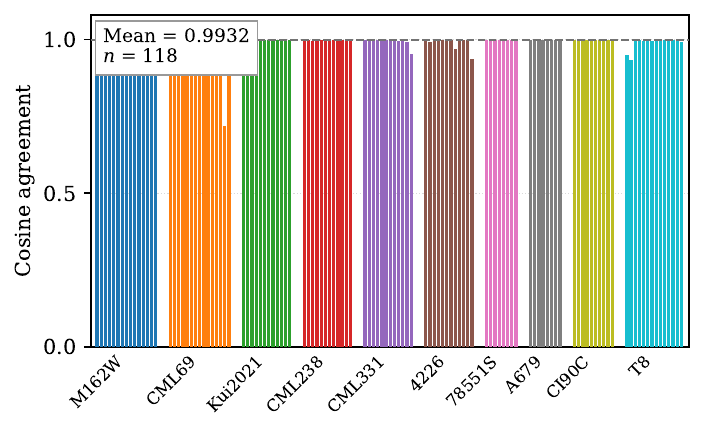}
        \caption{Vertical insertion angle}
        \label{fig:trait_vertical}
    \end{subfigure}
    \hfill
    \begin{subfigure}[t]{0.48\textwidth}
        \centering
        \includegraphics[width=\textwidth]{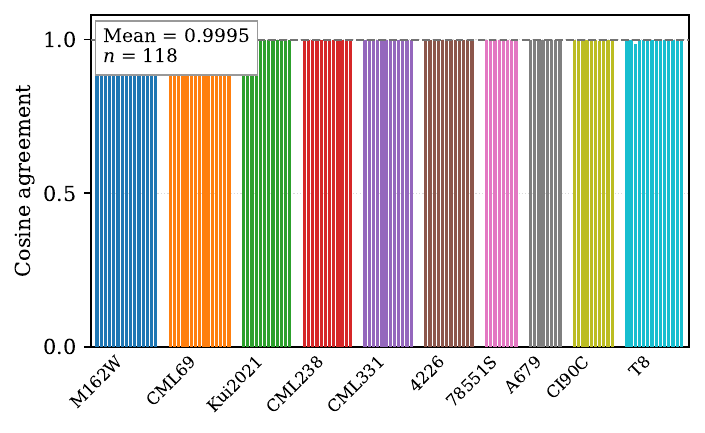}
        \caption{Azimuth}
        \label{fig:trait_azimuth}
    \end{subfigure}
    \caption{Agreement between traits measured on the exported surface and on
    the segmented blade cloud, over the $118$ leaves of the ten genotypes of \tabref{tab:plants}. \subref{fig:trait_length} Blade length and \subref{fig:trait_node} node height against the $1{:}1$ line, annotated with mean absolute error, coefficient of determination, and leaf count. \subref{fig:trait_vertical} Vertical insertion angle and \subref{fig:trait_azimuth} azimuth, as the per-leaf cosine of the angular disagreement, which is bounded above by one and remains well defined across the $0/360^\circ$ wrap. Bars are grouped by genotype, and the dashed line marks exact agreement.}
    \label{fig:trait_agreement}
\end{figure}

\paragraph{Agreement between traits measured on the surface and blade cloud}
The residuals above do not show whether traits derived from the exported surface agree with the same traits derived from the segmented blade cloud. We therefore applied the same trait definitions to both geometries. The cloud midrib was reconstructed from the blade points, whereas the surface midrib was taken from the middle row of the surface; both measurements used the same reconstructed stalk centerline. Blade length is the midrib arc length, and node height is the elevation on the stalk centerline nearest to the corresponding geometry. Vertical insertion angle was measured over the first $30\%$ of the midrib relative to the local stalk direction, and azimuth is the compass bearing from node to tip. Blade length agreed to a mean absolute difference of $2.8$\,cm against a mean of $68.8$\,cm ($4.1\%$, $R^2 = 0.962$), and node height agreed to $1.3$\,cm against a mean of $93.4$\,cm ($1.4\%$, $R^2 = 0.999$) (\figref{fig:trait_agreement}). The vertical insertion angle and azimuth agreed to mean cosines of $0.9932$ and $0.9995$, respectively, and the azimuth of all leaves agreed to within $10^\circ$. The mean surface-minus-cloud biases were $+0.43$\,cm for blade length and $+1.19$\,cm for node height.

The vertical insertion angle, however, had a noticeable tail. It disagreed by a median of $1.8^\circ$ but a 95th percentile of $14.8^\circ$, and its worst leaf disagreed by more than $40^\circ$. The largest disagreements occurred on short leaves near the top or bottom of the canopy. On these leaves, the first $30\%$ of the midrib spans a short segment, so the estimated direction is sensitive to the location of the reconstructed midrib start. Independent physical trait measurements were not available for these plants. The surface-to-cloud comparison therefore assesses internal consistency and not external accuracy.

\subsection{Results across the 100 plants}
\label{sec:hundred}

\begin{figure}[t!]
    \centering
    \includegraphics[width=\textwidth]{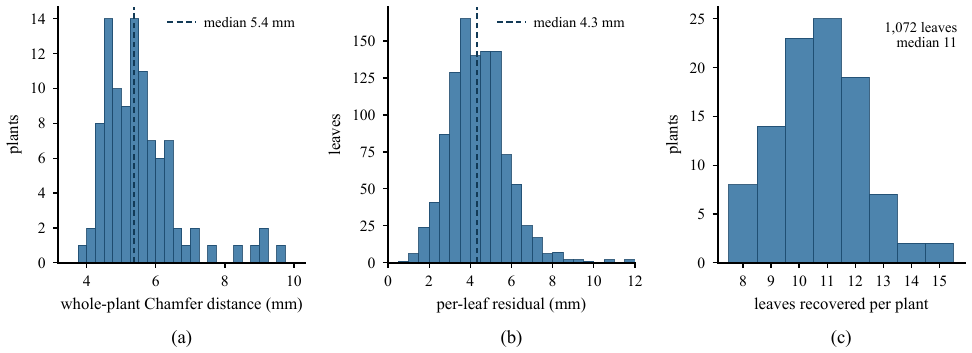}
    \caption{The pipeline over all $100$ evaluated plants, on the final
    refined geometry. \textbf{(a)} Whole-plant symmetric Chamfer distance against the input cloud, one entry per plant; the dashed line marks the median. \textbf{(b)} Blade-level residual over all $1{,}072$ reconstructed leaves, measured as the exact point-to-surface distance of \secref{sec:quantitative}. Three leaves lie beyond the axis. \textbf{(c)} Leaves recovered per plant.}
    \label{fig:hundred}
\end{figure}

The same fixed parameter set used for the ten plants above was applied to all $100$ plants of \secref{sec:dataset}, and \figref{fig:hundred} summarizes cohort-level leaf counts and fidelity metrics. The pipeline returned a refined model for all plants and recovered $1{,}072$ leaves in total, between $8$ and $15$ per plant with a median of $11$ (\figref{fig:hundred}(c)). The whole-plant symmetric Chamfer distance had a median of $5.4$\,mm and an interquartile range of $4.8$ to $6.0$\,mm, with the full set spanning $3.9$ to $9.7$\,mm (\figref{fig:hundred}(a)). Over all $1{,}072$ leaves, the blade-level residual had a median of $4.3$\,mm and an interquartile range of $3.5$ to $5.2$\,mm (\figref{fig:hundred}(b)). Resampling with $20{,}000$ bootstrap replicates gave a $95\%$ confidence interval (CI) of $5.13$ to $5.52$\,mm for the whole-plant median. Since leaves of one plant share a scan, a segmentation, and a stalk centerline, all leaf-level intervals that we report resample whole plants instead of individual leaves. On this basis, the interval for the blade-level median was $4.13$ to $4.47$\,mm. The segmentation left a median of $3.6\%$ of each plant's non-stalk points unlabeled. Of the $1{,}072$ attachments, $1{,}051$ ($98.0\%$) satisfied both conditions of \secref{sec:leafseg}, against the $98.3\%$ of \tabref{tab:nodes} measured the same way. Of the $21$ attachments that failed at least one condition, $17$ failed only the spacing floor, $3$ failed only the base-to-node gap, and one failed both. Across the $100$ plants, insufficient spacing between neighboring nodes was the more common attachment failure. By contrast, the two flagged attachments among the ten detailed plants exceeded the base-to-node gap limit.

The ten plants reported in detail had whole-plant distances of $5.8$ to $9.7$\,mm and lay toward the upper end of the full-cohort distribution. They were selected to span the panel's morphological range instead of as a random subset. Nine of the $100$ plants raised the plant-level attachment-condition flag, and the remaining $91$ did not. All $21$ attachments that failed at least one condition occurred in these nine plants. This flag is raised when at least one attachment still fails the gap or spacing condition and no admissible replacement is available. It summarizes the same criterion used in \tabref{tab:nodes}; it is not an additional test.

\subsection{Behavior of the geometric checks}
\label{sec:guards_results}

The checks of \secref{sec:nurbsdiff} were inactive on most leaves and engaged only where the initialization or the unchecked fit triggered them. Here we return to the ten plants of \secref{sec:dataset}, since the full per-leaf outcome of each check is retained only for them. The rigid-prealignment test evaluated a candidate rotation on all $118$ leaves and accepted it on $9$, spread over four plants: four on CML69, three on M162W, one on CML238, and one on 4226. These nine leaves were refit from the rotated control net with the attachment row anchored and the orientation check active throughout. The check backtracked on between $4$ and $500$ of the $500$ optimizer steps and rejected between $0$ and $490$ outright. All nine leaves finished with zero inverted surface cells. The rejection count records how often a proposed update violated the orientation constraints; it does not measure convergence or final fidelity.

The other $109$ leaves were fit from their unmodified procedural initialization, and the mid-blade fold diagnostic applied to these fits found no fold on $81$ and a fold on $28$. Refitting from the locally rolled initialization removed the fold on $14$ of the $28$ folded leaves. On a further $7$, refitting improved all fold measures together with the Chamfer term without eliminating the fold signal. All $21$ candidates were adopted. On the remaining $7$, the candidate failed its acceptance test and the original fit was retained, fold included. In total, $14$ of the $109$ screened leaves ($12.8\%$) still carried a residual mid-blade fold signal. The fold status of the $9$ prealigned leaves was not measured, since they were screened only by the narrower orientation check. This check tests for inverted cells but not for the transverse control-point swap or the backward stepping that the fold diagnostic also covers. These surfaces remain usable for area and length measurement. However, they are not guaranteed to be locally non-self-intersecting away from the node, and a simulation that requires a manifold surface should filter on the fold status saved for each leaf.

The base-collar repair was not triggered on any leaf. The path check of \secref{sec:params_ppg} flagged no mis-started midline, so no leaf met the precondition of the repair, and its status was reported as not applicable on all $118$. We keep the repair because the failure that it addresses follows from a mis-started midline, and the path check is the only signal that separates that case from a legitimately curved collar.

All leaves ran the full budget of $500$ iterations and completed on the first attempt, with no numerical retries on any of the ten plants. As a result, the iteration count was fixed and is not a measured convergence outcome. In addition, each check compared its candidate against the unchecked baseline of the same leaf and kept the baseline unless the candidate satisfied its own stated criterion. Thus, on the $109$ leaves fitted from an unmodified initialization, the fidelity reported in \secref{sec:quantitative} can fall below that of the unchecked pipeline by at most the $2\%$ margin allowed by the fold repair of \secref{sec:nurbsdiff}. This bound does not cover the $9$ prealigned leaves. Their rotation was accepted on the candidate-comparison metric of \secref{sec:nurbsdiff}, whereas the fidelity of \secref{sec:quantitative} is measured separately on the exported surface. Three of the $21$ adopted fold repairs used part of that margin to remove the fold, finishing $0.05$, $0.20$, and $0.27\%$ worse in Chamfer distance than the baselines that they replaced. The other $18$ improved on their baseline. The checks were designed to ensure geometric validity, not to improve accuracy. \tabref{tab:AllPlantsCD} lists the plants on which they engaged.

\paragraph{Sensitivity to the acceptance thresholds}
The checks above are governed by numeric thresholds. We swept the three that the blade-refinement stage exposes on its command line, namely the prealignment confidence margin, the local-roll Chamfer ratio, and the orientation area fraction of \tabref{tab:parameters}. We perturbed one threshold at a time in each direction, held all other parameters at their published values, and refit all $118$ leaves of the ten plants under each of the six resulting configurations. A seventh run with all thresholds at their published values served as a control.

The leaf inventory was insensitive to all three thresholds. All configurations exported the same $118$ leaves with the same per-plant distribution and no fit failures. The thresholds affected only the surface fit of each leaf, and their effects differed. The local-roll Chamfer ratio had no observed effect over the range tested: moving it from $1.02$ to $1.01$ and to $1.05$ changed no leaf. The orientation area fraction affected no more than two leaves in either perturbation. Halving it to $0.015$ improved a single leaf from $7.56$ to $6.91$\,mm, and doubling it to $0.06$ moved two leaves by no more than $0.03$\,mm in either direction. The prealignment confidence margin produced the largest changes, and its effect was asymmetric. Tightening it from $0.08$ to $0.16$ changed one leaf, by $0.04$\,mm in its favor. Loosening it to $0.04$ admitted two rotations that the published setting rejected. These two rotations degraded one leaf from $6.84$ to $14.87$\,mm and another from $3.53$ to $6.68$\,mm and raised the cohort mean from $5.30$ to $5.39$\,mm. The other $116$ leaves were unchanged. These results support retaining the published value of $0.08$. The sweep covered only the three exposed thresholds and none of the other constants in \tabref{tab:parameters}, and it measured the effect on fitted geometry over the ten plants.

\subsection{Comparison with the PSO+NURBS-Diff pipeline}
\label{sec:comparison}

Over the $100$ plants, the proposed pipeline reached a median whole-plant Chamfer distance of $5.4$\,mm against $9.2$\,mm for the historical baseline and was closer to the scan on $99$ of the $100$ plants. We compare against the PSO+NURBS-Diff pipeline of \citet{hadadi2025procedural}, which used Particle Swarm Optimization to initialize each leaf surface before NURBS-Diff refinement and relied on manual or semi-automated organ segmentation. That study searched each leaf with a global-best swarm of $300$ particles, using cognitive, social, and inertia coefficients of $2.5$, $0.5$, and $0.9$ with time-varying schedules. The search ran in two passes, a first pass over $33$ parameters that fixed the blade's azimuth and a second pass of $50$ iterations over the remaining $32$. The $100$ PSO surfaces compared here are the models released with that study and were built on the curated organ segmentation distributed with MaizeField3D~\citep{maizefield3d}. The proposed pipeline did not use these curated labels for task-specific training or as pipeline input. The historical baseline therefore benefits from curated organ labels, whereas the proposed pipeline recovers its organ segmentation from reviewed VLM annotations and geometric processing. The fidelity comparison is consequently an end-to-end comparison of the surfaces produced by the two workflows, not a controlled isolation of initialization alone. \tabref{tab:comparison_pso} summarizes the comparison axes that the two pipelines share. The principal methodological difference at the fitting stage is how the starting surface is obtained. Replacing the swarm with a surface generated from the measured plant descriptor of \secref{sec:params_ppg} removed the meta-heuristic search that dominated the previous runtime. Descriptor-based initialization also retained explicit organ traits alongside the fitted surface, whereas the earlier workflow optimized its control-point parameterization directly.

The timing comparison differs in scope between the two pipelines. The historical figures are those published in~\citep{hadadi2025procedural} on that paper's hardware, exclude its segmentation step, and were not re-measured here. Ours are from the run of \secref{sec:runtime} and exclude preprocessing, rendering, and annotation. With these scopes in mind, procedural initialization took $82$\,s against $2740 \pm 180$\,s for PSO, a speedup factor of $33$, and the refinement itself was roughly $5$ times faster. Both refinements used $500$ Adam iterations and the same learning-rate schedule. However, the hardware, control-net resolution, implementation, and timing scopes differ, so these ratios describe the reported workloads alone and do not measure controlled speedups.

\begin{figure}[t!]
    \centering
    \includegraphics[width=\textwidth]{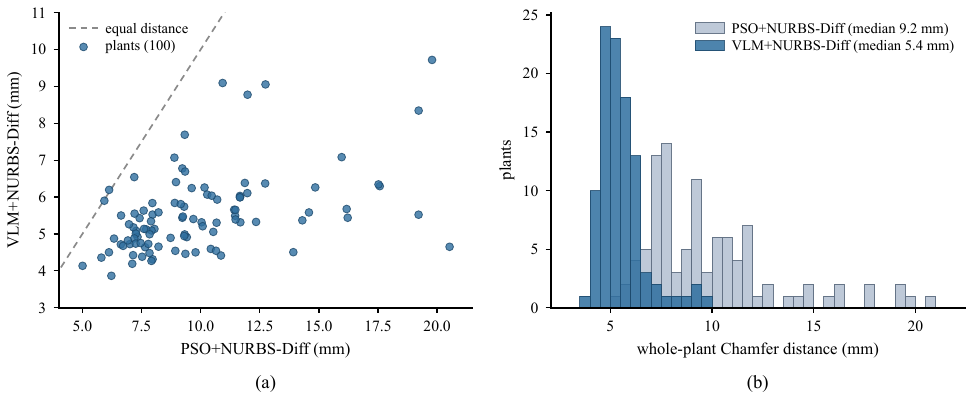}
    \caption{Whole-plant fidelity of the proposed pipeline against the
    PSO+NURBS-Diff pipeline of~\citep{hadadi2025procedural} on the same $100$ plants, both recomputed under one protocol. \textbf{(a)} One point per plant. Points below the dashed line ($99$ of the $100$) are plants that this work reconstructs closer to their scan. \textbf{(b)} The two distributions with their medians.}
    \label{fig:pso_comparison}
\end{figure}

We compared fidelity on the whole-plant distance, which applies to both pipelines regardless of how either defined its per-leaf objective, and recomputed this distance for both under one protocol. Each assembled surface was evaluated against a $10{,}000$-point representation of the corresponding scan, each surface was sampled at $200{,}000$ points, and the symmetric mean Chamfer distance was computed between them. Over the $100$ plants (\figref{fig:pso_comparison}), the proposed pipeline reached a median of $5.4$\,mm against $9.2$\,mm for PSO+NURBS-Diff, a median per-plant reduction of $40.9\%$. The proposed pipeline was closer to the scan on $99$ of the $100$ plants (paired Wilcoxon signed-rank test, normal approximation, $p < 10^{-17}$), and the single exception, at $6.20$\,mm against $6.13$\,mm, differed by under $0.1$\,mm. The Hodges-Lehmann estimate of the paired shift was $3.93$\,mm ($95\%$ distribution-free CI $3.42$ to $4.44$\,mm), and the matched-pairs rank-biserial correlation was $0.999$. Resampling the $100$ plants ($20{,}000$ bootstrap replicates) placed the median of the proposed pipeline at $5.4$\,mm ($95\%$ CI $5.13$ to $5.52$\,mm) against $9.2$\,mm ($8.13$ to $9.70$\,mm) for PSO+NURBS-Diff, and the median per-plant reduction at $40.9\%$ ($36.3$ to $44.9\%$). Decomposing the symmetric distance showed a reduction in both directions. The cloud-to-surface direction fell from a median of $11.2$ to $5.7$\,mm, and the surface-to-cloud direction fell from $7.0$ to $4.9$\,mm. We used the whole-plant metric because it can be recomputed consistently for both model sets; a corresponding per-leaf exported-surface metric was not available under one common protocol.

\begin{table}[t!]
    \centering
    \caption{Comparison of the proposed pipeline against the PSO+NURBS-Diff
    pipeline of~\citep{hadadi2025procedural}. Time values are the mean $\pm$ standard deviation of per-plant times over the ten plants of \secref{sec:dataset} for the proposed pipeline, and the values published in~\citep{hadadi2025procedural} for the previous one. No ratio is given on the summed compute row, since our $601$\,s includes $382$\,s of segmentation with no counterpart in the $3430$\,s. The two indented rows below it compare stages of matching scope. Historical uncertainties are reproduced as published. The two fidelity rows are recomputed here for both pipelines under one protocol over the same $100$ plants, as described in the text.}
    \label{tab:comparison_pso}
    \setlength{\extrarowheight}{3pt}
    \footnotesize
    \begin{tabular}{@{}lccc@{}}
        \toprule
        Metric & PSO+NURBS-Diff~\citep{hadadi2025procedural} & VLM+NURBS-Diff (this work) & Ratio \\
        \midrule
        Surface initialization                       & PSO over $32$ to $33$ variables & Measured plant descriptor                    & n/a \\
        Organ segmentation                           & Manual / semi-automated     & Reviewed VLM annotations plus geometric processing & n/a \\
        Whole-plant Chamfer distance, median (mm)    & $9.2$                       & $\mathbf{5.4}$                                  & $1.7\times$ \\
        \phantom{~}~plants with the lower distance   & $1$ of $100$                & $\mathbf{99}$ \textbf{of} $\mathbf{100}$        & n/a \\
        Per-plant geometric-core compute (s)         & $3430 \pm 490$              & $601 \pm 55$                                    & n/a \\
        \phantom{~}~of which initialization (s)      & $2740 \pm 180$ (PSO)        & $82 \pm 24$ (measured descriptor)               & $33\times$ \\
        \phantom{~}~of which differentiable refinement (s) & $690 \pm 310$        & $137 \pm 38$                                    & $5.0\times$ \\
        Interpretability of the output               & Optimized ctrl.\ pts.\ only & Azimuth, pitch, node height, width profile, ctrl.\ pts.\ & n/a \\
        Attachment guarantee                         & Not reported                & \textbf{Node row locked; conditions reported}   & n/a \\
        \bottomrule
    \end{tabular}
\end{table}

In addition to the refined NURBS surfaces, the proposed pipeline outputs a descriptor containing per-leaf pitch, azimuth, ordered control-point width profile, node position, stalk diameter, and midline control points. These quantities can be edited or perturbed downstream without re-running segmentation or fitting.

\subsection{Computational cost}
\label{sec:runtime}

\tabref{tab:runtime} gives the per-plant runtime of each local stage on the ten reported plants, measured on one workstation, with the annotation row pooled over a larger batch as stated in the caption. The workstation had an Intel Xeon E3-1240~v5 at $3.50$\,GHz with four cores and $16$\,GB of memory, and an NVIDIA Quadro K620 with $2$\,GB of memory. Only the differentiable refinement used the GPU, through CUDA, and all other stages ran on the CPU. The times for the four stages from mask extraction to parameter measurement are single-process wall-clock times. The NURBS-Diff row sums the optimizer time logged for each leaf and excludes setup, diagnostics, mesh export, and review images. The five local rows sum to $601 \pm 55$\,s, or about ten minutes of compute per plant. Mask extraction was the largest single term at $187 \pm 21$\,s, since skeletonization, junction splitting, and tangent-based re-merging ran once per view. Combining the pooled annotation mean with the local-stage mean gave an approximate total of $815$\,s per plant.

Each leaf is fitted against its own points with its own locked attachment row, independently of its neighbors. A parallel implementation could therefore distribute the leaf fits, although the reported runner processed them sequentially. The logged optimizer time was $11.6 \pm 4.2$\,s per leaf. The annotation stage requires nine independent tasks per plant, one leaf task per rendered view plus one stalk task on the front view. Validation failures can trigger retries, so nine tasks do not always correspond to nine model calls. The pooled $214$\,s value is the elapsed time of the sequential annotation stage and is the only reported component determined by a remote service instead of local hardware. Submitting the independent tasks concurrently could reduce elapsed time subject to the service's concurrency limits, but this option was not tested.

\begin{table}[t!]
    \centering
    \caption{Per-plant runtime over the ten plants of \secref{sec:dataset},
    reported as mean $\pm$ standard deviation. The annotation row is elapsed wall-clock time for eight leaf-view tasks and one stalk task per plant, pooled over $19$ plants annotated in batches. The annotation row is given without a deviation because the batches give totals and not per-plant times, and the batch means span $209$ to $233$\,s. Preprocessing, rendering, assembly, and reporting were not timed.}
    \label{tab:runtime}
    \setlength{\extrarowheight}{3pt}
    \footnotesize
        \begin{tabular}{@{}lr@{}}
        \toprule
        Stage & Time per plant (s) \\
        \midrule
        VLM annotation                                 & $214$ \\
        \midrule
        Mask extraction and leaf separation (8 views)  & $187 \pm 21$ \\
        Cross-view consensus                           & $47 \pm 3$ \\
        Blade segmentation and node policy             & $148 \pm 18$ \\
        Parameter measurement and procedural assembly  & $82 \pm 24$ \\
        NURBS-Diff refinement (optimization only)      & $137 \pm 38$ \\
        \midrule
        Geometric core                                 & $601 \pm 55$ \\
        Approximate annotation and geometric sum       & $815$ \\
        \bottomrule
    \end{tabular}
\end{table}

\subsection{Validation of the organ segmentation}
\label{sec:seg_validation}

The results above characterize the segmentation by internal measures, namely how many views support each leaf, how many points remain unassigned, and how many attachments satisfy the two conditions on node placement. We assess these measures through four further questions: (a) the dependence of the leaf set on the chosen viewpoints, (b) the reproducibility of an annotation drawn from a hosted generative model, (c) agreement with an independent organ count, and (d) the point-for-point correspondence of leaves with independently labeled organs. Question (d) provides external validation of the leaf-recovery results in \secref{sec:seg_results}.

\paragraph{Dependence on the individual viewpoints}
The pipeline saves the supporting views of each consensus leaf, and consensus retains a leaf only when at least two views support it (\secref{sec:backproject}). The effect of removing a viewpoint can therefore be evaluated from these support lists without re-running the pipeline. Across the $100$ plants, consensus established $1{,}073$ leaves and $1{,}072$ reached a refined surface. Two consensus leaves, one on each of two plants, back-projected to no points, and both rested on the two-view floor. A leaf with no points cannot be refined or exported, so both were dropped before blade refinement. These two leaves were the only consensus leaves that did not reach export, and dropping them removed no measurement, since they had no geometry. The fragment correction of \secref{sec:leafseg} then added back one leaf. Over the whole cohort, the fragment correction applied a single \textsc{move} and a single \textsc{split}, both on the same plant, and these corrections were the only change to the leaf inventory on any of the $100$ plants. Of the $1{,}073$ leaves, only $15$ ($1.4\%$) rested on the two-view floor itself and would have been lost if either supporting view had been removed. Deleting any single view removed between $1$ and $7$ leaves, a mean of $3.8$ over the eight views, or $0.3\%$ of the recovered organs. The most critical viewpoint was $+45^\circ$ at $7$ leaves, and the least critical was $+135^\circ$ at $1$. The front view resolved the most leaves on its own (\tabref{tab:consensus_support}), but removing it did not produce one of the largest losses, because most of the leaves that it saw were also seen in other views. At the plant level, $87$ of the $100$ plants lost no leaf when any single view was removed, $12$ lost exactly one leaf, and one lost two. Thus, no single view in the eight-view set was critical to the recovered leaf inventory under this support-list analysis. This analysis does not measure the effect of removing one view on blade extent or fitted-surface fidelity.

The support-list analysis avoids a re-run but bounds only the removal of one view at a time. To test the stronger case, we re-ran the downstream stages on the ten plants from the same annotations with two views withheld. We kept the front, back, $\pm 45^\circ$, and $\pm 135^\circ$ renders and dropped the $\pm 30^\circ$ pair. This choice left the six-view subset with the widest angular spacing. All thresholds were held fixed, and all stages from mask extraction through NURBS-Diff refinement were re-executed as in the eight-view run. Six views recovered $115$ of the $118$ leaves. Seven of the ten plants returned the same count, and the other three lost one leaf each: Kui2021 fell from $12$ to $11$, CI90C fell from $10$ to $9$, and T8 fell from $14$ to $13$. Removing a quarter of the viewpoints therefore reduced the recovered organs by $2.5\%$. The re-run recomputed consensus, ownership, repair, and refinement under the reduced view set, instead of estimating the loss from the original support lists.

\paragraph{Reproducibility of the annotation}
All stages after annotation are deterministic given their input. However, the annotation itself is a sample from a hosted model, and our requests specified neither a temperature nor a seed. To measure the resulting spread, we selected four plants that included an easily read plant and a crowded one, re-annotated the front view of each five times, and compared the returned annotation masks. Across the $40$ within-plant run pairs, the mean pairwise intersection over union of the colored annotation masks, including leaf curves and node markers, was $0.729$, with a minimum of $0.626$. \figref{fig:vlm_repeats} in \ref{sec:appendix:repeats} shows the twenty accepted annotations. Part of this spread reflected placement, not detection, since the curves are only a few pixels wide and a small lateral offset reduces their overlap.

For this reason, the pipeline does not treat a single view as a result. Consensus retains a leaf only when at least two views support it (\secref{sec:backproject}). A stroke that appears on one call and not the next changes the exported set only when it supplies the second support for its leaf. This condition applies to the $1.4\%$ of leaves identified above as resting on the two-view floor. The repeat-annotation experiment isolated variability in the stochastic annotation stage alone, leaving the complete eight-view workflow unmeasured. Because only $1.4\%$ of recovered leaves rested on the two-view floor, the support analysis above suggests that cross-view consensus prevents most single-view variability from changing the exported leaf set. Repeating all eight annotations and the complete downstream workflow would be required to measure the remaining end-to-end variability.

To test how much of the result depended on the particular model as opposed to the annotation stage itself, we repeated the same $20$-run design with a second model. This comparison was limited to the returned annotation images. Under the tested API configuration, the second model produced partial curves, misplaced magenta components, and substantial changes in thresholded image foreground. \ref{sec:appendix:second_model} reports this annotation-stage probe; the outputs were not run through the downstream reconstruction stages.

\paragraph{Agreement with a manual count}
To obtain an external reference, the leaves of the ten plants reported leaf by leaf were counted by hand from their rendered views, without consulting the pipeline output. The manual count totaled $125$ leaves against the $118$ exported by the pipeline, so the exported total was $94.4\%$ of the manual total (\tabref{tab:manual_count}). Four plants agreed exactly, nine agreed to within one leaf, and the largest disagreement was two leaves, on CI90C. On no plant did the pipeline return more leaves than the manual count.

The seven-count difference was concentrated near the top of the plants. Manual inspection identified three recurring conditions: (a) an uppermost leaf small enough to be read as part of the tassel, (b) a blade whose scan is too sparse to trace as one continuous stroke, and (c) a blade that twists around the tassel and breaks its stroke in most views. The same conditions occurred in three of the four two-view support cases of \secref{sec:seg_results}. The internal support measures and the external count thus identified the upper canopy as the weakest region.

\begin{table}[t!]
    \centering
    \caption{Recovered leaves against a manual count of the same scans.
    ``Manual'' is the number of leaves counted by hand from the rendered views. ``Consensus'' is the number established by cross-view consensus and ``Exported'' the number that reached a refined surface. The two differ on CML331, where the fragment correction of \secref{sec:leafseg} recovered a twelfth leaf that consensus had not established. ``Difference'' is the manual count less the exported count.}
    \label{tab:manual_count}
    \setlength{\extrarowheight}{2pt}
    \footnotesize
    \begin{tabular}{@{}lrrrr@{}}
        \toprule
        Genotype & Manual & Consensus & Exported & Difference \\
        \midrule
        M162W   & 15 & 15 & 15 & 0 \\
        CML69   & 16 & 15 & 15 & 1 \\
        Kui2021 & 13 & 12 & 12 & 1 \\
        CML238  & 13 & 12 & 12 & 1 \\
        CML331  & 13 & 11 & 12 & 1 \\
        4226    & 12 & 12 & 12 & 0 \\
        78551S  & 8  & 8  & 8  & 0 \\
        A679    & 9  & 8  & 8  & 1 \\
        CI90C   & 12 & 10 & 10 & 2 \\
        T8      & 14 & 14 & 14 & 0 \\
        \midrule
        All ten & 125 & 117 & 118 & 7 \\
        \bottomrule
    \end{tabular}
\end{table}

\paragraph{Candidates discarded by the support floor}
Across the ten plants, consensus discarded $41$ candidate leaves supported by only one view, and the position of each candidate was saved. Twenty-one had candidate heights at or above the highest retained consensus anchor, and $20$ lay below it. Position alone does not establish whether a candidate represents a leaf or a tassel stroke, and plant-level counts do not establish correspondence. Accepting all single-view candidates would therefore admit $41$ candidates without cross-view corroboration. We retain the two-view requirement and save the discarded candidates for review.

One limitation of the manual count is that it was made from the same rendered views that the annotator saw. A leaf invisible in all of them is invisible to both, so the $94.4\%$ ratio measures count agreement for the rendered scans, not completeness against the plant. The comparison below addresses the other limitation, namely that a count of organs establishes no correspondence between them.

\paragraph{Leaf-by-leaf agreement with published organ labels}
MaizeField3D~\citep{maizefield3d} distributes a curated per-point organ segmentation of the same scans, with one fixed color per leaf, numbered from the base of the plant upward, and a black stalk. This segmentation was produced by the manual and semi-automated labeling used in the earlier study of \citet{hadadi2025procedural}, independently of all stages of this pipeline. We placed these labeled clouds in the working frame of the pipeline and confirmed that each coincides with its working cloud to within the $4.1$\,mm median point spacing of that cloud (\secref{sec:dataset}). Palette colors were decoded with a per-channel tolerance of $20$, and a color class was retained as a leaf only if it contained at least $20$ points. We then transferred labels and pipeline assignments onto the working cloud by nearest neighbor, accepting transfers within $2.0$\,cm, and matched each exported leaf one to one to a labeled leaf by maximum intersection over union (IoU).

The curated labels mark $1{,}023$ leaves across all $100$ plants. The pipeline recovered $1{,}017$ of them at an intersection over union of at least $0.5$, a detection rate of $99.4\%$ ($95\%$ CI $98.9$ to $99.8\%$), and it recovered all labeled leaves on $94$ of the $100$ plants. Matched leaves agreed closely in extent, with a mean intersection over union of $0.948$ pooled over all $1{,}017$ matched leaves.
Aggregated by plant (the level at which the intervals below resample), the median plant had a mean intersection over union of $0.954$ ($95\%$ CI $0.948$ to $0.959$), a per-point recall over labeled blade tissue of $97.3\%$ ($96.8$ to $97.5\%$), and a per-point precision of $96.7\%$ ($96.2$ to $97.7\%$). The released palette and the consensus stage of \secref{sec:backproject} both number leaves from bottom to top, so ordinal agreement can also be evaluated. Across all $1{,}017$ matched leaves, $898$ ($88.3\%$) agreed in ordinal position. On the $60$ plants where the two methods recovered equal leaf counts, $598$ of $627$ matched leaves ($95.4\%$) agreed in ordinal position as well as extent. These rates were measured on the reviewed annotation set of \secref{sec:vlm} and therefore describe the pipeline together with that review.

The two external references have different coverage. Fifty-five exported leaves had no labeled counterpart, and on $39$ plants the pipeline returned more leaves than the curated segmentation labels. At least some unmatched exports corresponded to leaves omitted from the curated labels. On T8, the curated segmentation labels $11$ leaves, whereas the manual count of \tabref{tab:manual_count} and the pipeline both gave $14$. Conversely, the manual total exceeded the pipeline total on six of the ten detailed plants.

\paragraph{Scope of the supervised-segmentation and PSO baselines}
The preceding comparison with the curated MaizeField3D labels is the benchmark for the organ-segmentation stage. Those labels were not used for task-specific training or as pipeline input, and the pipeline recovered $1{,}017$ of $1{,}023$ reference leaves at IoU $\ge 0.5$, with a mean matched-leaf IoU of $0.948$ and per-point recall and precision of $97.3\%$ and $96.7\%$, respectively. Segmentation quality was therefore measured directly against a pipeline-independent organ reference, not inferred from the fidelity of the final mesh. We do not claim an accuracy advantage over supervised networks trained on MaizeField3D; within their training distribution such networks can segment accurately. We claim that the evaluated workflow obtained organ-level output without using target-domain organ labels for task-specific training or as pipeline input. Generalization to other species, growth stages, or acquisition setups remains to be tested.

A supervised segmenter provides a label field and would serve as an alternative segmentation front end. An end-to-end reconstruction baseline must additionally reach the fitted-surface stage evaluated here. For that comparison we used the pipeline of \citet{hadadi2025procedural}, the prior method that starts from the same class of plant scan and produces fitted NURBS plant surfaces. A complementary study could retrain a published maize segmentation model on part of MaizeField3D and score both segmentation approaches on held-out plants under one protocol. That experiment would quantify the accuracy-versus-annotation tradeoff of the segmentation front end; it is distinct from the end-to-end scan-to-procedural-model comparison made here.

\subsection{Future directions}
\label{sec:future_work}

The trait comparison of \secref{sec:quantitative} measured agreement between the exported surfaces and the segmented blade clouds. Independent physical trait measurements were not available for this dataset. Future studies should collect co-registered physical measurements to evaluate the external accuracy of blade length, node height, insertion angles, and blade width, while separating acquisition, registration, and reconstruction errors.

Annotation is the only stage that depends on a hosted model, and the repeat runs of \secref{sec:seg_validation} showed its run-to-run variability. The second-model probe in \ref{sec:appendix:second_model} also showed that a replacement model must preserve the input plant geometry and pixel registration. Future work could fine-tune a plant-specific vision-language model on expert-reviewed plant annotations, with structured segmentation outputs registered to the rendered views. An agentic execution layer could coordinate the stages, inspect their outputs using the geometric tests and saved per-leaf diagnostics, rerun a stage when validation fails, and route uncertain cases to human review. Before deployment, its validation policies should be calibrated and evaluated against expert-labeled successful and failed cases, including their false-acceptance and false-rejection rates.

Evaluation should also extend to other plant architectures and acquisition conditions. Other grasses with a single culm and alternating leaves may require parameter changes, whereas branching or compound-leaved species would require changes to the one-node-per-leaf model of \secref{sec:leafseg}. Tests on plants scanned in place should evaluate the effects of background clutter, occlusion, and field acquisition conditions on annotation and segmentation. These studies would determine whether the pipeline can produce editable organ-level models across plant architectures and under field conditions.

\section{Conclusions}
\label{sec:conclusion}

In this paper, we presented an end-to-end pipeline that uses vision-language models to reconstruct an editable procedural model of a field-grown maize plant from its point cloud. Multi-view VLM annotations provided candidate leaf midlines and nodes, together with stalk landmarks from the front view. These annotations were back-projected onto the 3D point cloud, and the resulting view-specific leaves were reconciled by cross-view consensus. Their midlines were then expanded to full blades by an orientation-weighted surface graph, leaving outlier points unlabeled. Each leaf attachment to the stalk was evaluated using a constrained node rule against stated gap and spacing conditions. We measured the reconstructed stalk and full-blade clouds, wrote the results to the plant descriptor (the per-organ parameter file used to generate the procedural surface), and fitted each blade using a differentiable NURBS fitting procedure. The pipeline recovered $1{,}072$ leaves across $100$ field-grown plants and reached a median whole-plant Chamfer distance of $5.4$\,mm. The pipeline's reconstructions were closer to the scan, on $99$ of the $100$ plants, than those of the earlier PSO-initialized pipeline. The pipeline matched $1{,}017$ of the $1{,}023$ ($99.4\%$) reference leaves at IoU $\ge 0.5$ without using the curated labels for task-specific training or as pipeline input. The pipeline exported $118$ leaves from the ten plants reported leaf by leaf, with a median blade-level residual of $5.2$\,mm. This residual is under $1\%$ of the $70.2$\,cm median blade length, indicating that the fitted surfaces closely tracked the corrected scan geometry.

These results indicate that zero-shot VLM annotations become reliable organ-level measurements once they are reconciled, tested, and corrected by downstream geometric stages. In practice, the pipeline converted individually uncertain proposals into reliable organ-level measurements. Because some attachments failed at least one test, the pipeline recorded the outcome of every applicable geometric test for each returned organ, letting a user identify the specific attachments or surfaces that still need manual review. As further evidence of this reliability, leaf traits measured on the exported surfaces also agreed closely with those measured on the segmented blade clouds, but validation against co-registered physical measurements remains future work. Together, the results demonstrate a traceable end-to-end workflow for editable maize models while identifying the outputs that remain uncertain.

\section*{CRediT authorship contribution statement}
\textbf{Mozhgan Hadadi:} Data curation, Formal analysis, Investigation, Software, Validation, Visualization, Writing: original draft. \textbf{Talukder Z. Jubery:} Conceptualization, Methodology, Project administration, Supervision, Writing: review \& editing. \textbf{Adarsh Krishnamurthy:} Conceptualization, Methodology, Project administration, Supervision, Writing: review \& editing. \textbf{Baskar Ganapathysubramanian:} Conceptualization, Methodology, Project administration, Supervision, Writing: review \& editing.

\section*{Declaration of competing interest}
The authors declare that they have no known competing financial interests or personal relationships that could have appeared to influence the work reported in this paper.

\section*{Declaration of generative AI and AI-assisted technologies in the writing process}
During the preparation of this work the authors used Claude (Anthropic) in order to improve the language, grammar, and readability of the manuscript. After using this tool the authors reviewed and edited the content as needed and take full responsibility for the content of the publication.

Two further uses of these technologies fall outside the scope of the above declaration and are stated here for completeness. The vision-language model of \secref{sec:vlm} is a component of the reported methodology, not a writing aid. Its provider, version, and prompts are reported in full in that section. AI coding assistants (Claude, Anthropic; GPT, OpenAI) were used during implementation of the pipeline software, and all algorithms, parameter settings, and reported results were specified, verified, and validated by the authors.

\section*{Data availability}
The code used to generate the procedural models will be available upon acceptance. The raw point clouds used in this study are available through the publicly released MaizeField3D dataset~\citep{maizefield3d}.

\section*{Acknowledgements}
This work was supported by the AI Institute for Resilient Agriculture (USDA-NIFA 2021-67021-35329), NSF 2412929, and Iowa State University's Plant Science Institute.

\bibliographystyle{elsarticle-num-names}
\bibliography{references}

\clearpage
\appendix

\setcounter{figure}{0} \setcounter{table}{0}

\section{Additional methodological details}
\label{sec:appendix:methods}

This appendix gives the exact VLM prompts, the standard NURBS basis and differentiation formulas, the conditional base-collar repair, and the parameter table for the reported pipeline.

\subsection{Exact VLM prompts}

\begin{figure}[p]
    \centering
    \begin{tcolorbox}[colback=gray!5, colframe=gray!60,fontupper=\scriptsize\ttfamily, left=4pt, right=4pt, top=2pt, bottom=2pt, boxrule=0.5pt, title={\scriptsize\sffamily\bfseries Prompt 1: leaf midlines and node dots (applied to every rendered view)}]
    On the point cloud data in this plant image, do the following:

    1. Keep the background EXACTLY as it is -- do not change any white area or the corner annotations (text labels, axis arrows, scale bar).

    2. NODE IDENTIFICATION: At every node where a leaf connects to the central stalk, place a bright magenta solid dot (\textasciitilde{}10\,px diameter) on the stalk at the leaf-stalk junction (node) so the nodes are clearly visible against the black stalk.

    3. SEGMENTATION: Identify ALL main leaves and trace the midline (central vein) of each main leaf from the stalk junction (node) to the leaf tip. Draw each midline as a solid colored line (8\,px thick). Use a UNIQUE color for each leaf so they are distinguishable (e.g., red, pea green, green, orange, cyan, coral, blue, brown, dark purple, yellow, golden, azure, aquamarine, lime, khaki, etc. but not magenta). Make SURE to include all of the leaves. ALL leaves, even small upper ones near the tassel. Don't miss any upper or lower leaves. Don't use any color more than once.

    4. The stalk, tassel, AND every leaf body must remain SOLID BLACK. Do NOT keep the original leaf colors and do NOT fill any leaf with a color. The ONLY colored marks on a leaf are the 8-px-thick midline (one unique color per leaf, drawn on top of the black leaf) and the magenta node dot at its base. Everything else on the plant stays black.

    5. Keep the same image size and plant position.

    Return ONLY the image.
    \end{tcolorbox}

    \vspace{0.6em}

    \begin{tcolorbox}[colback=gray!5, colframe=gray!60,fontupper=\scriptsize\ttfamily, left=4pt, right=4pt, top=2pt, bottom=2pt, boxrule=0.5pt, title={\scriptsize\sffamily\bfseries Prompt 2: stalk centerline (front view only)}]
    On the point cloud data in this plant image, do the following:

    1. Keep the background EXACTLY as it is -- do not change any white area or the corner annotations (text labels, axis arrows, scale bar).

    2. Every leaf and the tassel must remain SOLID BLACK. Do NOT draw any midline, node dot, or any annotation on any leaf or on the tassel.

    3. STALK CENTERLINE: Identify the central stalk. Trace its midline following the stalk's natural shape and curvature from the very BASE (where it meets the ground) up to the TOP (top of the tassel). Draw this midline as a solid RED line (\textasciitilde{}8\,px thick) that follows the stalk's curve in the image -- not a straight line, but the actual centerline as the stalk appears.

    4. STALK ENDPOINTS: Place ONE solid magenta square (\textasciitilde{}12\,px) centered on the BOTTOM endpoint of the red line (base of the stalk) and ONE solid magenta square (\textasciitilde{}12\,px) centered on the TOP endpoint of the red line (at the end of the tassel). Only TWO magenta squares total -- no other magenta marks anywhere.

    5. Keep the same image size and plant position.

    Return ONLY the image.
    \end{tcolorbox}
    \caption{The two prompts used in the VLM stage (\secref{sec:vlm}). Prompt~1 is applied to each of the eight rendered views and produces the leaf midlines and stalk-node dots shown in \figref{fig:vlm_output}. Prompt~2 is applied to the front view only and produces the red stalk stroke and the two magenta endpoint squares used to reconstruct the 3D stalk centerline (\secref{sec:leafseg}).}
    \label{fig:prompt}
\end{figure}

\subsection{NURBS basis and differentiation}
\label{sec:appendix:nurbs}

\paragraph{Overview of NURBS}
NURBS are widely used to represent curves and surfaces in computer-aided design (CAD) and computer graphics. They are built on B-splines, which are piecewise polynomial curves constructed as linear combinations of B-spline basis functions~\citep{piegl2012nurbs}. These basis functions are defined over a knot vector $\mathbf{U}=\{u_0,\dots,u_{n+p+1}\}$, where $p$ is the degree of the B-spline and $n$ is the highest control-point index. A curve thus has $n+1$ control points and $n+p+2$ knots. The knot vector contains non-decreasing real numbers that partition the parametric space of the curve. The B-spline basis functions $N_{i,p}(u)$ are defined recursively, starting with piecewise constant functions for $p=0$,
\begin{equation}
N_{i,0}(u)=
    \begin{cases}
      1 & \text{if $u_{i}\leq u<u_{i+1}$}\\
      0 & \text{otherwise}
    \end{cases},
\end{equation}
and, for higher degrees $p=1,2,3,\dots$, computed using the recursive formula
\begin{equation}
N_{i,p}(u)=\frac{u-u_i}{u_{i+p}-u_i}N_{i,p-1}(u)+\frac{u_{i+p+1}-u}{u_{i+p+1}-u_{i+1}}N_{i+1,p-1}(u).
\end{equation}
A term with a zero denominator is defined as zero. At the clamped endpoint, we use the standard convention $N_{n,p}(1)=1$.
B-spline surfaces extend B-spline curves to two dimensions using tensor products. Given two knot vectors $\mathbf{U}=\{u_0,\dots,u_{n+p+1}\}$ and $\mathbf{V}=\{v_0,\dots,v_{m+q+1}\}$ and an $(n+1) \times (m+1)$ grid of control points $\mathbf{P}_{i,j}$, the B-spline surface is
\begin{equation}
\mathbf{S}(u,v)=\sum_{i=0}^{n}\sum_{j=0}^{m}N_{i,p}(u)N_{j,q}(v)\mathbf{P}_{i,j}.
\end{equation}
NURBS extend B-splines by assigning a weight $w_{i,j}$ to each control point $\mathbf{P}_{i,j}$, and the surface is defined as
\begin{equation}
\mathbf{S}(u,v)=\frac{\sum\limits_{i=0}^{n}\sum\limits_{j=0}^{m}N_{i,p}(u)N_{j,q}(v)w_{i,j}\mathbf{P}_{i,j}}{\sum\limits_{i=0}^{n}\sum\limits_{j=0}^{m}N_{i,p}(u)N_{j,q}(v)w_{i,j}}, \quad 0\leq u,v\leq 1.
\end{equation}
The rational basis functions give NURBS additional flexibility, allowing them to represent both standard geometric shapes (such as circles and ellipses) and complex free-form surfaces.

\paragraph{Differentiable NURBS}
NURBS-Diff~\citep{prasad2022nurbs} is a differentiable spline module that integrates NURBS CAD models with deep learning tools. It defines the derivatives of NURBS curves and surfaces with respect to their input parameters (control points, weights, and knot vectors) in closed form, and these derivatives are used to construct an approximate Jacobian for gradient-based optimization. Following \citet{hadadi2025procedural}, we hold the weights and knot vectors fixed and optimize only the control point positions. The derivative of the NURBS surface with respect to the control points is
\begin{equation}
\label{eq:S,P_ij}
\mathbf{S}_{,\mathbf{P}_{i,j}}(u,v) = \frac{N_{i,p}(u)N_{j,q}(v)w_{i,j}}{\sum\limits_{k=0}^{n}\sum\limits_{l=0}^{m}N_{k,p}(u)N_{l,q}(v)w_{k,l}}\,\mathbf{I}_3.
\end{equation}

\subsection{Conditional base-collar repair}
\label{sec:appendix:collar}

The fixed node-side row preserves the attachment, but the downstream rows remain free. When the extracted path begins behind the trusted blade base, optimization can fold these rows beneath the fixed row. We test a post-fit collar repair only for leaves whose path-start repair in \secref{sec:params_ppg} was applied. For control rows $1$ to $3$, the first candidate places each row center on a Hermite curve between the fixed node-row center and the fitted center of row $4$, with tangent scale $0.65$. Its transverse offsets are blended with a smoothstep function from the procedural offsets at that row to the fitted offsets at row $4$. Row $4$ and all subsequent rows remain unchanged. If the Hermite-center candidate fails, a linear center path with the same transverse-offset blend is tested. A candidate is accepted only if the row centers progress positively along the chord from the node row to row~4, every adjacent transverse edge has a positive dot product with its procedural counterpart, and no sampled collar cell with nondegenerate candidate and reference areas has an orientation cosine below $-0.05$. Otherwise, the original fit is retained. The post-repair Chamfer value is saved for auditing but is not used as an additional acceptance gate.

\subsection{Pipeline parameters}
\label{sec:appendix:parameters}

\tabref{tab:parameters} summarizes the principal numeric parameters stated in the Methods section, together with the stage that applies each parameter and the source of its value. The table distinguishes four provenance sources. \textsf{P} denotes values saved in the run configuration, \textsf{L} denotes values saved in the per-plant blade-refinement configuration, \textsf{A} denotes refinement-stage arguments, and \textsf{M} denotes module constants fixed in the source code. The tabulated values were checked against the saved configuration and source code used for the reported runs. Three parameters (the prealignment confidence margin and the two local-roll acceptance limits) were varied in \secref{sec:guards_results}.

\begin{table}[p]
\centering
\footnotesize
\caption{Numeric parameters of the pipeline, grouped by the stage that applies them. The Src column gives the provenance code (\textsf{P}, \textsf{L}, \textsf{A}, or \textsf{M}) defined in the text. In the Unit column, ``n/a'' indicates that no physical unit applies.}
\label{tab:parameters}
\begin{tabular}{@{}llrl c@{}}
\toprule
Stage & Role & Value & Unit & Src \\
\midrule
Preprocessing & Unit conversion applied to all input clouds & $100$ & cm\,m$^{-1}$ & \textsf{M} \\
 & Radial percentile defining the stalk-base column & $20$ & \% & \textsf{M} \\
Rendering & Metric scale bar drawn in each view & $10$ & cm & \textsf{M} \\
Annotation & Aspect-ratio tolerance accepting a returned canvas & $0.05$ & n/a & \textsf{M} \\
 & Silhouette IoU floor for a leaf annotation & $0.60$ & n/a & \textsf{M} \\
 & Silhouette IoU floor for a stalk annotation & $0.40$ & n/a & \textsf{M} \\
 & Hue window isolating the magenta stroke & $[145, 162]$ & n/a & \textsf{M} \\
 & Saturation and value floor for that stroke & $165$ & n/a & \textsf{M} \\
Back-projection & Additional normalized-depth slack above the calibrated splat offset & $0.008$ & normalized depth & \textsf{M} \\
Consensus & Shared points required to link two per-view leaves & $40$ & points & \textsf{M} \\
 & Vertical separation admitting a leaf split & $6$ & cm & \textsf{M} \\
 & Cluster span triggering re-clustering & $8$ & cm & \textsf{M} \\
 & Distance for merging an embedded leaf into its host & $2$ & cm & \textsf{P} \\
 & Embedded-leaf point fraction required & $0.5$ & n/a & \textsf{P} \\
 & Host share defining a dominant host & $0.7$ & n/a & \textsf{P} \\
 & Smallest surviving connected component & $50$ & points & \textsf{P} \\
 & Distance beyond which a component is speckle & $45$ & cm & \textsf{P} \\
Stalk & Stray hits discarded around a node square & $4$ & cm & \textsf{P} \\
 & Neighbors in the centerline graph & $100$ & n/a & \textsf{M} \\
 & Radial penalty weight on graph edges & $0.05$ & n/a & \textsf{M} \\
 & Samples in the resampled centerline & $400$ & n/a & \textsf{M} \\
 & Subsampling ceiling before the graph is built, inert at $10^{4}$ & $10^{5}$ & points & \textsf{M} \\
Blade refinement & Bounded stalk radius below 70\% of height & $3$ & cm & \textsf{M} \\
 & Quantile determining the adaptive radius above it & $0.7$ & n/a & \textsf{M} \\
 & Floor on the adaptive radius & $0.5$ & cm & \textsf{M} \\
 & Neighbors defining the local surface frame & $20$ & n/a & \textsf{L} \\
 & Neighbors in the propagation graph & $20$ & n/a & \textsf{L} \\
 & Edge-length cap in the propagation graph & $4$ & $\times$ median & \textsf{L} \\
 & Euclidean weight added to the geodesic cost & $0.25$ & n/a & \textsf{L} \\
 & Hysteresis margin before a label switches & $8$ & cm & \textsf{L} \\
 & Node-to-base distance admitting an attachment & $6$ & cm & \textsf{M} \\
 & Minimum spacing between adjacent nodes & $4$ & cm & \textsf{M} \\
 & Smallest retained leaf & $20$ & points & \textsf{P} \\
 & Percentile determining the envelope half-axes & $90$ & \% & \textsf{M} \\
 & Cap on per-patch envelope width & $1.5$ & $\times$ median & \textsf{M} \\
 & Moving-median window smoothing the envelope & $3$ & patches & \textsf{M} \\
 & Neighbors in the recomputed midline graph & $12$ & n/a & \textsf{M} \\
 & Edge cap in that graph & $5$ & $\times$ median & \textsf{M} \\
 & Samples in the recomputed midline & $200$ & n/a & \textsf{M} \\
Descriptor & Neighbors in the parameter-path graph & $50$ & n/a & \textsf{P} \\
 & Edge cap in the parameter-path graph & $6$ & $\times$ median & \textsf{P} \\
 & Control points sampled along each path & $8$ & n/a & \textsf{P} \\
 & Half-width of the transverse width window & $1$ & cm & \textsf{M} \\
 & Maximum leading portion removed by path repair & $20$ & \% & \textsf{M} \\
 & Gap required before a midline start is repaired & $2$ & cm & \textsf{M} \\
 & Backtracking required alongside that gap & $1$ & cm & \textsf{M} \\
 & Arc length over which the repair tapers out & $5$ & cm & \textsf{M} \\
 & Control-net rows along the blade, degree 3 & $11$ & n/a & \textsf{P} \\
 & Control-net columns across the blade, degree 2 & $3$ & n/a & \textsf{P} \\
Refinement & Blade points sampled as the fitting target & $4000$ & points & \textsf{P} \\
 & Optimizer iterations per leaf & $500$ & n/a & \textsf{P} \\
 & Initial learning rate & $0.01$ & n/a & \textsf{P} \\
 & Weight on the transverse second-difference term & $0.01$ & n/a & \textsf{A} \\
 & Weight on the mixed-difference term & $0.0001$ & n/a & \textsf{A} \\
 & Weight on the transverse-compactness term & $0.0008$ & n/a & \textsf{A} \\
 & Minimum accepted cell-orientation cosine & $0.02$ & n/a & \textsf{M} \\
 & Minimum cell-area ratio to the initialization & $0.02$ & n/a & \textsf{M} \\
 & Improvement required to accept a prealignment & $0.08$ & n/a & \textsf{A} \\
 & Base and tip fractions anchoring that test & $0.02$ & n/a & \textsf{A} \\
 & Fidelity ratio admitting a fold repair & $1.02$ & n/a & \textsf{P} \\
 & Bad middle-cell fraction flagging a fold & $0.03$ & n/a & \textsf{A} \\
 & Blade fraction excluded at each end & $0.20$ & n/a & \textsf{A} \\
 & Tangent scale blending the repaired collar & $0.65$ & n/a & \textsf{P} \\
 & Row at which the collar rejoins the surface & $4$ & n/a & \textsf{P} \\
 & Attachment rows held fixed during optimization & $1$ & n/a & \textsf{P} \\
\bottomrule
\end{tabular}
\end{table}

\section{Per-plant qualitative results}
\label{sec:sample:plants}
\setcounter{figure}{0} \setcounter{table}{0}

\tabref{tab:AllGenotypesTable} presents intermediate and final pipeline outputs for the ten plants reported leaf by leaf.

\newcommand{\galcell}[1]{%
    \raisebox{-0.5\height}{\includegraphics[width=0.155\textwidth]{#1}}}
\newcommand{\galname}[1]{\scriptsize #1}
\newcommand{\galhead}{%
    \scriptsize Genotype & \scriptsize Input scan & \scriptsize Consensus midlines
    & \scriptsize Blade segmentation & \scriptsize Initial procedural model
    & \scriptsize Refined procedural model \\
    & \scriptsize (stage 1) & \scriptsize (stage 5) & \scriptsize (stage 6)
    & \scriptsize (stage 7) & \scriptsize (stage 8) \\}
\setlength{\LTcapwidth}{\textwidth}
\setlength{\tabcolsep}{1.5pt}
\renewcommand{\arraystretch}{0.9}

\begin{longtable}{@{}>{\raggedright\arraybackslash}m{0.09\textwidth}
                   @{\hspace{3pt}}ccccc@{}}
    \caption{Per-plant qualitative outputs for the ten maize plants reported
    leaf by leaf, in the genotype order of \tabref{tab:plants}. The pipeline stage from \figref{fig:pipeline} is listed below each column heading. Each row shows the canonicalized input point cloud, the consensus leaf midlines, the blade-level segmentation after geometric correction, the initial procedural model generated from the measured descriptor, and the procedural model after NURBS-Diff refinement. Panels in each row use the same camera. Each plant is framed separately, so apparent size is not comparable across rows.}
    \label{tab:AllGenotypesTable} \\
    \toprule \galhead \midrule
    \endfirsthead
    \multicolumn{6}{@{}l@{}}{\scriptsize\emph{Table \thetable, continued}} \\
    \toprule \galhead \midrule
    \endhead
    \bottomrule
    \endlastfoot
    \galname{M162W} & \galcell{dataset_9009.png}
        & \galcell{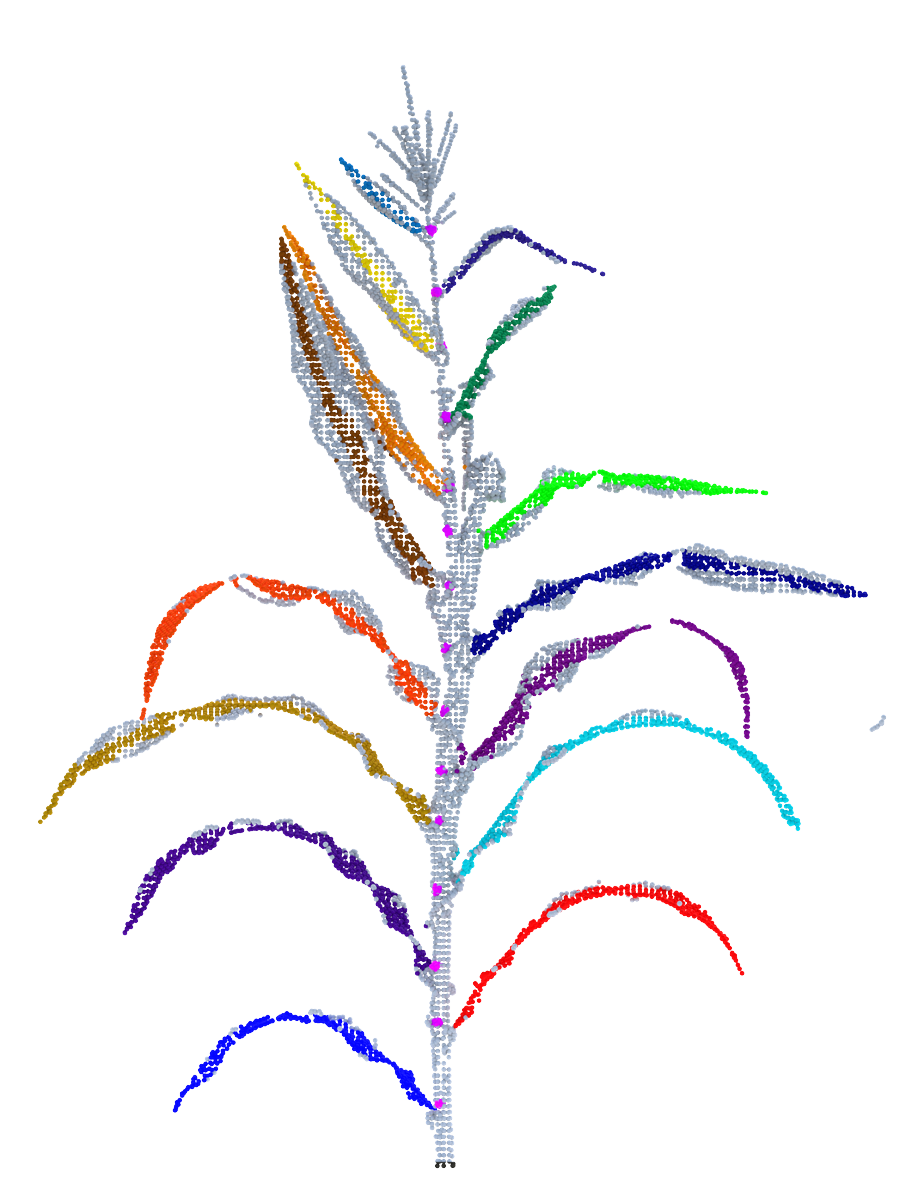} & \galcell{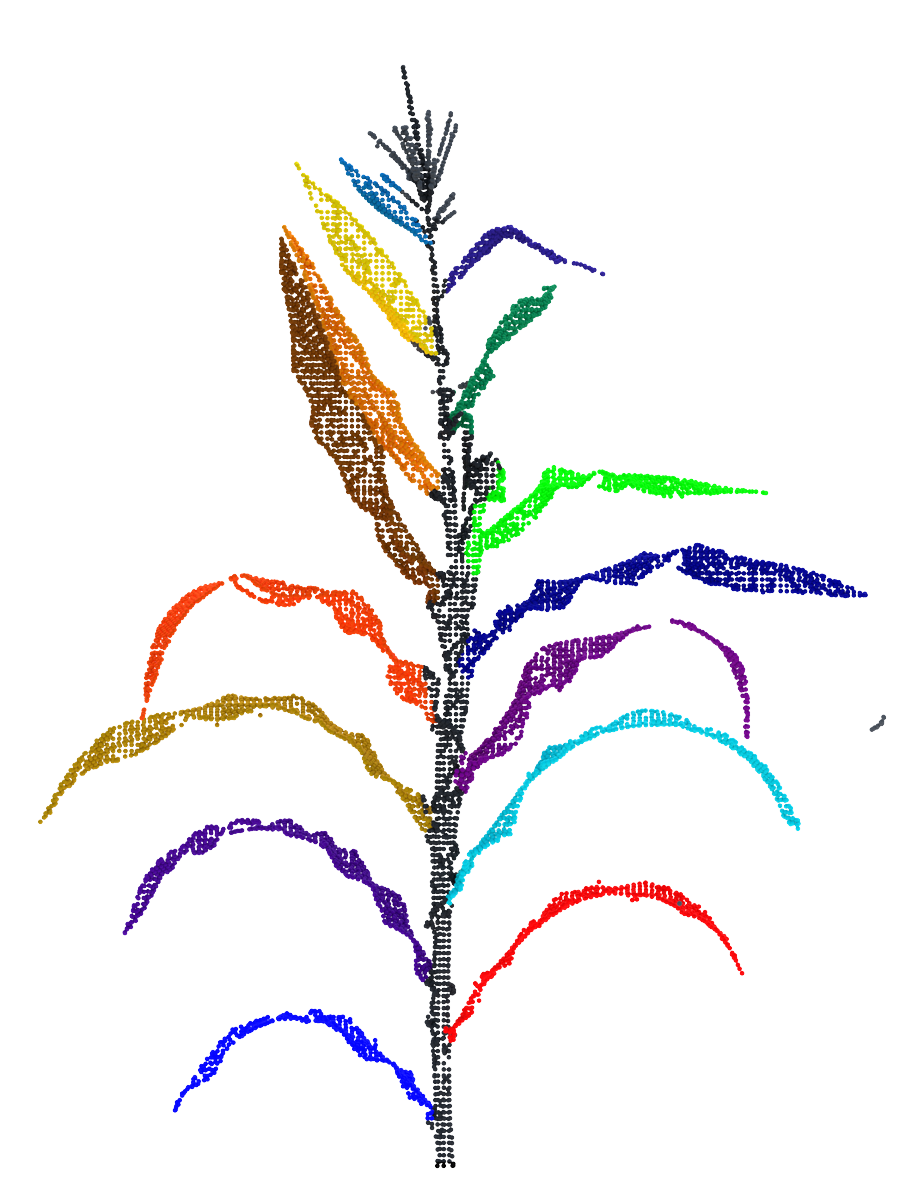}
        & \galcell{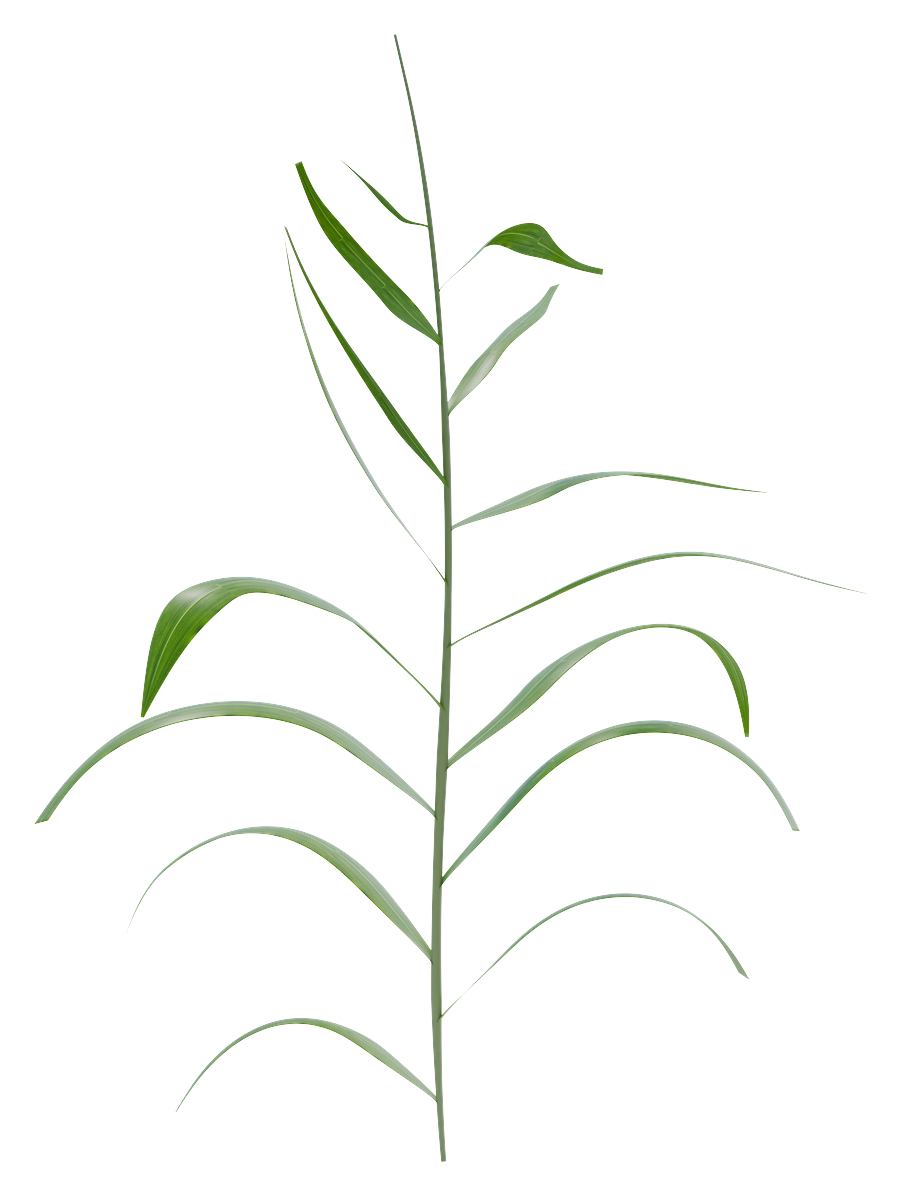} & \galcell{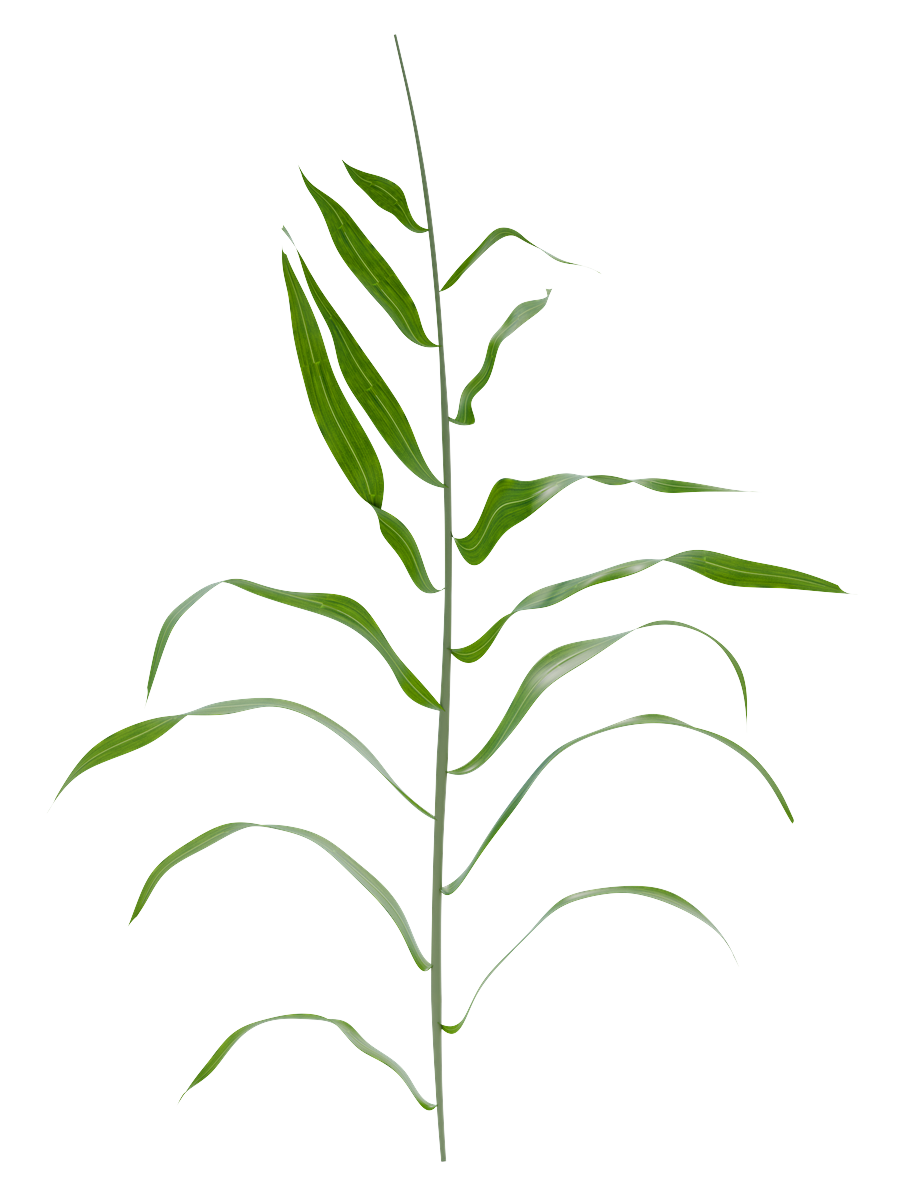} \\
    \galname{CML69} & \galcell{dataset_9005.png}
        & \galcell{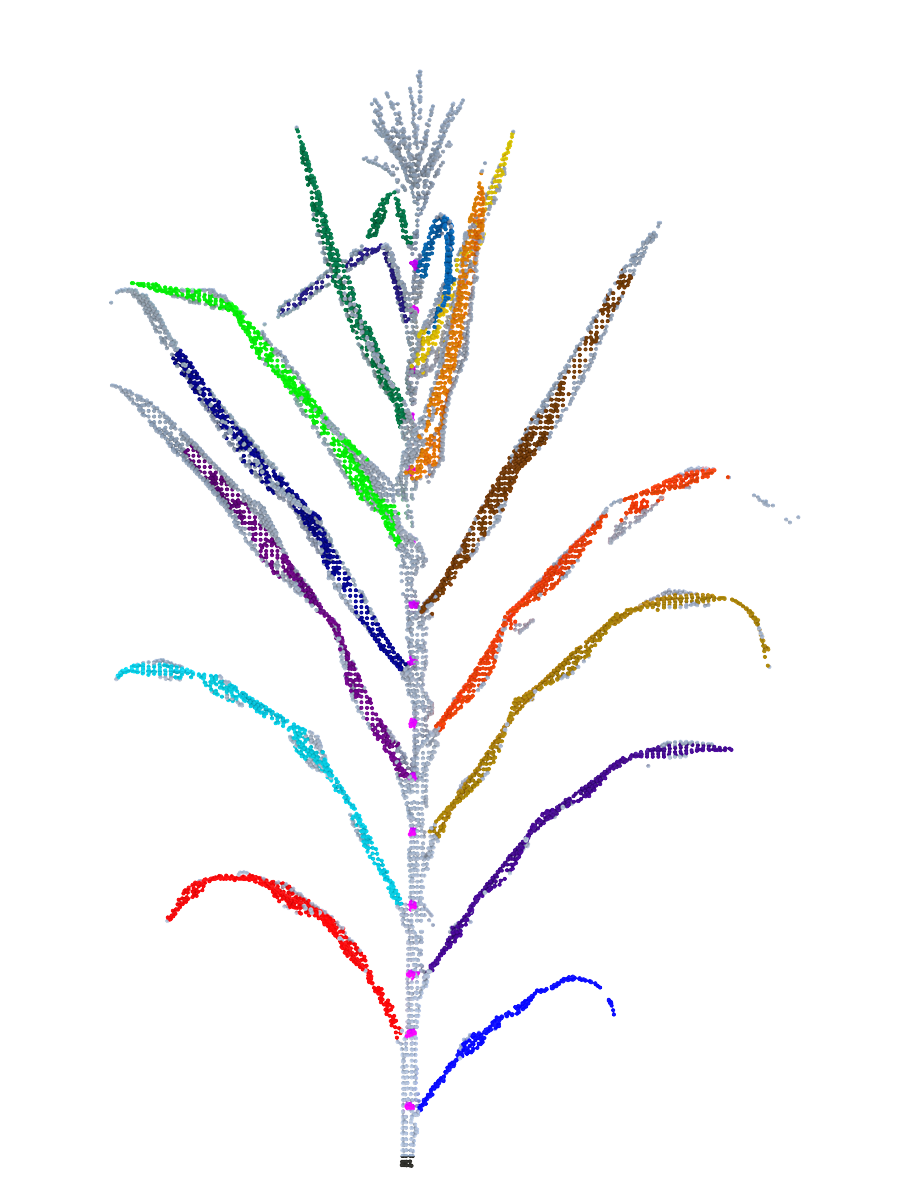} & \galcell{stage_9005_leafseg.png}
        & \galcell{stage_9005_ppg.png} & \galcell{stage_9005_nurbs.png} \\
    \galname{Kui2021} & \galcell{dataset_9008.png}
        & \galcell{stage_9008_midline.png} & \galcell{stage_9008_leafseg.png}
        & \galcell{stage_9008_ppg.png} & \galcell{stage_9008_nurbs.png} \\
    \galname{CML238} & \galcell{dataset_9006.png}
        & \galcell{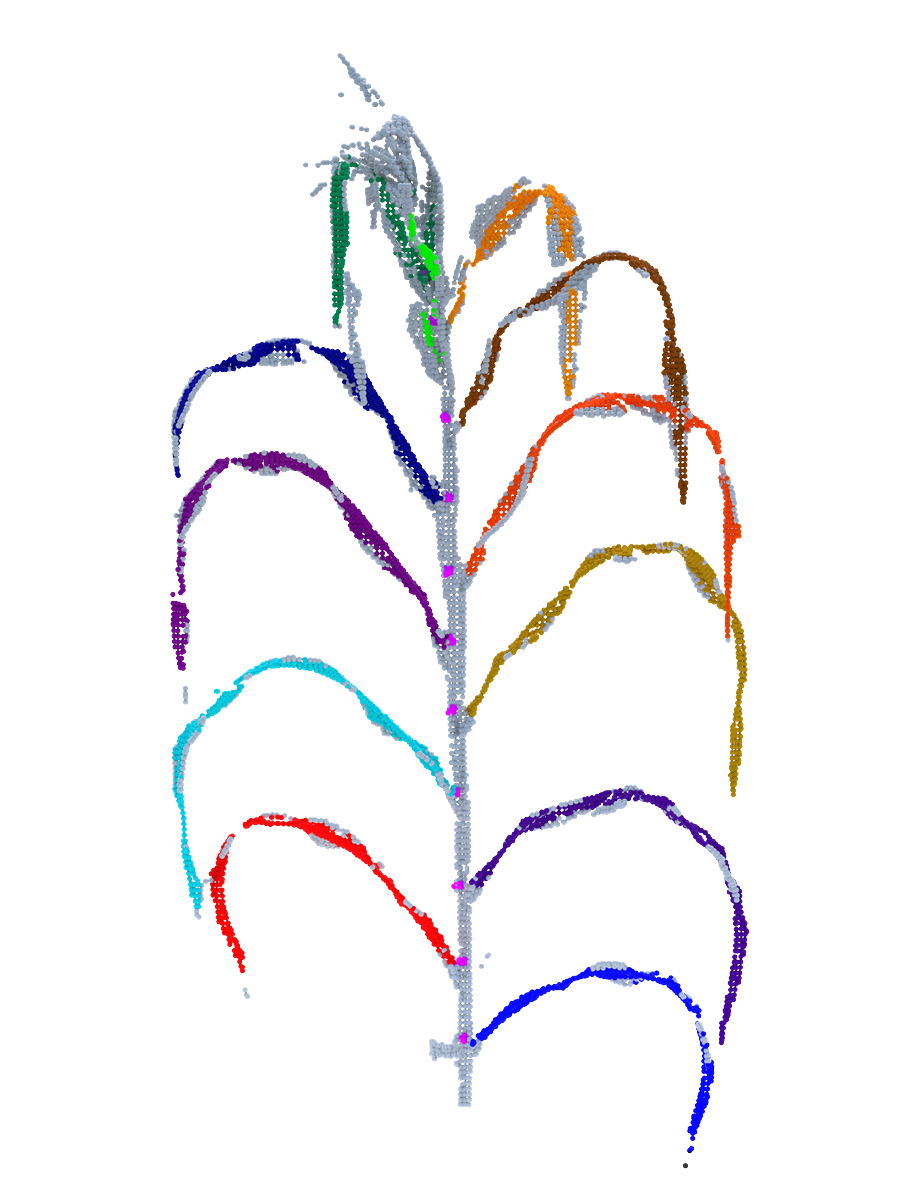} & \galcell{stage_9006_leafseg.png}
        & \galcell{stage_9006_ppg.png} & \galcell{stage_9006_nurbs.png} \\
    \galname{CML331} & \galcell{dataset_9007.png}
        & \galcell{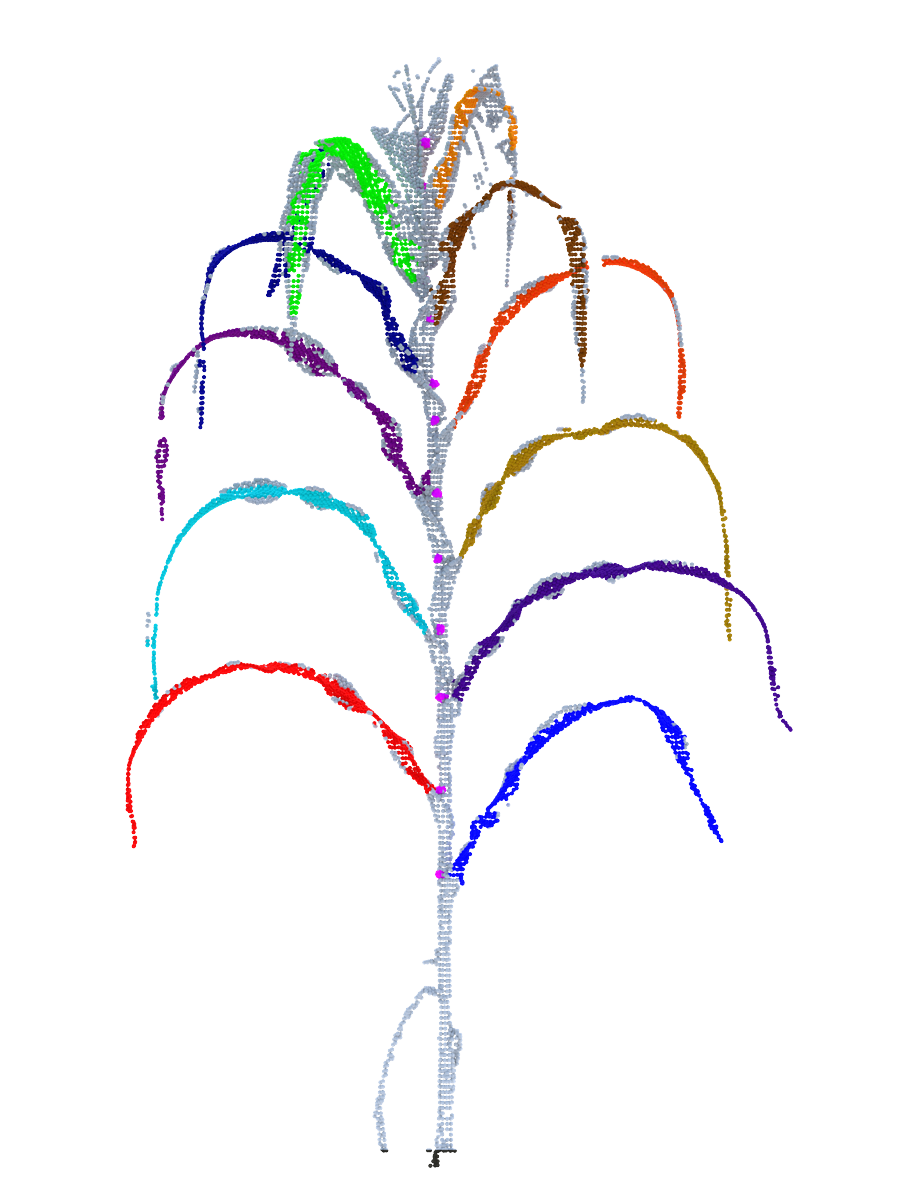} & \galcell{stage_9007_leafseg.png}
        & \galcell{stage_9007_ppg.png} & \galcell{stage_9007_nurbs.png} \\
    \galname{4226} & \galcell{dataset_9000.png}
        & \galcell{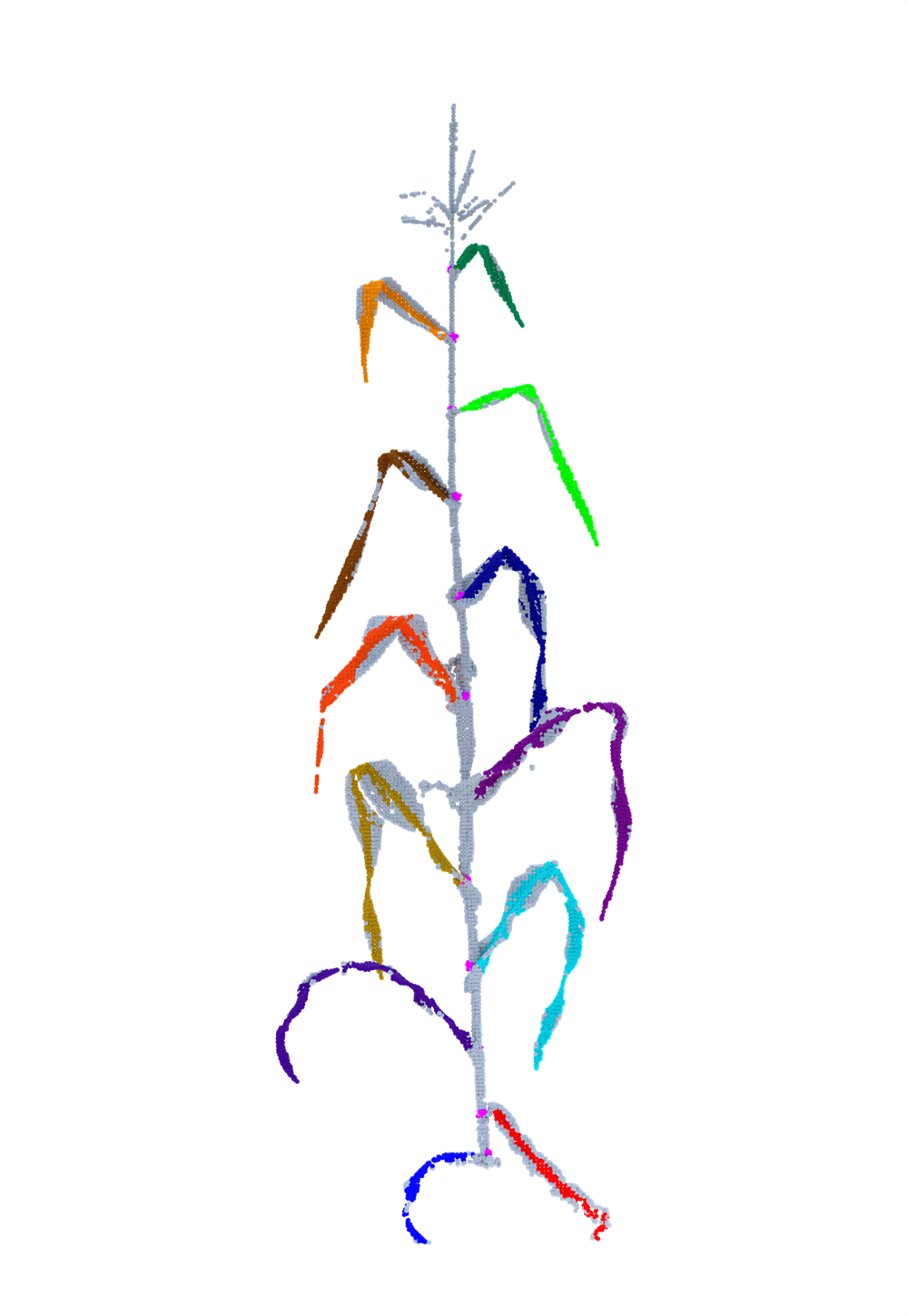} & \galcell{stage_9000_leafseg.png}
        & \galcell{stage_9000_ppg.png} & \galcell{stage_9000_nurbs.png} \\
    \galname{78551S} & \galcell{dataset_0335.png}
        & \galcell{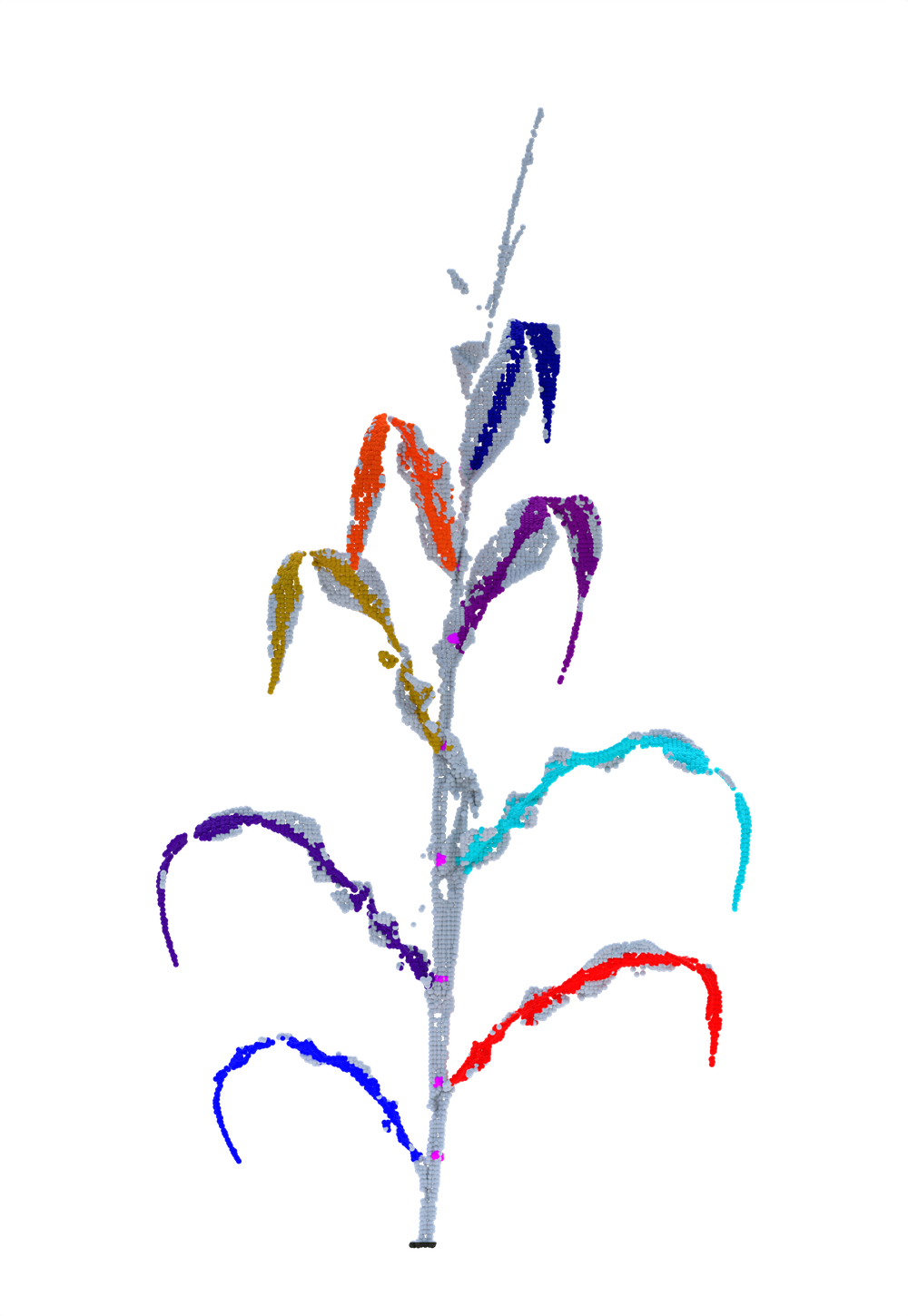} & \galcell{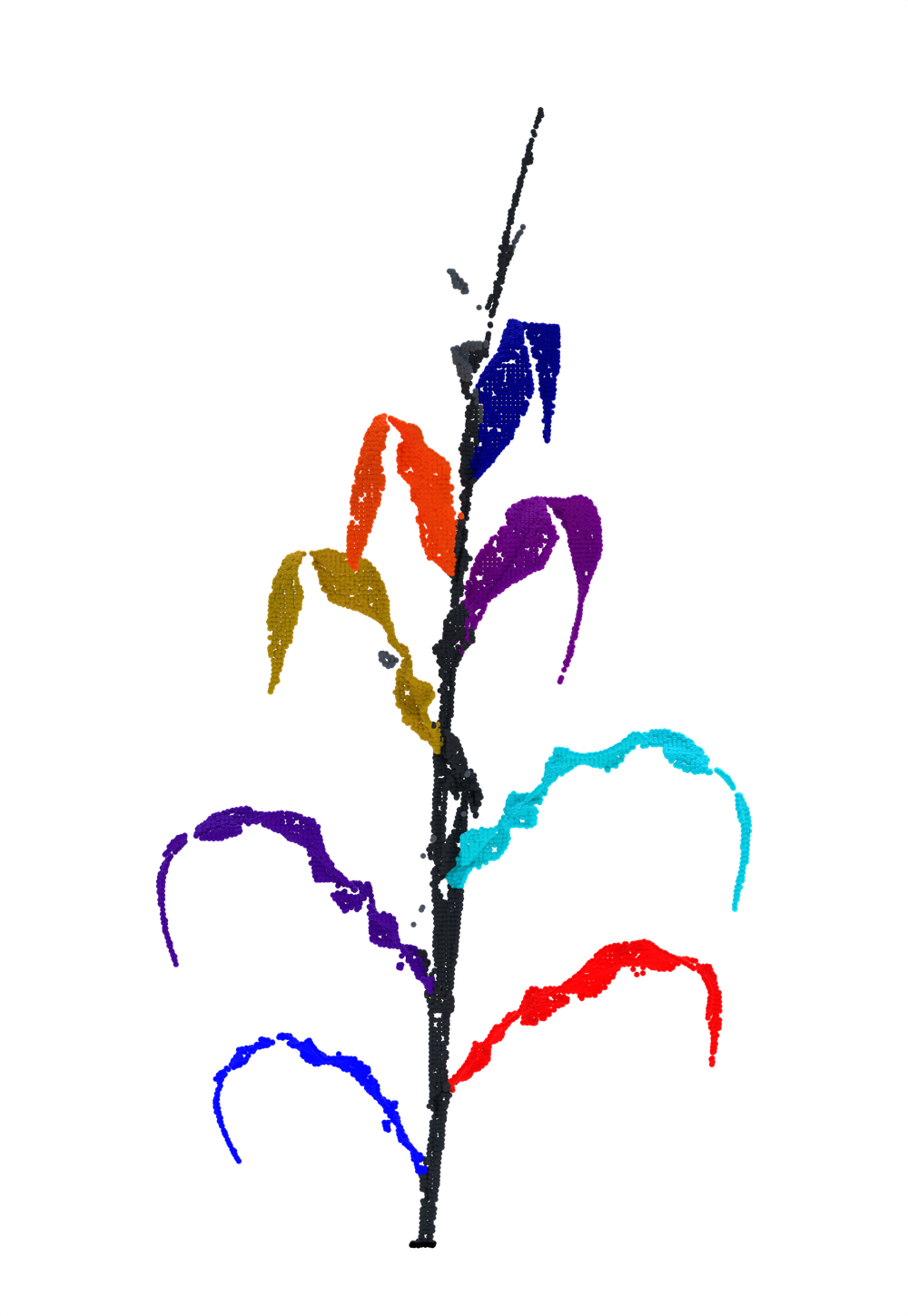}
        & \galcell{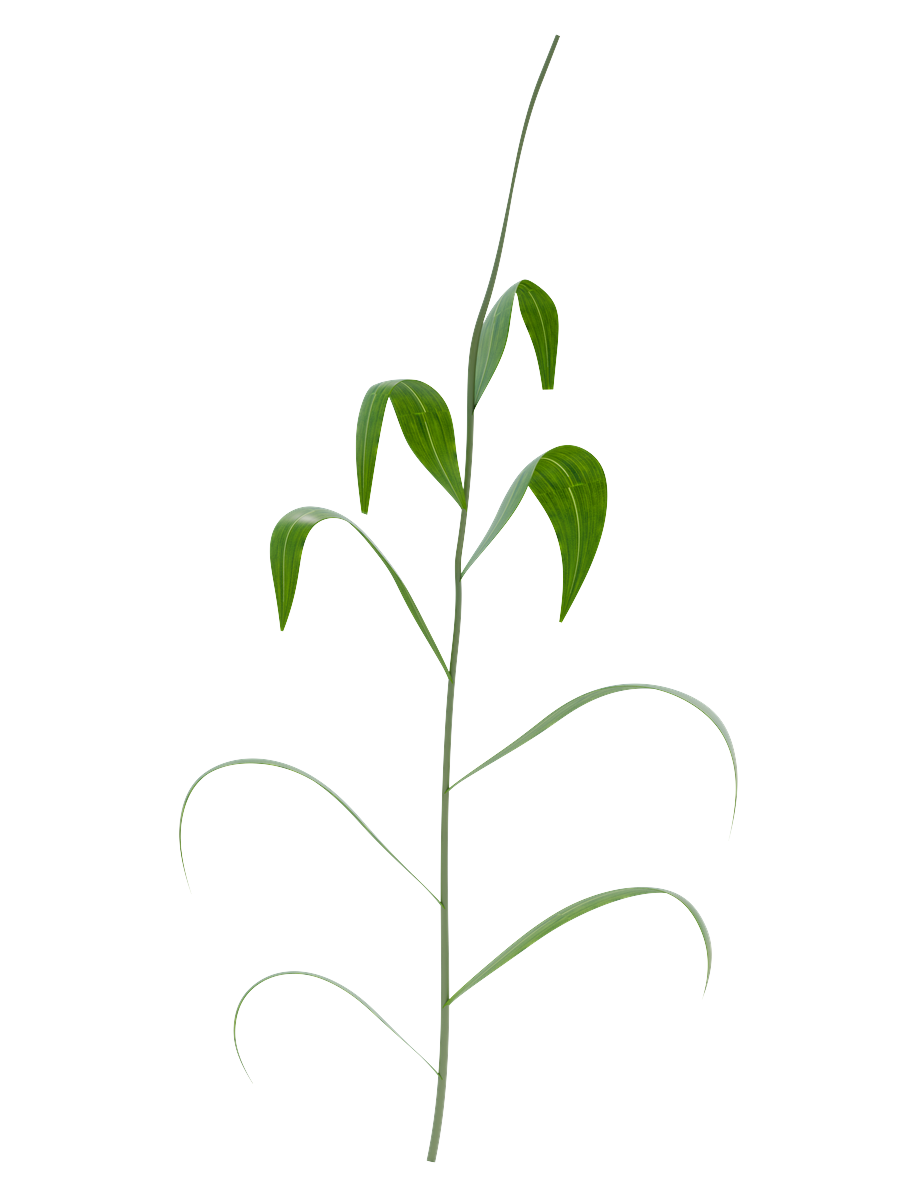} & \galcell{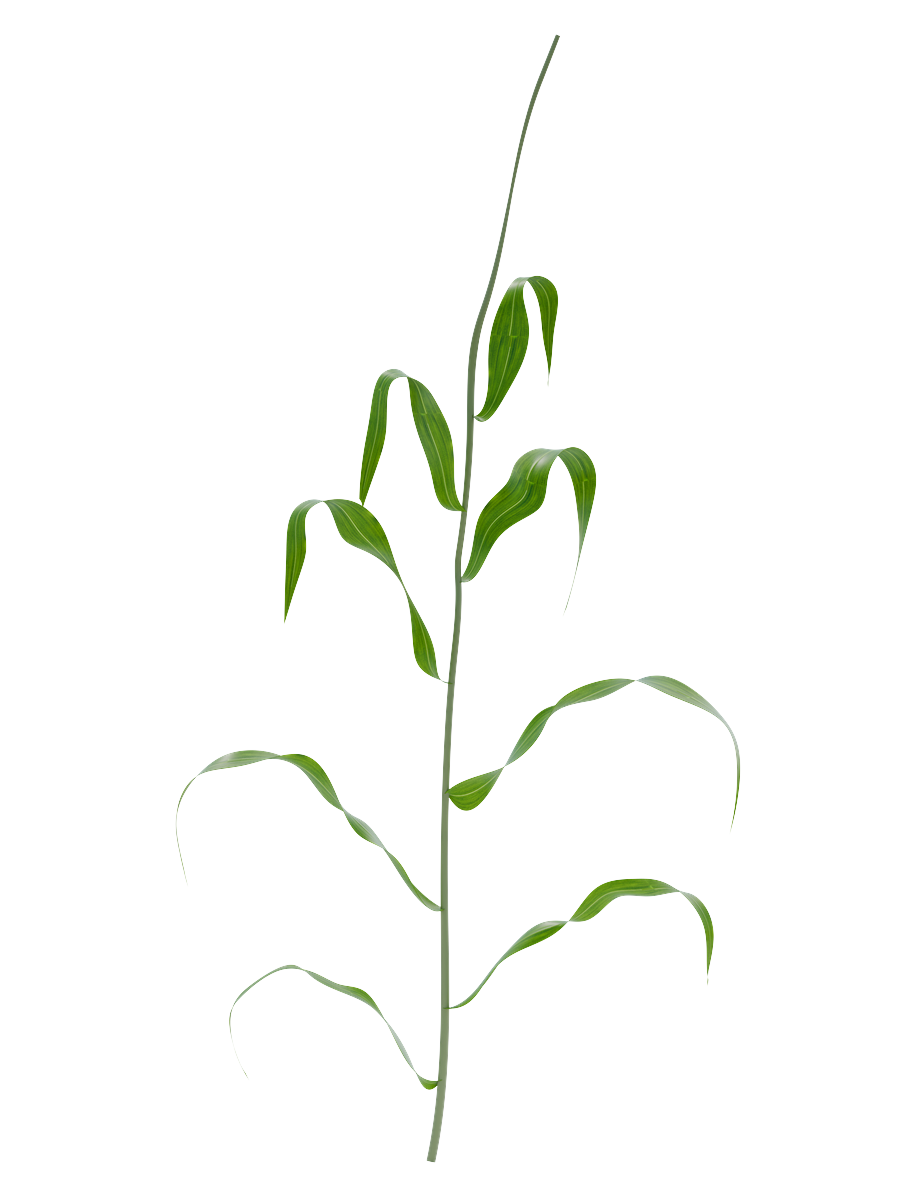} \\
    \galname{A679} & \galcell{dataset_9002.png}
        & \galcell{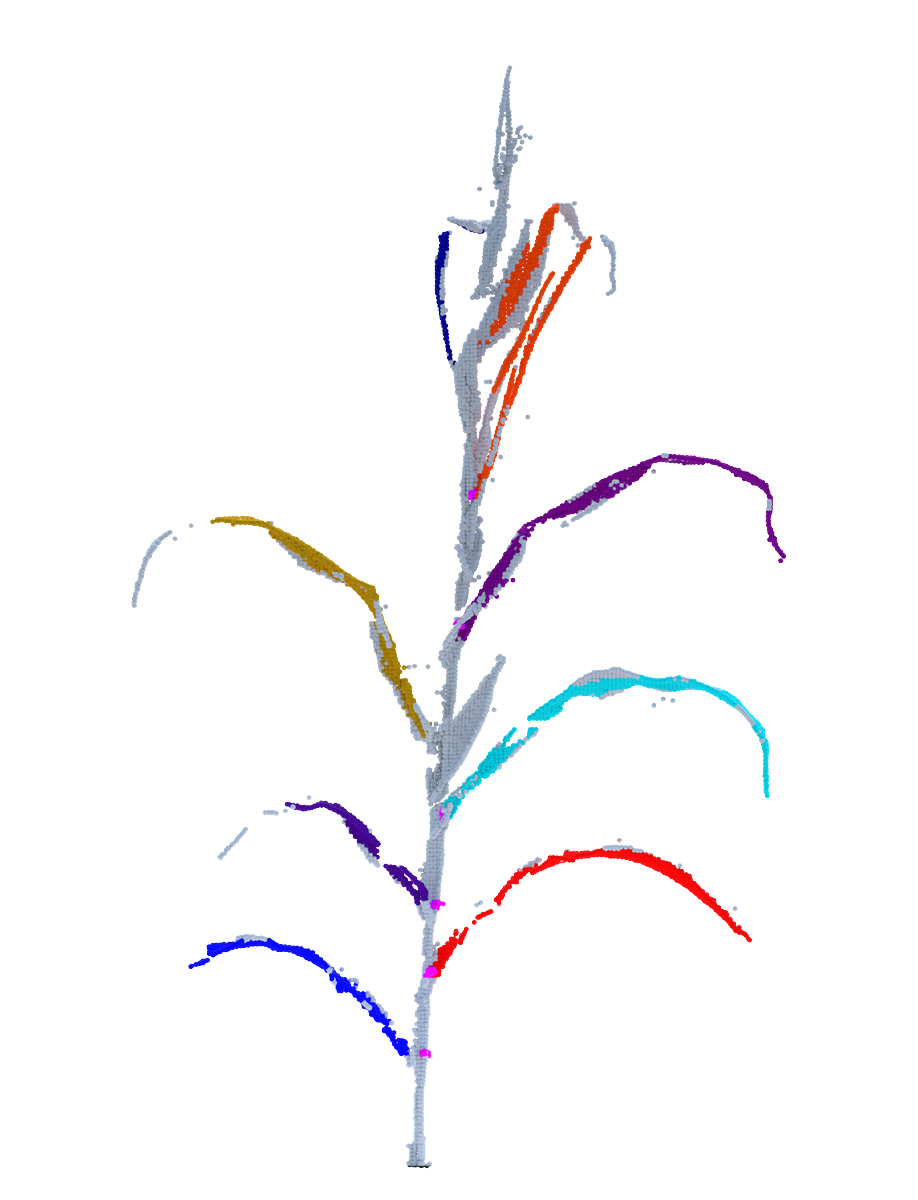} & \galcell{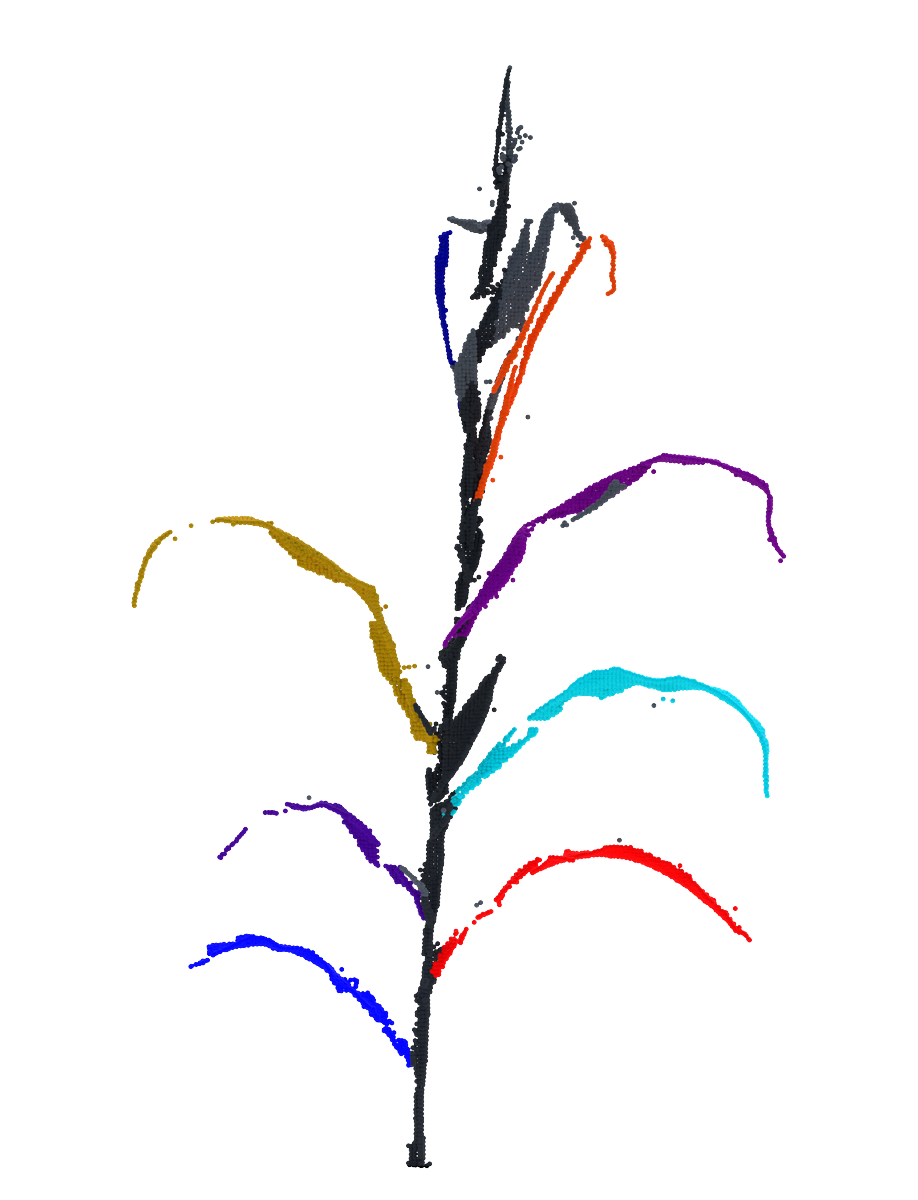}
        & \galcell{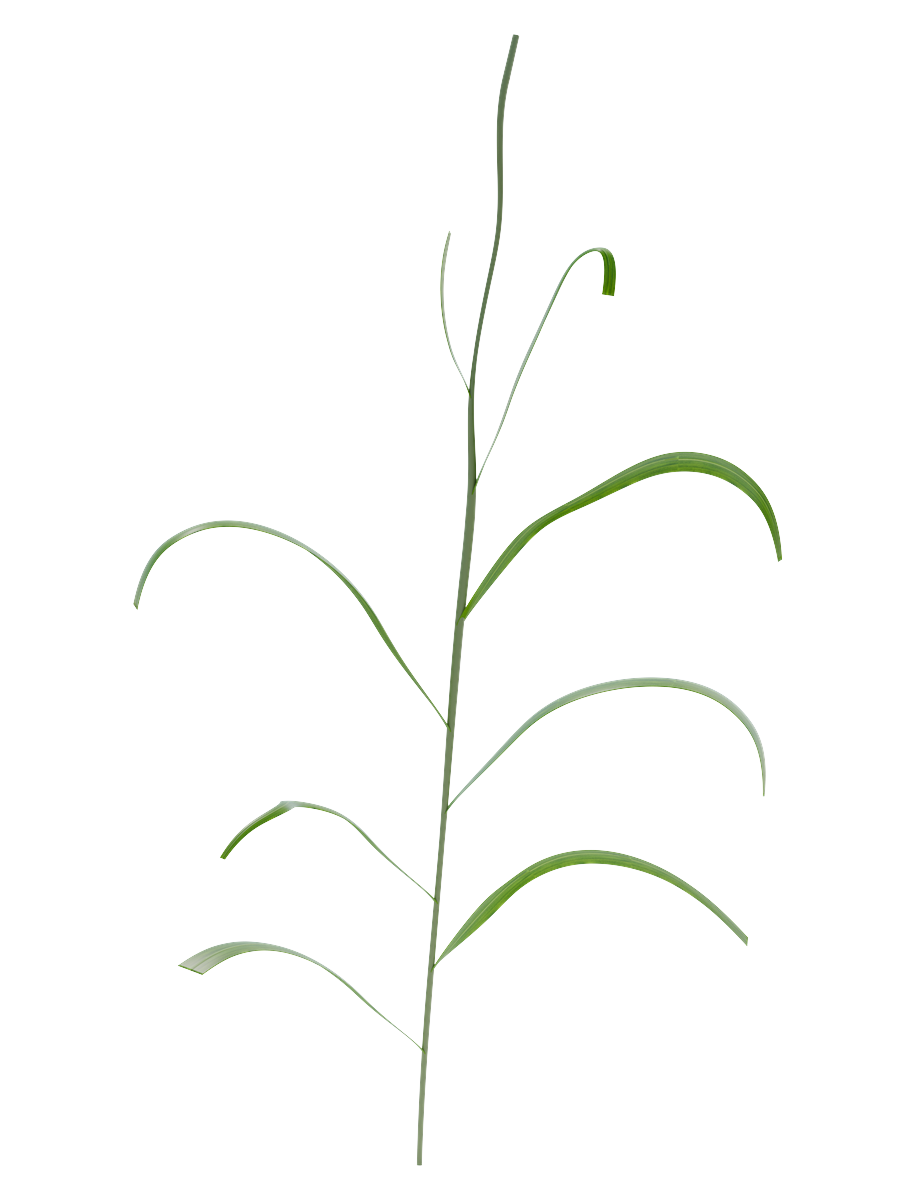} & \galcell{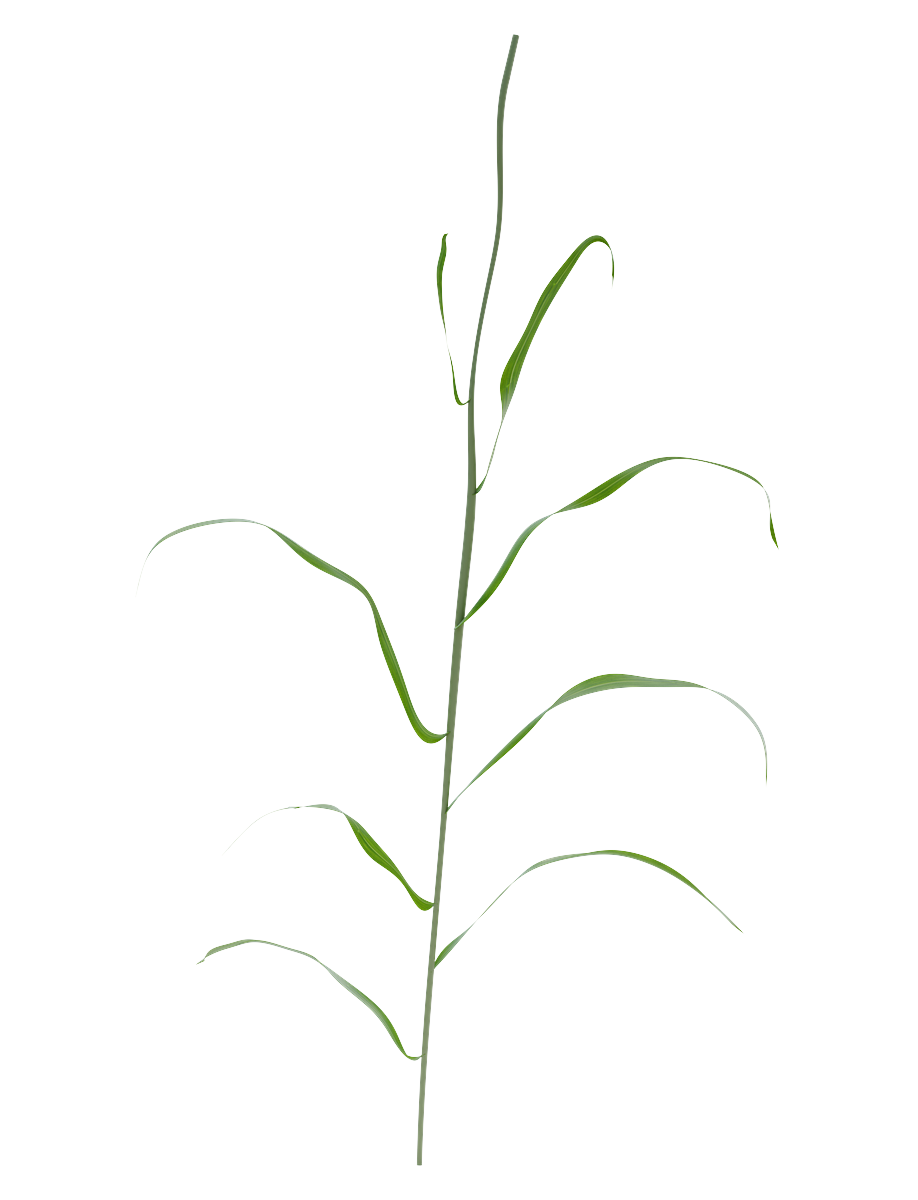} \\
    \galname{CI90C} & \galcell{dataset_9004.png}
        & \galcell{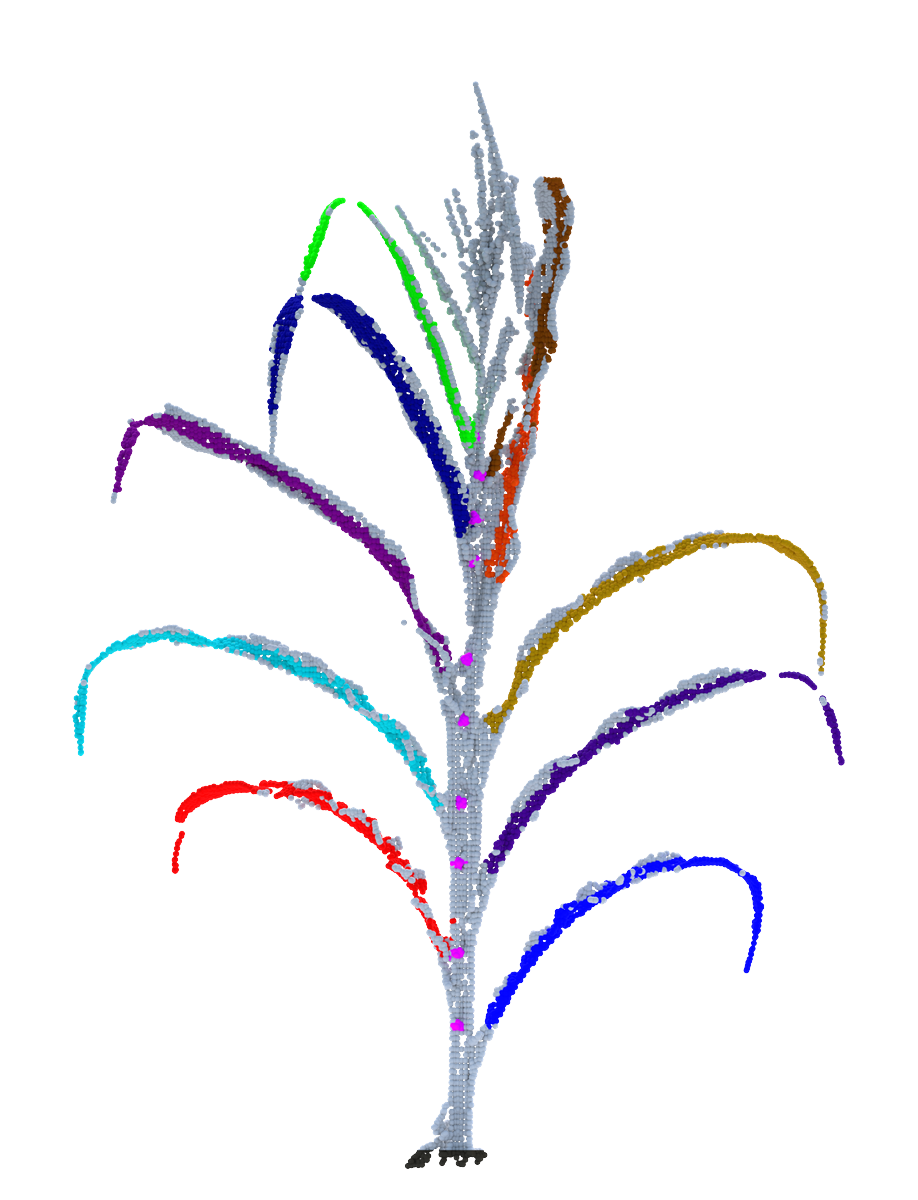} & \galcell{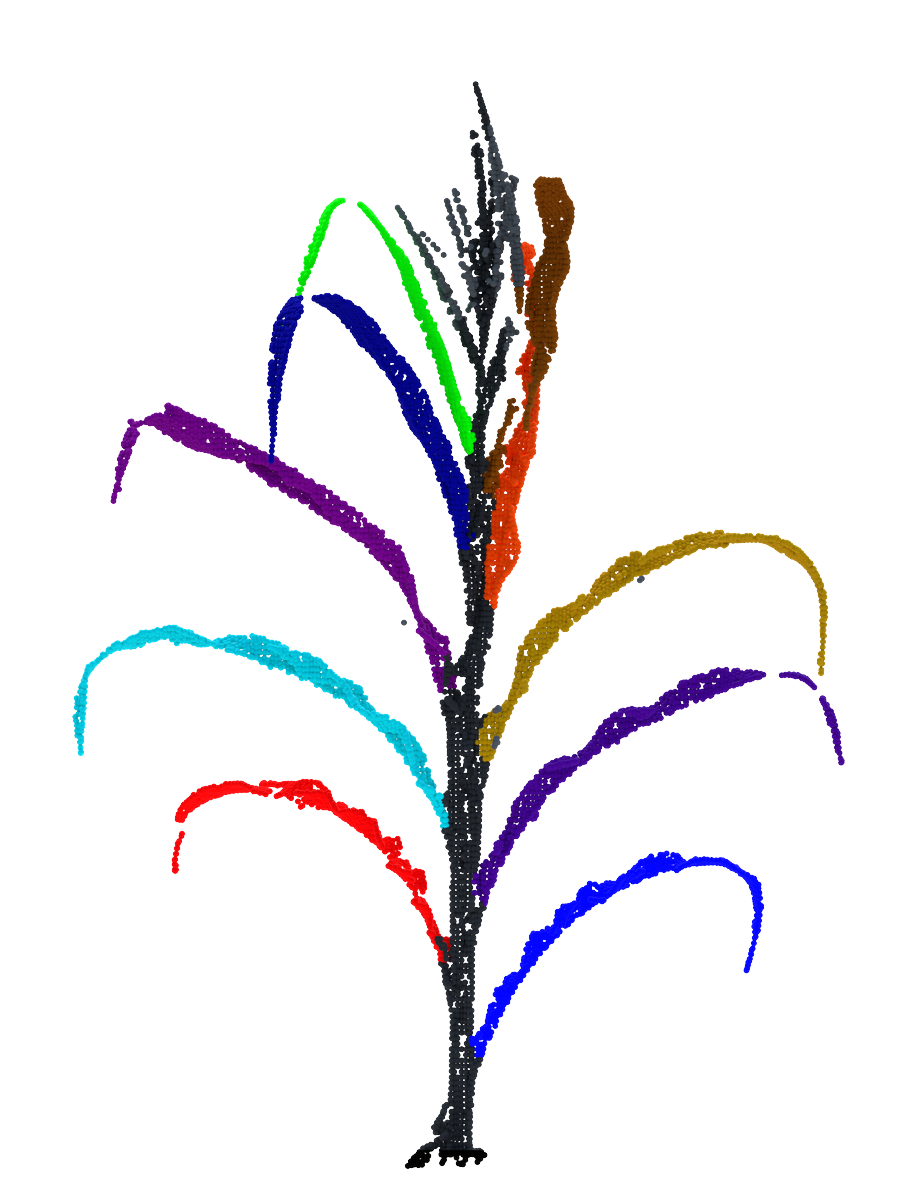}
        & \galcell{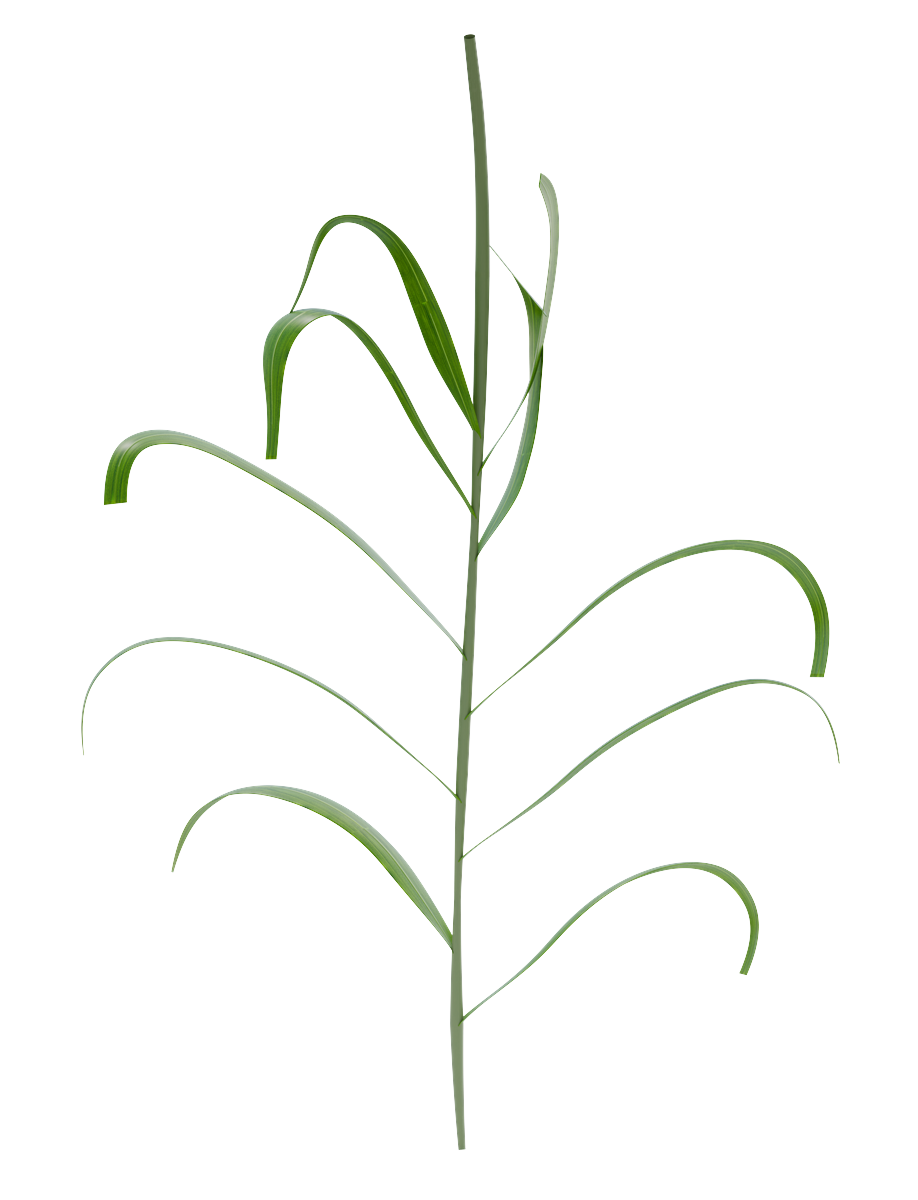} & \galcell{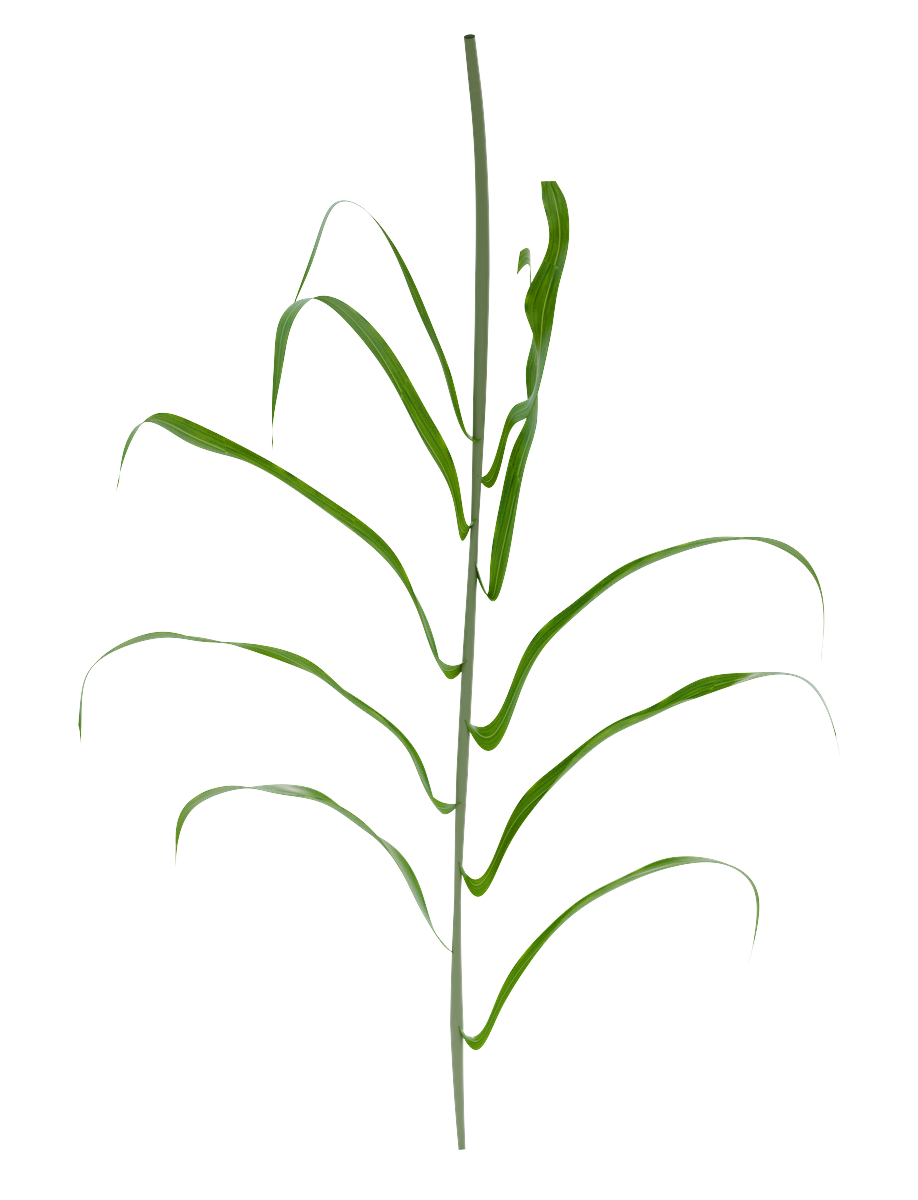} \\
    \galname{T8} & \galcell{dataset_9011.png}
        & \galcell{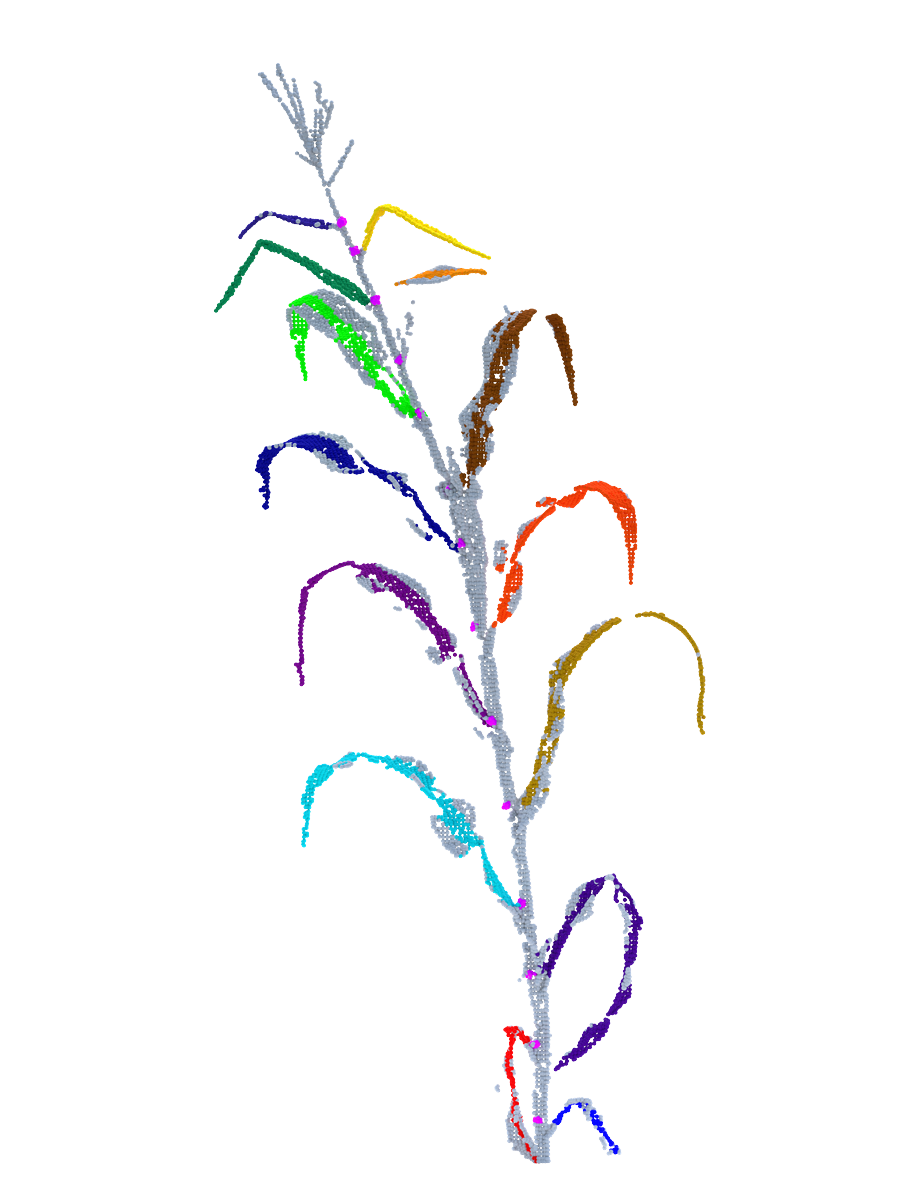} & \galcell{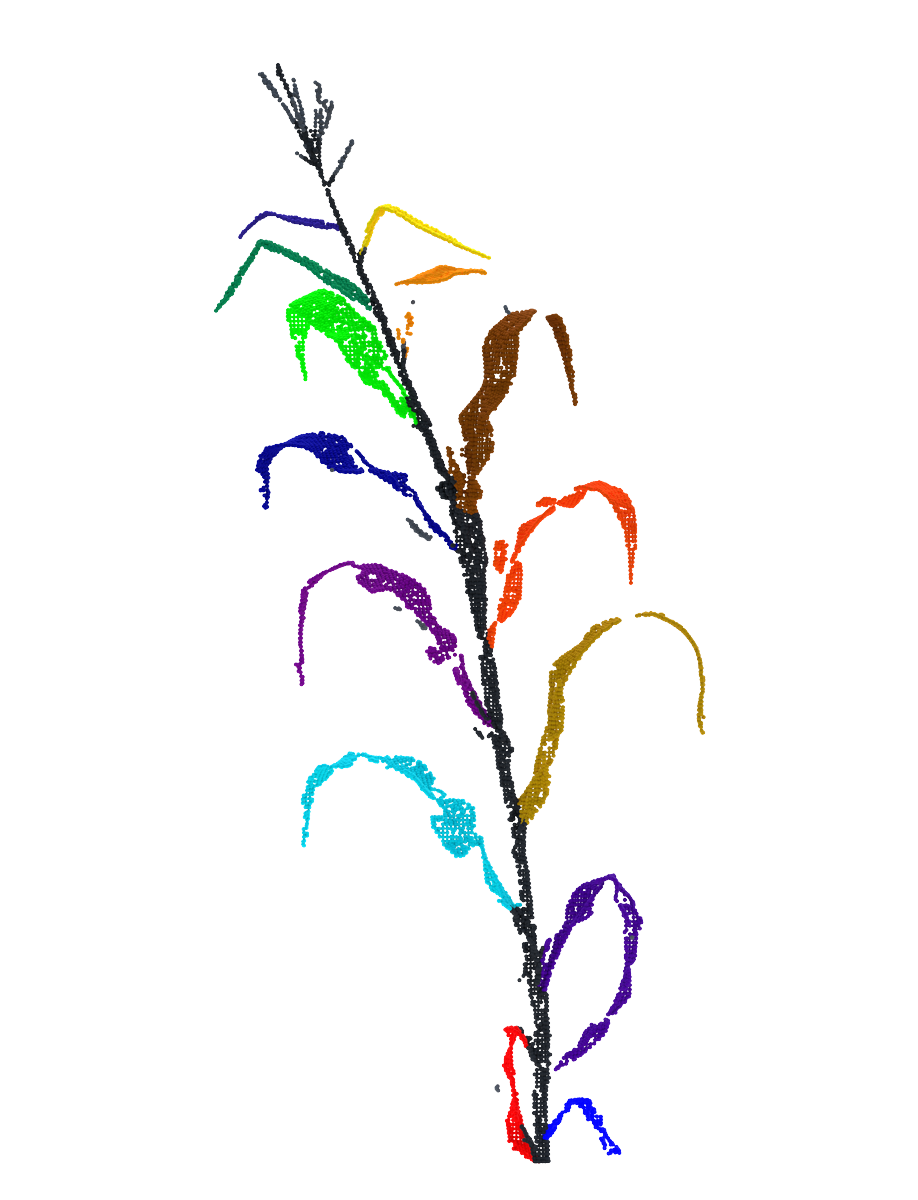}
        & \galcell{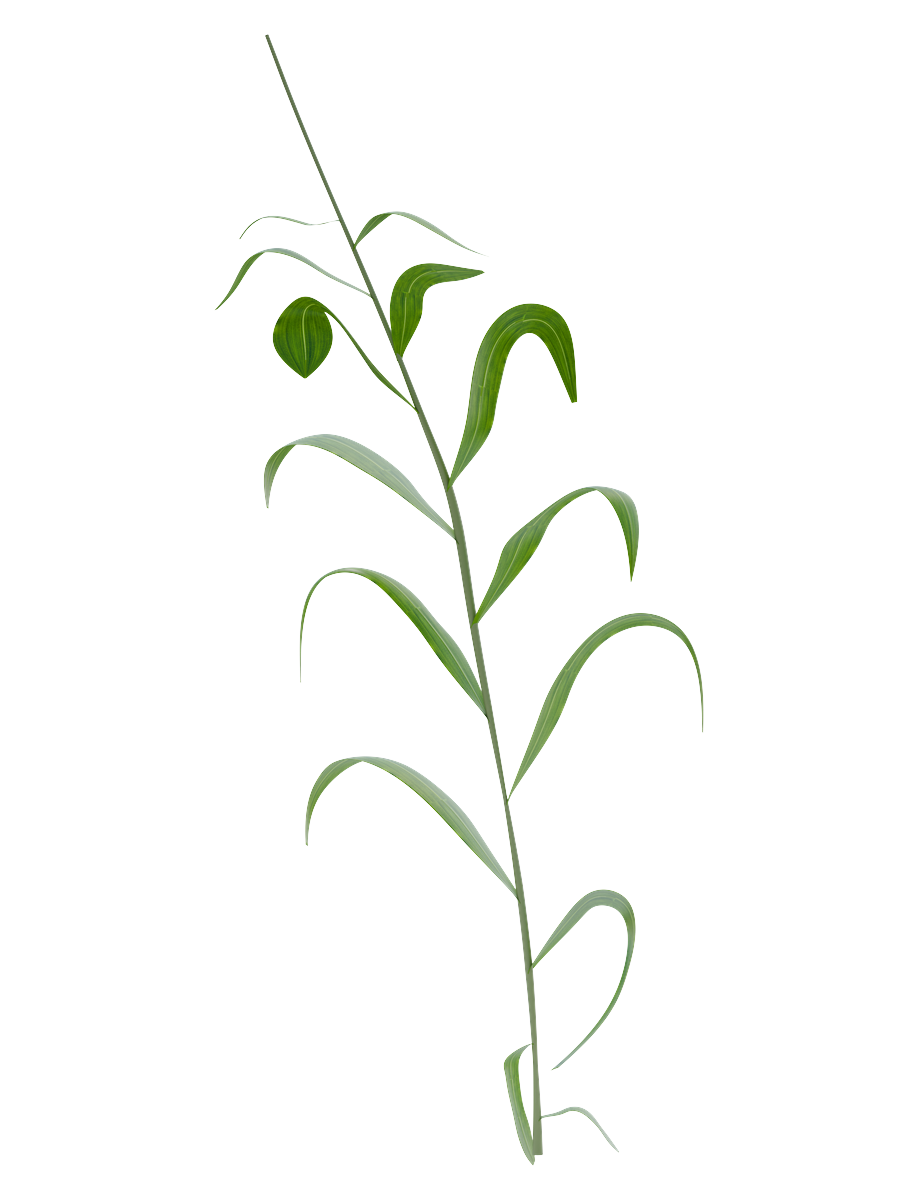} & \galcell{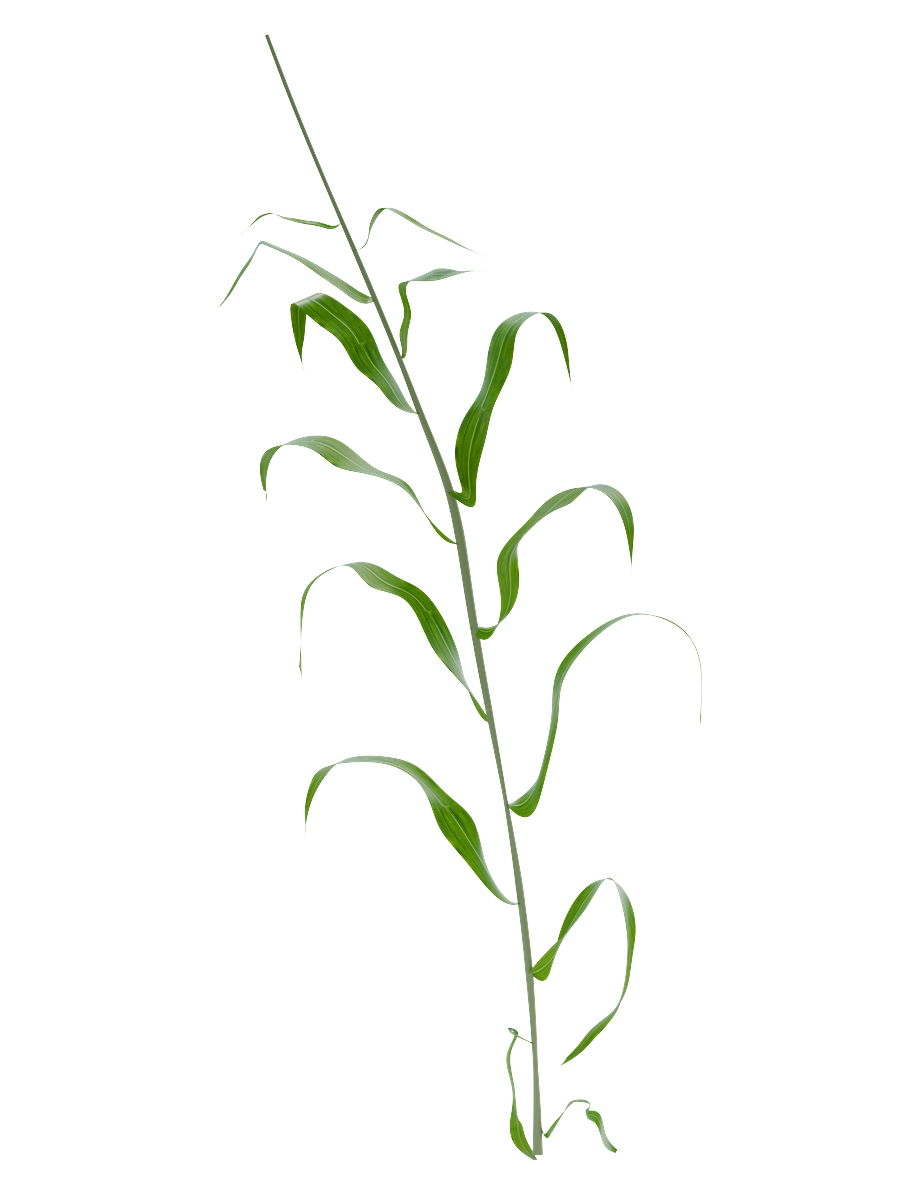} \\
\end{longtable}

\section{Per-plant fidelity and check activity}
\label{sec:appendix_cd}
\setcounter{figure}{0} \setcounter{table}{0}

\tabref{tab:AllPlantsCD} lists, for each evaluated plant, the median blade-level distance from the observed blade points to the final fitted surfaces, the whole-plant symmetric Chamfer distance of the assembled mesh, and the number of leaves on which each geometric modification of \secref{sec:nurbsdiff} was adopted. The table reports selected modifications, not all diagnostic evaluations. Rigid prealignment was accepted on $9$ leaves. The fold diagnostic flagged $28$ of the $109$ eligible fits, and $21$ fold repairs were adopted. CML69 and M162W account for $7$ of the $9$ accepted rotations. Neither modification was adopted on Kui2021. No base-collar repair was selected for the ten plants.

The two fidelity columns ranked individual plants differently, most visibly for A679, which had the lowest blade-level residual of the ten plants but one of the higher whole-plant distances. The blade-level value measured corrected blade points against their fitted leaf surfaces, whereas the whole-plant distance measured the assembled mesh against the complete scan. Unmodeled structures such as tassels and sheaths can contribute to the whole-plant distance, but these summary metrics do not identify which structures cause the difference for an individual plant. We therefore report both measures.

\begin{table}[h!]
    \centering
    \caption{Per-plant blade-level and whole-plant fidelity, with counts of
    adopted geometric modifications. Blade RMS uses the exact point-to-surface measure of \secref{sec:quantitative}, and Plant CD is computed against the input cloud. ``Rot.'' denotes the gated rigid rotation, and ``Fold'' denotes the mid-blade fold repair. The final row gives totals for the counts and medians for the distances.}
    \label{tab:AllPlantsCD}
    \setlength{\extrarowheight}{2pt}
    \footnotesize
    \begin{tabular}{@{}lrrrrr@{}}
        \toprule
        Genotype & Leaves & Blade RMS (mm) & Plant CD (mm) & Rot. & Fold \\
        \midrule
        M162W   & 15 & $5.8$ & $6.7$  & 3 & 3 \\
        CML69   & 15 & $5.5$ & $7.7$  & 4 & 3 \\
        Kui2021 & 12 & $6.0$ & $9.7$  & 0 & 0 \\
        CML238  & 12 & $5.0$ & $8.3$  & 1 & 0 \\
        CML331  & 12 & $6.1$ & $8.8$  & 0 & 3 \\
        4226    & 12 & $4.5$ & $5.9$  & 1 & 2 \\
        78551S  & 8  & $5.6$ & $5.8$  & 0 & 1 \\
        A679    & 8  & $3.8$ & $9.1$  & 0 & 3 \\
        CI90C   & 10 & $4.8$ & $6.2$  & 0 & 4 \\
        T8      & 14 & $4.4$ & $6.3$  & 0 & 2 \\
        \midrule
        All ten & 118 & $5.2$ & $7.2$ & 9 & 21 \\
        \bottomrule
    \end{tabular}
\end{table}

\section{Repeated annotation of a single view}
\label{sec:appendix:repeats}
\setcounter{figure}{0} \setcounter{table}{0}

\figref{fig:vlm_repeats} shows the five retained annotation outputs used for the reproducibility measurement of \secref{sec:seg_validation}. Each plant was annotated five times from the same rendered image with the same prompt and model. Per-plant mean (minimum) pairwise intersection over union of the complete colored annotation masks, including leaf curves and node markers, was $0.792$ ($0.764$) for Plant~A, $0.772$ ($0.739$) for Plant~B, $0.682$ ($0.626$) for Plant~C, and $0.670$ ($0.635$) for Plant~D. The narrow curves make this measure sensitive to small lateral displacements.

The panels show both lateral displacement of the curves and differences in annotation content; the mask measure does not separate these two sources. The displayed panels are common-cropped and downsampled copies, whereas the intersection-over-union values were computed from the raw returned images. These repeats quantified variability in one annotated view. Cross-view consensus requires support from at least two views, but this experiment did not measure the effect of single-view variability on the complete eight-view workflow or validate the two-view threshold. Measuring end-to-end variability would require repeating all eight annotations and the downstream stages.

\newcommand{\repeatpanel}[1]{%
    \begin{subfigure}[t]{0.185\textwidth}
        \centering
        \includegraphics[width=\textwidth]{#1}
    \end{subfigure}%
}

\begin{figure}[p]
    \centering
    \makebox[0.185\textwidth]{Run 1}\hfill
    \makebox[0.185\textwidth]{Run 2}\hfill
    \makebox[0.185\textwidth]{Run 3}\hfill
    \makebox[0.185\textwidth]{Run 4}\hfill
    \makebox[0.185\textwidth]{Run 5}

    \vspace{0.3em}

    \repeatpanel{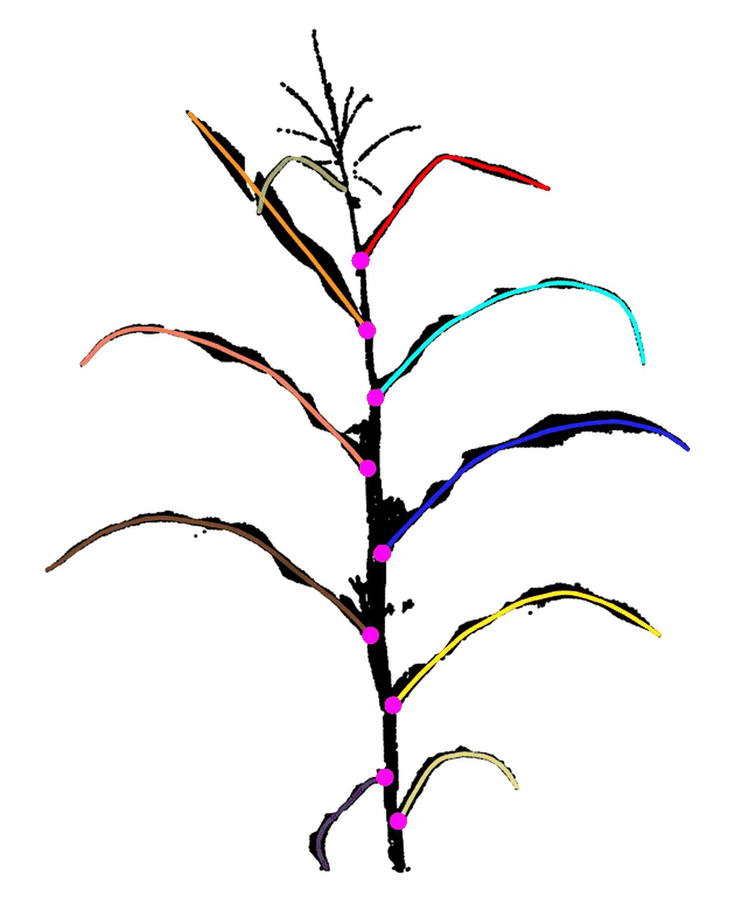}\hfill
    \repeatpanel{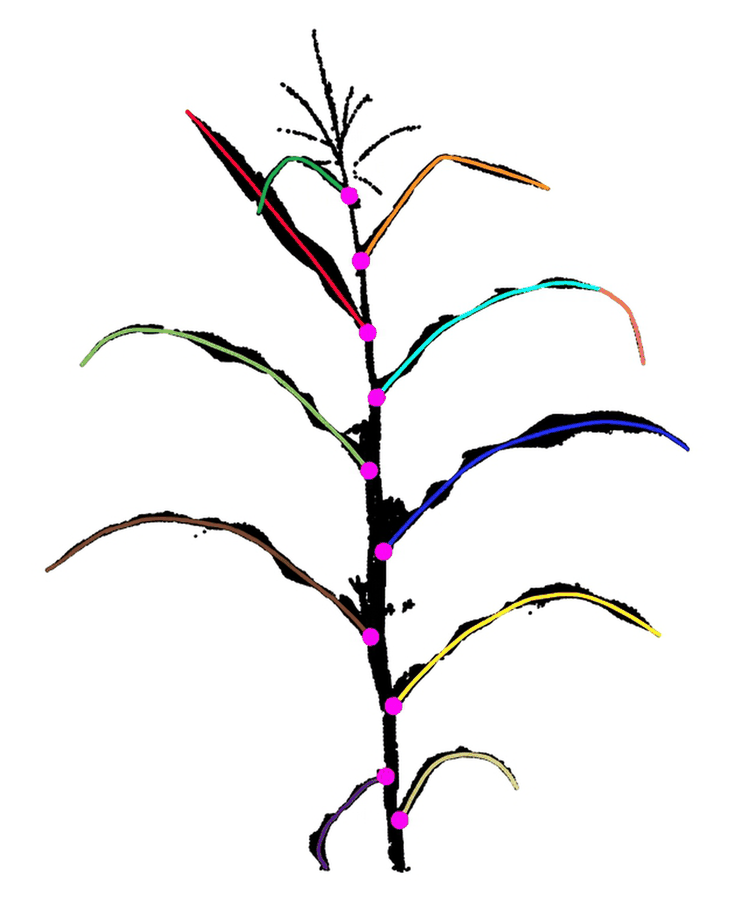}\hfill
    \repeatpanel{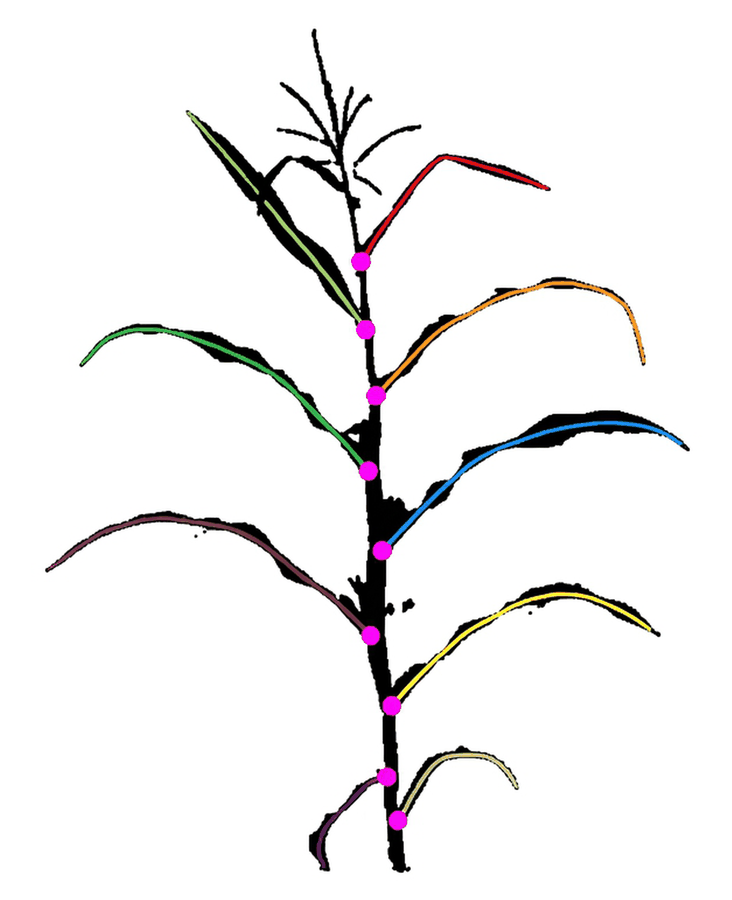}\hfill
    \repeatpanel{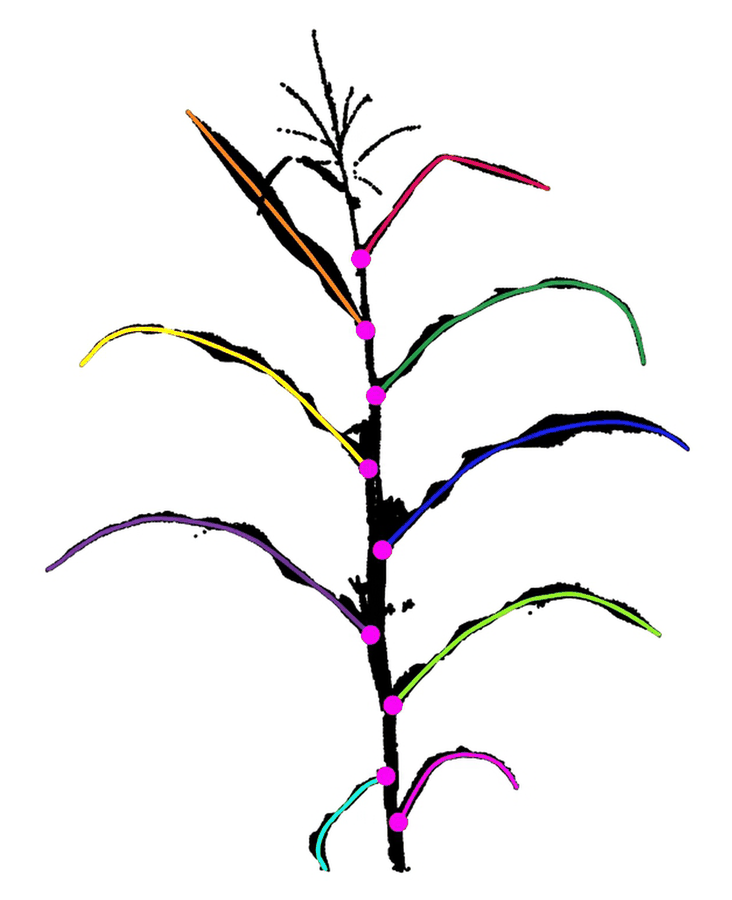}\hfill
    \repeatpanel{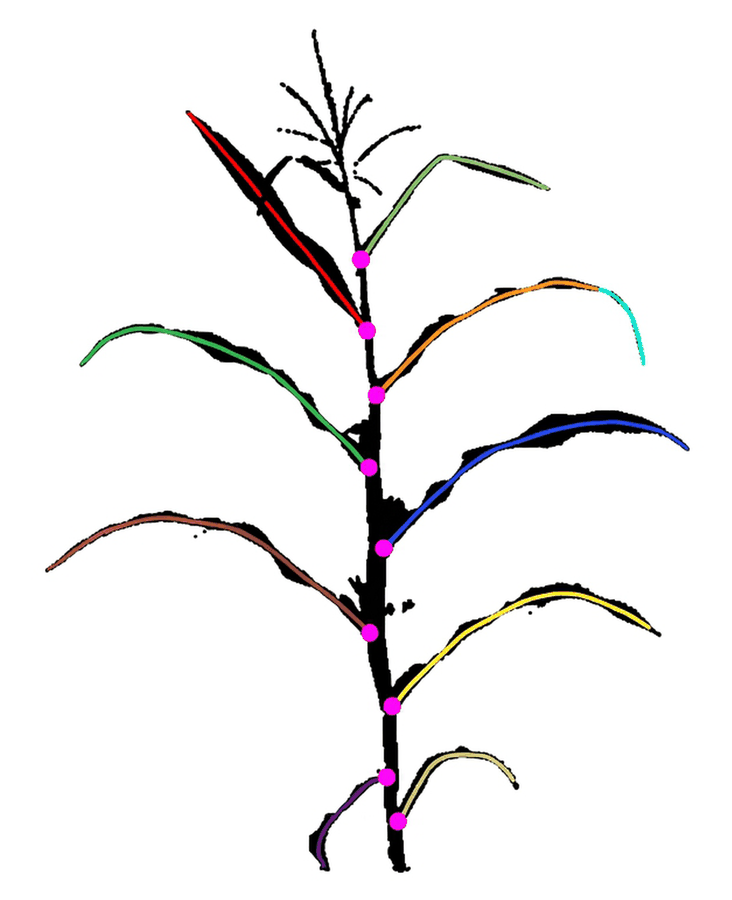}\par
    \vspace{0.25em}{\footnotesize Plant A (0003)}

    \vspace{0.4em}

    \repeatpanel{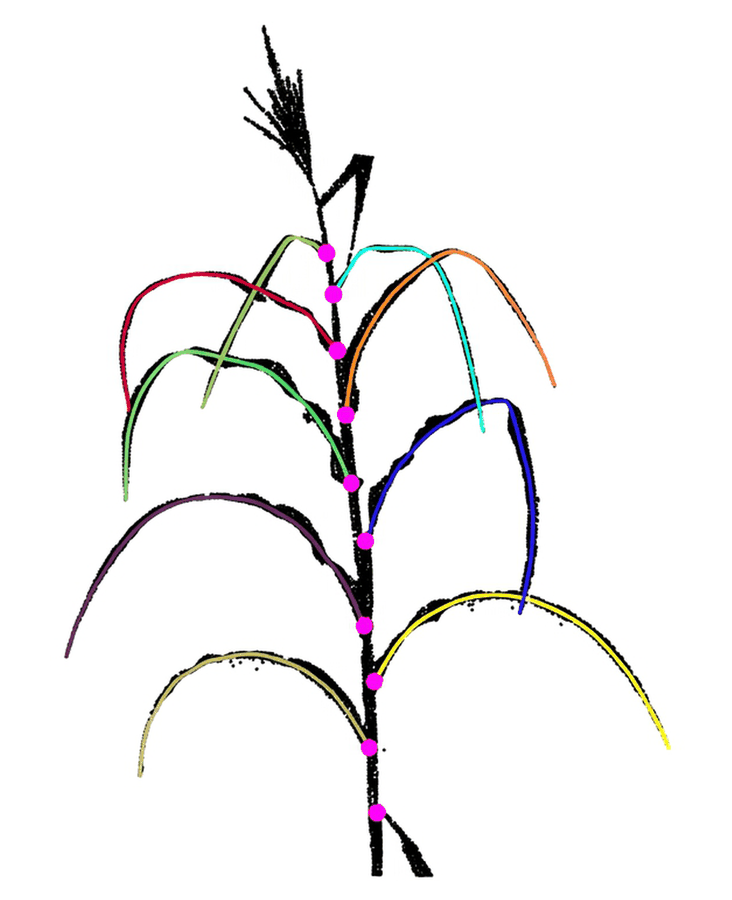}\hfill
    \repeatpanel{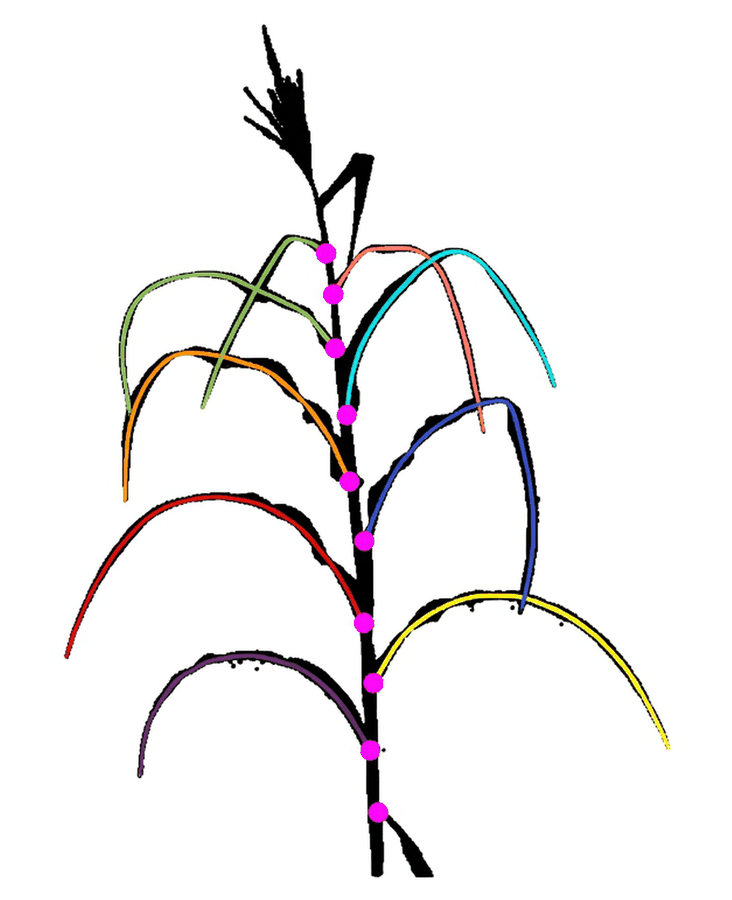}\hfill
    \repeatpanel{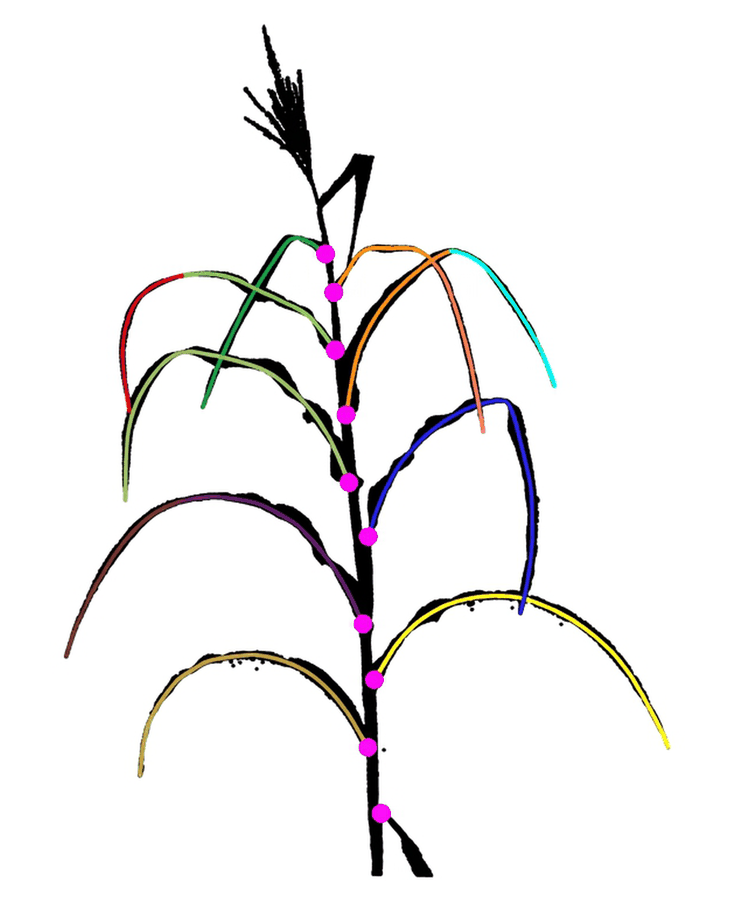}\hfill
    \repeatpanel{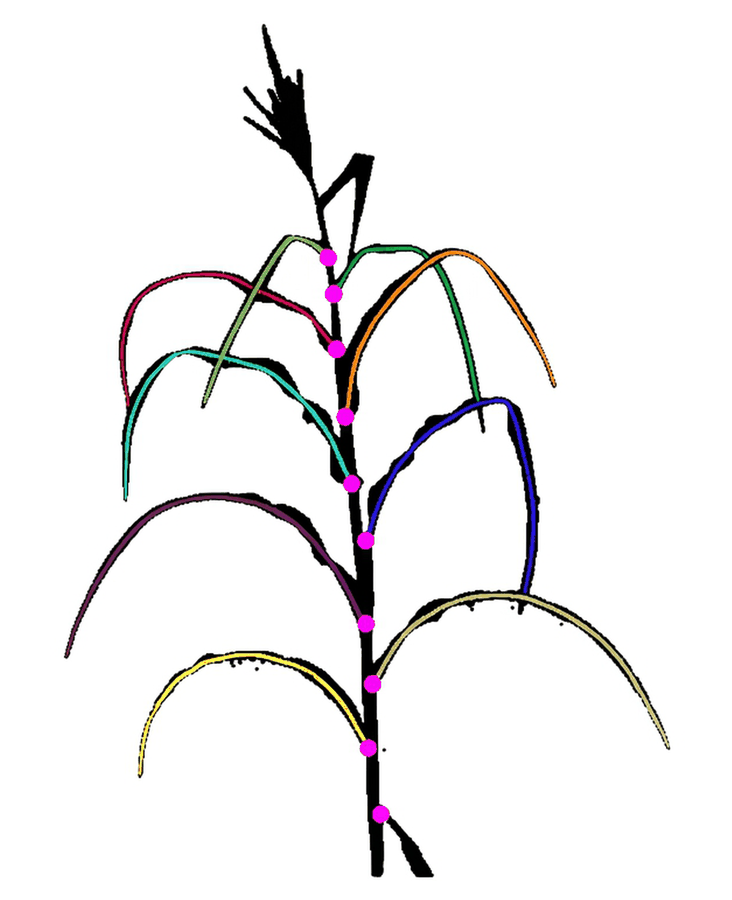}\hfill
    \repeatpanel{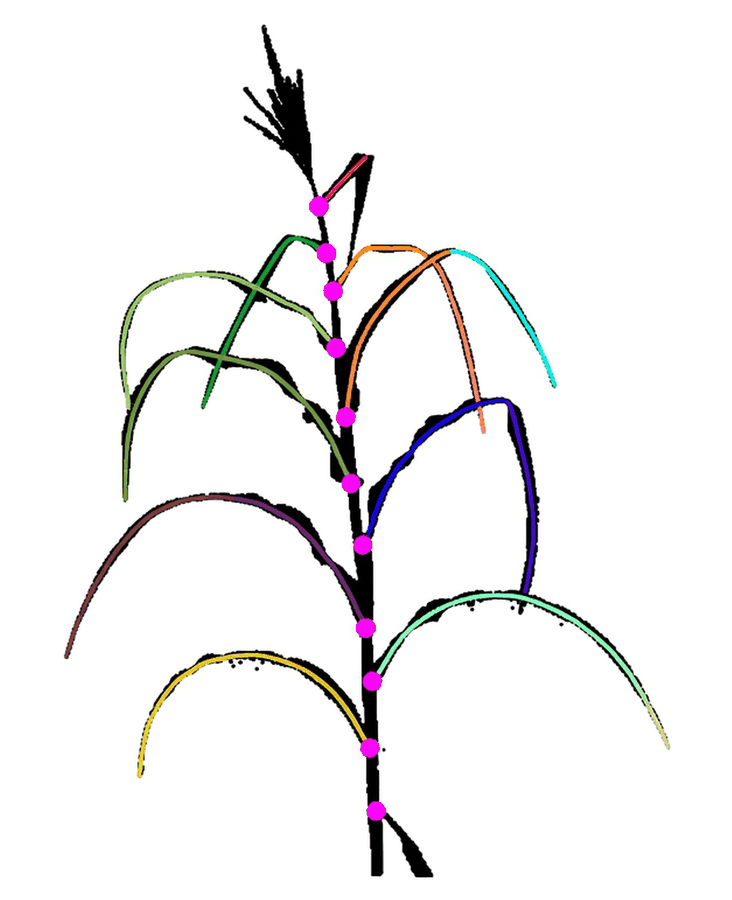}\par
    \vspace{0.25em}{\footnotesize Plant B (0139)}

    \vspace{0.4em}

    \repeatpanel{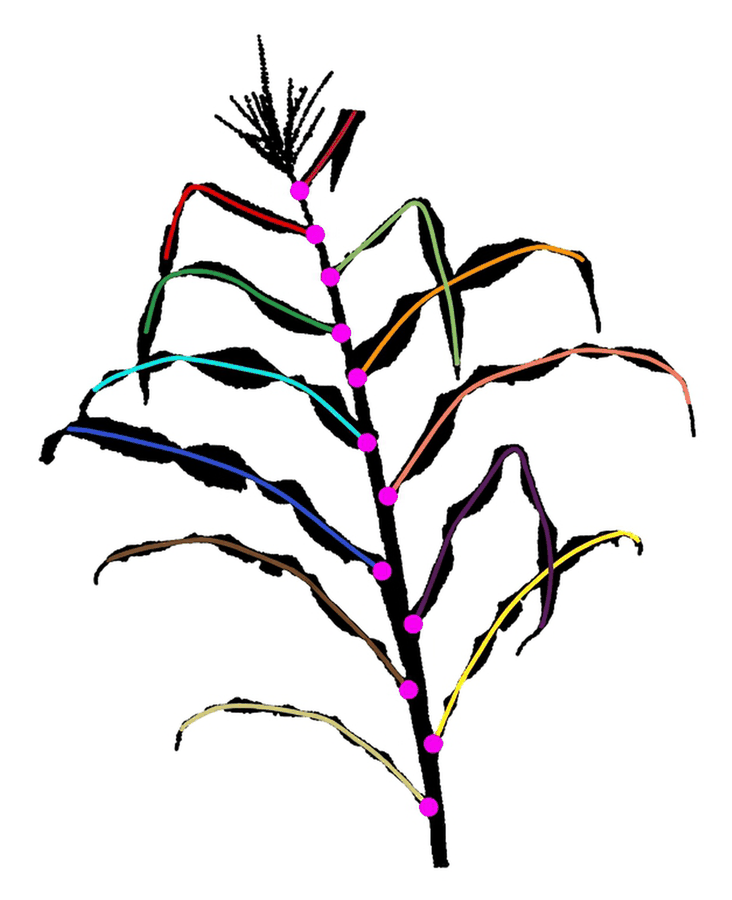}\hfill
    \repeatpanel{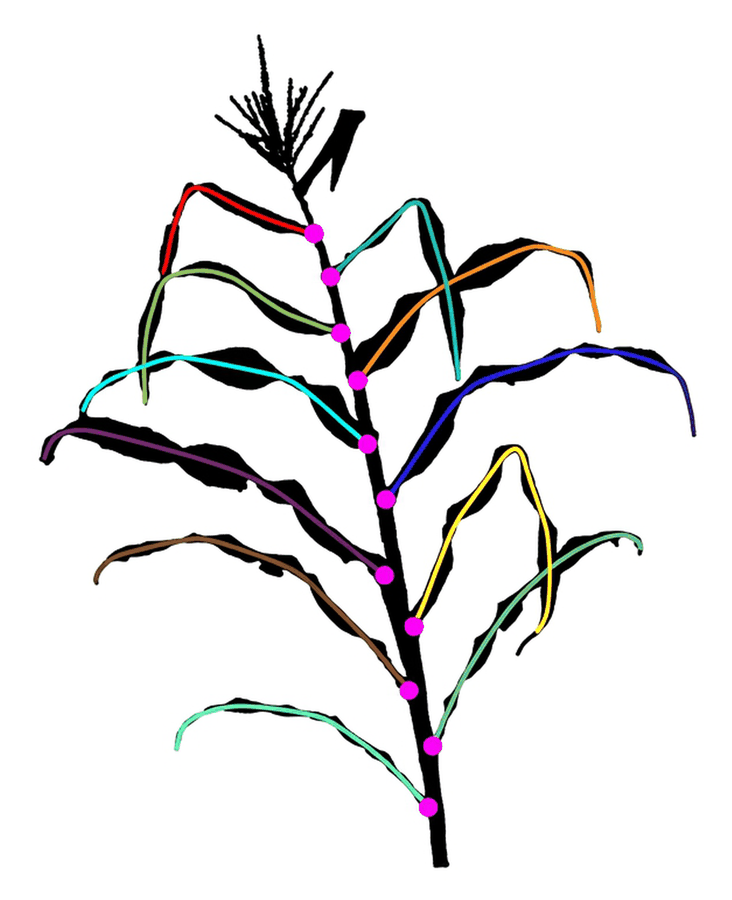}\hfill
    \repeatpanel{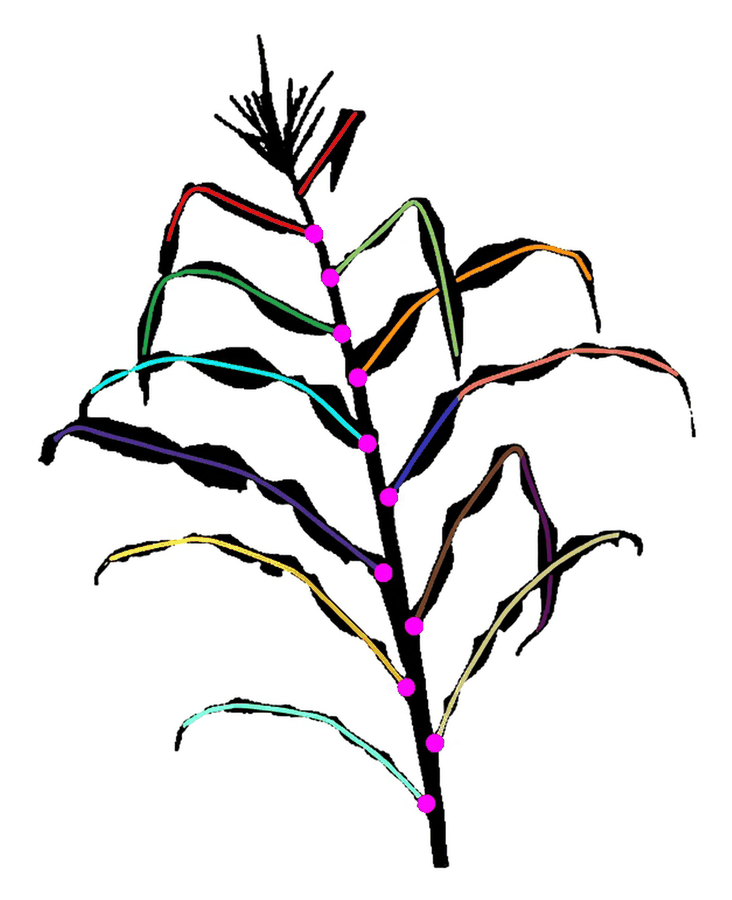}\hfill
    \repeatpanel{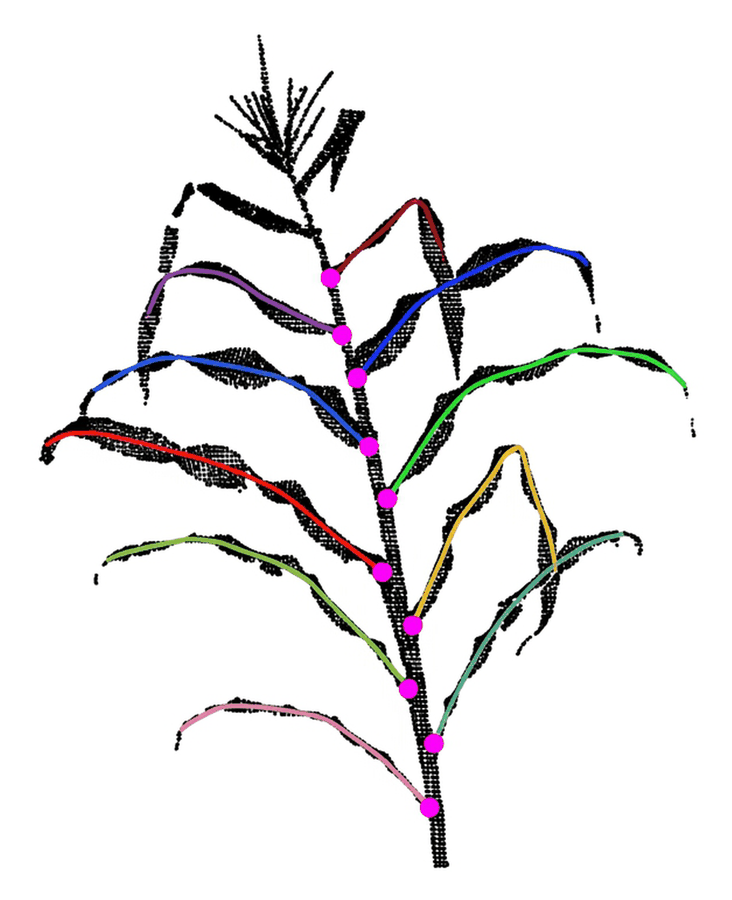}\hfill
    \repeatpanel{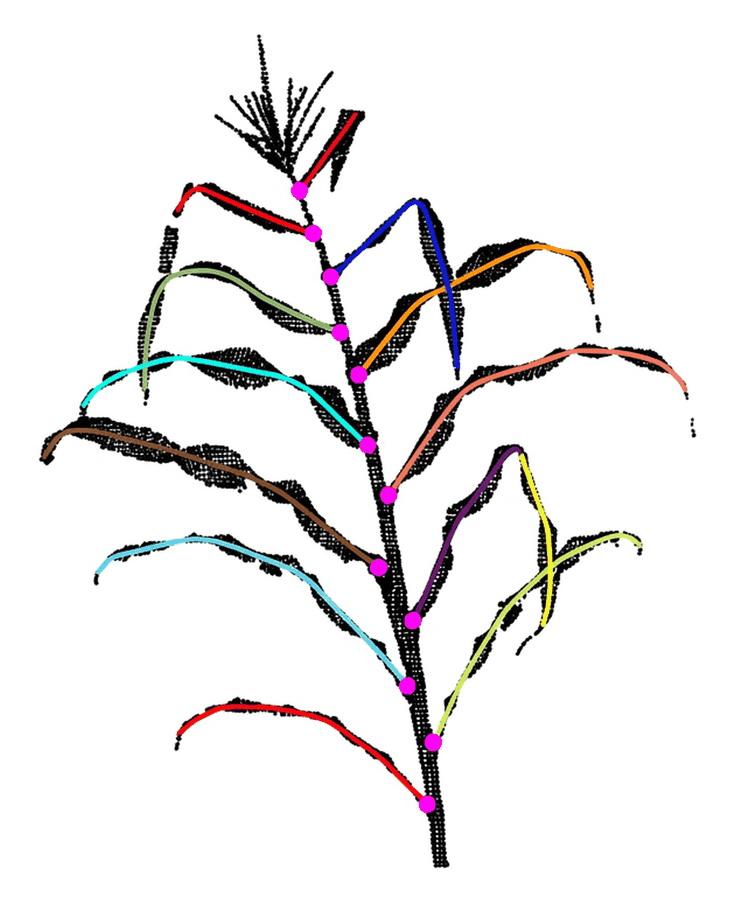}\par
    \vspace{0.25em}{\footnotesize Plant C (0181)}

    \vspace{0.4em}

    \repeatpanel{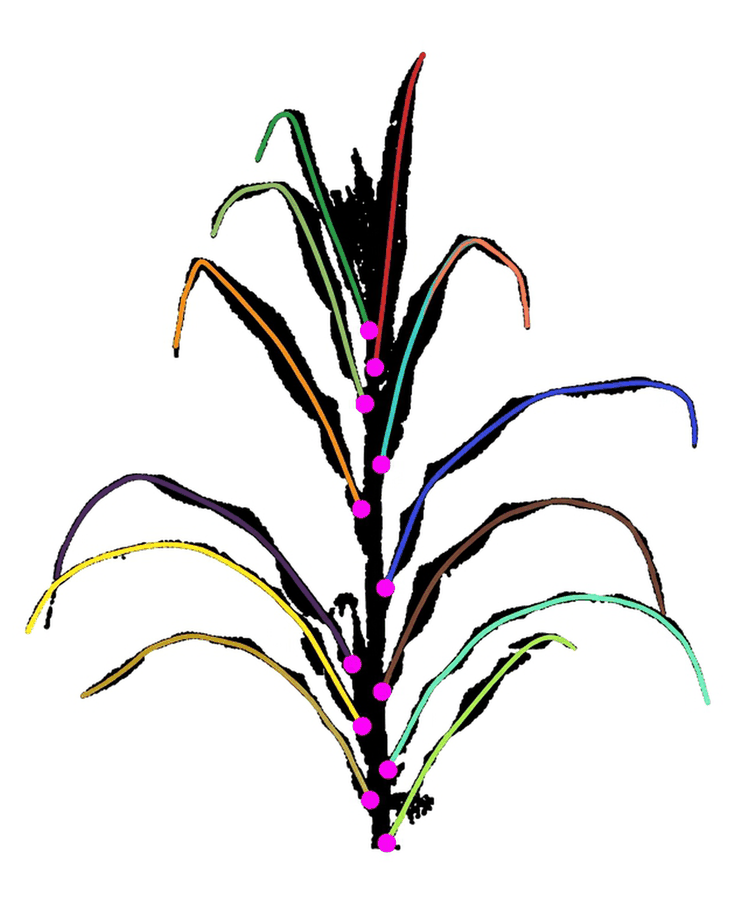}\hfill
    \repeatpanel{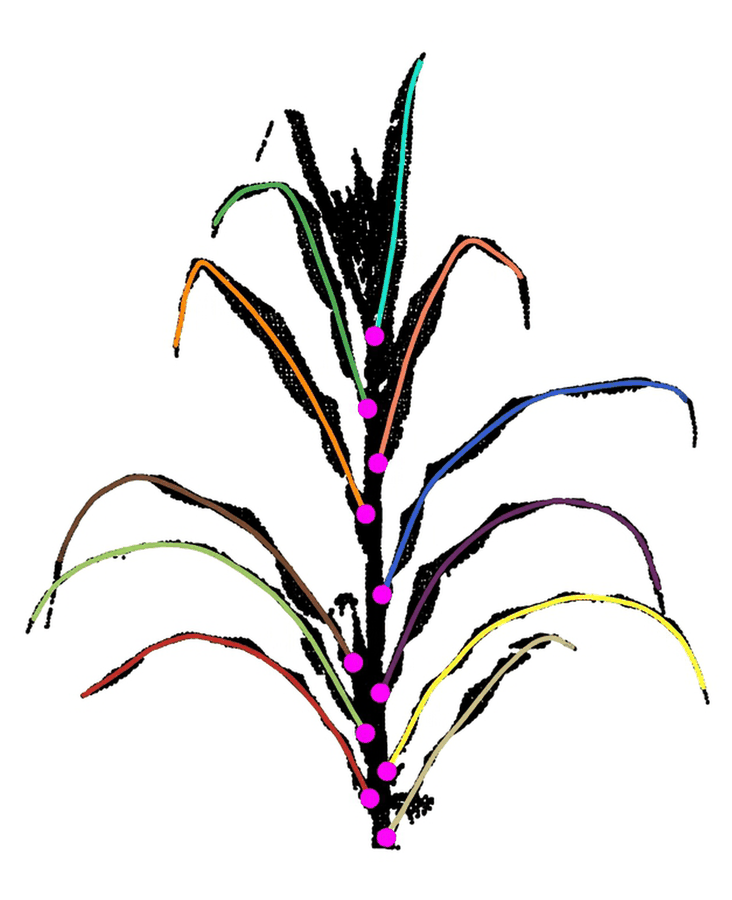}\hfill
    \repeatpanel{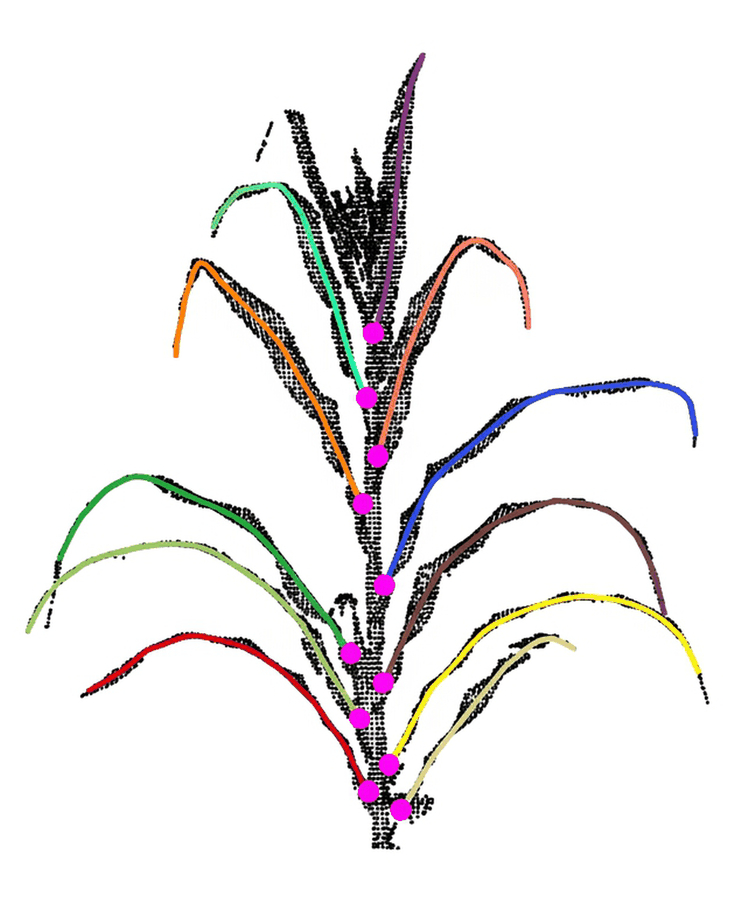}\hfill
    \repeatpanel{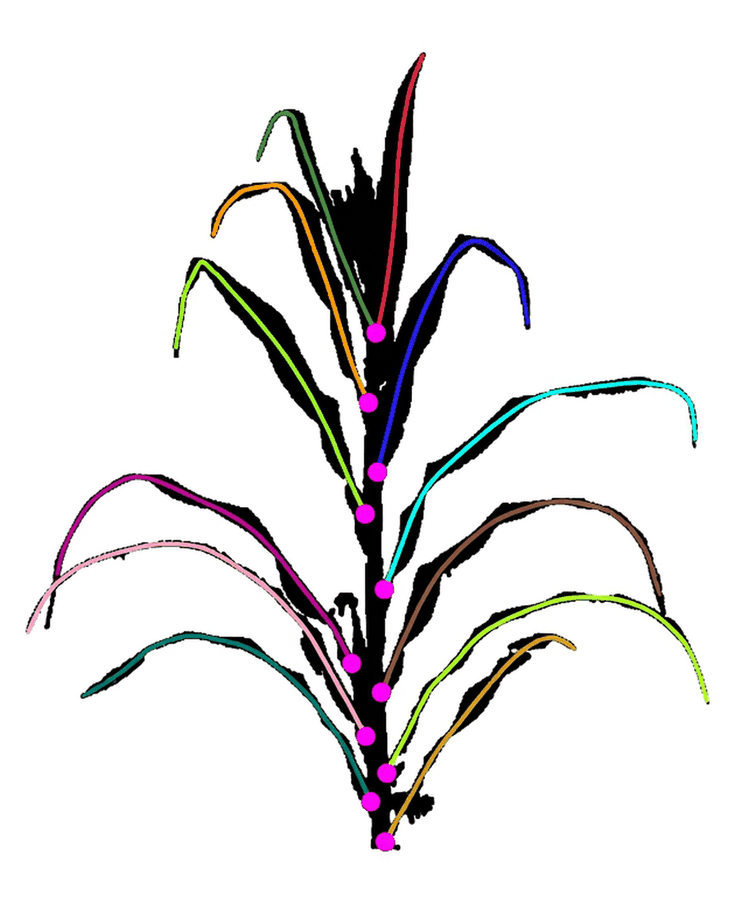}\hfill
    \repeatpanel{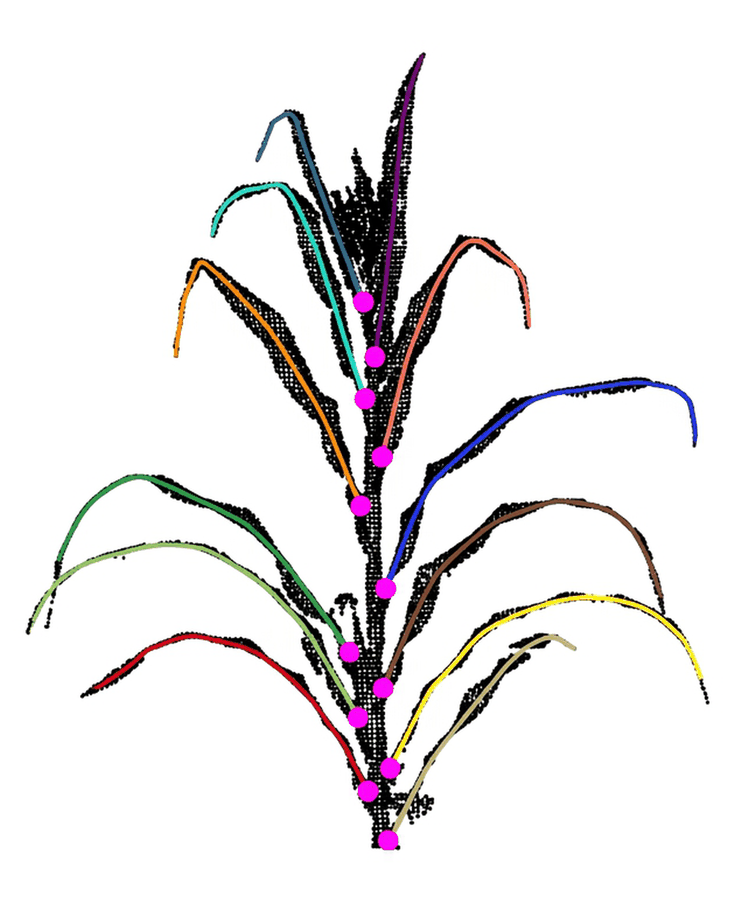}\par
    \vspace{0.25em}{\footnotesize Plant D (0089)}

    \caption{Five repeated front-view annotations of each of four plants, drawn
    from the same rendered image on all calls. Rows are plants, ordered by decreasing mean pairwise intersection over union of the complete colored annotation masks across the ten within-plant pairs, and columns are replicate annotation tasks. Colored curves are the returned leaf midlines, and magenta markers are candidate node positions.}
    \label{fig:vlm_repeats}
\end{figure}

\section{Probe of a second image-output model}
\label{sec:appendix:second_model}
\setcounter{figure}{0} \setcounter{table}{0}

Annotation is the only stage of the pipeline that calls a hosted model. To examine model dependence at this stage, we repeated the experiment above with a second image-output model, \textsf{gpt-image-2}~\citep{openai2026gptimage2}, using the same four plants, front-view renders, prompt, and five calls per plant. \figref{fig:vlm_second} shows the returned images. The probe ended at the returned annotation images; the outputs were not passed through mask extraction or the downstream reconstruction stages.

All $20$ calls returned colored curves and magenta components in the encoding used by the extractor. To quantify image-foreground overlap, we resized each input render to the returned resolution of $1024\times1024$ pixels and classified a pixel as foreground when any RGB channel was below $235$. We computed foreground-mask recall as $|M_{\mathrm{input}}\cap M_{\mathrm{return}}|/|M_{\mathrm{input}}|$. Foreground-mask recall averaged $78.1\%$ over the $20$ returns and ranged from $28.7\%$ to $98.5\%$. For Plant~B, foreground-mask recall was $98.0\%$ on run~1, $34.8\%$ on run~2, and $28.7\%$ on run~4. This measure counts all thresholded image foreground, including the render overlays, and therefore does not establish exact pixel preservation or 3D point registration.

Visual inspection also showed partial midlines, blades divided between two colors, and magenta components placed away from the stalk. The detected magenta-component count ranged from $10$ to $18$ per call for Plant~A, compared with $9$ to $10$ for the primary model, and from $15$ to $21$ for Plant~C, compared with $11$ to $12$. Because the downstream stages were not run on the second-model outputs, these observations do not establish errors in the exported leaves or nodes. Partial or split curves and misplaced components could weaken the per-view evidence. Mask regrouping, cross-view consensus, blade propagation, and final-node repair could correct or discard some of these defects.

We report this comparison to characterize the requirements of the annotation stage under the tested API configuration, not to rank two services. A substitute model must preserve the pixel-to-point correspondence sufficiently for back-projection and return annotations that the extractor can process. The retained primary-model repeat outputs passed the pipeline validation checks, but exact pixel preservation was not measured. The reconstruction results in this paper were produced with the model named in \secref{sec:vlm}.

\newcommand{\secondpanel}[1]{%
    \begin{subfigure}[t]{0.185\textwidth}
        \centering
        \includegraphics[width=\textwidth]{#1}
    \end{subfigure}%
}

\begin{figure}[p]
    \centering
    \makebox[0.185\textwidth]{Run 1}\hfill
    \makebox[0.185\textwidth]{Run 2}\hfill
    \makebox[0.185\textwidth]{Run 3}\hfill
    \makebox[0.185\textwidth]{Run 4}\hfill
    \makebox[0.185\textwidth]{Run 5}

    \vspace{0.3em}

    \secondpanel{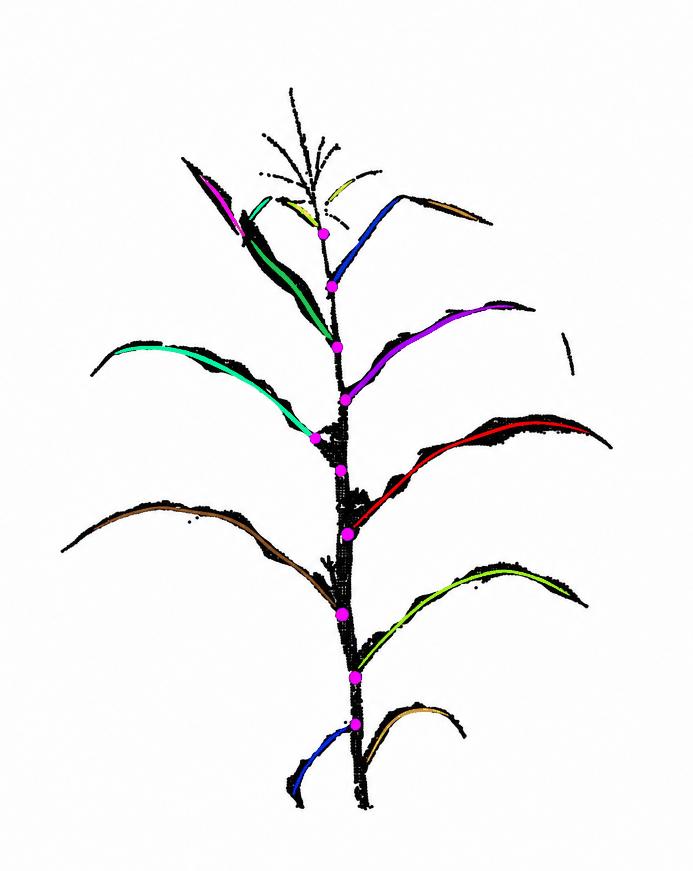}\hfill
    \secondpanel{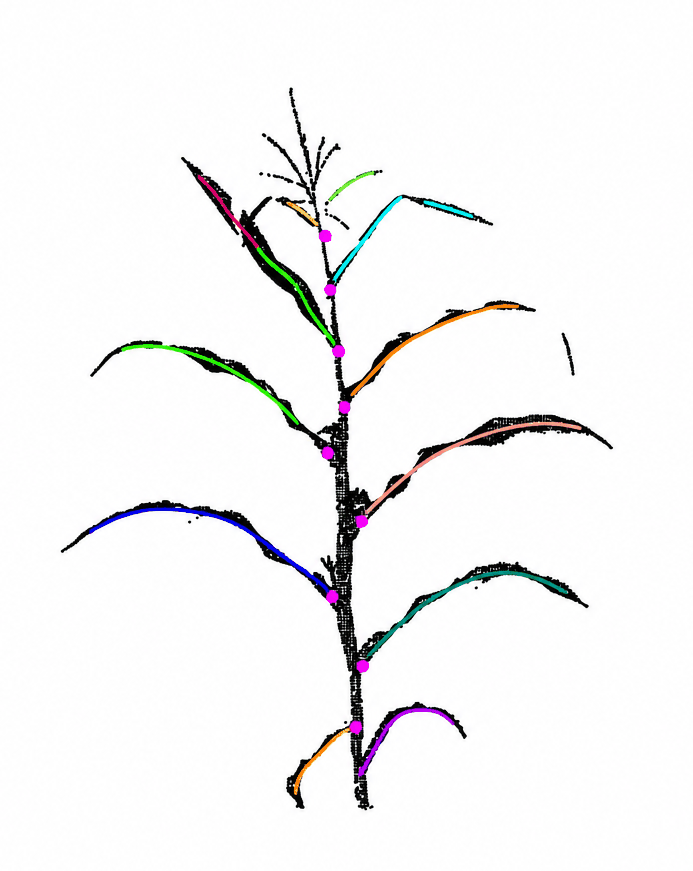}\hfill
    \secondpanel{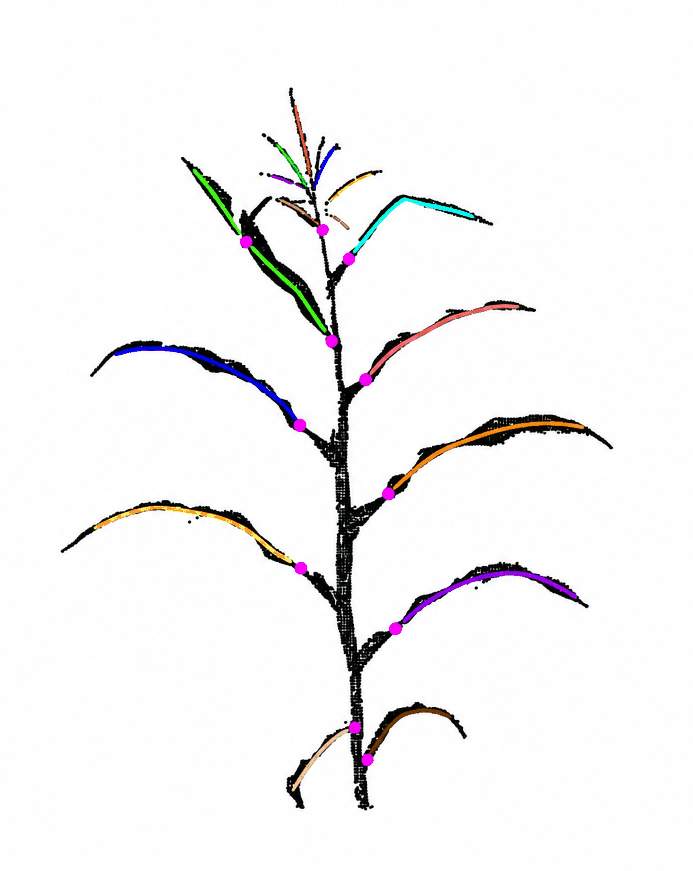}\hfill
    \secondpanel{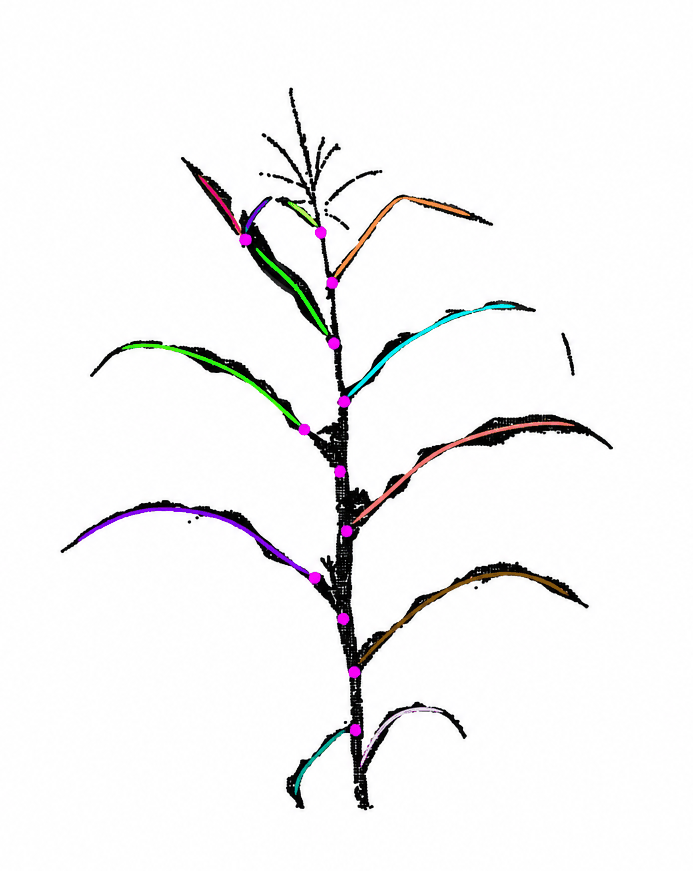}\hfill
    \secondpanel{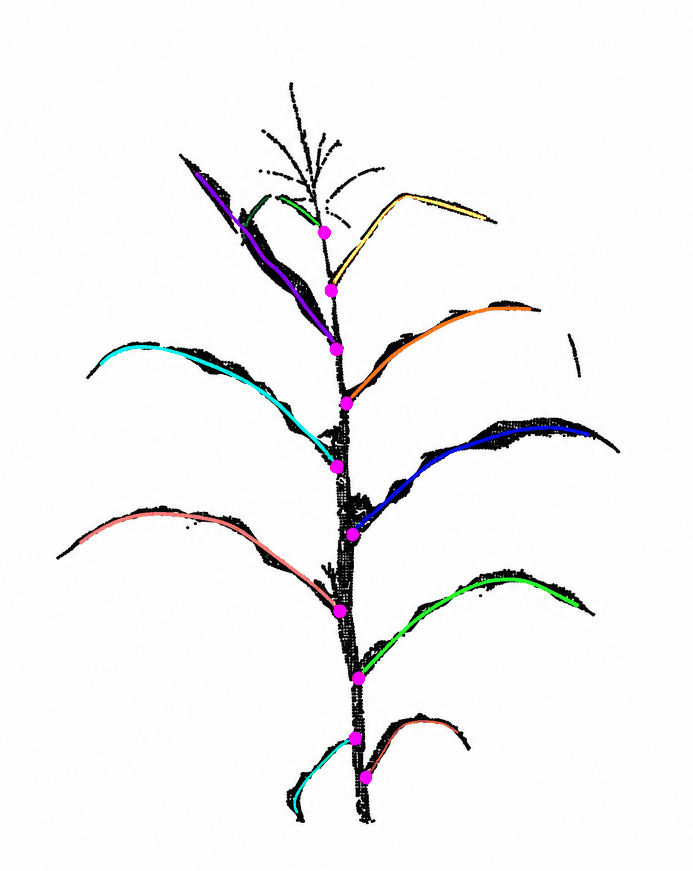}\par
    \vspace{0.25em}{\footnotesize Plant A (0003)}

    \vspace{0.4em}

    \secondpanel{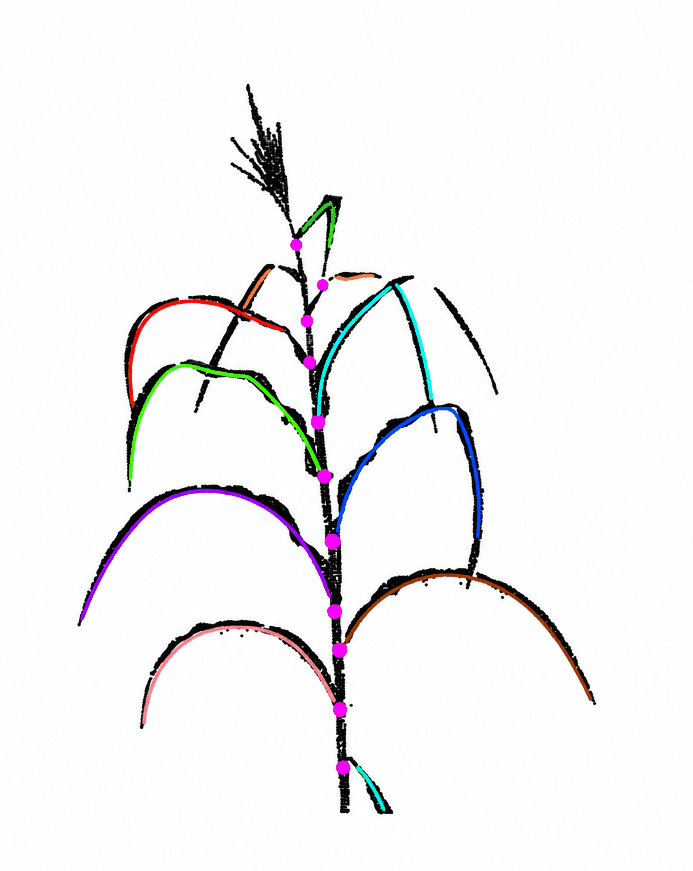}\hfill
    \secondpanel{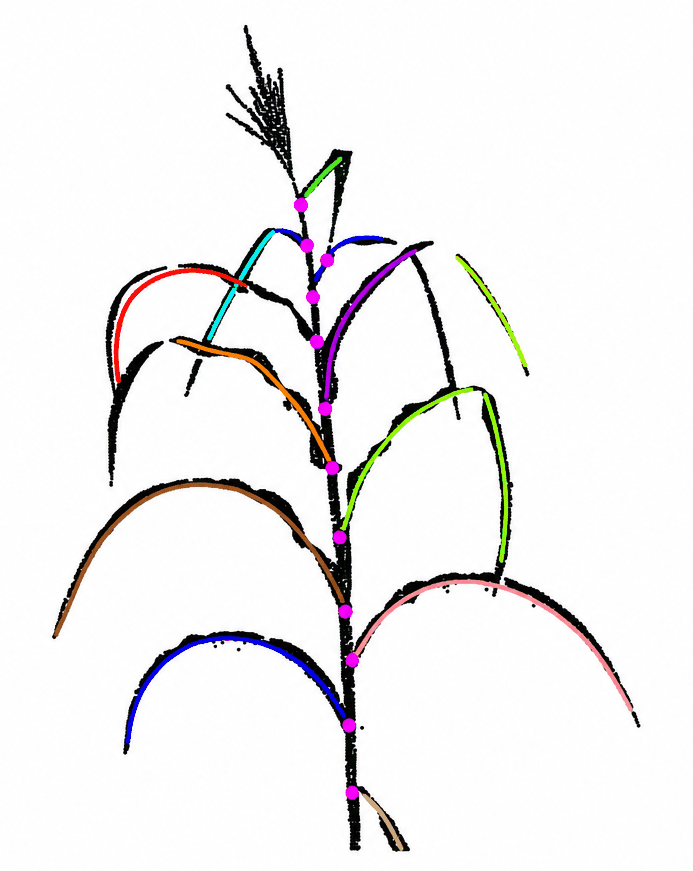}\hfill
    \secondpanel{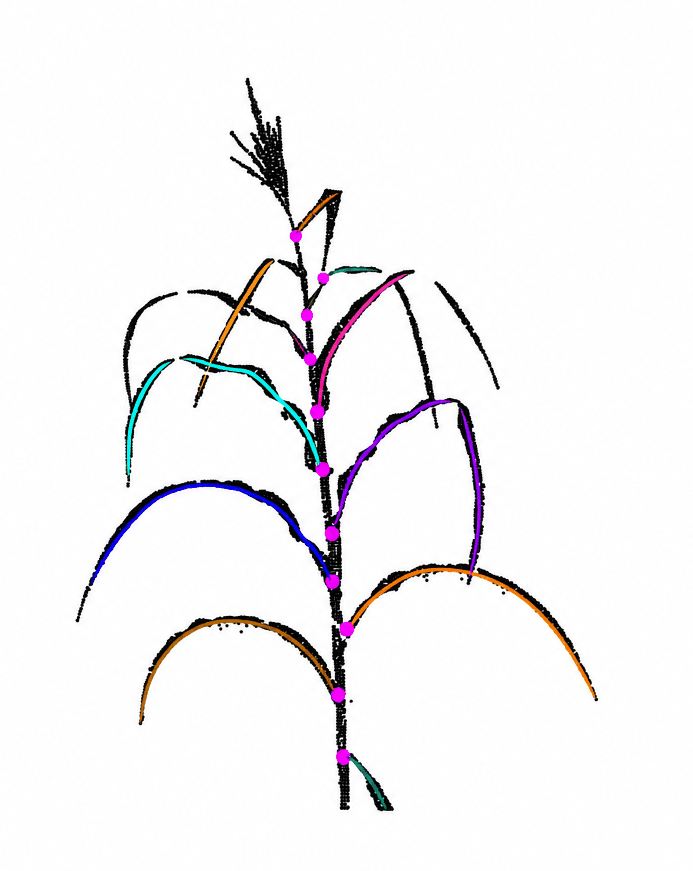}\hfill
    \secondpanel{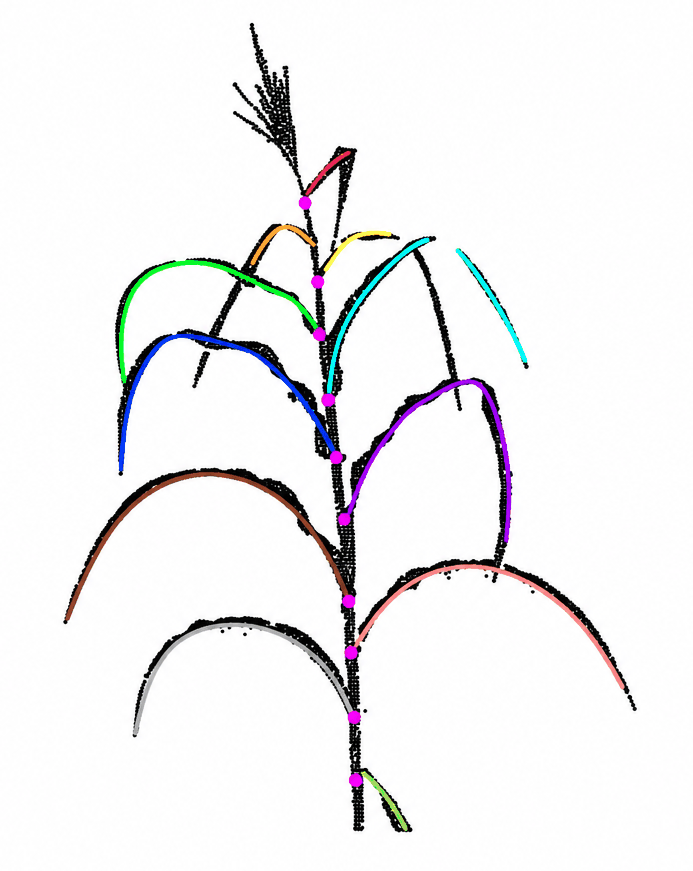}\hfill
    \secondpanel{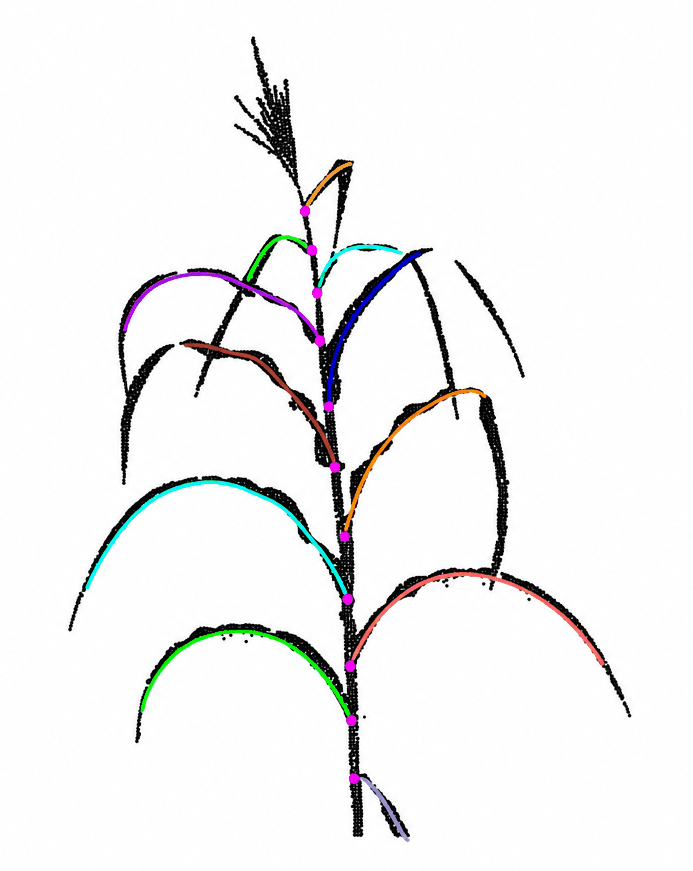}\par
    \vspace{0.25em}{\footnotesize Plant B (0139)}

    \vspace{0.4em}

    \secondpanel{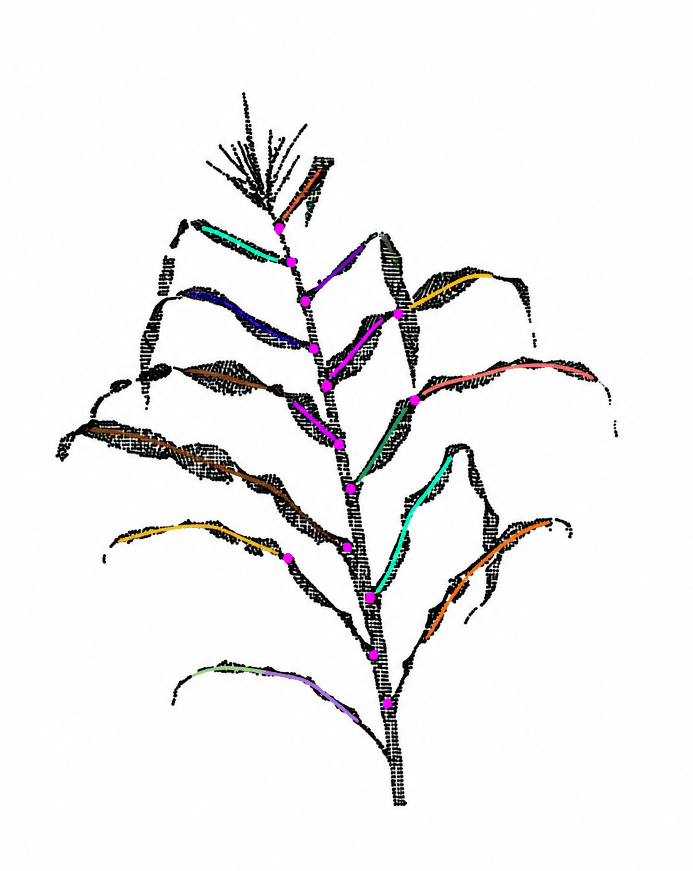}\hfill
    \secondpanel{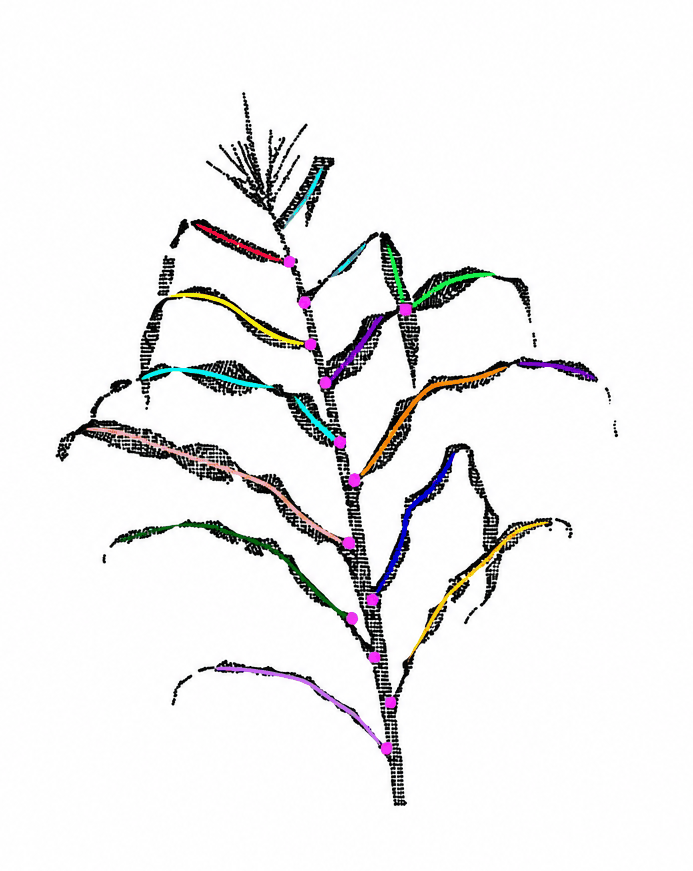}\hfill
    \secondpanel{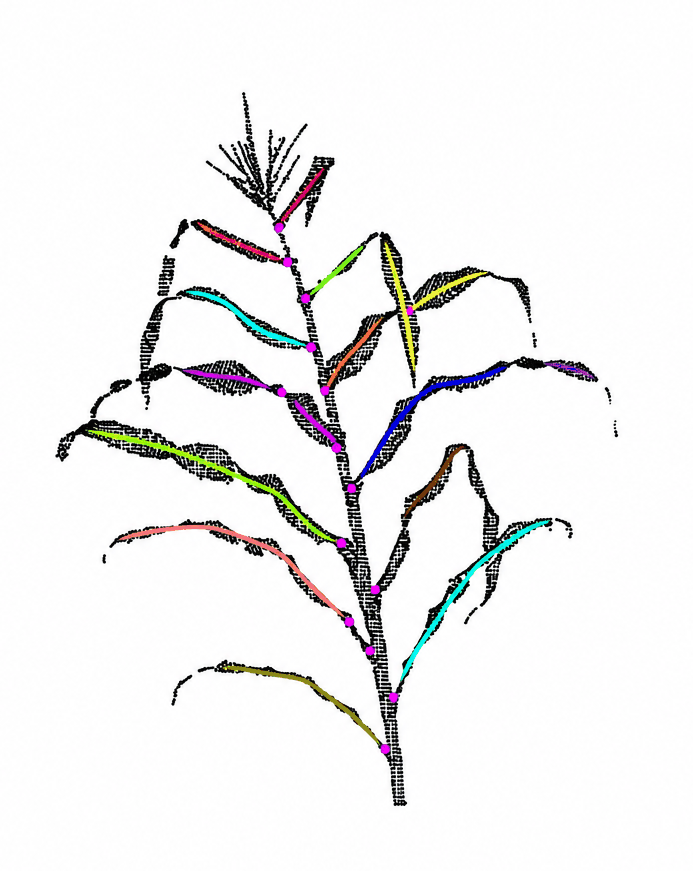}\hfill
    \secondpanel{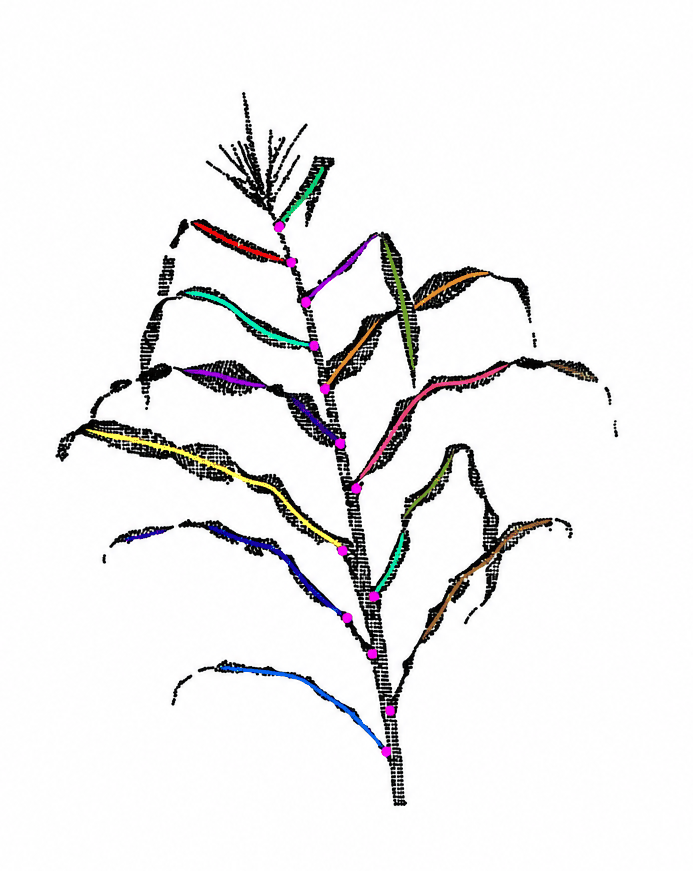}\hfill
    \secondpanel{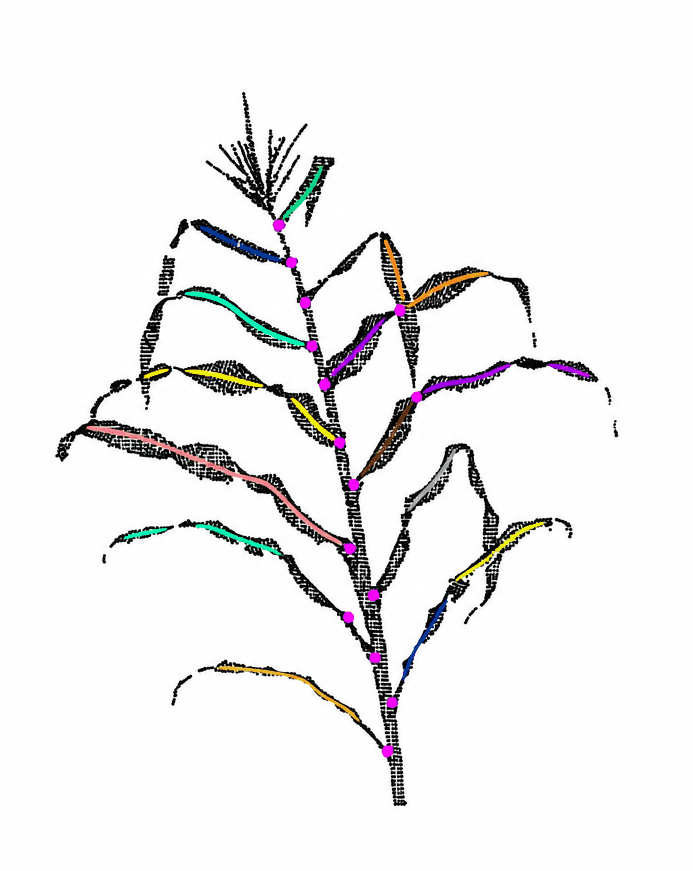}\par
    \vspace{0.25em}{\footnotesize Plant C (0181)}

    \vspace{0.4em}

    \secondpanel{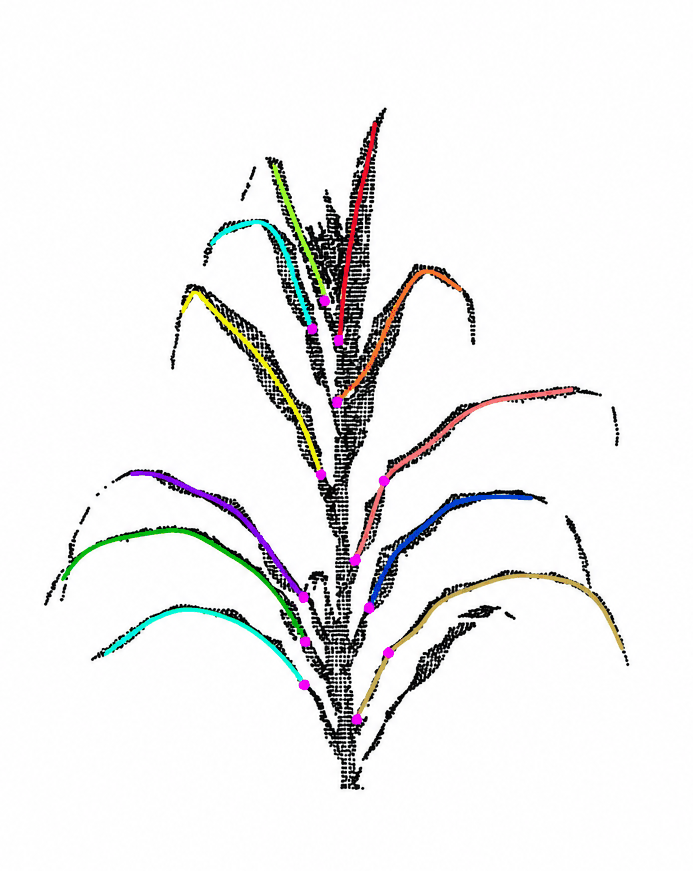}\hfill
    \secondpanel{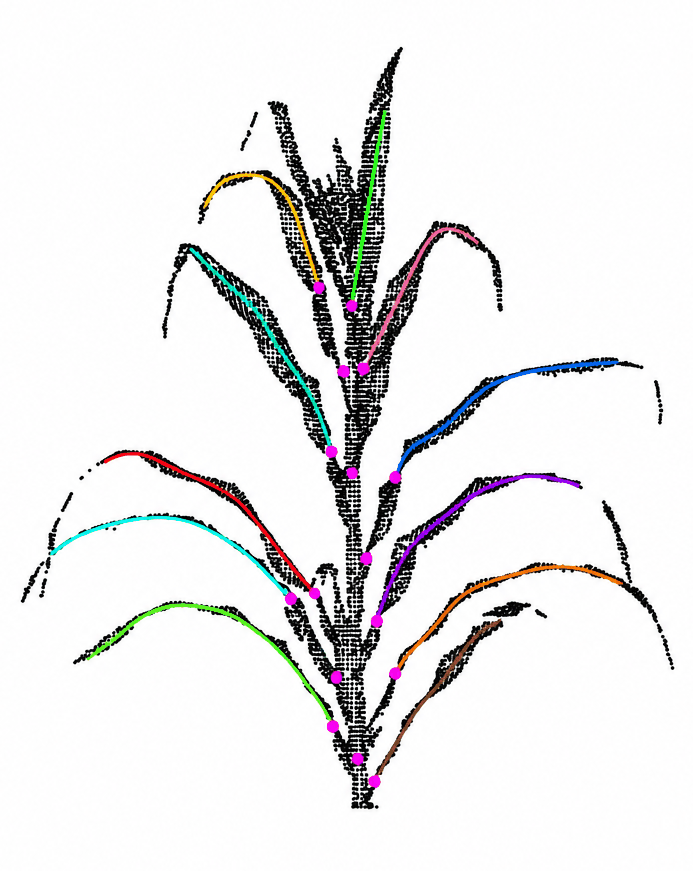}\hfill
    \secondpanel{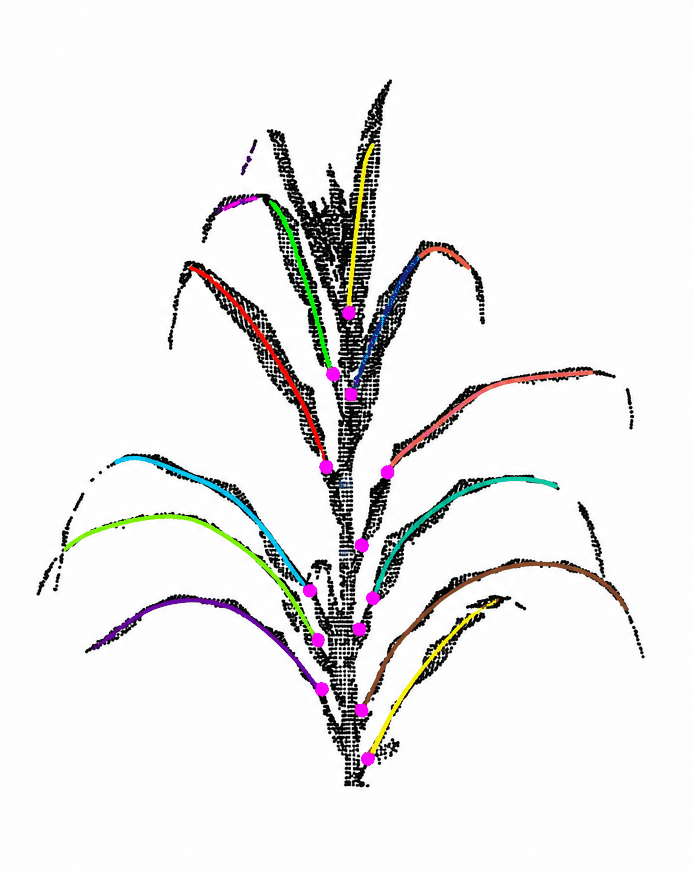}\hfill
    \secondpanel{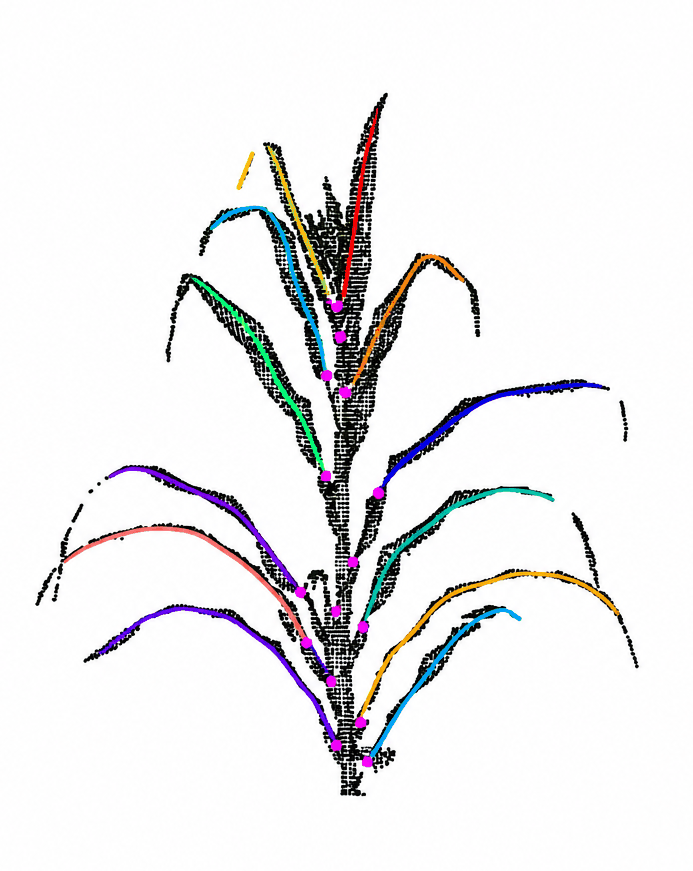}\hfill
    \secondpanel{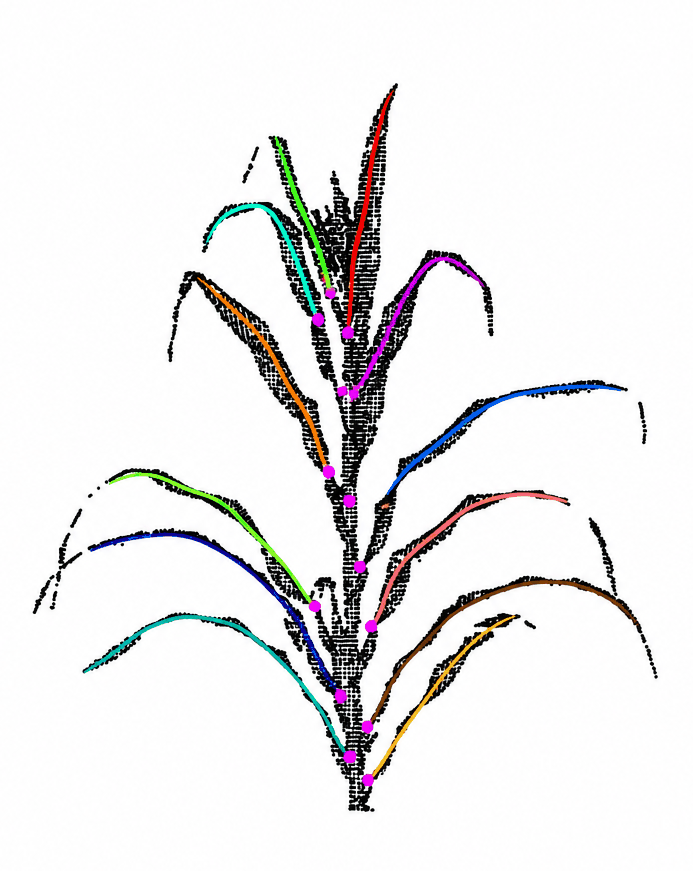}\par
    \vspace{0.25em}{\footnotesize Plant D (0089)}

    \caption{Returned annotation images from the second-model probe with \textsf{gpt-image-2} on the same four plants, renders, prompt, and replicate order as \figref{fig:vlm_repeats}.}
    \label{fig:vlm_second}
\end{figure}

\end{document}